\documentclass[10pt,journal,compsoc]{IEEEtran}
\ifCLASSOPTIONcompsoc
  \usepackage[nocompress]{cite}
\else
  \usepackage{cite}
\fi
\usepackage{multirow}
\usepackage{bm}
\usepackage{algorithm}
\usepackage{algorithmic}
\usepackage{balance}
\usepackage{fancyhdr}
\usepackage{subfigure}
\usepackage{rotating}
\usepackage{amsmath}
\usepackage{color}
\usepackage{hyperref}

\ifCLASSINFOpdf
\else
\fi
\begin{document}
%
\title{A Task-guided, Implicitly-searched  and Meta- initialized Deep Model for Image Fusion}
%
%
%
%

\author{
	Risheng~Liu,~\IEEEmembership{Member,~IEEE,}
	Zhu~Liu, Jinyuan~Liu,  Xin~Fan,~\IEEEmembership{Senior Member,~IEEE,} Zhongxuan Luo
\IEEEcompsocitemizethanks{\IEEEcompsocthanksitem Risheng. Liu is with DUT-RU International School of Information Science  $\&$
	Engineering, Dalian University of Technology, Dalian, 116024, China and is also with Peng Cheng Laboratory, Shenzhen, 518055, China.
	(Corresponding author, e-mail: rsliu@dlut.edu.cn).
	
	\IEEEcompsocthanksitem Zhu Liu is with the School of Software Technology, Dalian University of
	Technology, Dalian, 116024, China. (e-mail: liuzhu@mail.dlut.edu.cn).
	\IEEEcompsocthanksitem Jinyuan Liu is with the School of Software Technology, Dalian University of
	Technology, Dalian, 116024, China. (e-mail: atlantis918@hotmail.com).
	\IEEEcompsocthanksitem Xin. Fan is with the DUT-RU International School of Information Science
	  $\&$ Engineering, Dalian University of Technology, Dalian, 116024, China.
	(email: xin.fan@dlut.edu.cn).
	\IEEEcompsocthanksitem Zhongxuan. Luo is with the School of Software Technology, Dalian University of
	Technology, Dalian, 116024, China. (email: zxluo@dlut.edu.cn)
	}
\thanks{Manuscript received April 19, 2005; revised August 26, 2015.}}

%
%

\markboth{Journal of \LaTeX\ Class Files,~Vol.~14, No.~8, August~2015}%
{Shell \MakeLowercase{{et al.}}: Bare Demo of IEEEtran.cls for Computer Society Journals}
%



\IEEEtitleabstractindextext{%
\begin{abstract}
Image fusion plays a key role in a variety of multi-sensor-based vision systems, especially for enhancing visual quality and/or extracting aggregated features for perception. However, most existing methods just consider image fusion as an individual task, thus ignoring its underlying relationship with these downstream vision problems. Furthermore, designing proper fusion architectures often requires huge engineering labor. It also lacks mechanisms to improve the flexibility and generalization ability of current fusion approaches. To mitigate these issues, we establish a Task-guided, Implicit-searched and Meta-initialized (TIM) deep model to address the image fusion problem in a challenging real-world scenario. Specifically, we first propose a constrained strategy to incorporate information from downstream tasks to guide the unsupervised learning process of image fusion. Within this framework, we then design an implicit search scheme to automatically discover compact architectures for our fusion model with high efficiency. In addition, a pretext meta initialization technique is introduced to leverage divergence fusion data to support fast adaptation for different kinds of image fusion tasks. Qualitative and quantitative experimental results on different categories of image fusion problems and related downstream tasks (e.g., visual enhancement and semantic understanding) substantiate the flexibility and effectiveness of our TIM. The source code will be  available at \url{https://github.com/LiuZhu-CV/TIMFusion}.
\end{abstract}

}

\maketitle

\IEEEdisplaynontitleabstractindextext

\IEEEpeerreviewmaketitle

\section{Introduction}\label{sec:introduction}
\IEEEPARstart{I}mage fusion is a fundamental technique for visual perception and facilitates wide vision applications, e.g., visual enhancement~\cite{TarDAL,liu2020bilevel,liu2023holoco,jiang2022towards} and semantic understanding~\cite{ma2022pia,takumi2017multispectral,zhou2021edge,zhougm}.
Over the past few years, deep learning technology has increasingly energized image fusion approaches, achieving state-of-the-art performance. Unfortunately, three aspects of these approaches can be improved.~(i) Most of them focus on promoting the visual effects of the fused images rather than considering the downstream vision tasks, placing an obstacle to scene understanding applications. (ii)~Current fusion methods design handcrafted architectures with an increase of depth or width, which rely on verbose dedicated adjustments; thus, they inevitably cause time-consuming structural engineering. (iii)~These methods are learned with specific training data, which cannot acquire the generalization ability for various fusion scenarios. In this part, we first briefly discuss their major shortcomings of  learning-based methods and then propose our core contribution. 


First,  existing methods mostly concentrate on image fusion individually and few consider the underlying relationship of downstream vision tasks with fusion.  There are two categories of learning-based fusion methods, including conventional framework with plugging learnable mechanisms and end-to-end learning. In detail,  the first category of fusion first utilizes learnable auto-encoders to extract the features and leverages traditional fusion rules (e.g., $\ell_{1}$ norm~\cite{li2018densefuse}, weighted average~\cite{zhao2021efficient} and maxing choose~\cite{ZhaoDIDFuse2020}) to fuse the features of different modalities. The fused images are reconstructed by the corresponding decoders.  These handcrafted fusion rules actually realize the simple information aggregation. These  methods are limited by handcrafted strategies, which cannot accomplish adaptive modal feature preservation. On the other hand,  end-to-end learning-based schemes have been proposed to straightforwardly fuse diverse modalities by versatile architectures~\cite{xu2020fusiondn,zhang2021sdnet,xu2020u2fusion,abs-2211-14461} or generative adversarial networks~\cite{TarDAL,ma2019fusiongan,ma2020ddcgan}. Existing fusion schemes mostly focus on the improvement of fusion quality supervised by statistic measurements (e.g., modal gradients~\cite{zhang2021sdnet} and image quality assessment~\cite{xu2020fusiondn}), where these statistic measurements provide the guideline to fuse images containing the information as close as modal images.
It is worth pointing out that without comprehensive modeling, existing fusion methods are easy to neglect the representative features for underlying vision tasks and deteriorate their performance. 

Second, current methods, either plugging learnable modules or end-to-end learning widely rely on handcrafted architectures. However, the manual design is easy to induce feature redundancy, causing to generate edge artifacts, and cannot sufficiently utilize the distinct characteristics of modal information. Furthermore, designing highly performed architectures are with huge labor and ample handcrafted experience. For instance, as for plugged learnable modules, dense blocks~\cite{li2018densefuse}, multi-scale module~\cite{liu2020bilevel}, spatial attention~\cite{li2020nestfuse} and feature decomposition~\cite{ZhaoDIDFuse2020} are utilized to cascade depth for extracting modal features. As for end-to-end learning, dense connection~\cite{xu2020fusiondn,xu2020u2fusion,TarDAL}, and residual modules~\cite{zhang2021sdnet,zhang2020rethinking} are proposed to aggregate the modal feature jointly.
Meanwhile, there are few works that leverage the differentiable architecture search~\cite{liu2018darts} to discover the suitable architectures for image fusion~\cite{liu2021searching,liu2021smoa}. Although these methods achieve remarkable performance,  the mainstream search strategies are always leveraged for the large-scale dataset,  sacrificing the accurate gradient estimation. This would cause the unstable architecture search procedure under  small-scale fusion data, partially damaging the final performance of fusion.

Third, most fusion methods are trained on specific training data. Unfortunately, due to the distribution of diverse fusion tasks varying significantly, these methods cannot acquire the fast adaptation ability and flexibly transfer these solutions to other fusion scenes. Specifically, schemes of plugged learnable modules~\cite{li2018densefuse,liu2021learning2} are trained with a large dataset, such as MS-COCO~\cite{lin2014microsoft}, in order to sufficiently learn the ability of encoding and reconstructing features. However, these methods cannot effectively extract the salient and typical information from multi-modal images because of the difference in data distribution. As for end-to-end learning, there actually lacks an effective practice to investigate the intrinsic feature representative among diverse fusion tasks. Though several methods introduce versatile architectures~\cite{zhang2021sdnet,xu2020u2fusion}, the feature learning is still based on the specific data.

\subsection{Our Contributions}
To partially overcome these critical limitations, we  propose a Task-guided, Implicitly-searched, and Meta-initialized (TIM) image fusion model. Specifically, we first formulate the task-guided image fusion as a constrained strategy, to aggregate the information from downstream vision tasks to assist the unsupervised learning procedure of fusion. Then, rather than leveraging  differentiable search directly, we develop an Implicit Architecture Search (IAS) to investigate the structure of the fusion model with high efficiency and stability. In order to acquire the generalization ability,  we propose the Pretext Meta Initialization (PMI) to learn  the general feature extraction, endowing the fast adaptation ability for fusion model to address various fusion scenarios. The main contributions of this work  can be summarized as:
\begin{itemize}
	\item Targeting to incorporate the task-related guidance into  the learning  of image fusion, we establish a constrained strategy to model image fusion with downstream tasks, in order to break down the bottleneck of ignoring vision tasks information of most fusion approaches.
	\item As for architecture construction, we propose an implicit  search strategy to automatically discover of fusion model with high efficiency, avoiding the verbose adjustment and huge structural engineering of mainstream design methodologies.
	\item As for parameter training, we develop the pretext meta initialization strategy to learn the intrinsic  feature extraction among different  fusion data, thus makes the fusion model has the capability to realize the fast adaptation for various scenarios, only using few amounts of data.
	\item We successively  apply our fusion method to various downstream vision perception tasks. Objective and subjective comparisons on  enhancement and semantic understanding tasks with sufficient evaluations demonstrate our superiority and the effectiveness of proposed mechanisms. 
\end{itemize}

\section{Related Work}
In this section, we  briefly review the development of image fusion methods and discuss the limitations of current image fusion schemes.

Conventional fusion frameworks include three aspects, i.e., feature extraction and handcrafted rule-based fusion, feature  reconstruction. Typically, multi-scale transform schemes are broadly used for multi-modality fusion tasks, such as discrete wavelet~\cite{petrovic2004gradient,lewis2007pixel}, contourlet transform~\cite{da2006nonsubsampled}, non-subsampled transform~\cite{bhatnagar2013directive}, curvelet transform~\cite{nencini2007remote} and discrete cosine scheme~\cite{ahmed1974discrete}. These algorithms firstly transpose the images into various scales, and design suitable
fusion rules to generate results.
 Moreover, methods based on sparse representation attempt to learn the comprehensive dictionary from different multi-modality images. For example, Liu~\emph{et.al.}~\cite{liu2014simultaneous} propose an adaptive sparse representation to learn the structural information from image patches.
Furthermore, Principal Component Analysis (PCA)~\cite{he2010multimodal} and Independent Component Analysis (ICA)  are two effective tools to preserve the intrinsic and compact features for subspace analysis-based fusion. Optimization models, e.g., total variation  minimization~\cite{ma2016infrared} are used for IVIF task, which fuses images on the perspectives from  pixel intensities and gradient distribution of different modalities.
 However,  certain manually designed rules are not inapplicable for various multi-modality fusion scenes and tasks. On the other hand, classic image fusion methods cannot fully characterize the typical properties of each modality image. Complex  strategies limit  performance and  decrease  the inference efficiency.

Since the strong feature fitting ability, plentiful learning-based image fusion methods are proposed, which are composited by two categories: conventional frameworks with learnable mechanisms and end-to-end learning. 
The schemes of introducing learnable mechanisms are basically composited by an encoder, fusion rules and the decoder. At the training phase, the encoder and decoder are jointly trained for each modality, aiming to fully characterize the  representative features and reconstruct the source images. At the test phase, diverse fusion rules such as  summation, weighted average and $\ell_1$ norm operations are applied, based on the vital features from each encoder, to obtain the
comprehensive fused feature. For instance, Li~\emph{et. al.}~\cite{li2018densefuse} propose a Densenet as encoder and the $\ell_1$ norm as rules for IVIF task. Liu~\emph{et. al.}~\cite{liu2021learning} utilize the coarse-to-fine fusion architecture as the encoder and the edge-attention mechanism as the fusion rules without well-aligned datasets.  Liu~\emph{et. al.}~\cite{liu2021smoa} utilize  architecture search to perform the construction
of the encoder and decoder. Liu~\emph{et. al.}~\cite{liu2020bilevel} integrate the flexible plugged priors and data-driven modules as the bi-level optimization based on different characteristics for Infrared-Visible Image Fusion (IVIF) and Medical Image Fusion (MIF) tasks. 
However, similar with the conventional fusion schemes, these hybrid methods are limited by the complicated strategies design, which is easy to induce artifacts and blurs and always ignore the salient modal features.

In recent years, designing end-to-end fusion models  has been received widespread attention.
 Generative Adversarial Networks (GAN) ~\cite{TarDAL,ma2020ddcgan} attempt to control generator output the fused image with thermal radiation in infrared images and gradient details in visible images by the reinforcement from  different discriminators. For the first time, Ma~\emph{et. al.} introduce several generative adversarial schemes~\cite{ma2019fusiongan,ma2020ddcgan} to  transfer diverse modal distribution for the fused images. Liu~\emph{et. al.}~ proposes the target-ware dual adversarial learning to preserve the structural information of infrared images and texture details from visible images, which is also benefit for subsequent detection task. 
Moreover, the universal frameworks ({{e.g.},}  FusionDN~\cite{xu2020fusiondn},  U2Fusion~\cite{xu2020u2fusion}, SDNet~\cite{zhang2021sdnet}) based on image quality assessment, continuous learning and squeeze with decomposition mechanisms are widely performed on digital image  (e.g., multi-focus and exposure), multi-spectral and medical  fusion tasks. Though, these schemes can realize diverse fusion tasks based on information aggregation with statistic metrics, lacking a concrete fusion standard to define the meaningful information. Lastly, there are several works~\cite{TarDAL,tang2022image,sun2022detfusion} to connect image fusion with semantic understanding tasks.  However, these works lack the investigation of inner task-guided relationship, efficient architecture construction and rely on the complex training strategies, which are easy to focus on one task and neglect the optimization of others.

\begin{figure*}[thb]
	\centering
	\includegraphics[width=0.9\textwidth]{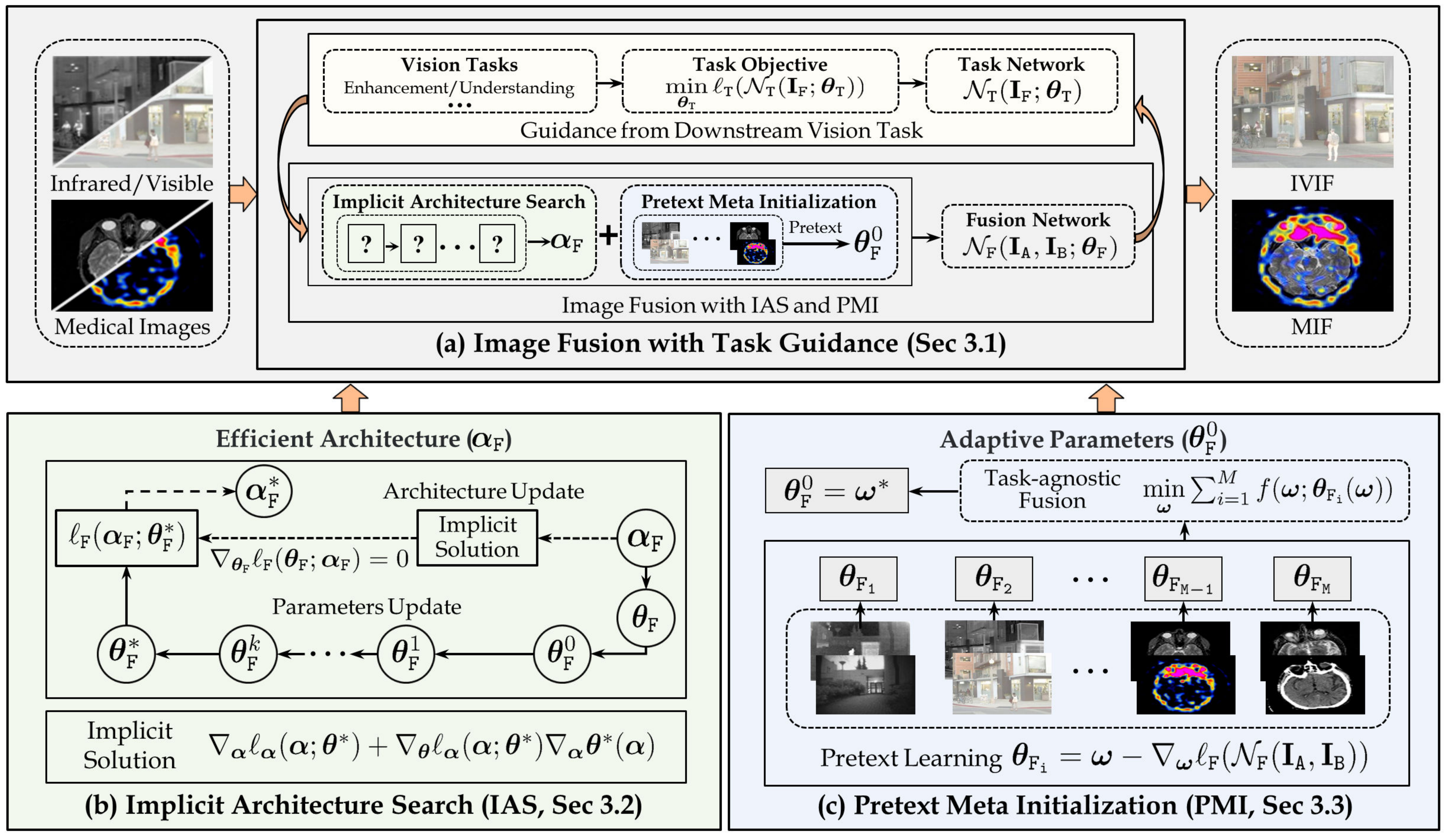}
	\vspace{-0.5em}
	\caption{{ 
					Schematic of the main components of the TIM scheme. We  propose 
				a task-guided image fusion, which introduces  the task guidance for image fusion at  (a). 
			The concrete procedure of the implicit architecture search strategy for the fusion network construction is shown in (b). 	Pretext meta initialization to learn the inhere fusion principle for fast adaptation of fusion is shown at  (c).
			}}
	
	\label{fig:illustration}
\end{figure*}
\section{The Proposed Method}

In essence, the mainstream deep learning-based methods for image fusion ~\cite{zhang2021sdnet,zhao2021efficient,TarDAL,li2018densefuse} are to perform end-to-end training with network, so as to establish the mapping between multi-modal inputs and fused images directly. In this part, we first propose a image fusion network, which can be formulated as 
 $\mathbf{I}_\mathtt{F} = \mathcal{N}_\mathtt{F}(\mathbf{I}_\mathtt{A},\mathbf{I}_\mathtt{B};\bm{\theta}_\mathtt{F})$. $\mathbf{I}_\mathtt{A},\mathbf{I}_\mathtt{B}$ and $\mathbf{I}_\mathtt{F}$ denote
 the multi-modal inputs and fused images, respectively. Introducing the constraint of loss function $\ell_{\mathtt{F}}$, we can utilize $\ell_{\mathtt{F}}(\mathcal{N}_\mathtt{F}(\mathbf{I}_\mathtt{A},\mathbf{I}_\mathtt{B};\bm{\theta}_\mathtt{F}) )$ to train the fusion network.\footnote{In this part, we only introduce the formal  representations of the fusion network and  loss function. Concrete network construction and loss function are presented in the implementation details of Sec.~4.} However, the straightforward training of fusion lacks the consideration of  following vision tasks to integrate task-preferred information, which cannot effectively promote the task performance. Thus, we aims to leverage the task guidance to establish task-oriented image fusion. 
 The overview of   paradigm is shown at the Fig.~\ref{fig:illustration}.

\subsection{Image Fusion with Task Guidance}


In this part, we introduce a task-specific objective for image fusion from the nested optimization perspective, which decouples the while framework into two parts, including image fusion network $\mathcal{N}_\mathtt{F}$ and vision task network $\mathcal{N}_\mathtt{T}$. Thus, the holistic network and parameters are decoupled as $\mathcal{N} = \mathcal{N}_\mathtt{T}\circ  \mathcal{N}_\mathtt{F} $ and $\bm{\theta} = \{\bm{\theta}_\mathtt{T},\bm{\theta}_\mathtt{F}\}$.  The goal of  vision task is to realize the vision perception with generating the task-oriented outputs $\mathbf{y}$  based on  one fusion image $\mathbf{I}_\mathtt{F}$. Similarly, the learning procedure can be defined as 
$\mathbf{y} = \mathcal{N}_\mathtt{T}(\mathbf{I}_\mathtt{F};\bm{\theta}_\mathtt{T})$.
This framework can  effectively transfer the general solution  of single-modal vision task into our framework, which can composite
a highly efficient $\mathcal{N}_\mathtt{T}$.  In this way, we bridge  vision tasks into image fusion procedure, where the optimization of image fusion is constrained by losses of informative richness $\ell_{\mathtt{F}}$ and task-specific maintenance proportion  $\ell_{\mathtt{T}}$.
Taking the effective feedback of task performance as the fusion standard, the task-oriented image fusion can be achieved. The whole task-guided objective to bridge visual perception task and image fusion is shown at Fig.~\ref{fig:illustration} (a).
\begin{align}
&	\min\limits_{\bm{\theta}_\mathtt{T}}  \ell_\mathtt{T}(\mathcal{N}_\mathtt{T}(\mathbf{I}_\mathtt{F};\bm{\theta}_\mathtt{T})),\label{eq:main}\\
&	\mbox{ s.t. } \left\{
	\begin{aligned}
	\mathbf{I}_\mathtt{F} &= \mathcal{N}_\mathtt{F}(\mathbf{I}_\mathtt{A},\mathbf{I}_\mathtt{B};\bm{\theta}_\mathtt{F}^{*}), \\
	\bm{\theta}_\mathtt{F}^{*} & =   \arg\min_{\bm{\theta}_\mathtt{F}} \ell_{\mathtt{F}}(\mathcal{N}_\mathtt{F}(\mathbf{I}_\mathtt{A},\mathbf{I}_\mathtt{B};\bm{\theta}_\mathtt{F}) ).
	\end{aligned}
	\right.	
	\label{eq:constraint}
\end{align}

The constrained formulation is shown at Eq.~\eqref{eq:main} and Eq.~\eqref{eq:constraint}. Specifically, as for the given vision task, we introduce the standard loss functions $\ell_\mathtt{T}$ to train  $\mathcal{N}_\mathtt{T}$  based on a single fusion image $\mathbf{I}_\mathtt{F}$.  Meanwhile, we consider the image fusion process as a  constraint, which is expressed at Eq.~\eqref{eq:constraint} and reveals the procedure of obtaining the fused image $\mathbf{I}_\mathtt{F}$ based on the optimal network parameters $\bm{\theta}_\mathtt{F}^{*}$.  Directly solving this nested optimization is challenging, due to the complex  coupled formulation. Specifically, the gradient of task-specific objective 
can be formulated as $\frac{\partial \ell_\mathtt{T}}{\partial \bm{\theta}_\mathtt{T}} = \frac{\partial \ell_\mathtt{T}(\bm{\theta}_\mathtt{T};\bm{\theta}_\mathtt{T}(\bm{\theta}_\mathtt{F}^{*}))}{\partial \bm{\theta}_\mathtt{T}} + G(\bm{\theta}_\mathtt{T}(\bm{\theta}_\mathtt{F}^{*})) $, where $G(\bm{\theta}_\mathtt{T}(\bm{\theta}_\mathtt{F}^{*})) $ represents the indirect gradient based the response from image fusion $\bm{\theta}_\mathtt{F}^{*}$. Noting that, rather than providing vision tasks with more fusion response,
we aims to strengthen the image fusion with task guidance. Thus, instead of straightforwardly addressing this task-specific objective using exact solutions~\cite{liu2022general,liu2021bilelvel,liu2022optimization,liu2022task}, we streamline a gradual stage-wise  procedure to aggregate  task preference for  fusion.

In order to investigate the relationship between image fusion and downstream vision tasks, a straightforward way is  joint learning.
 Joint learning from scratch may leads to the hard convergence without well initialized $\mathbf{I}_\mathtt{F}$. Thus, we firstly put more attention to addressing single image fusion  constraint (i.e., Eq.~\eqref{eq:constraint}). In detail,
one major obstacle is to obtain efficient architecture, which should be effective for feature extraction.
We present the Implicit Architecture Search (IAS) to discover  effective architectures to composite $\mathcal{N}_\mathtt{F}$. Exploring further, facing with different data distribution of vision tasks,  well 
initialized parameters of image fusion can realize the flexible adaptation. Thus, we propose the Pretext Meta Initialization (PMI) to learn  generalizable parameters (denoted as  $\bm{\theta}_\mathtt{F}^{0}$) to investigate the task-agnostic fusion ability. Based on IAS and PMI, we can utilize the gradient descent $\bm{\theta}_\mathtt{F}^{k} - \nabla_{\bm{\theta}_\mathtt{F}} \ell_{\mathtt{F}}(\mathcal{N}_\mathtt{F}(\mathbf{I}_\mathtt{A},\mathbf{I}_\mathtt{B};\bm{\theta}_\mathtt{F}^{k}) )$ to obtain fundamental fused images, which is shown at the bottom  of Fig.~\ref{fig:illustration} (a).

Then we put the constraint of image fusion into the optimization of vision tasks to jointly optimize the network of  fusion and downstream tasks. The  composited objective can be written as $\min_{\bm{\theta}_{\mathtt{T}},\bm{\theta}_{\mathtt{F}}} \ell_{\mathtt{T}}(\mathcal{N}_\mathtt{T}(\mathbf{I}_\mathtt{F};\bm{\theta}_\mathtt{T})) + \eta \ell_{\mathtt{F}}(\mathcal{N}_\mathtt{F}(\mathbf{I}_\mathtt{A},\mathbf{I}_\mathtt{B};\bm{\theta}_\mathtt{F}) )$, where $\eta$ is the balanced weight. Obviously, this formulation reveals that the gradient of $\bm{\theta}_\mathtt{F}$ is composited by the measurement of information richness from $\ell_{\mathtt{F}}$ and task guidance from $\ell_{\mathtt{T}}$.
Noting that, this learning strategy is mutually beneficial for both two networks. On the one hand, the nested optimization of image fusion with $\mathbf{I}_\mathtt{F}$ can guide the learning of vision task. On the other hand, the backward  feedback of specific vision with $\mathbf{y}$ can facilitate the task-related information into image fusion to finally realize the task-oriented learning, as shown by the cycled yellow arrows in Fig.~\ref{fig:illustration} (a).

\subsection{Implicit  Architecture Search }

As shown in   Fig.~\ref{fig:illustration} (a), we utilize architecture search to discover efficient architecture for image fusion. 
Currently, there are two popular methodologies to design
the architecture of image fusion, i.e., manual design and general architecture search.  However, handcrafted architectures of fusion are mostly based on existing mechanisms, limited in the heavy labor and design experiences.  On the other hand,  the mainstream differentiable search strategies~\cite{liu2021searching,liu2018darts} have been introduced with large-scale datasets, which cannot estimate the accurate gradients due to the one-step approximation considering the efficiency. Thus, these methods are easy to generate unstable architectures, especially for the insufficient data of image fusion.
Thus, we propose the implicit architecture search, which can effectively support the solving procedure of Eq.~\eqref{eq:constraint} towards stable architectures.

 The whole procedure is plotted at Fig.~\ref{fig:illustration} (b).  Following with differentiable relaxation~\cite{liu2018darts,liu2021smoa}, we introduce $\bm{\alpha}_\mathtt{F}$ to denote the architecture weights of  $\mathcal{N}_\mathtt{F}$.   Then we introduce the search objective $\ell_{\bm{\alpha}_\mathtt{F}}$ to measure the influence of $\bm{\alpha}_\mathtt{F}$. 
The goal of  implicit  strategy is to avoid the insufficient learning of $\bm{\theta}_\mathtt{F}$ and large computation, which is more suitable for the unsupervised  fusion tasks.  Noting that, we omit the subscript $\mathtt{F}$ for brief representation.  As for the solving procedure, by substituting
$\bm{\theta}$, the concrete gradient $\mathbf{G}_{\bm{\alpha}} $ of $\ell_{\bm{\alpha}}$ can be generally written as:
\begin{equation}\label{eq:bsg3}
   \mathbf{G}_{\bm{\alpha}} =\nabla_{\bm{\alpha}}\ell_{\bm{\alpha}}(\bm{\alpha};\bm{\theta}) + \nabla_{\bm{\theta}}\ell_{\bm{\alpha}}(\bm{\alpha};\bm{\theta})\nabla_{\bm{\alpha}} \bm{\theta}(\bm{\alpha}).
\end{equation}
 Based on the assumption that lower-level sub-problem have one single optimal solution and referring to implicit function theory, the optimal parameters $\bm{\theta}$ characterizes that $\nabla_{\bm{\theta}}\ell(\bm{\theta};\bm{\alpha}) = 0$ and
 $
\nabla_{\bm{\alpha}}\bm{\theta}(\bm{\alpha}) = -\nabla_{\bm{\alpha},\bm{\theta} }^{2}\ell(\bm{\alpha};\bm{\theta})  \nabla_{\bm{\theta},\bm{\theta}}^{2}\ell(\bm{\alpha};\bm{\theta})^{-1}$. In this way, we can obtain the preciser gradient estimation than general search strategies, avoiding the insufficiency of  one-step update. 
 Inspired by Gauss-Newton (GN) method, we leverage the outer product of first-order gradient to approximate second-order derivative.
Based on the least square method, the implicit approximation of architecture gradient can be formulated as:
\begin{equation}\label{eq:search_strategy}
\mathbf{G}_{\bm{\alpha}} = \nabla_{\bm{\alpha}}\ell_{\bm{\alpha}}(\bm{\alpha};\bm{\theta})  - \frac{\nabla_{\bm{\theta}}\ell(\bm{\alpha};\bm{\theta})^{\top} \nabla_{\bm{\theta}}\ell_{\bm{\alpha}}(\bm{\alpha};\bm{\theta})}{\nabla_{\bm{\theta}}\ell(\bm{\alpha};\bm{\theta})^{\top} \nabla_{\bm{\theta}}\ell(\bm{\alpha};\bm{\theta})}
\nabla_{\bm{\alpha}}\ell(\bm{\alpha};\bm{\theta}).
\end{equation}

Furthermore, we discuss the advantages of proposed methods. Firstly, this strategy is based on the requirement of sufficiently learned network parameters.  The optimal parameters can provide the accurate gradient estimation.
Secondly, compared with the general differentiable search, since it is not required to update once of each iteration, it has the  search stability for  architectures. Moreover, image fusion task is a unsupervised task without abundant data. IAS actually has more efficient for this task.

Then we introduce the concrete search objective. We first propose a operation-sensitive regularization $\mathtt{Reg}$ into the search objective, in order to indicate the basic properties of operations (e.g., computational cost and compactness of architecture). For instance, $\mathtt{Reg}$ can be considered as the weighted summation of latency based on all operations, which is used to constrain the parameter volume.  We also can  control the compactness to define  $\mathtt{Reg}$ with the total number of skip connections. Thus, the search objective is formulated as:
$\ell_{\bm{\alpha}_\mathtt{F}}  = \ell_\mathtt{F} + \lambda (\mathtt{Reg} (\bm{\alpha}_\mathtt{F}))$.
where $\lambda$ represents the trade-off coefficient to balance the fusion quality and operation-sensitive properties. 
\subsection{Pretext Meta Initialization}
 Obviously, $\bm{\theta}_\mathtt{F}$ plays a critical role to bridge the information aggregation of image fusion and following vision tasks. Well initialization
$\bm{\theta}_\mathtt{F}$  should reveal the intrinsic
fusion principles  and is as an intermediary for fast adaptation. On the other hand, $\bm{\theta}_\mathtt{F}$  should merge the stylized
domain information to strengthen the generalization ability for unseen fusion data.
However,  existing image fusion methods seldom digest the inhere fusion principles. These approaches design  specific fusion rules with models for specific fusion tasks. More importantly,
 fusion tasks are varied widely and have distinct intensity distributions.
It is untoward
to obtain the generalizable $\bm{\theta}_\mathtt{F}^{0}$ by directly pre-training on the hybrid fusion datasets, which cannot sufficiently store meta knowledge of fusion tasks and is without  consistent  representations. 

 Therefore, as shown at Fig.~\ref{fig:illustration} (c), we present the pretext meta initialization strategy to learn  the fast adaptation ability, which can assist the framework adapt to specific fusion task fast to learn task-oriented $\bm{\theta}_\mathtt{F}^{*}$, associated with informative fusion and downstream vision perception tasks. 
  We denote $\bm{\omega}$ as the  weights learning from the pretext task among diverse fusion scenes. In fact, we introduce an additional constraint into Eq.~\eqref{eq:main} and Eq.~\eqref{eq:constraint}, which is defined as follows:
  \begin{equation}~\label{eq:meta}
	\bm{\theta}_\mathtt{F}^{0} = \bm{\omega}^{*}, \text{with} \quad \bm{\omega}^{*} = \arg\min_{\bm{\omega}}\sum_{i=1}^{M} f(\bm{\omega};\bm{\theta}_{\mathtt{F}_\mathtt{i}}(\bm{\omega})),
  \end{equation}
where  $M$ denotes the fusion tasks.

Thus, we construct a pretext meta initialization-constraint for image fusion-based vision optimization. It is actually another optimization problem based on  image fusion constraint, i.e., Eq.~\eqref{eq:constraint}, which brings  challenging computation difficulties. Thus, we propose a hierarchical solving procedure~\cite{finn2017model,liu2021towards,jin2021bridging}. We  consider this solution under the solution of image fusion constraint. 
In details, we define  $f$ as the feature-level informative richness measurement, aiming to weight the generalization ability of $\bm{\omega}$,  following with~\cite{xu2020u2fusion}. 
The solving procedure of pretext objective Eq.~\eqref{eq:meta} can be divided into two  steps, i.e., optimizing $\bm{\theta}_{\mathtt{F}_\mathtt{i}}$ with specific fusion scenes and minimizing the meta objective among diverse scenes. As for each scene, we can obtain specific $\bm{\theta}_{\mathtt{F}_\mathtt{i}}$ by several gradient steps, which can be formulated as $\bm{\theta}_{\mathtt{F}_\mathtt{i}} \leftarrow \bm{\omega} -  \nabla_{\bm{\omega}} \ell_{\mathtt{F}}(\mathcal{N}_\mathtt{F}(\mathbf{I}_\mathtt{A},\mathbf{I}_\mathtt{B}) )$. Then we measure the performance of these task-specific weights $\bm{\theta}_{\mathtt{F}_\mathtt{i}}$  to learn the common latent 
distribution and essential fusion principles of image fusion tasks. The computation procedure is $\bm{\omega} \leftarrow \bm{\omega} - \nabla_{\bm{\omega}}\sum_{i=1}^{M} f(\bm{\omega};\bm{\theta}_{\mathtt{F}_\mathtt{i}}(\bm{\omega}))$. This objective can reflect the generalizable ability of $\bm{\omega}$.
We perform two steps iteratively until achieving  $\bm{\omega}^{*}$. Then we assign the values of $\bm{\omega}$ for $\bm{\theta}_\mathtt{F}^{0}$  and continue to solve other constraints of Eq.~\eqref{eq:main}. The concrete details  are reported in the Alg.~\ref{alg:framework}. Related ablation studies to demonstrate the effectiveness of two strategies are performed on Section 5.3. It worth to point out that, based on the well initialization, we can utilize few training data and small iterations to achieve the significant results compared with direct training. 
 
 To sum up,  we provide other two important supports to endow the effective architecture construction principle for $\mathcal{N}_\mathtt{F}$ and establish the pretext meta initialization to  learn the adaptive parameters among different data. Thus, these techniques effectively support the optimization of image fusion constraints, i.e., Eq.~\eqref{eq:constraint}.
We summarize the complete scheme as Alg.~\ref{alg:framework}. Noting that, in order to simplify the 
representation, we omit the concrete learning rates.
 
 \begin{algorithm}[t] 
 	\caption{Task-Oriented Image Fusion}\label{alg:framework}
 	\begin{algorithmic}[1] 
 		\REQUIRE The search space $\mathcal{O}$, loss functions $\ell_\mathtt{F}$, $\ell_\mathtt{T}$ and other necessary hyper-parameters.
 		\ENSURE The optimal architectures and parameters.	
 		\STATE \% Implicit architecture search for image fusion.
 		\WHILE {not converged}
 	    \STATE Updating  parameters $\bm{\theta}_\mathtt{F}$  using Eq.~\eqref{eq:constraint} with $T$ steps.
 	   \STATE  Using  implicit approximation (i.e., Eq.~\eqref{eq:search_strategy})  to update architecture.    $\bm{\alpha}_\mathtt{F}\leftarrow \bm{\alpha}_\mathtt{F}- \mathbf{G}_{\bm{\alpha}} $.
 		\ENDWHILE
 		 \STATE \% Pretext meta initialization.
 		
 		 	\WHILE {not converged}
 		 	\FOR {each fusion task $\mathtt{i}$  with $K$ steps }
 		 	\STATE $\bm{\theta}_{\mathtt{F}_\mathtt{i}} \leftarrow \bm{\omega} -  \nabla_{\bm{\omega}} \ell_{\mathtt{F}}(\mathcal{N}_\mathtt{F}(\mathbf{I}_\mathtt{A},\mathbf{I}_\mathtt{B}) )$.
 		 	\ENDFOR 
 		 \STATE $\bm{\omega} \leftarrow \bm{\omega} - \nabla_{\bm{\omega}}\sum_{i=1}^{M} f(\bm{\omega};\bm{\theta}_{\mathtt{F}_\mathtt{i}}(\bm{\omega}))$.
		\ENDWHILE
 		 \STATE $\bm{\theta}_\mathtt{F}^{0} = \bm{\omega}^{*}$.
 		 \STATE \% Task-guided learning of whole network.
 		 \STATE $\bm{\theta}_\mathtt{T}^{*},\bm{\theta}_\mathtt{F}^{*}  = \arg\min_{\bm{\theta}_{\mathtt{T}},\bm{\theta}_{\mathtt{F}}}  \ell_\mathtt{T}(\mathcal{N}_\mathtt{T}(\bm{\theta}_\mathtt{T})) + \eta \ell_{\mathtt{F}}(\mathcal{N}_\mathtt{F}(\bm{\theta}_\mathtt{F}^{0}) ) $.
 		\RETURN $\bm{\alpha}_\mathtt{F}, \bm{\theta}_\mathtt{T}^{*}$ and $\bm{\theta}_\mathtt{F}^{*} $.
 	\end{algorithmic}
 \end{algorithm}

\section{Applications}\label{sec:apps}
In this section,  we will elaborately
illustrate the implementation details for image fusion. Considering two vision tasks, including image fusion for visual enhancement and semantic understanding, we extend the architecture design for these tasks and report the necessary training details. 
\subsection{Implementation Details} In this part,  implementation details for image fusion network $\mathcal{N}_\mathtt{F}$ including architecture construction and parameter training are introduced.  

\textbf{Search Configurations.}  The search space of image fusion is introduced from~\cite{liu2021searching}, which provide various
image fusion-oriented cells and operators. In~\cite{liu2021searching}, it provides
the details of cells (i.e.,  successive cell $\mathbf{C}_\mathtt{SC}$, decomposition cell $\mathbf{C}_\mathtt{DC}$ and multi-scale fusion cell $\mathbf{C}_\mathtt{MS}$).
The search space of operators including Channel Attention (CA), Spatial Attention (SA), Dilated Convolution (DC), Residual Block (RB), Dense Block (DB) and Separable Convolution (SC) with different kernel size ($3\times 3$ and $5\times 5$) are  provided at~\cite{liu2021searching,liu2021smoa}.
We define the regularization $\mathtt{Reg}$  as the weighted summation of GPU latency, aiming to obtain the lightweight efficient architecture. It is computed by linear summation, i.e., $\mathtt{Reg}(\bm{\alpha}) = \sum_{l}\sum_{\mathbf{o}\in\mathcal{O}}\alpha_{l}\text{LAT}(\mathbf{o})$. $l$ denotes the layer index. $\text{LAT}(\mathbf{o})$ represents the latency of operation $\mathbf{o}$.
As for the search of $\mathcal{N}_\mathtt{F}$, we set 20 and 80 epochs to optimize the  weights for cells and  weights  for operators respectively. Simultaneously, we collect 200 training data from IVIF and MIF tasks respectively, and update each epoch using one kind task-specific dataset alternately. In detail, we divided the whole
dataset equally into the network parameters updating and architecture optimizing. In search phase, $\mathcal{N}_\mathtt{F}$ is composed by two candidate cells, each cell has two  blocks. Utilizing SGD optimizer with initial learning rate $1e^{-3}$, $T$ = 20 iterations and performing cosine annealing strategy, we search the whole architecture 100 epochs. 

\textbf{Training Configurations.}
At the pretext meta initialization stage of image fusion, we utilize 400 pairs from multiple tasks to optimize well initialization $\bm{\omega}^{*}$.  In details, we consider four fusion tasks  including IVIF (e.g., {TNO}, {RoadScene}) and MIF (e.g., MRI, CT, PET and SPECT fusion) tasks. The learning rate for  single-task (step 9)  and  multi-tasks updating (step 11) are set as $1e^{-3}$ and $1e^{-4}$ respectively. As for single-task learning, we conduct 4 gradient updates, with the Adam optimizer. Furthermore, we prepare amounts of patches with size $64\times64$ and generate corresponding saliency maps. As for RGB images, we transform them to YCbCr channels and take the Y channel for fusion. Data augmentation, e.g., flip and rotation are utilized. All search and training experiments of image fusion are performed on a NVIDIA GeForce GTX 1070 GPU and 3.2 GHz Intel Core i7-8700 CPU.

\subsection{Image Fusion for  Visual Enhancement}
Designing suitable image fusion schemes to sufficiently incorporate different characteristics is a vital component. 
As analyzed in~\cite{liu2021searching}, the image fusion should preserve complete but discrepant information, i.e., structural target information and abundant texture details. Therefore, we formulate both two objects as the parallel fusion structure as $\mathcal{N}_\mathtt{T}$ to investigate these discrepancies, i.e., target  extraction and detail enhancement. In this way, we conduct the implicit search for the whole framework.
In order to constrain the computation burden, we utilize two successive cells (with two candidate operators)  to composite the outer representation of this parallel enhancement network. Introducing different losses, the principled objects can be achieved. Finally, by introducing spatial attention and three $3\times 3$ convolutions, the final fused images can be obtained, where the goal of spatial attention is  to generate a  map 
to fuse these hierarchical   features.

In order to simplify the training procedure, we consider two kind of losses, intensity loss and SSIM metric. We utilize the Mean Square Error (MSE) loss $\ell_\mathtt{int} =\|\mathbf{I}_\mathtt{1} - \mathbf{I}_\mathtt{2}\|^{2}_{2}$ to measure the difference of pixel intensity. Structural similarity, denoted $\ell_\mathtt{ssim}$, which is defined as $\ell_\mathtt{ssim} = 1- \mathtt{SSIM}(\mathbf{I}_\mathtt{1}, \mathbf{I}_\mathtt{2})$. Therefore, the whole loss is written as: $\ell = \ell_\mathtt{int} + \mu \ell_\mathtt{ssim}$. 
We introduce two weight formulations to measure the information preservation.
On the one hand, targeting to  extract rich features information in $\mathcal{N}_\mathtt{F}$ module, we introduce to estimate weight maps computed by the shallow and deep-level features  from VGG network, following with~\cite{xu2020u2fusion}.  To simplify, we denote it as $\ell_{\mathtt{F}}$.

On the other hand, focusing on the visual quality of specific fusion tasks, we introduce the spatial saliency map estimation to weight the  proportional information based on pixel distribution. Firstly, we calculate the histogram map of source images, denoted as $\mathbf{H}_\mathtt{his}$. Inspired by fusion rule~\cite{liu2021smoa}, the contrast ratio of each pixel can be computed by $\mathbf{M}(\mathtt{i}) = \sum^{255}_{\mathtt{j}=0} \mathbf{H}_\mathtt{his}(\mathtt{i}) |\mathtt{j}-\mathtt{i}|$, where $\mathtt{i}$, $\mathtt{j}$ denote the intensity value.  Then, we obtain the final estimation map by softmax function to constrain the range between 0 and 1.
In this way, given two modality-based images $\mathbf{I}_\mathtt{A}$, $\mathbf{I}_\mathtt{B}$, fusion image $\mathbf{I}_\mathtt{F}$ and saliency-guided weights $\mathbf{M}_\mathtt{A}$, $\mathbf{M}_\mathtt{B}$, we can obtain the weighted loss function, i.e.,  $\ell_\mathtt{int}^{\mathtt{V}} = \| \mathbf{M}_\mathtt{A} \otimes (\mathbf{I}_\mathtt{F} - \mathbf{I}_\mathtt{A})\|^{2}_{2} + \| \mathbf{M}_\mathtt{B} \otimes (\mathbf{I}_\mathtt{F} - \mathbf{I}_\mathtt{B})\|^{2}_{2}$ and $\ell_\mathtt{ssim}^{\mathtt{V}} = 1- \text{SSIM}(\mathbf{M}_\mathtt{A} \otimes \mathbf{I}_\mathtt{A},\mathbf{M}_\mathtt{F} \otimes  \mathbf{I}_\mathtt{A}) + 1- \text{SSIM}(\mathbf{M}_\mathtt{B} \otimes \mathbf{I}_\mathtt{F}, \mathbf{M}_\mathtt{B} \otimes  \mathbf{I}_\mathtt{B})$. To simplify, we denote this formulation as $\ell_{\mathtt{T}}= \ell_\mathtt{int}^{\mathtt{V}} + \mu \ell_\mathtt{ssim}^{\mathtt{V}}$.
As for the parallel outputs of $\mathcal{N}_\mathtt{T}$, $\ell_\mathtt{int}$ and  $\ell_\mathtt{ssim}$ are utilized to constrain the similarity of different modalities to realize the target extraction and detail enhancement respectively.

 Utilizing the hybrid datasets in {TNO}\footnote{https://figshare.com/articles/TNOImageFusionDataset/1008029} and {RoadScene}~\cite{xu2020u2fusion}, we search the parallel fusion structure  based on the searched $\mathcal{N}_\mathtt{F}$ for IVIF. Moreover, collecting 150 pairs of multi-modal medical data from Harvard website\footnote{http://www.med.harvard.edu/AANLIB/home.html}, we can search three task-specific networks for MRI-CT, MRI-PET and MRI-SPECT fusion tasks. SGD optimizer with learning rate $1e^{-3}$ and consine annealing strategy is utilized to train with 100 epochs. 
Then  inserted subsequent $\mathcal{N}_\mathtt{T}$, we train the whole network jointly. 
Furthermore, we also illustrate  details about the enhancement of training strategy for visual fusion. We set the learning rate as $1e^{-4}$ and introduce the Adam Optimizer to train the whole network for 100 epochs for infrared-visible and medical fusion tasks.

\subsection{Image Fusion for Semantic Understanding}
Based on the results from $\mathcal{N}_\mathtt{F}$, we can strengthen diverse $\mathcal{N}_\mathtt{T}$ for  semantic Understanding tasks (i.e., multi-spectral object detection and segmentation) by proposed architecture search. 
It should be emphasized that our goal is not to completely design the entire semantic perception network, but to search for the core feature expression  to improve the performance of perception tasks.

Targeting to obtain the efficient feature fusion for semantic perception, we improve the directed acyclic graph-type cell with feature distillation mechanism for flexible representations, which is denoted as $\mathbf{C}_\mathtt{FD}$. The graph cell contains several nodes, where the edges represent the relaxation of operators. At the final node, this cell performs the feature distillation mechanisms~\cite{liu2021ruas} by concatenating the features from other nodes.
To be specific, we utilize the cascaded feature distillation cell to constitute the modular feature fusion parts (e.g., neck part in object detection and feature decoding in segmentation), allowing a  seamless way for changing different backbones.  Considering the low-weighted and efficient feature representations for these high-level perception tasks, we introduce diverse single-layer  convolutions to constitute the search space, including normal convolution with  $k\times k$,  $k \in \{1,3,5,7\}$, dilated convolution with  $k\times k$,  $k \in \{3,5,7\}$ and dilate rate of 2, residual convolution with kernel size $k\times k$,  $k \in \{3,5,7\}$ and skip connection. 

\subsubsection{Object Detection}
In this paper, we utilize RetinaNet~\cite{lin2017focal} as the baseline scheme. 
Recently, a series of NAS-based detection schemes~\cite{ghiasi2019fpn,xu2019auto,wang2021fcos} are proposed to discover the neck part, including searching the connection  modes  from top-down and bottom-up  perspectives, or operators for multi-scale feature fusion. Following bottom-up principle, we utilize the feature distillation cell to fuse feature progressively. In detail, focusing on two features with diverse scales from backbones, we first resize the features with lower resolution and concatenate them into the cell at three-levels, where the cells contains four nodes.

We introduce the  {MultiSpectral} dataset proposed by Takumi et.al~\cite{takumi2017multispectral} for experiments. This dataset is captured by RGB, FIR, MIR and NIR cameras. Due to the low resolution (256 $\times$ 256) and blurred imaging, we re-partitioned and filtered the dataset. In detail, we select 2550 pairs for training and 250 pairs for testing. Five categories of objects are consisted in this dataset, including color cone, car stop, car, person and bike. In order to impose the principles of detection, we adopt the widely-used RetinaNet~\cite{ross2017focal} as the baseline for comparison. The major improvement comes from the FPN re-design by automatic search and pretext meta initialization.
Utilizing {MultiSpectral} dataset and plugging the  $\mathcal{N}_\mathtt{F}$, we search the whole architecture with proposed search strategy from  scratch progressively. More concretely, the batch size, architecture learning rate and search epochs are 1, 3$\times e^{-4}$ and 120 respectively. Targeting to  fast convergence, we firstly train the fusion module with 40 epochs for well initialization. As for the training procedure, we train the whole architecture 160000 steps and set the learning rate as 2$\times e^{-3}$ and delay it with a cosine annealing  to 1 $\times e^{-8}$.

\subsubsection{Semantic Segmentation}
As for semantic segmentation, we introduce ResNet18 as the encoder to conduct feature extraction. Compared with existing RGB-T segmentation schemes~\cite{zhang2021abmdrnet}, which leverage two backbones to encoding different modal features, our segmentation scheme is lightweight based on the nested formulation with image fusion. As for the decoder part, we utilize the similar fusion strategy to integrate the features from high and low-level feature maps. We first utilize the residual upsampling mechanism~{} to resize the low-resolution feature as large as high-level ones with same channels. Then we concatenate them as the inputs.
A residual connection is utilized for the output of cells. Similarly, we also utilize three-level features and put forward two cells to fuse them. Each cell has two nodes. Finally, the estimated  map is generated from $\frac{1}{8}$ size.   

Coupling with  $\mathcal{N}_\mathtt{F}$ and searched segmentation module, we further investigate the jointly learning between image fusion and semantic perception. As for segmentation task, we leverage the widely used {MFNet} dataset, including 1083 pairs for training and  361 pairs for testing. This dataset is composited by various scenarios (e.g, poor light, glare, daytime) with size 640$\times$480 and nine categories (i.e., background, bump, color cone, guardrail, curve, bike, person and car).  Plugged the pre-searched $\mathcal{N}_\mathtt{F}$, we search the segmentation network specifically. Widely-used Cross-Entropy loss computed in $\frac{1}{8}$ and $\frac{1}{16}$ is introduced as the search and training loss $\ell_{\mathtt{T}}$. With batch size of 2 and initial learning rate 1e$^{-2}$ and data augmentation (random clip and rotation), we search the decoder part for 100 epochs. Utilizing SGD optimizer, we decay the learning rate from 1e$^{-2}$ to $1e^{-8}$
within 240 epochs with all training data to train the network.

\section{Experimental Results}
In this section, we first perform the task-guided image fusion on two categories of applications including visual improving and semantic understanding. Then we  conduct comprehensive technique analyses  to illustrate the effectiveness of two  mechanisms (i.e., IAS and PMI).

\begin{table*}[bht]
	\renewcommand{\arraystretch}{1.3}
	\caption{Numerical results with representative image fusion methods on {TNO} dataset.}
	\label{tab:irvis}
	\centering
	\setlength{\tabcolsep}{1.6mm}{
		\begin{tabular}{ c c c c c c c c c c c c c }
			\hline
				Metrics  & DDcGAN & RFN& DenseFuse & FGAN & DID & MFEIF &  SMOA &TARDAL& SDNet  &U2Fusion	& TIM$_{\text{w/o L}}$ & TIM$_{\text{w/ L}}$ \\\hline
		 MI& 1.861&2.108&2.402&2.194&2.439&2.616&2.273&2.783&2.092&1.934&{3.285}&\textbf{3.416}\\
			 FMI& 0.761&0.809&0.817&0.401&0.779&0.817&\textbf{0.818}&0.812&0.806&0.796&\textbf{0.818}&0.816\\
			 VIF& 0.693&0.806&0.802&0.631&0.828&0.805&0.732&0.850&0.753&{0.792}&{0.861}&\textbf{0.884}\\
			 $\mathrm{Q^{AB/F}}$ &0.355&0.322&0.416&0.222&0.406&0.452&0.369&0.410&0.413&0.416&\textbf{0.458}&{0.456}\\\hline
			Parameters (M)& 1.097&2.733&{0.074}&{0.925}&0.373&0.705&0.223&0.297&\textbf{0.067}&0.659&1.232&{0.127}\\
			FLOPs (G)& 896.84&727.74&48.96&{497.76}&103.56&195.82&61.869&82.37&37.35&366.34&194.63&\textbf{35.39}\\
			Time (s)& 0.211&5.912&0.251&0.124&0.118&0.141&8.071&\textbf{0.001}&0.045&0.123&{0.145}&\textbf{0.001}\\
			\hline 
		\end{tabular}	
	}
\end{table*}

\begin{table*}[bht]
	\renewcommand{\arraystretch}{1.3}
	\caption{Numerical comparison with representative image fusion methods on  {RoadScene} dataset.}
	\label{tab:irvis_latency}
	\centering
	\setlength{\tabcolsep}{1.6mm}{
		\begin{tabular}{ c c c c c c c c c c c c c }
			\hline
			Metrics  & DDcGAN & RFN& DenseFuse & FGAN & DID & MFEIF &  SMOA &TARDAL& SDNet  &U2Fusion	& TIM$_{\text{w/o L}}$ & TIM$_{\text{w/ L}}$ \\\hline
	
			MI& 2.631&2.866&3.093&2.859&3.116&3.303&3.004&3.480&3.415&2.917&{3.675}&\textbf{3.896}\\
			FMI& 0.780&0.783&0.788&0.774&0.767&{0.792}&0.785&0.769&0.781&0.777&\textbf{0.797}&0.786\\
			VIF& 0.619&0.773&0.792&{0.614}&0.824&0.811&0.760&0.779&0.811&0.765&\textbf{0.838}&0.775\\
			$\mathrm{Q^{AB/F}}$ & 0.318&0.306&0.503&{0.275}&0.481&0.472&0.461&0.445&0.515&0.519&\textbf{0.526}&0.430\\\hline
FLOPs (G)& 549.68&432.10&30.01&{301.96}&63.48&120.03&37.92&50.49&22.89&224.53&62.10&\textbf{21.69}\\
Time (s)& 0.401&4.115&0.043&0.089&0.077&0.088&7.421&\textbf{0.001}&0.041&0.126&{0.028}&\textbf{0.001}\\
\hline 
		\end{tabular}	
	}
\end{table*}

\begin{figure*}[!htb]
	\centering \begin{tabular}{c@{\extracolsep{0.05em}}c@{\extracolsep{0.05em}}c@{\extracolsep{0.05em}}c@{\extracolsep{0.05em}}c@{\extracolsep{0.05em}}c@{\extracolsep{0.05em}}c@{\extracolsep{0.05em}}c@{\extracolsep{0.05em}}@{\extracolsep{0.05em}}c}

		\includegraphics[width=0.12\textwidth]{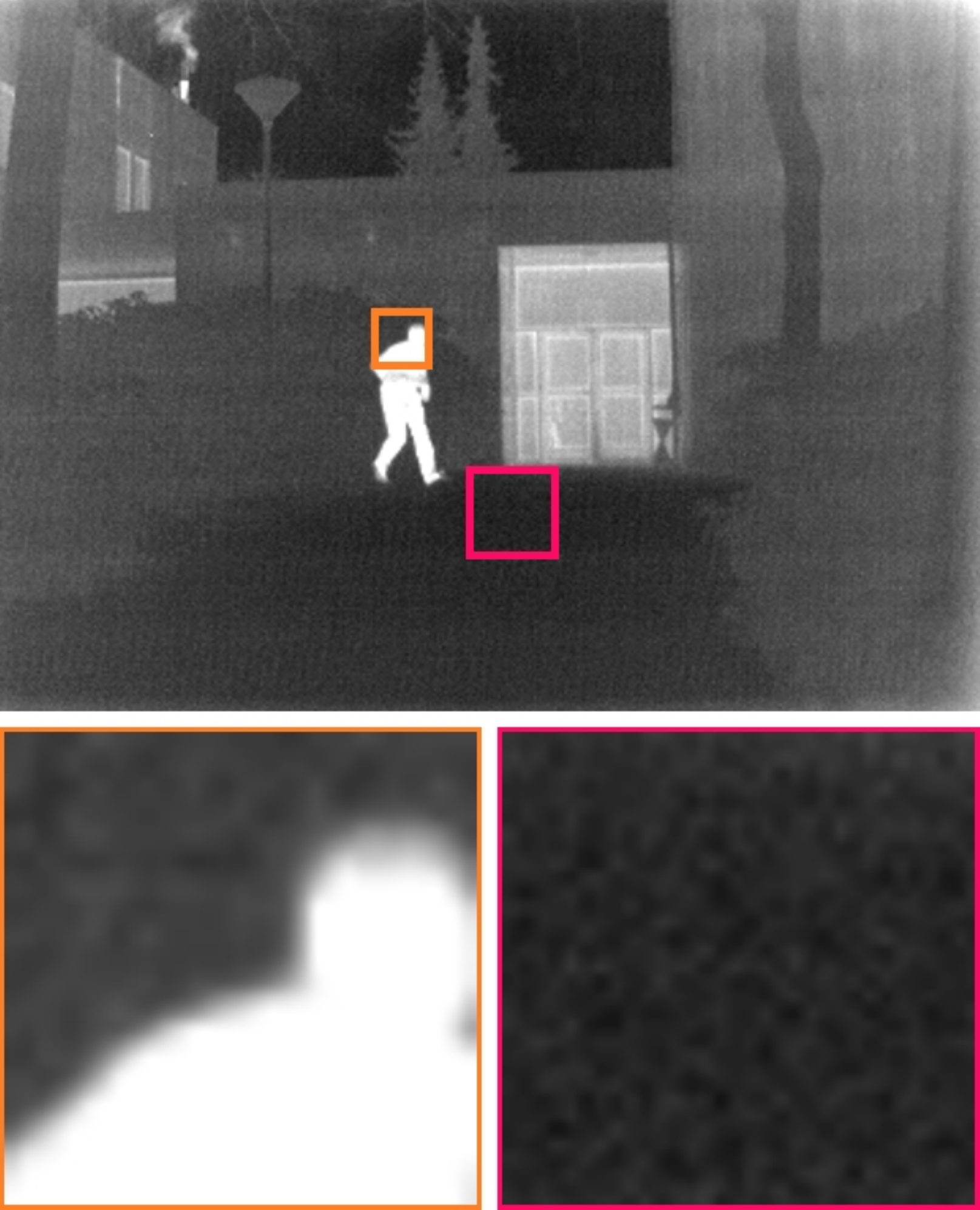}
		&\includegraphics[width=0.12\textwidth]{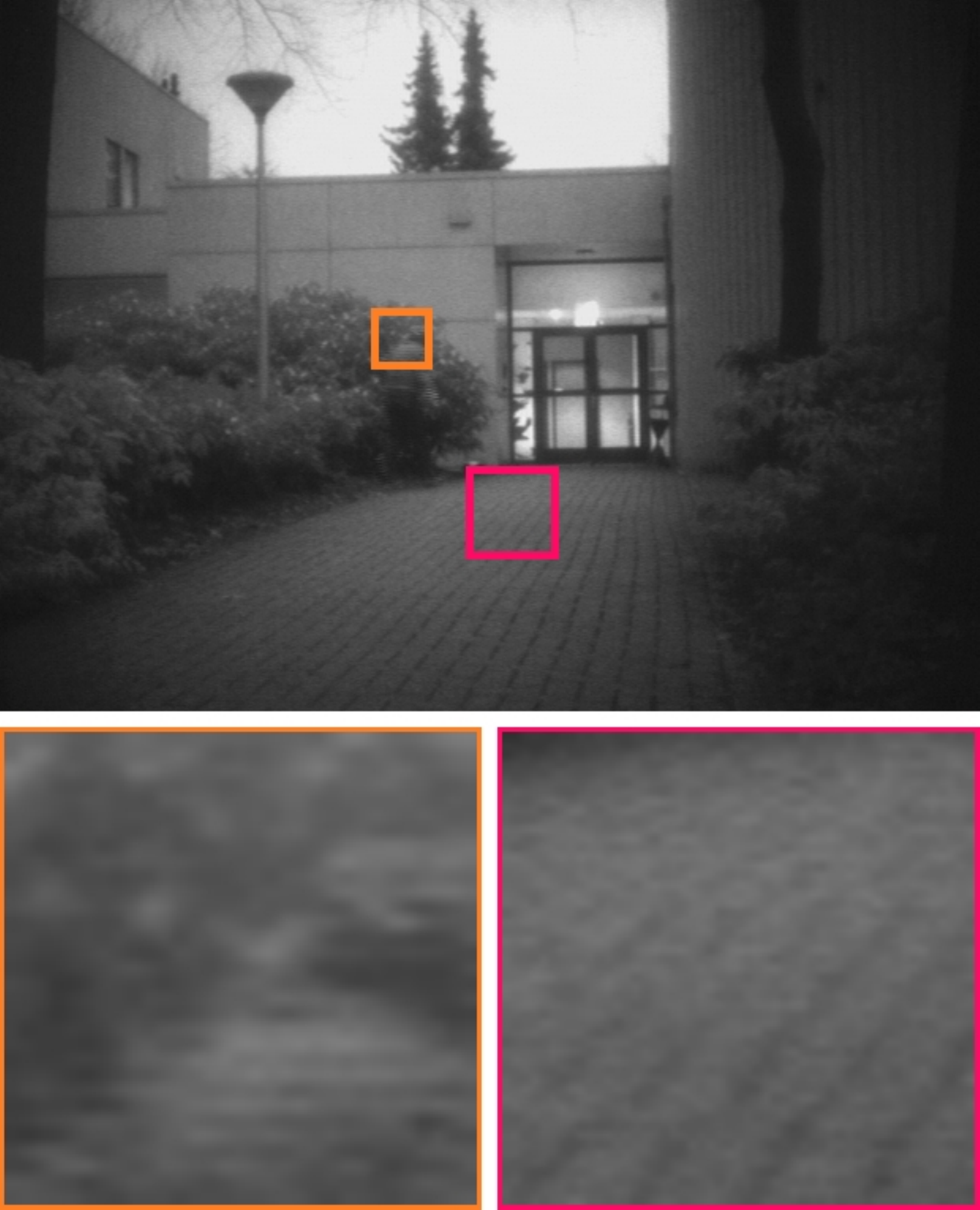}
		&\includegraphics[width=0.12\textwidth]{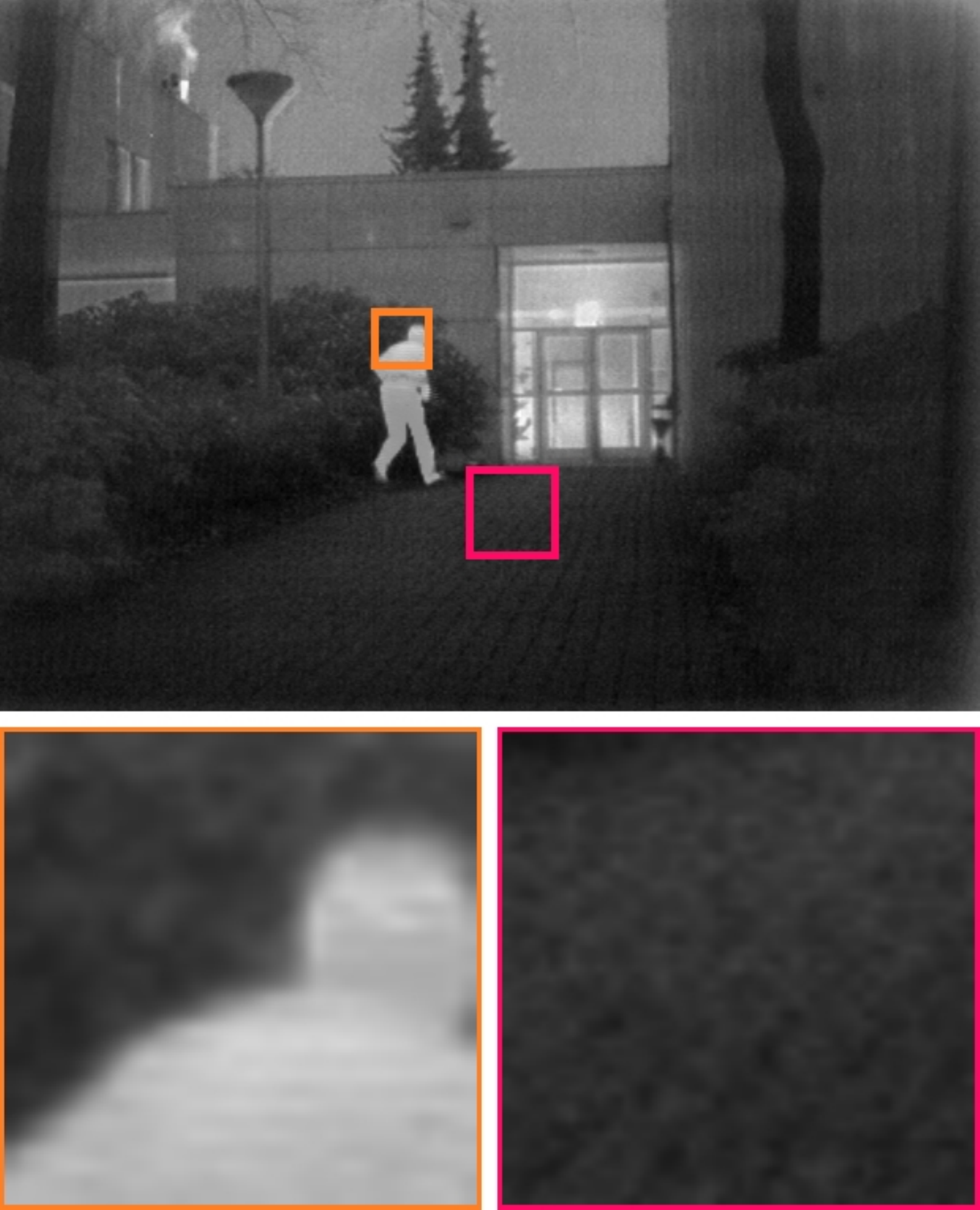}
		&\includegraphics[width=0.12\textwidth]{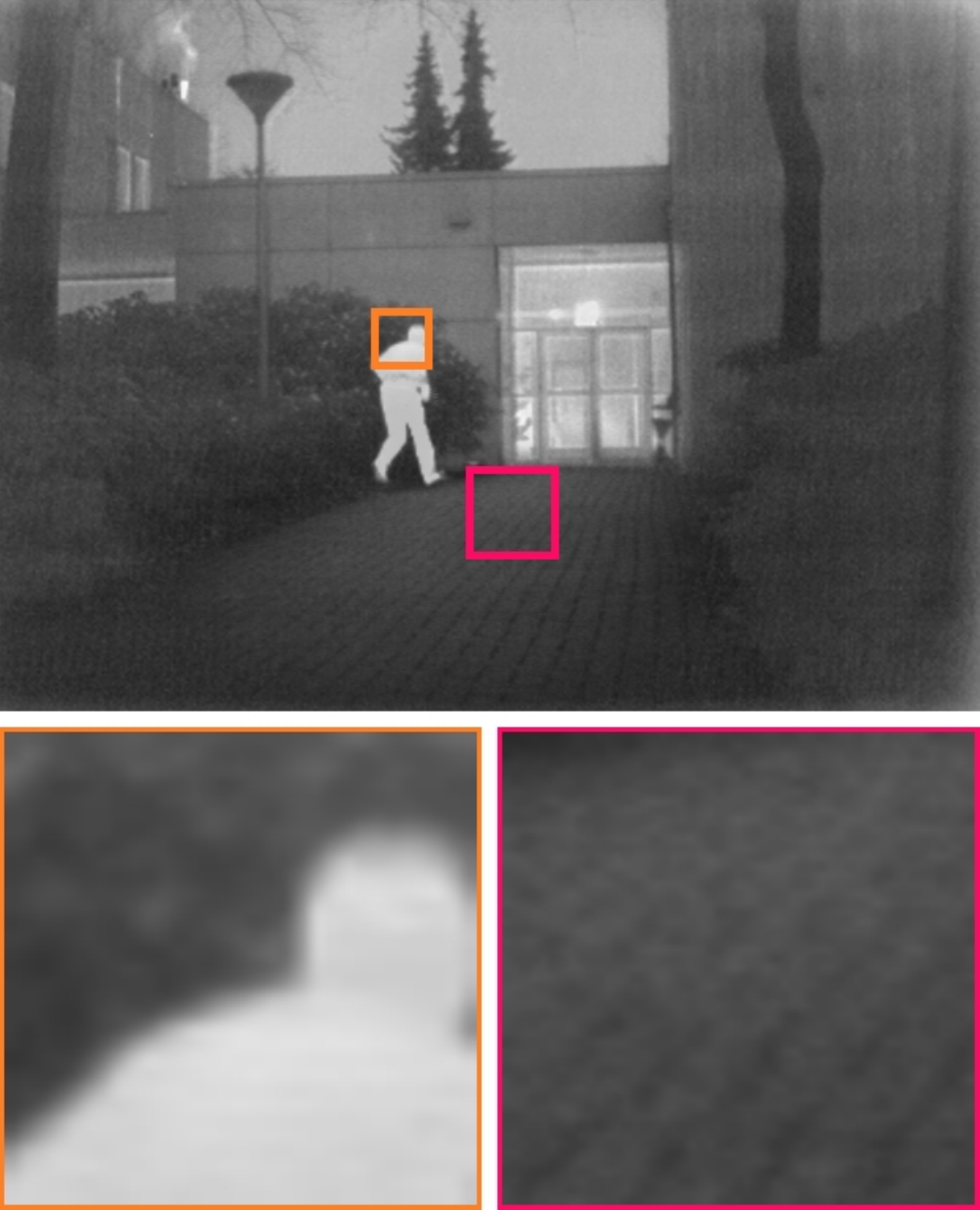}
		&\includegraphics[width=0.12\textwidth]{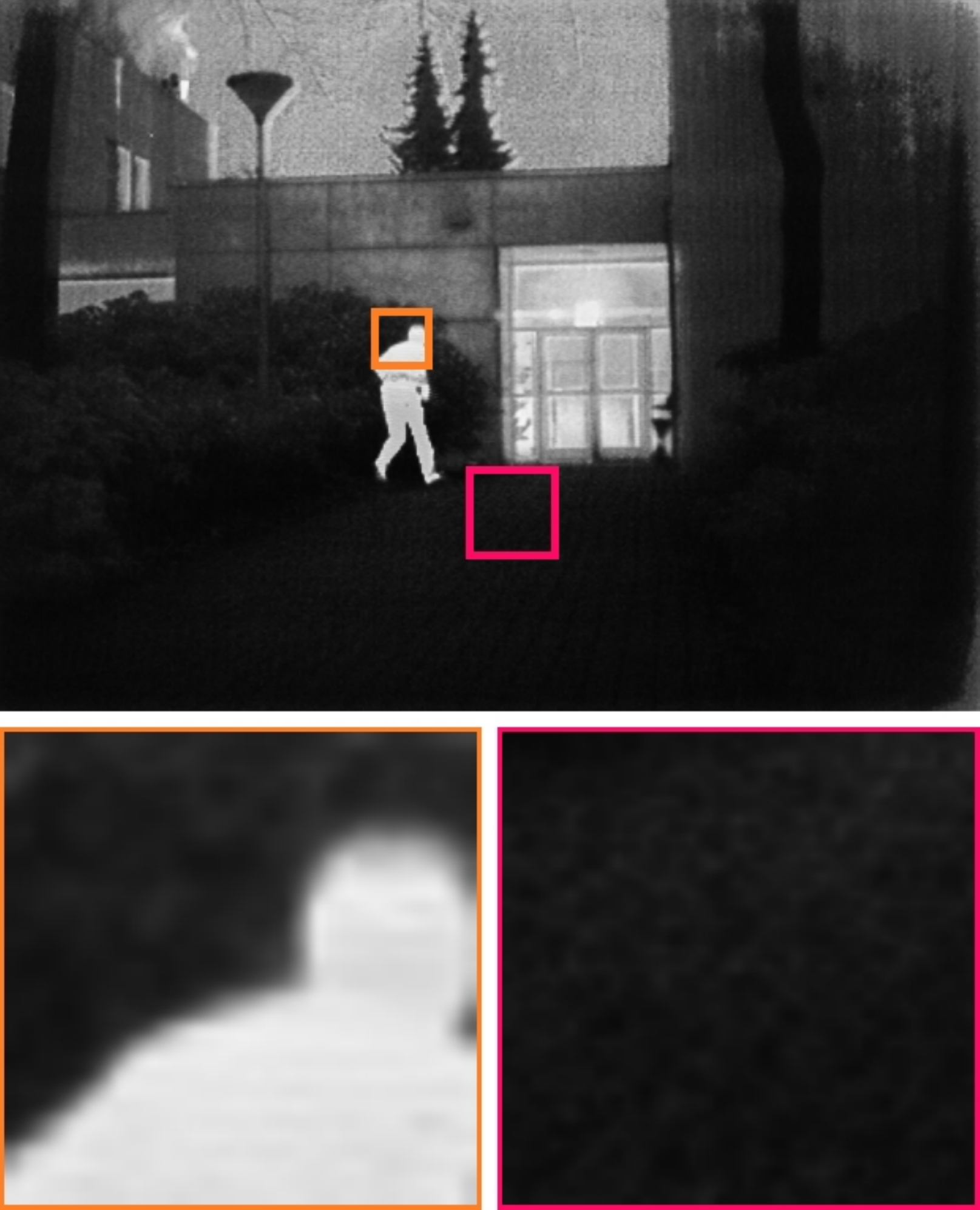}
		&\includegraphics[width=0.12\textwidth]{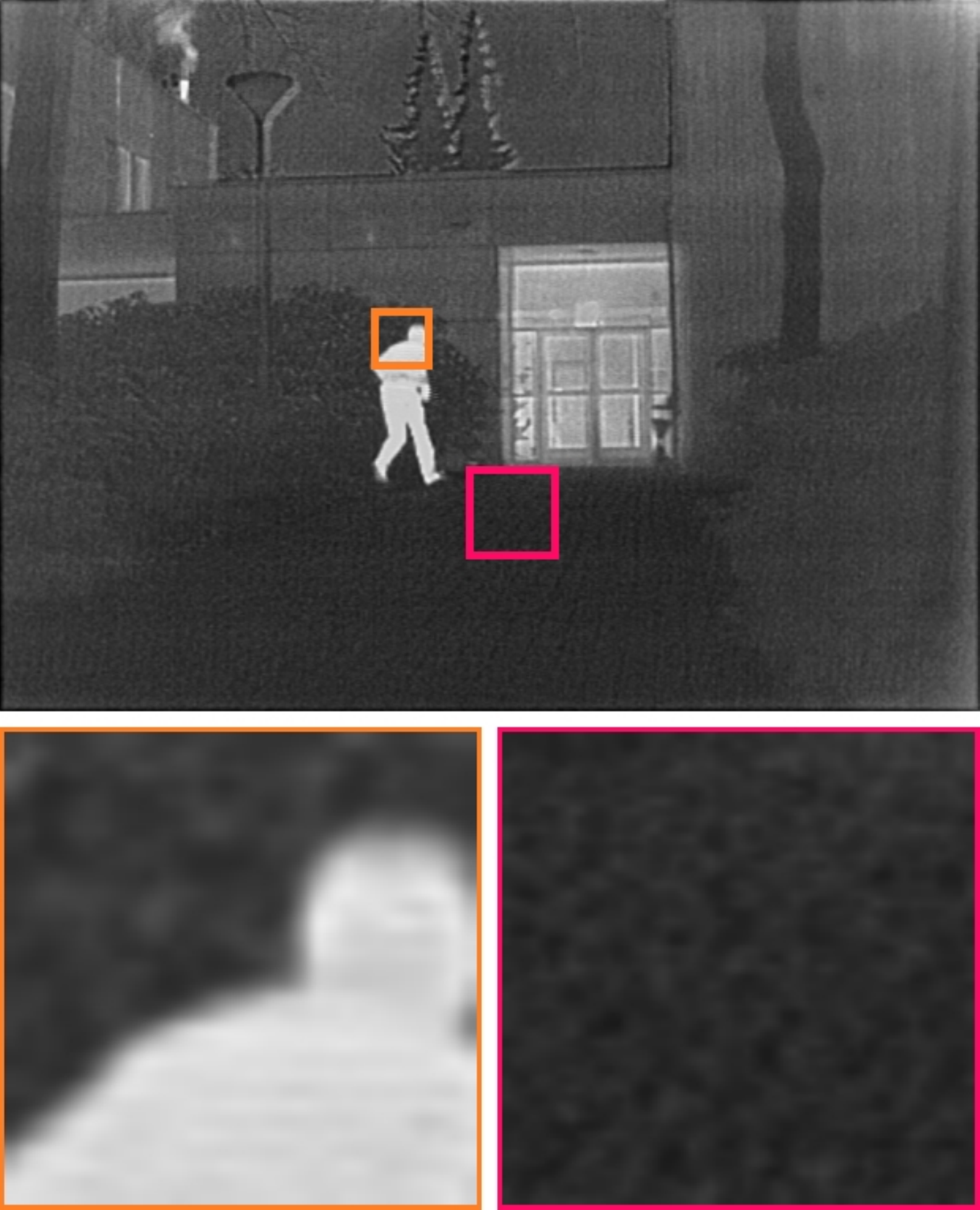}
		&\includegraphics[width=0.12\textwidth]{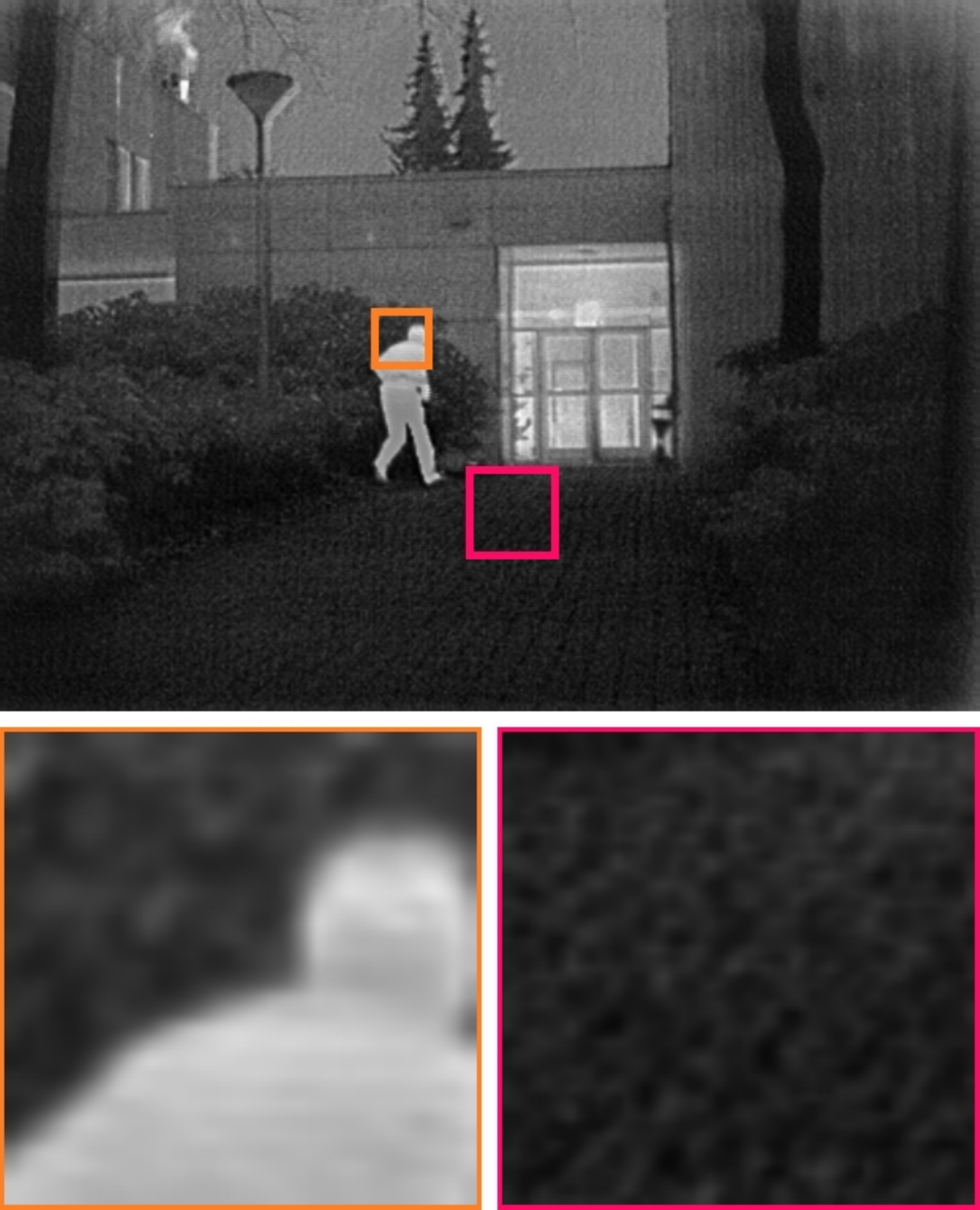}
		&\includegraphics[width=0.12\textwidth]{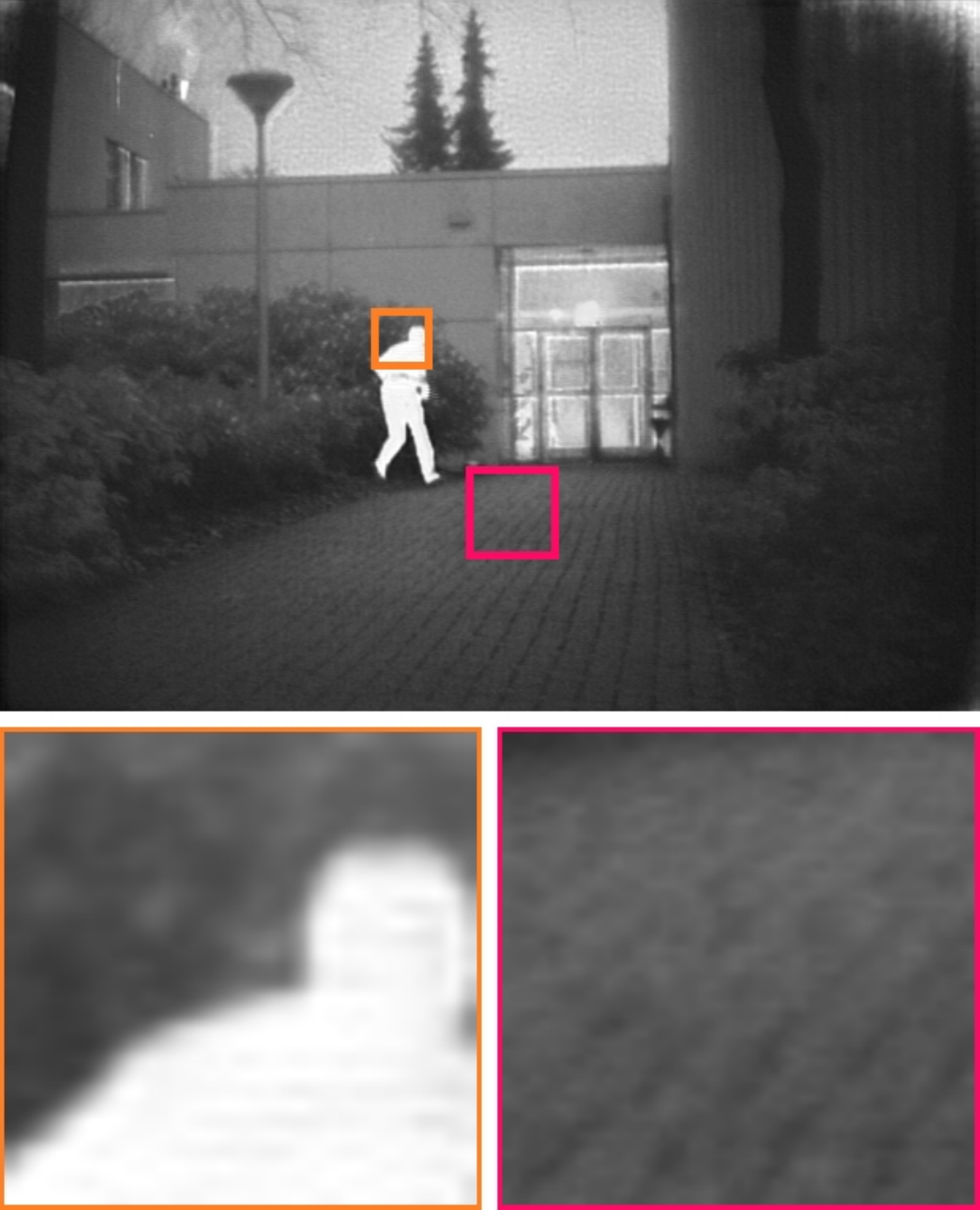}\\
		\includegraphics[width=0.12\textwidth]{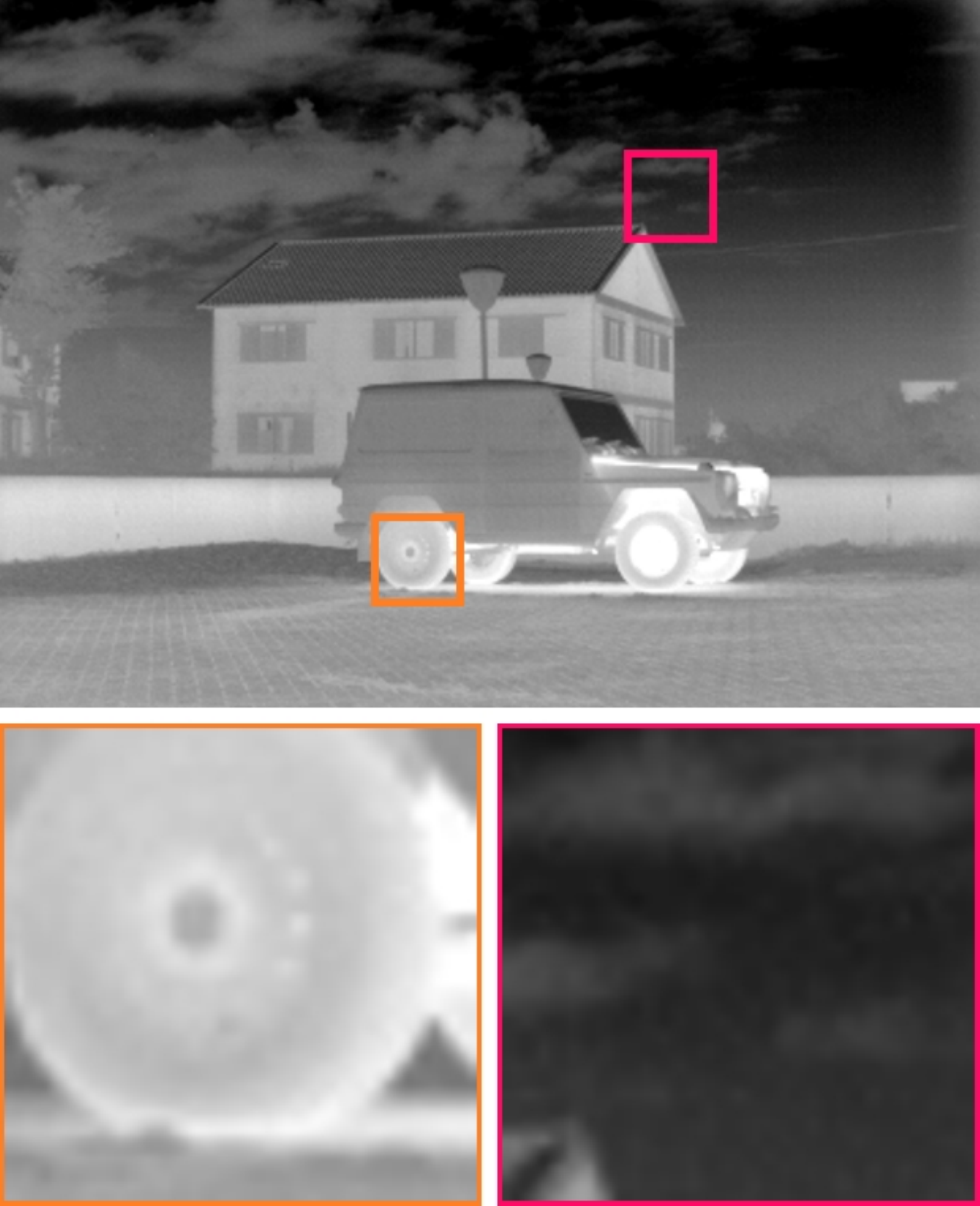}
		&\includegraphics[width=0.12\textwidth]{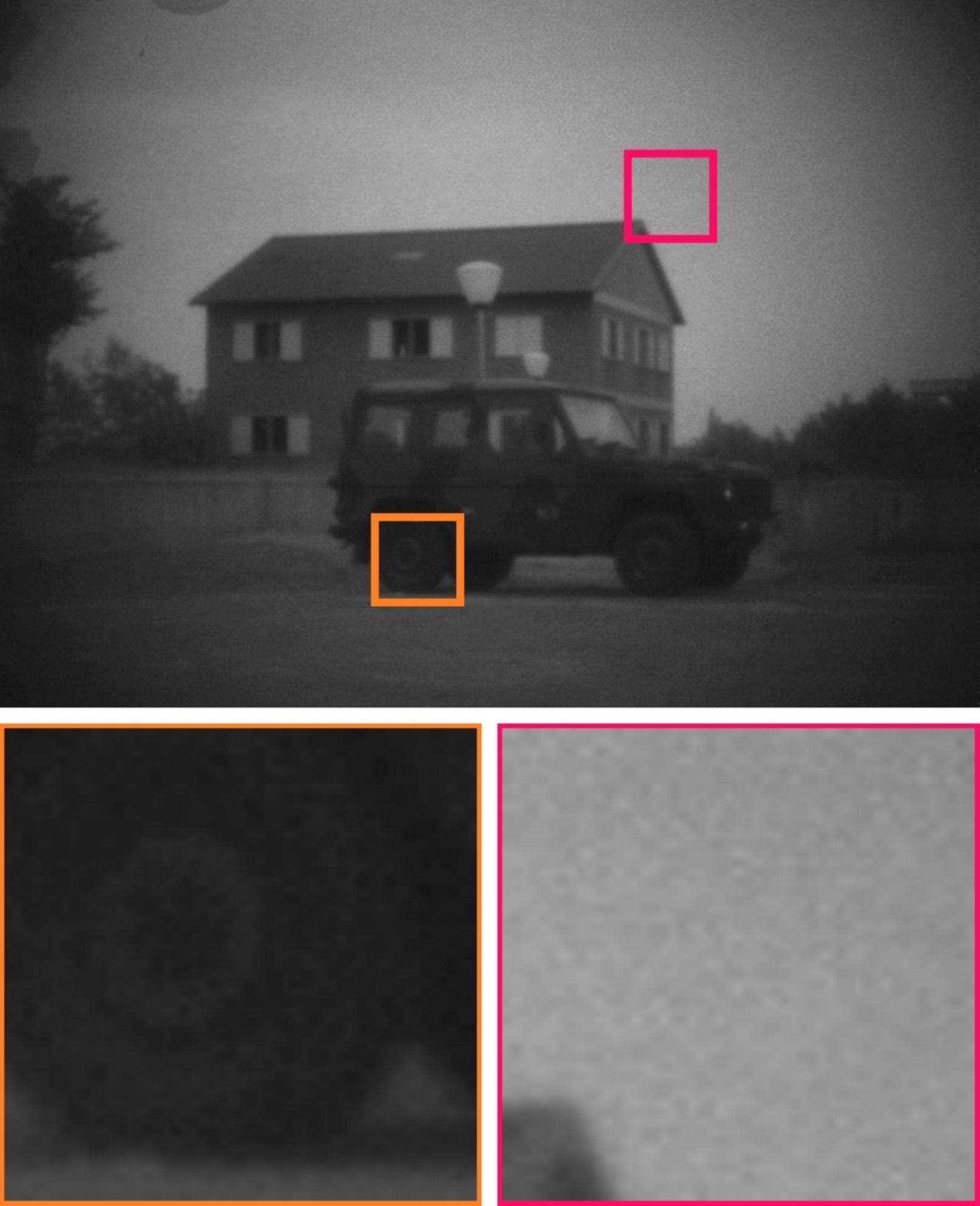}
		&\includegraphics[width=0.12\textwidth]{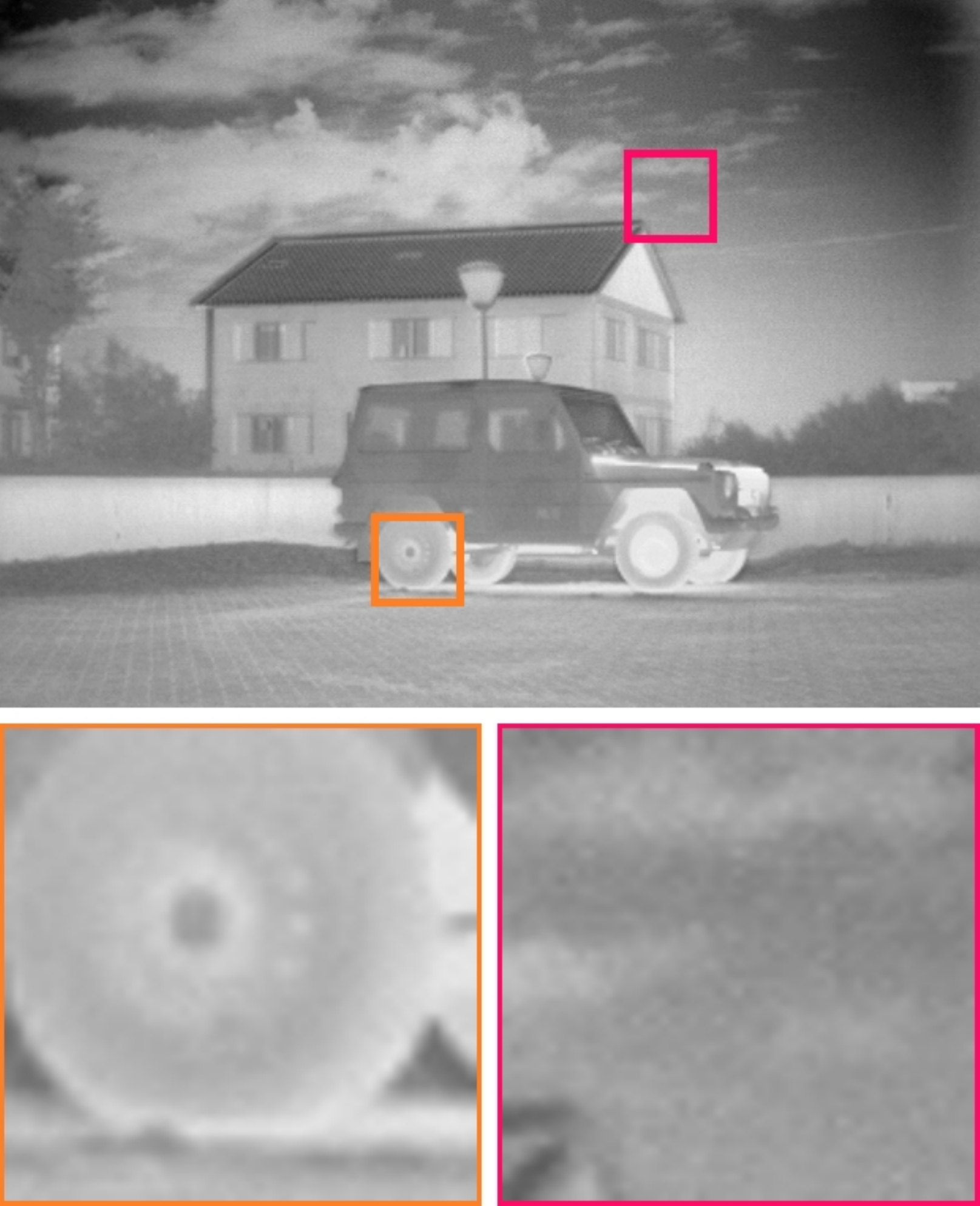}
		&\includegraphics[width=0.12\textwidth]{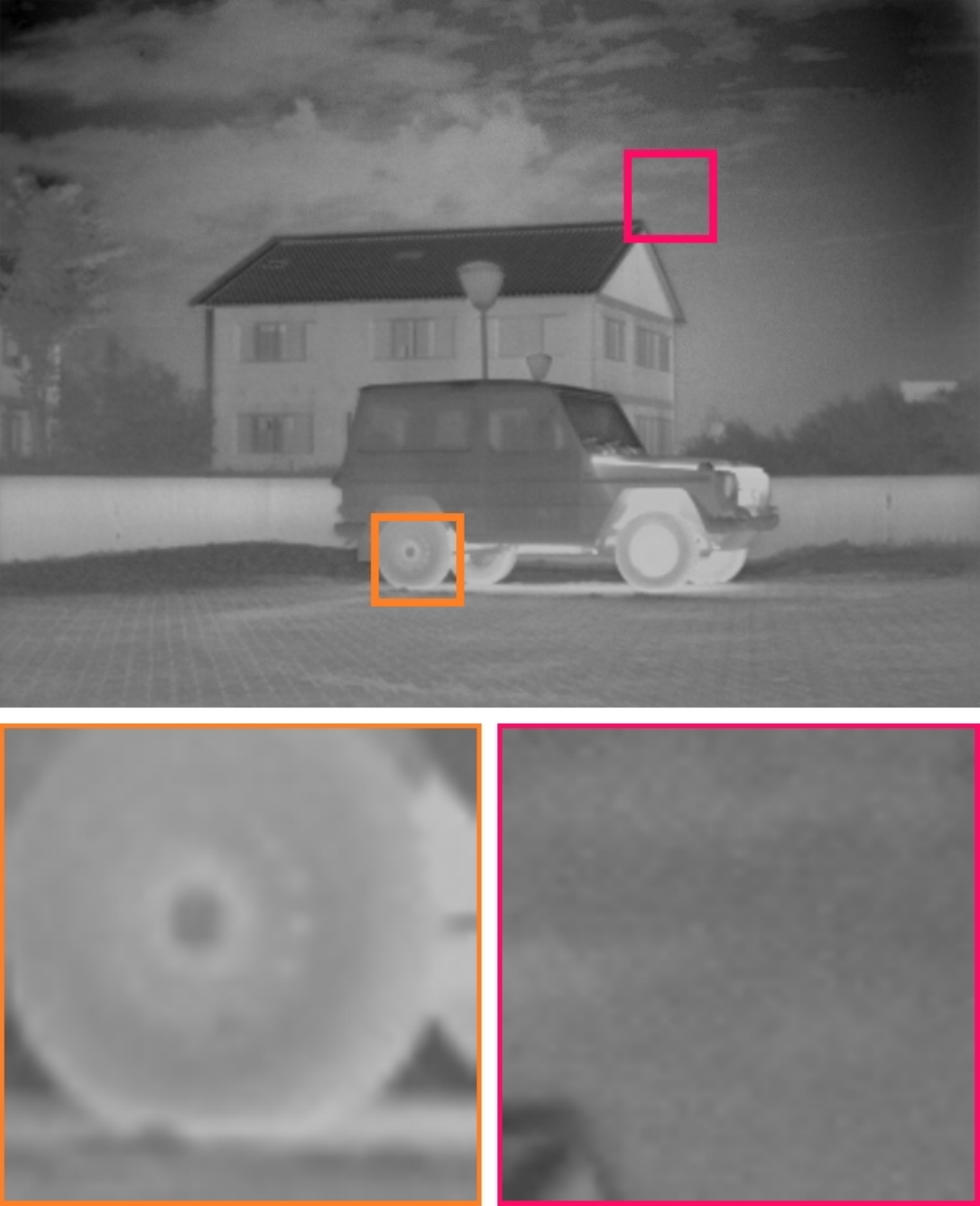}
		&\includegraphics[width=0.12\textwidth]{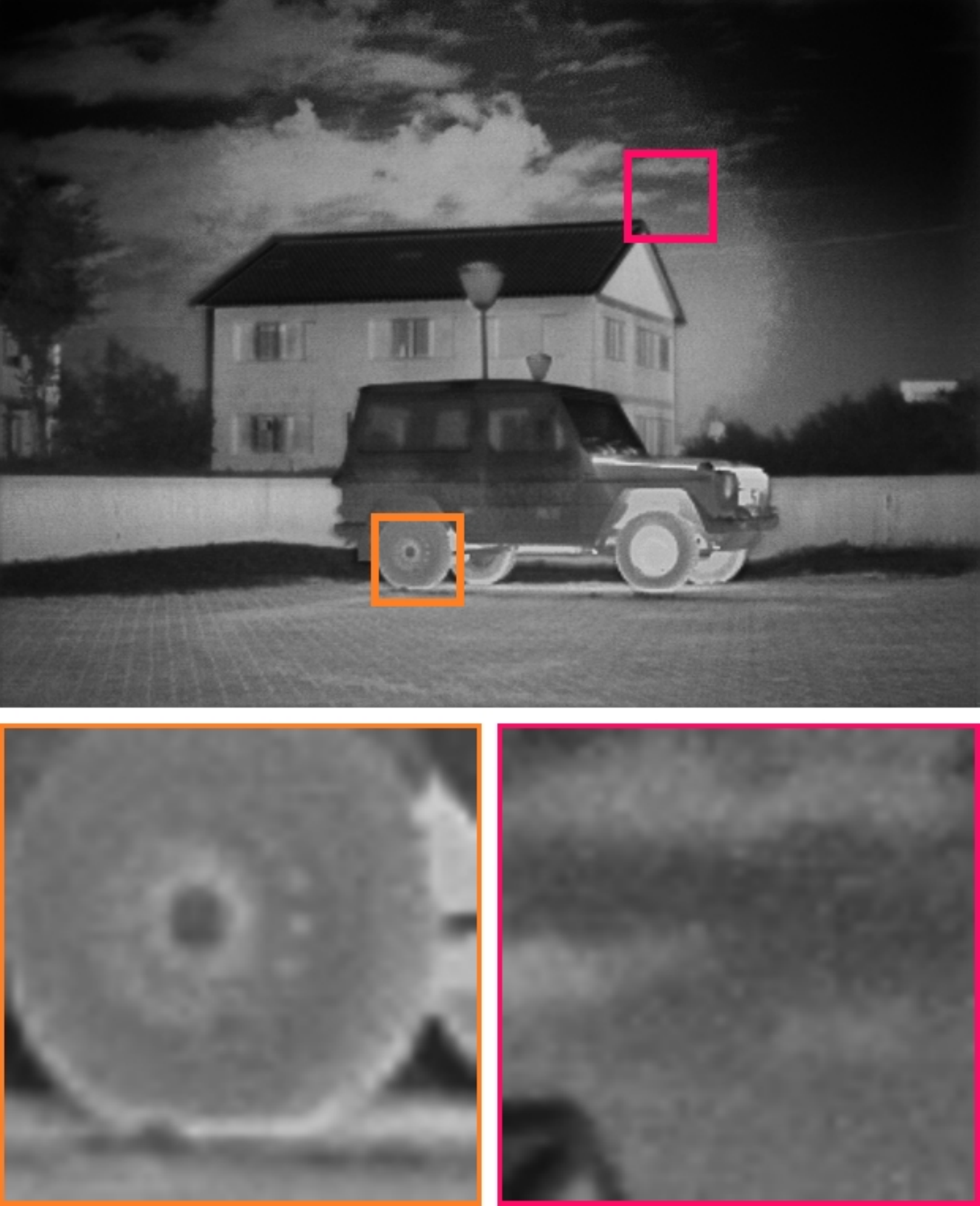}
		&\includegraphics[width=0.12\textwidth]{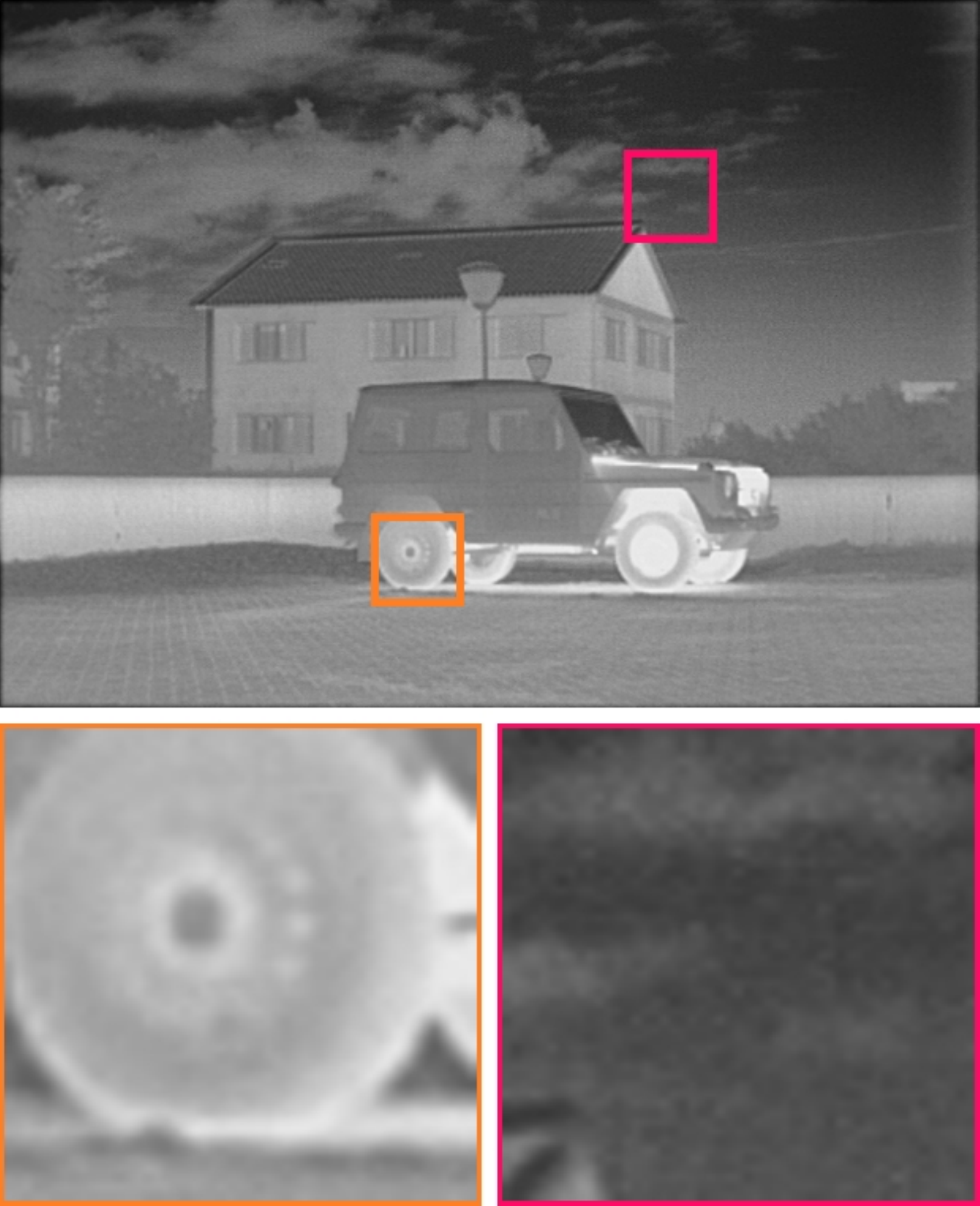}
		&\includegraphics[width=0.12\textwidth]{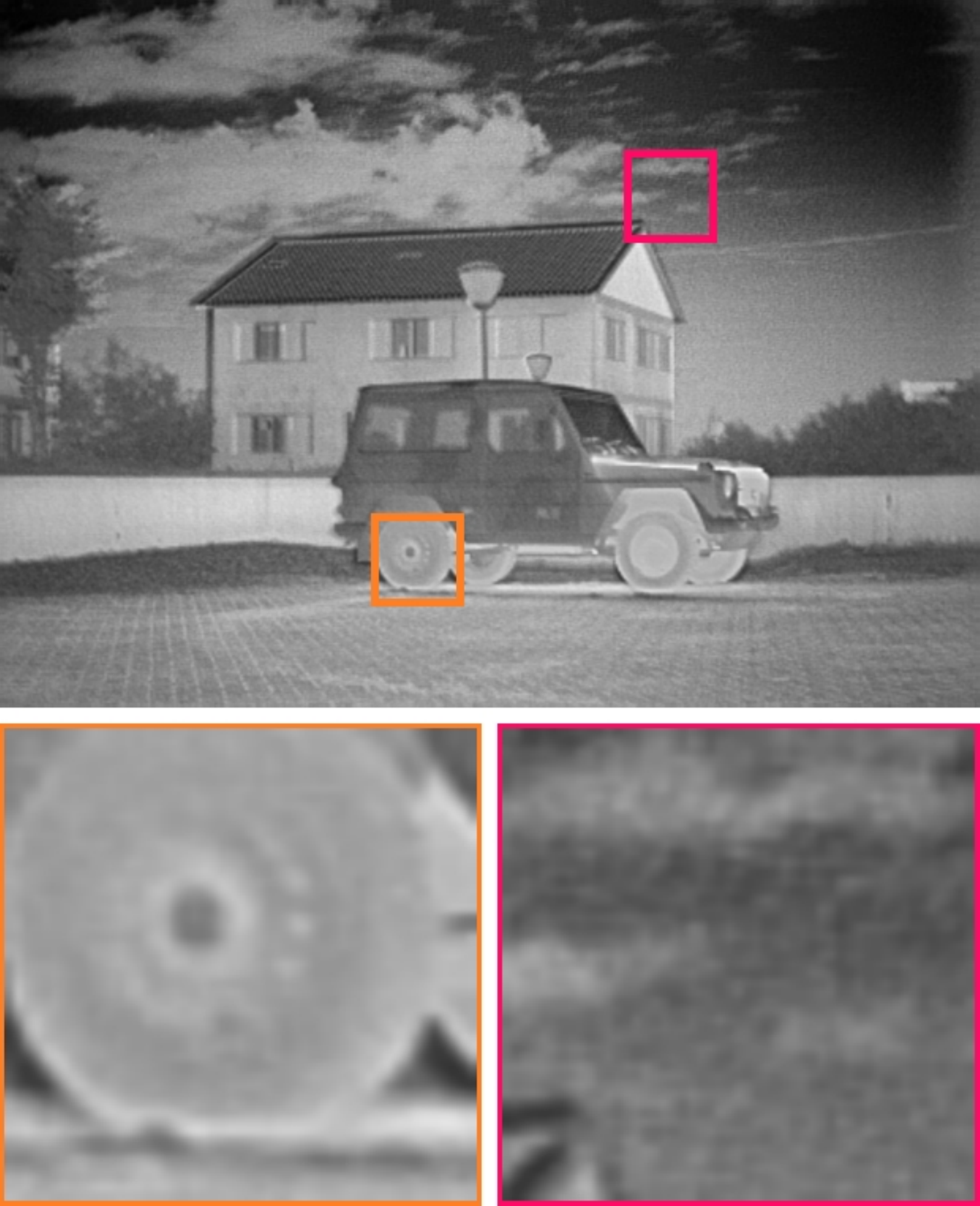}
		&\includegraphics[width=0.12\textwidth]{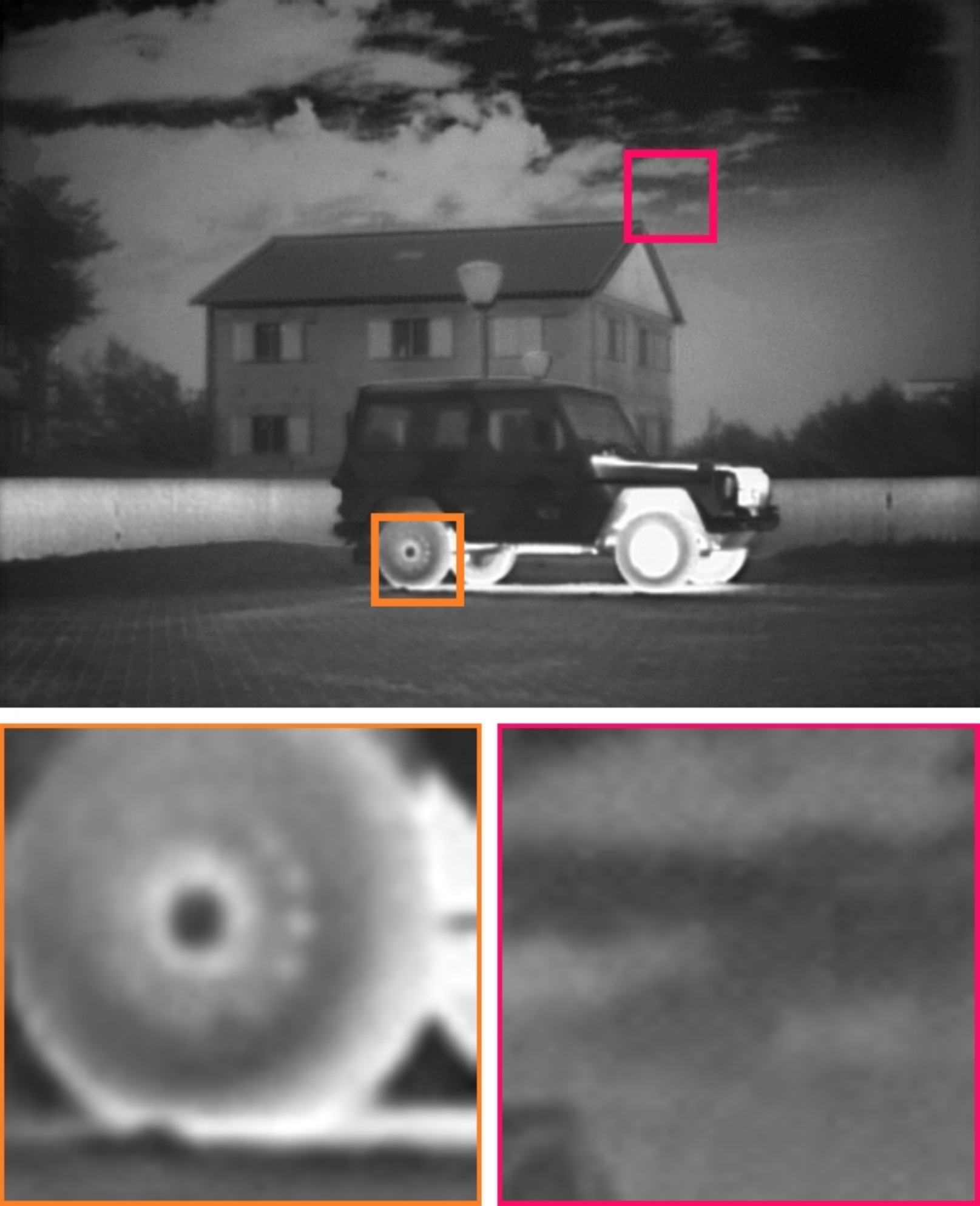}\\
		\includegraphics[width=0.12\textwidth]{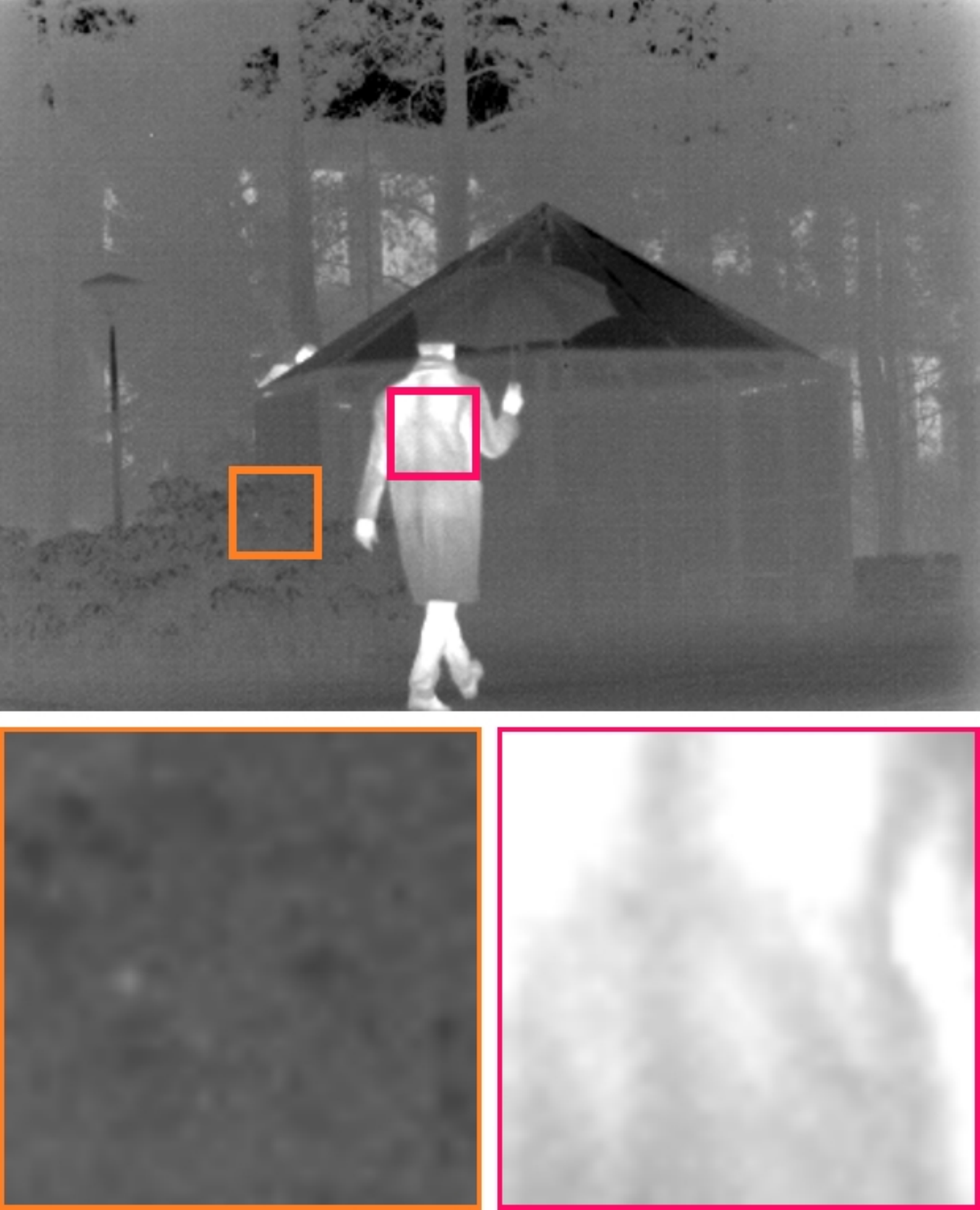}
		&\includegraphics[width=0.12\textwidth]{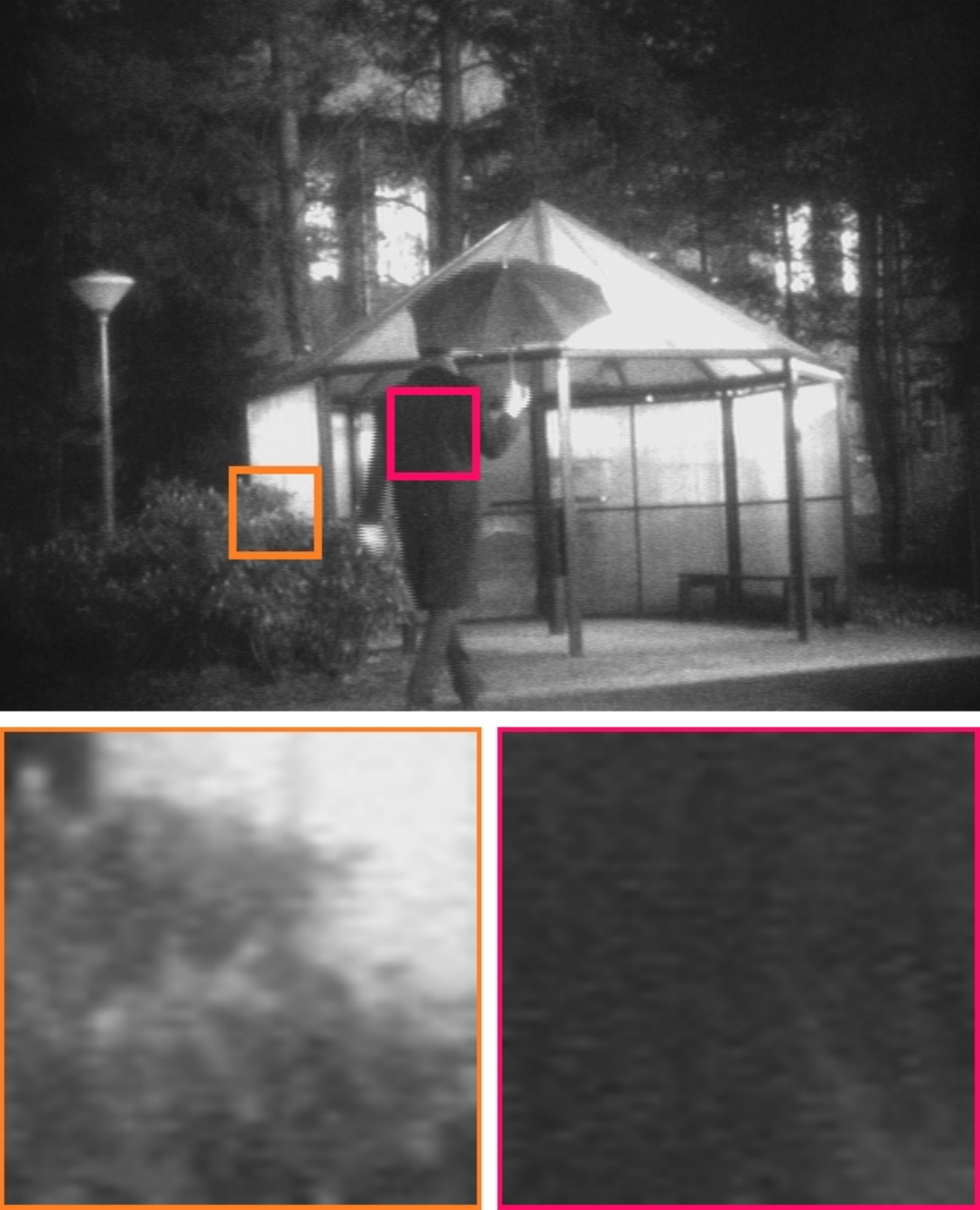}
		&\includegraphics[width=0.12\textwidth]{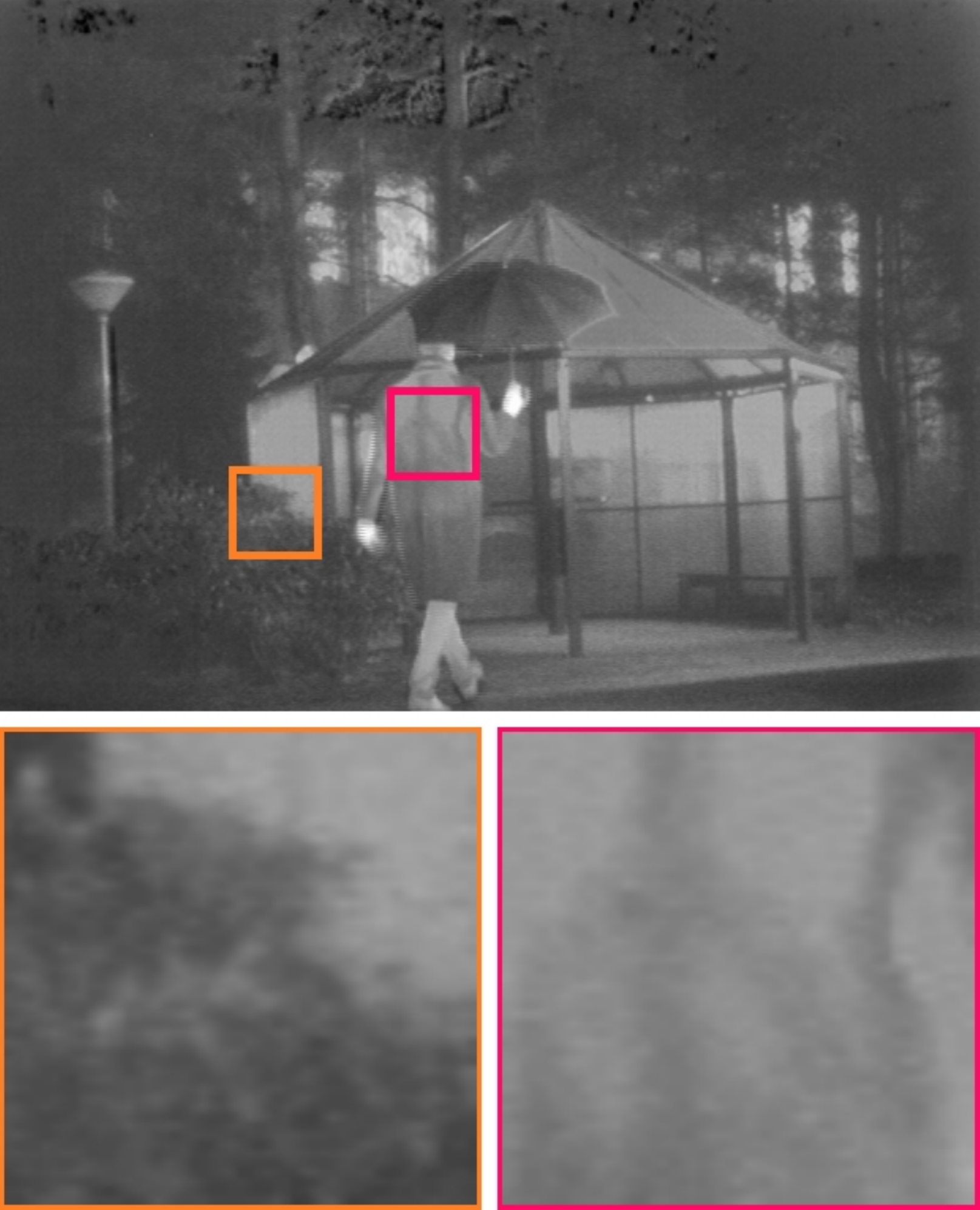}
		&\includegraphics[width=0.12\textwidth]{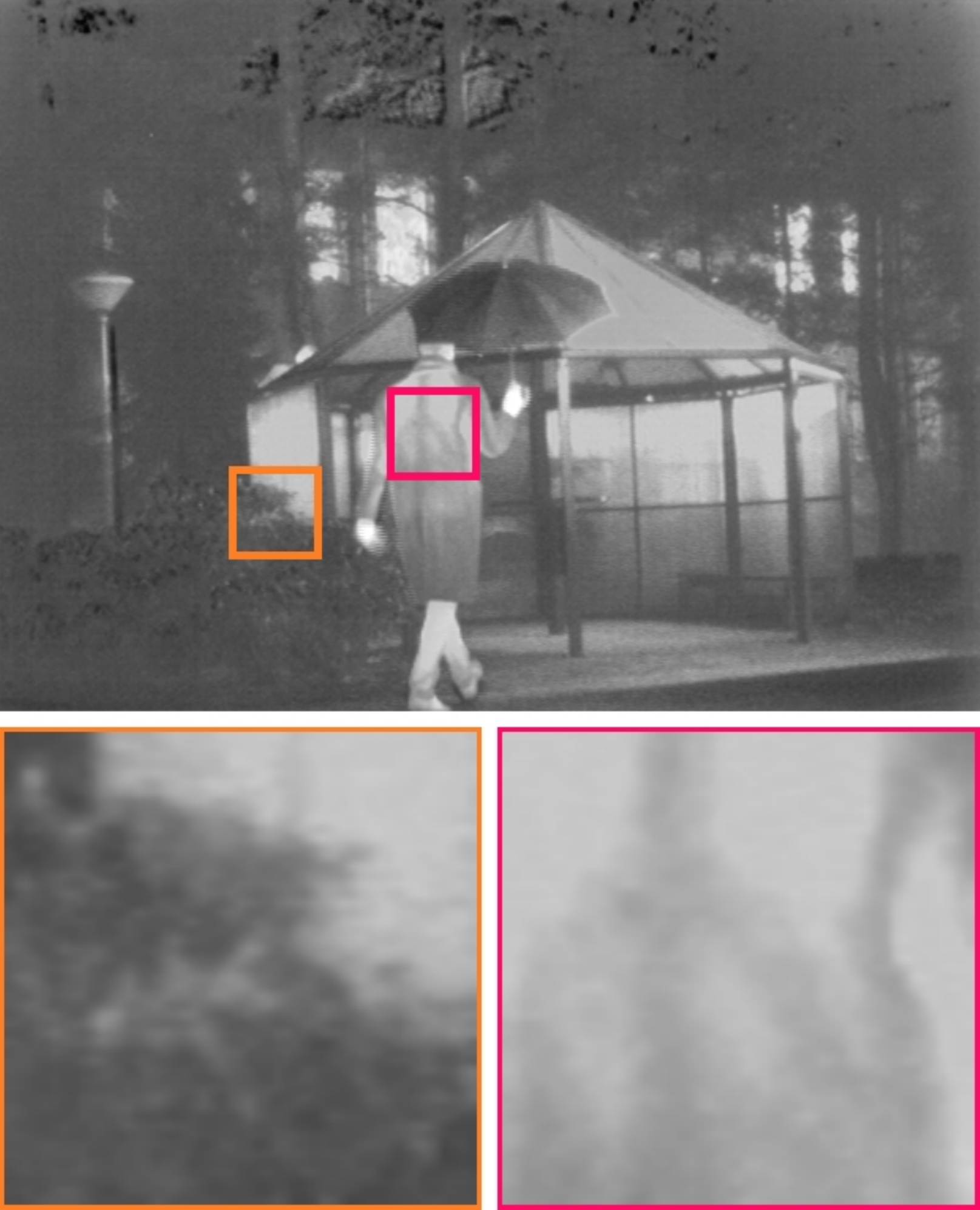}
		&\includegraphics[width=0.12\textwidth]{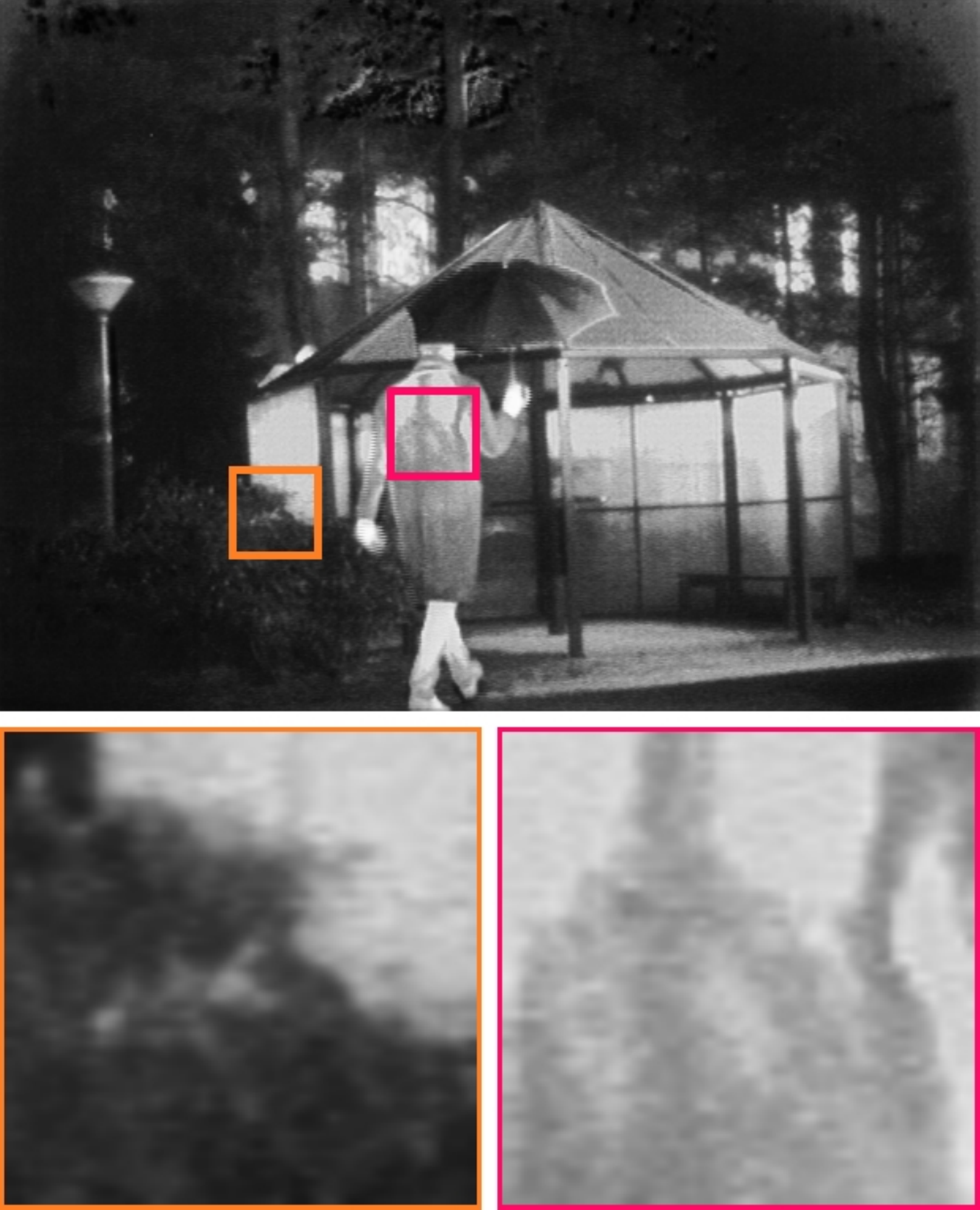}
		&\includegraphics[width=0.12\textwidth]{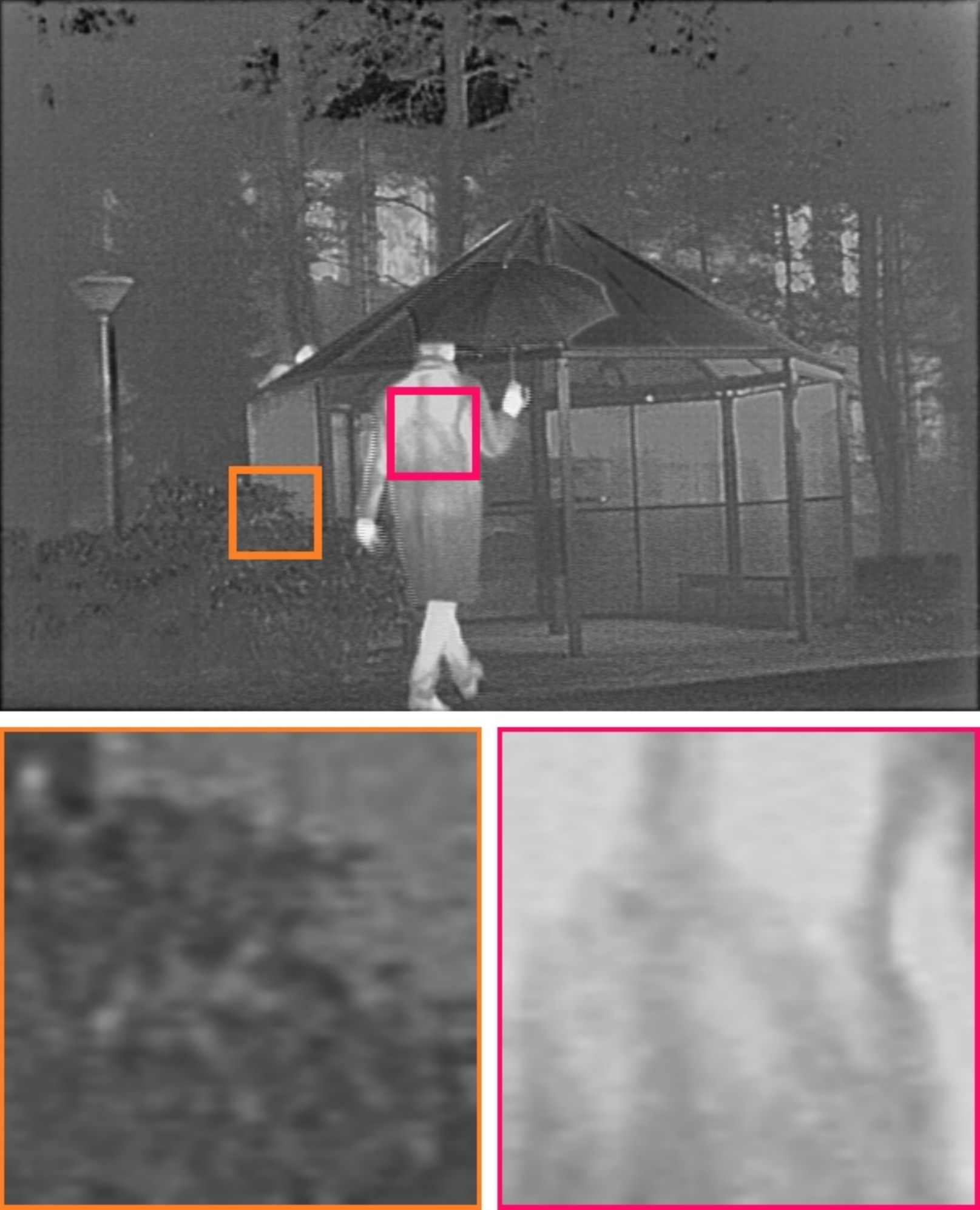}
		&\includegraphics[width=0.12\textwidth]{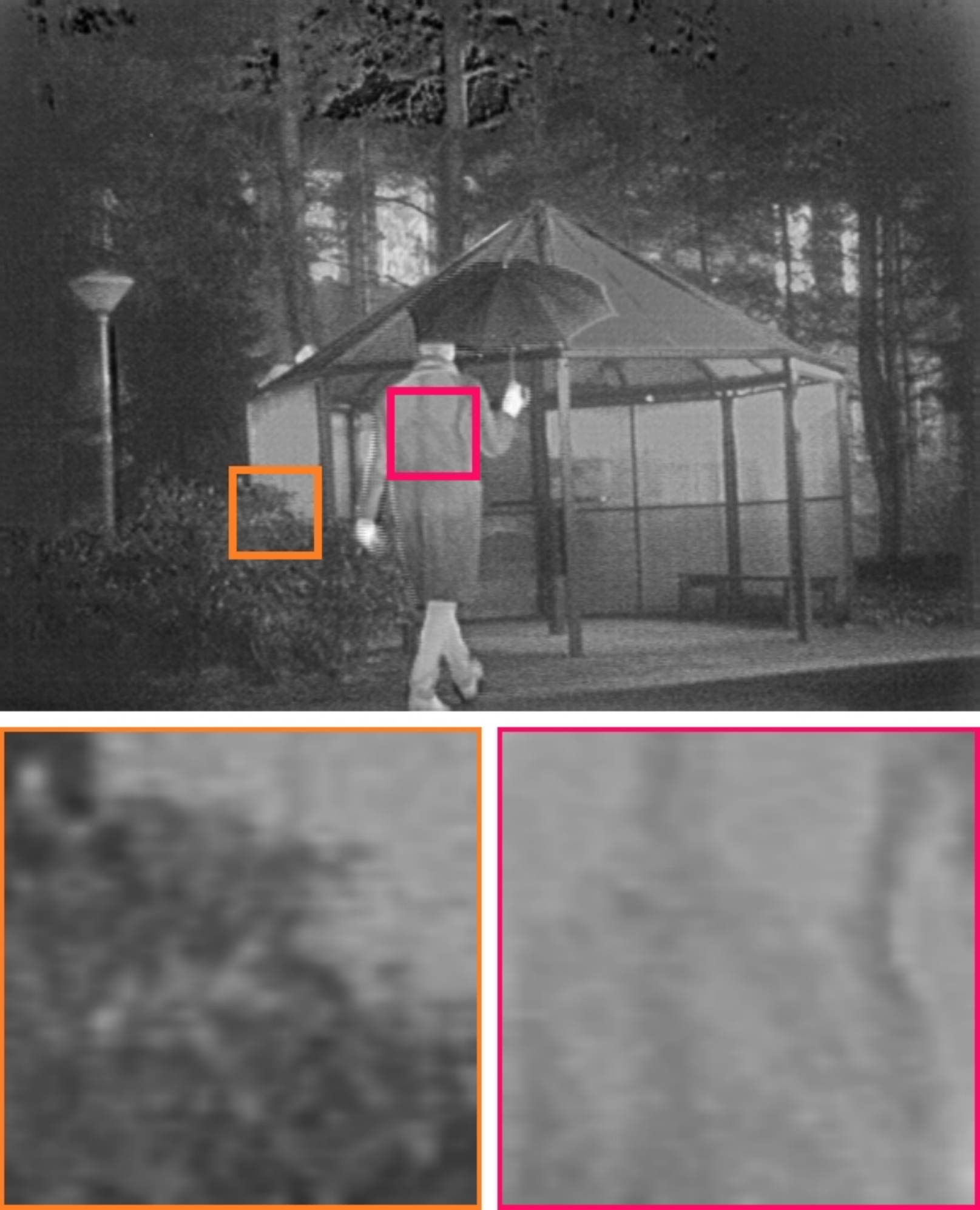}
		&\includegraphics[width=0.12\textwidth]{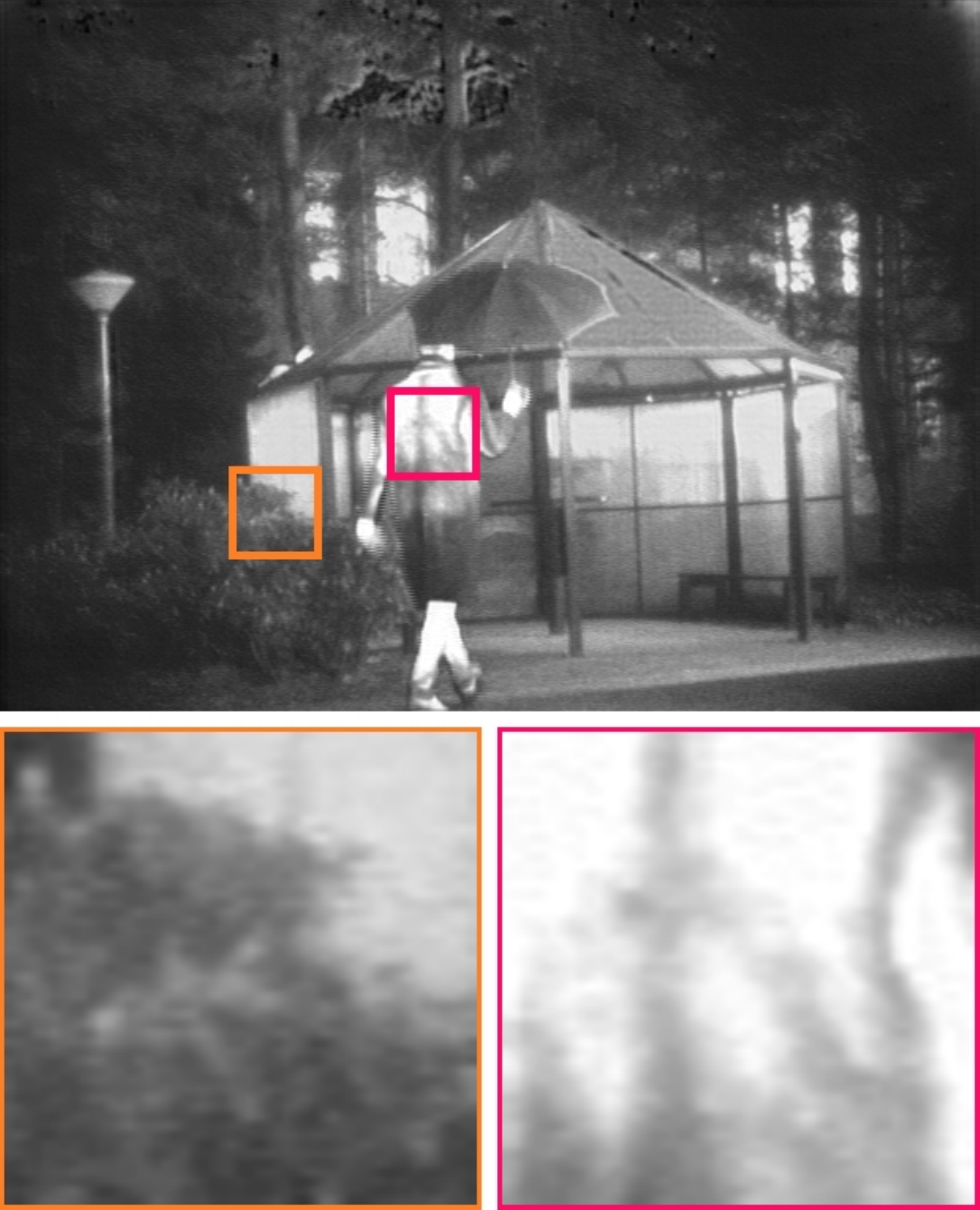}\\
		\footnotesize	Infrared&\footnotesize Visible & \footnotesize DenseFuse    &\footnotesize MFEIF & \footnotesize DID & \footnotesize SDNet & \footnotesize U2Fusion & \footnotesize TIM \\
	\end{tabular}
	\caption{Qualitative comparison of our method with five state-of-the-arts fusion methods on {TNO} dataset.}
	\label{fig:result_ir_vis_tno}
\end{figure*}
\begin{figure*}[!htb]
	\centering \begin{tabular}{c@{\extracolsep{0.05em}}c@{\extracolsep{0.05em}}c@{\extracolsep{0.05em}}c@{\extracolsep{0.05em}}c@{\extracolsep{0.05em}}c@{\extracolsep{0.05em}}c@{\extracolsep{0.05em}}c@{\extracolsep{0.05em}}@{\extracolsep{0.05em}}c}

		\includegraphics[width=0.12\textwidth]{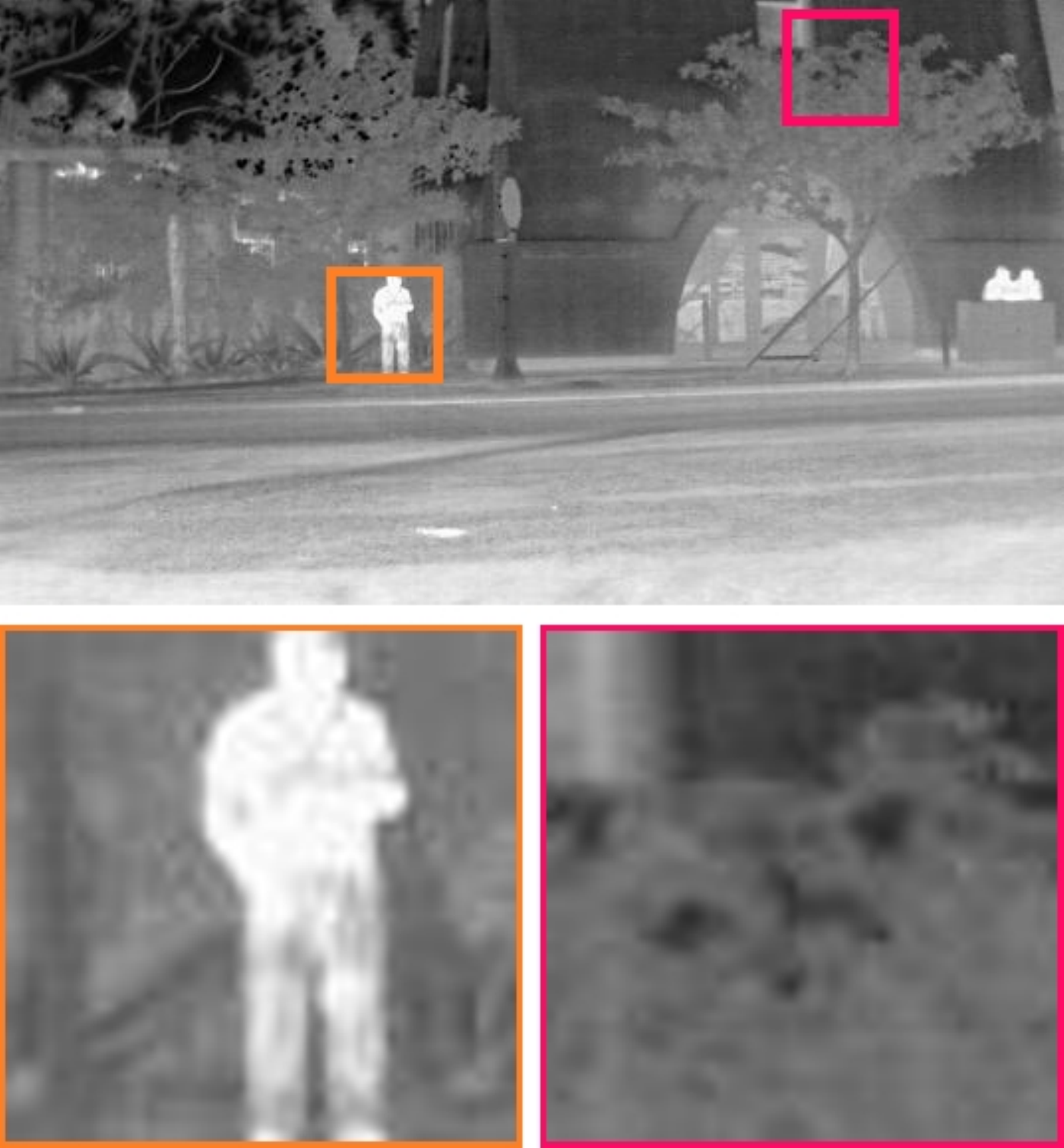}
		&\includegraphics[width=0.12\textwidth]{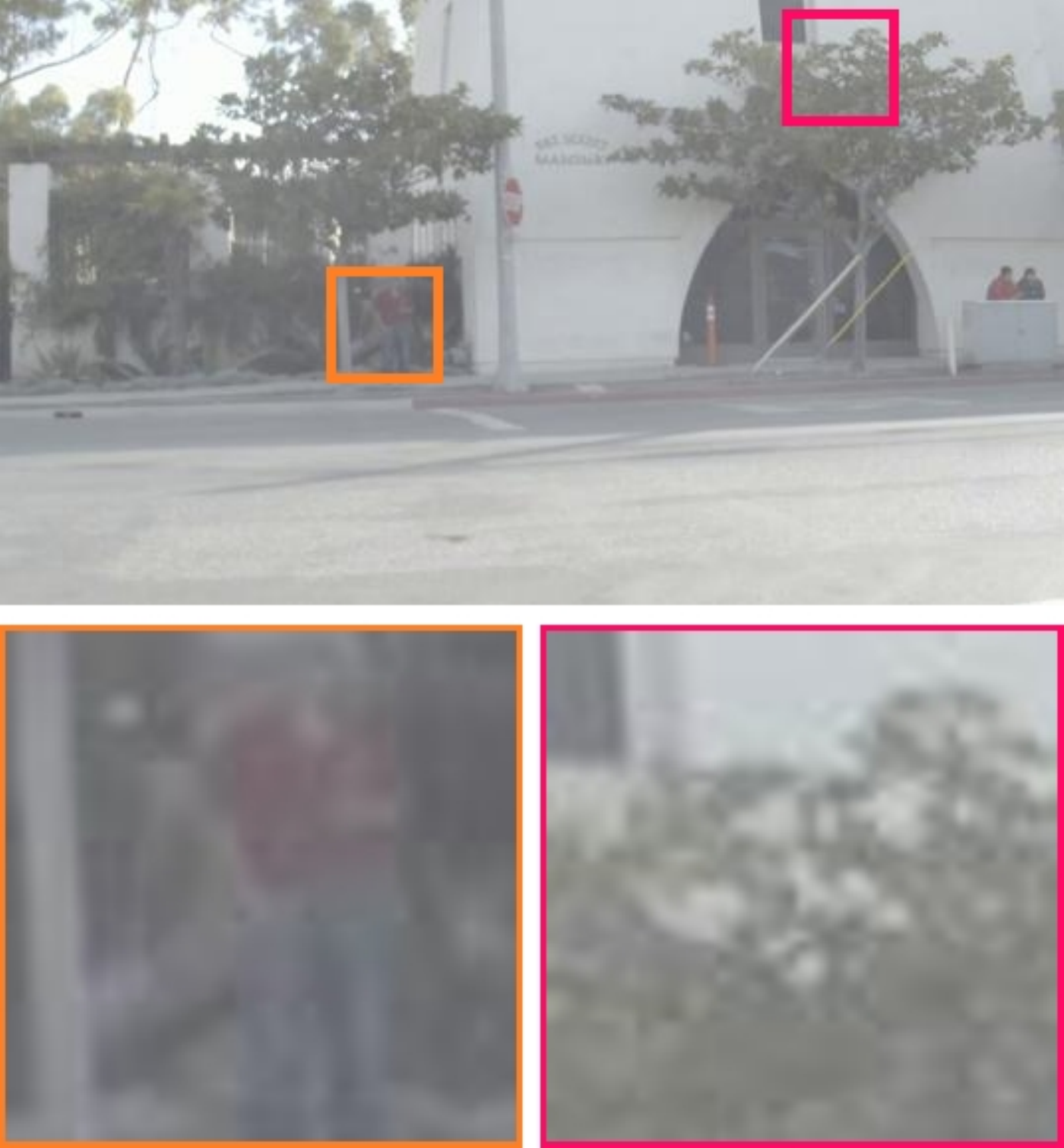}
		&\includegraphics[width=0.12\textwidth]{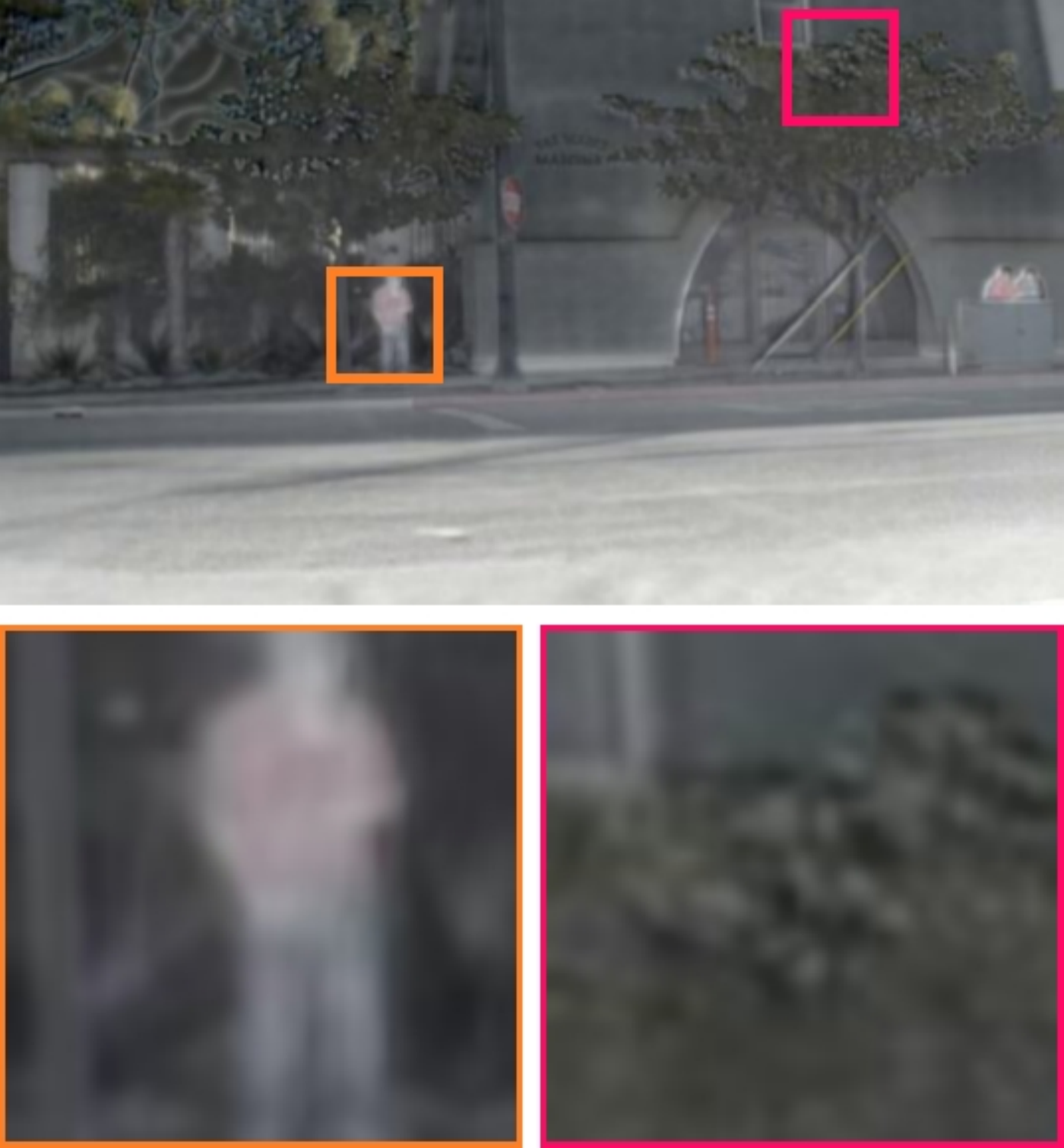}
		
		&\includegraphics[width=0.12\textwidth]{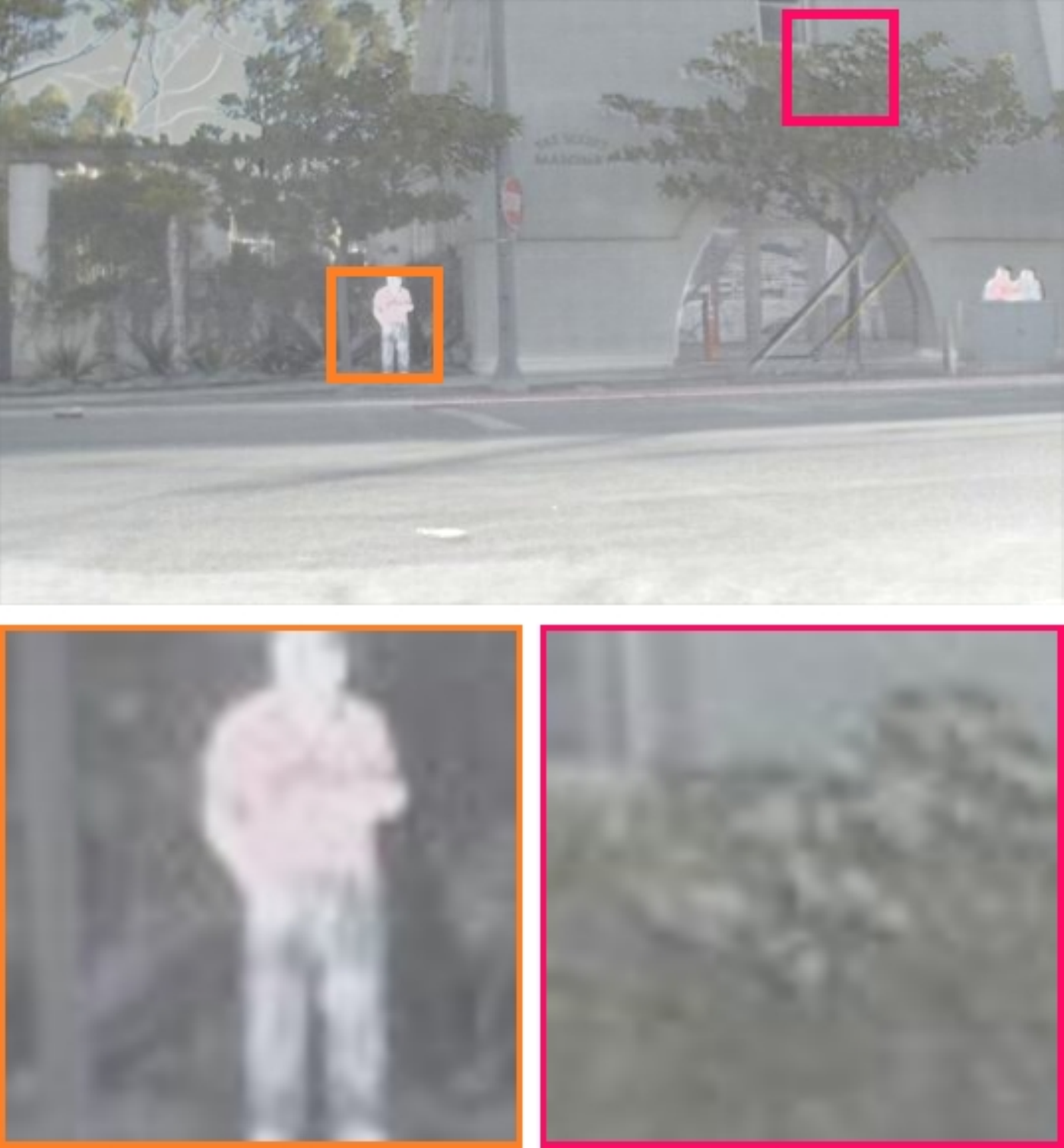}
		&\includegraphics[width=0.12\textwidth]{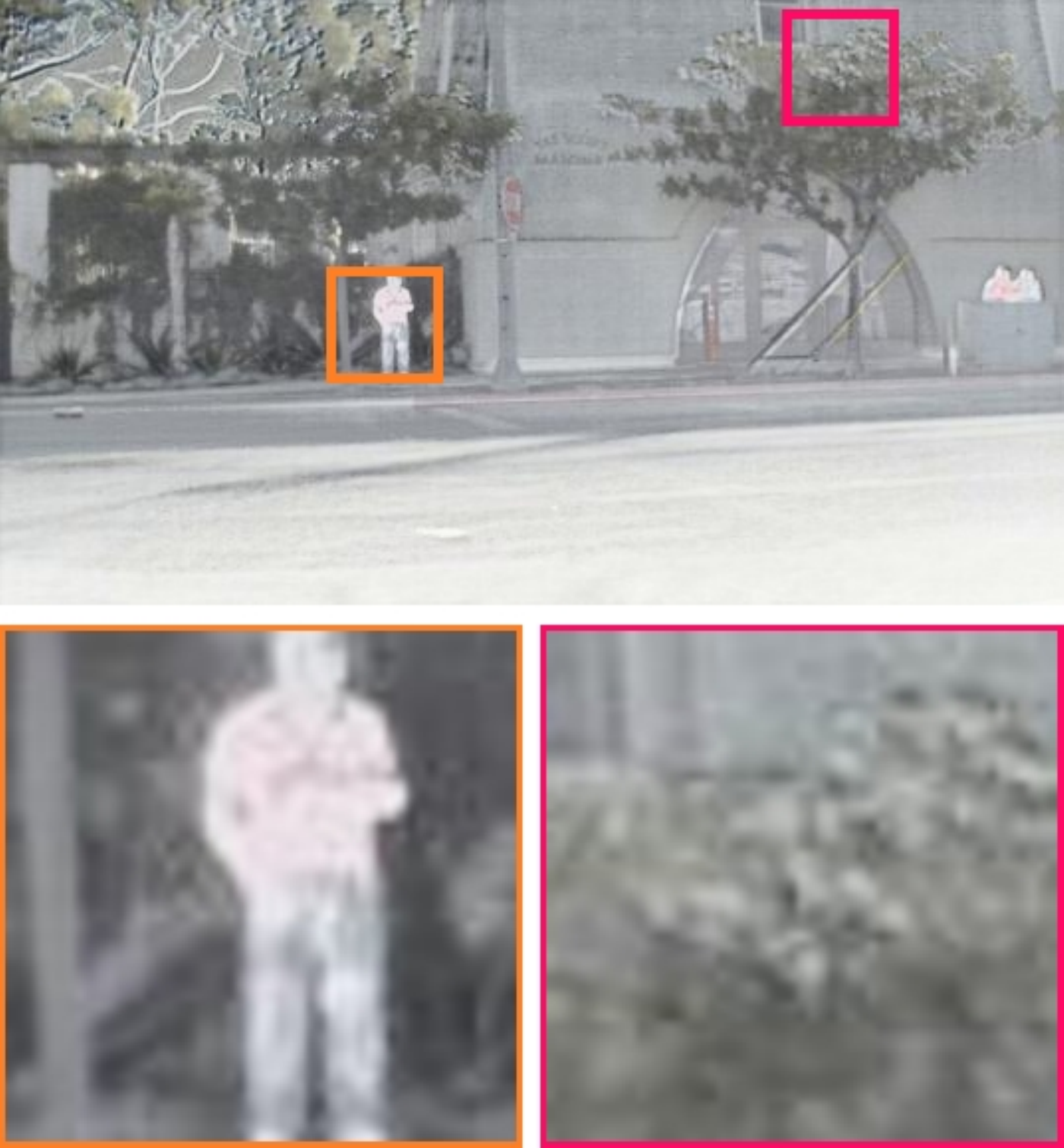}
		&\includegraphics[width=0.12\textwidth]{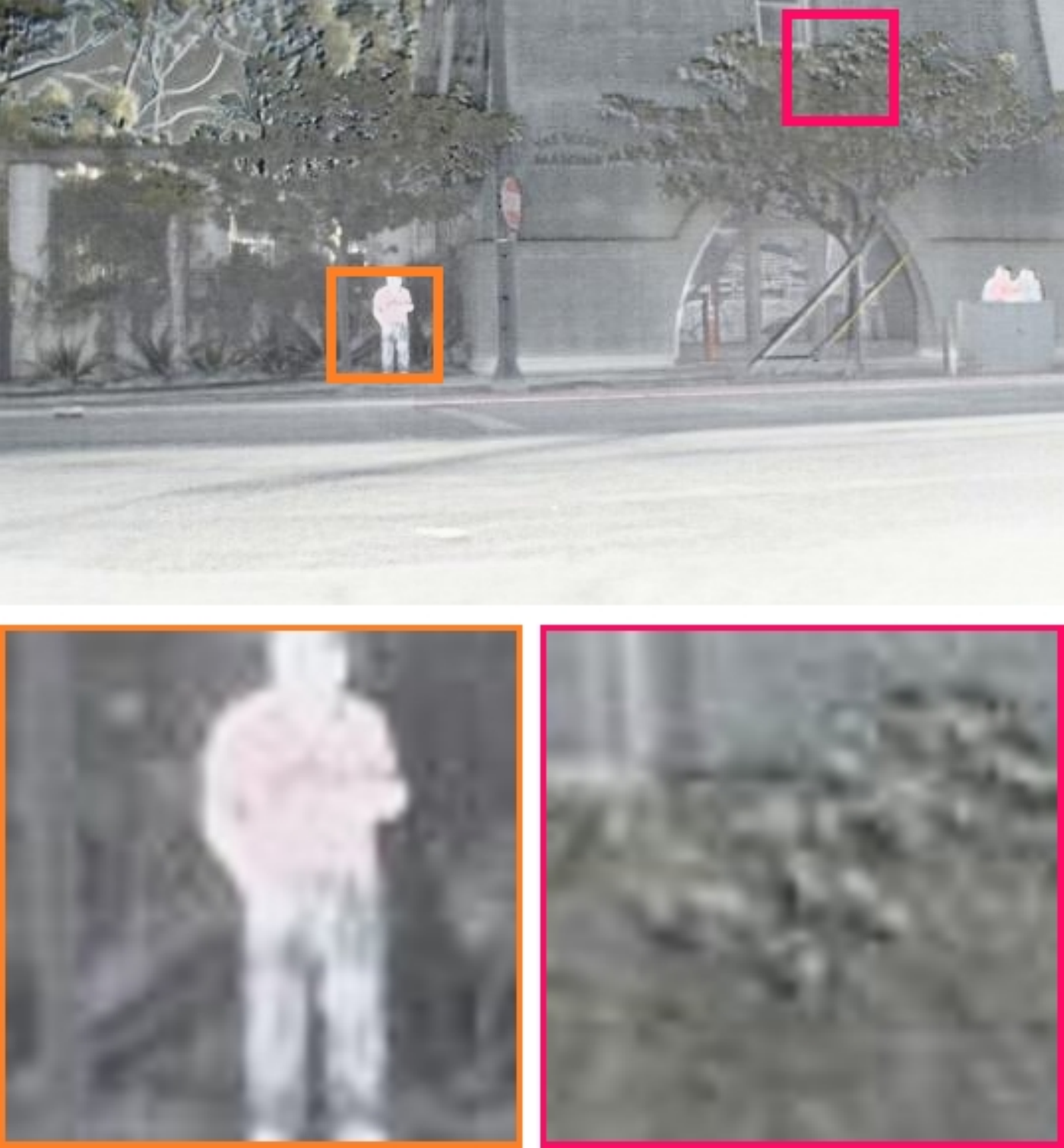}
		&\includegraphics[width=0.12\textwidth]{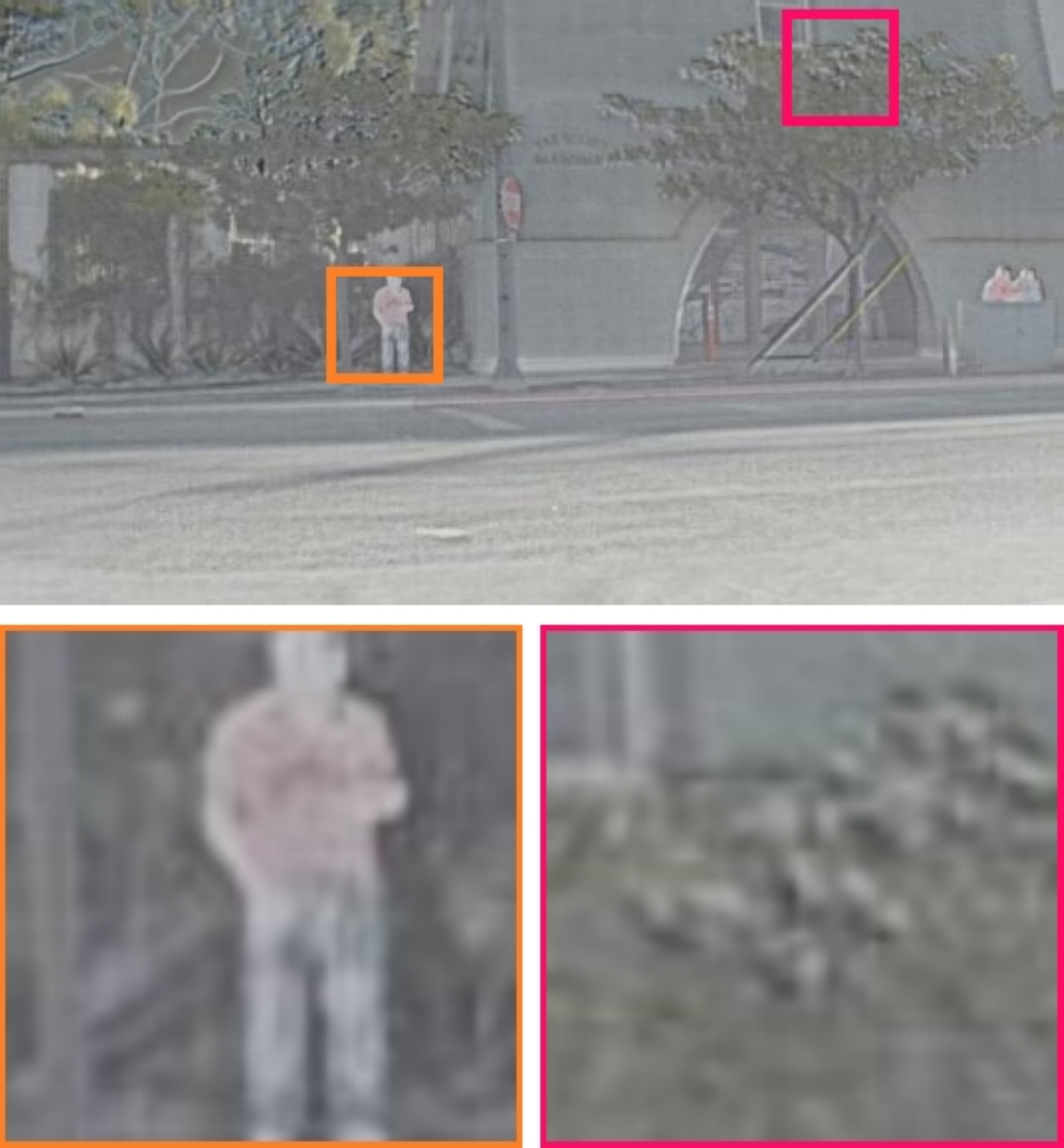}
		&\includegraphics[width=0.12\textwidth]{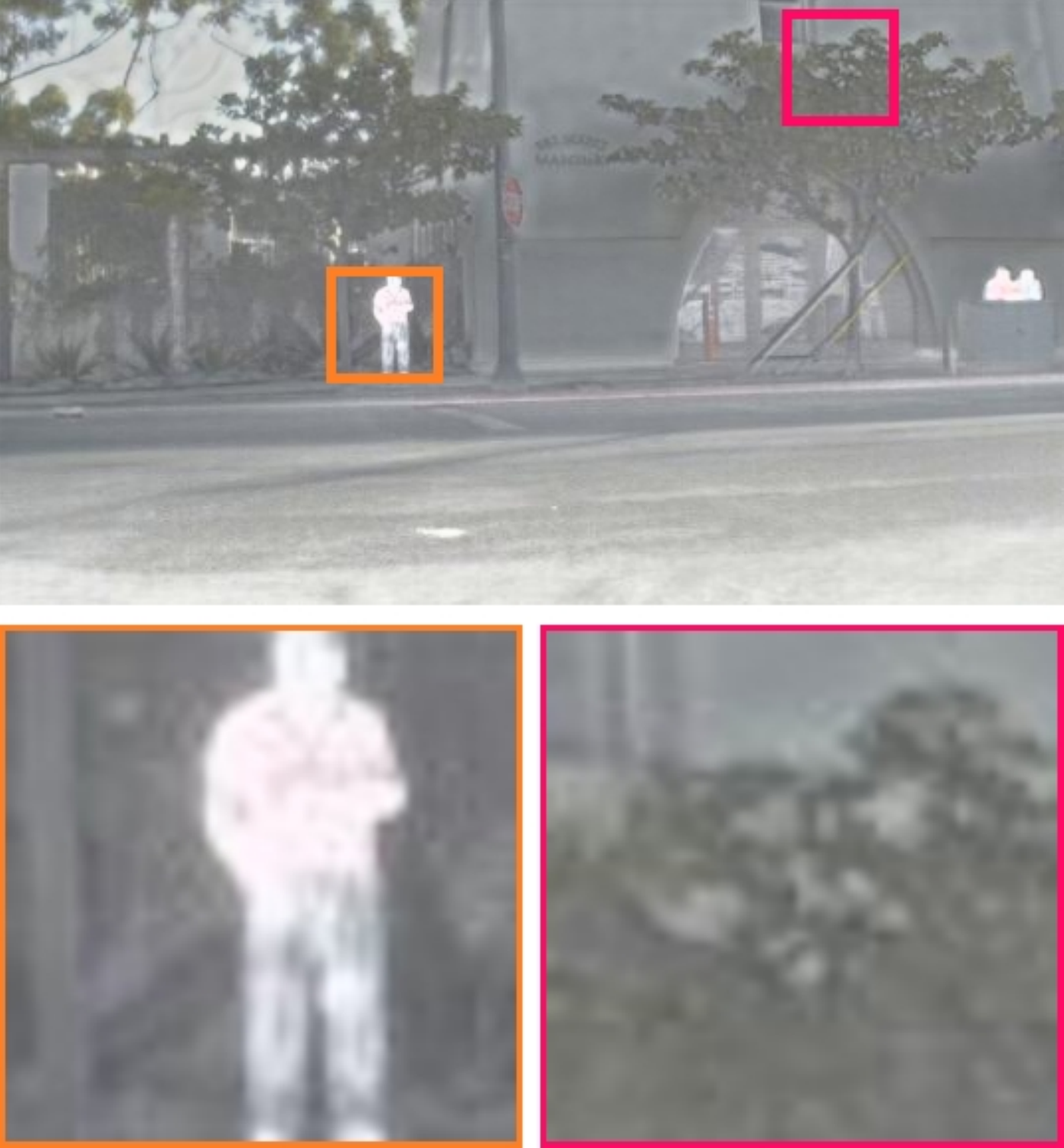}\\
		\includegraphics[width=0.12\textwidth]{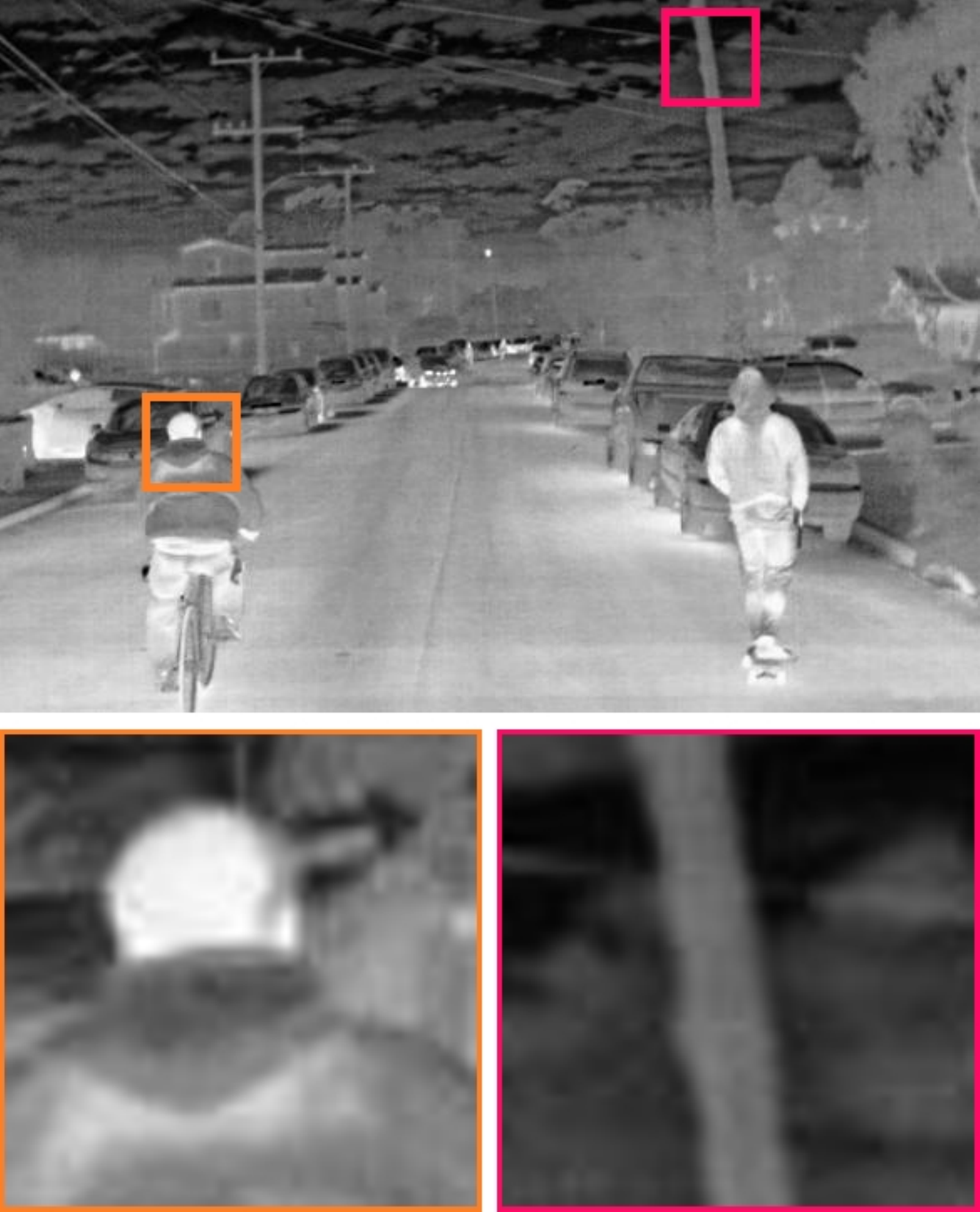}
		&\includegraphics[width=0.12\textwidth]{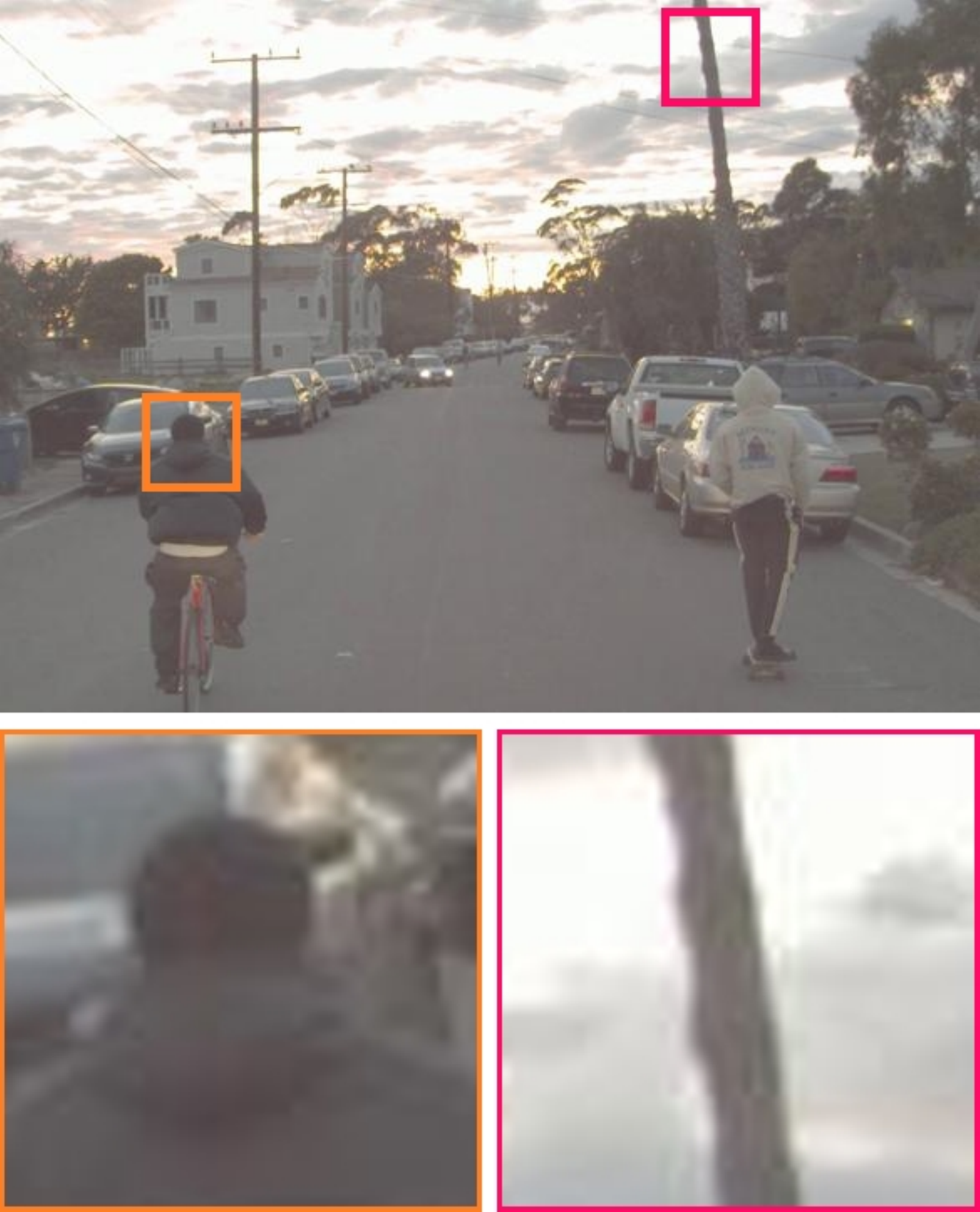}
		&\includegraphics[width=0.12\textwidth]{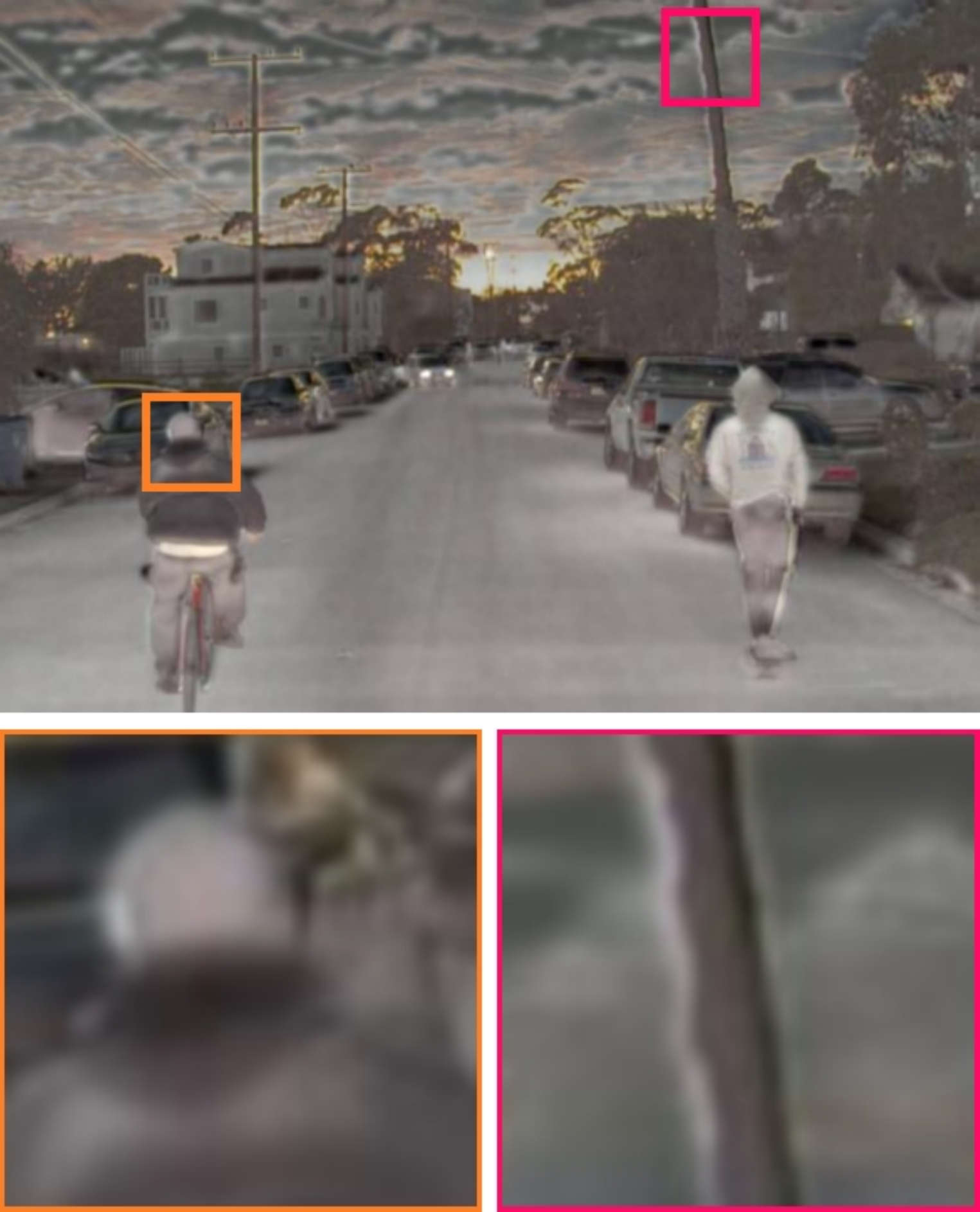}
		
		&\includegraphics[width=0.12\textwidth]{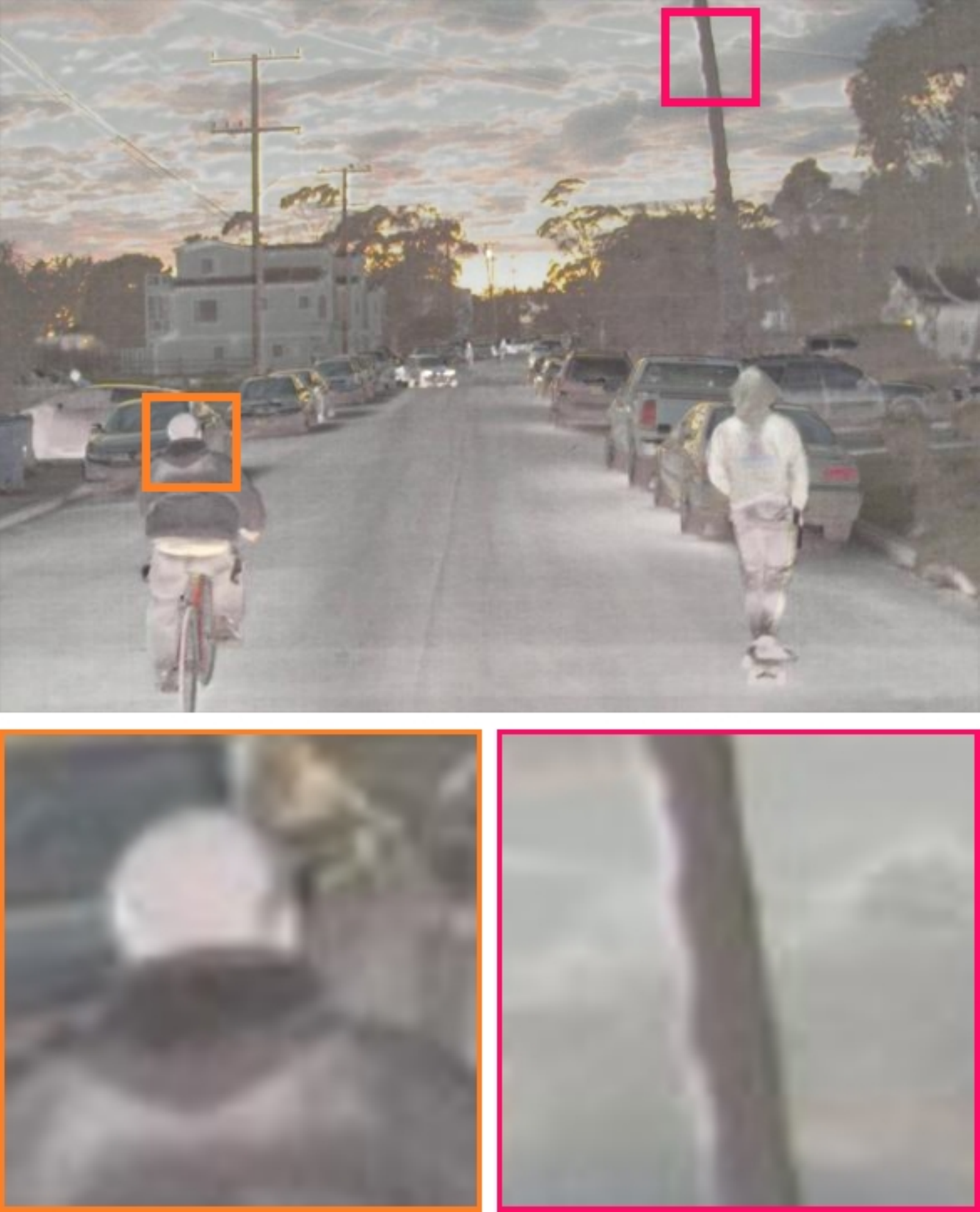}
		&\includegraphics[width=0.12\textwidth]{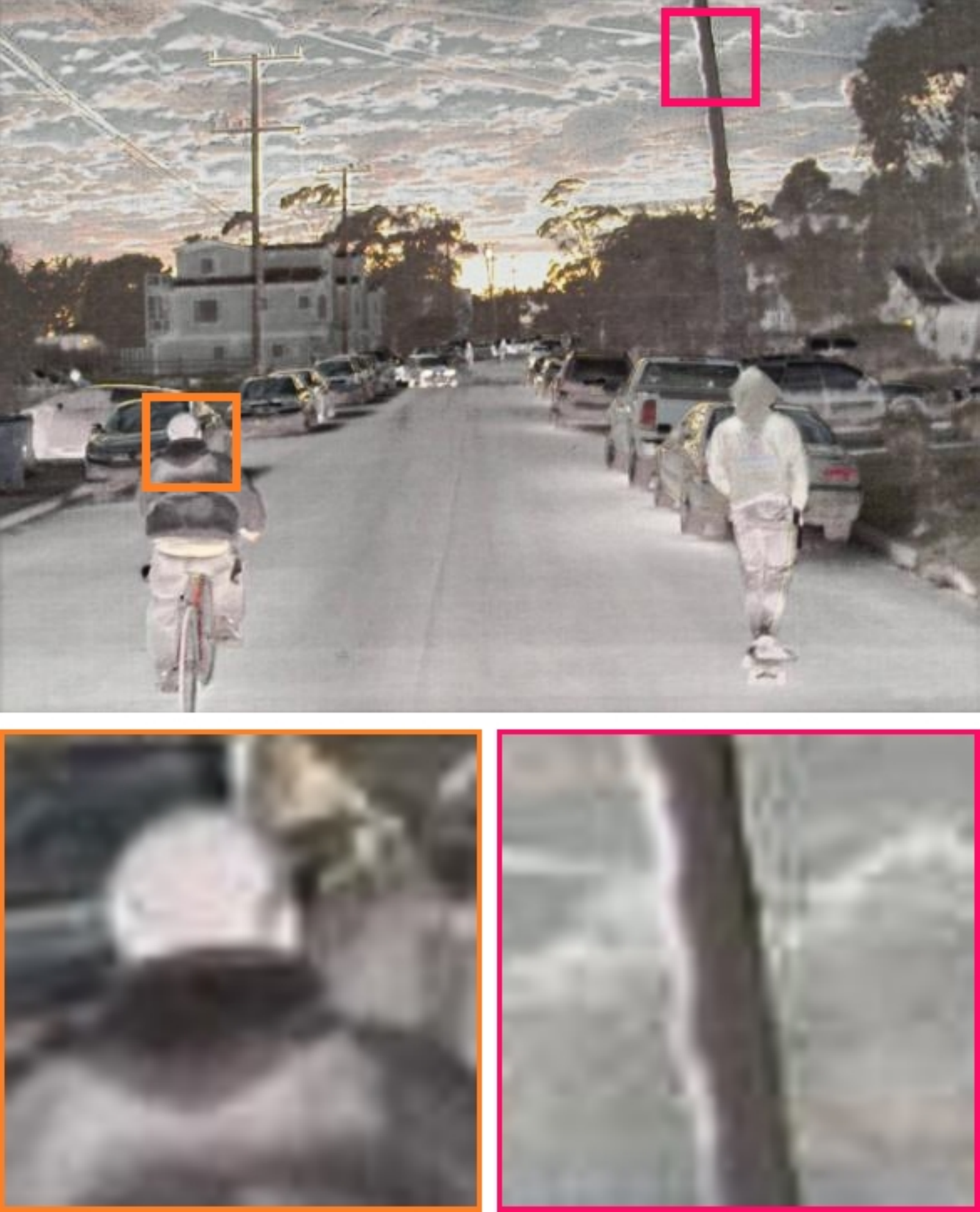}
		&\includegraphics[width=0.12\textwidth]{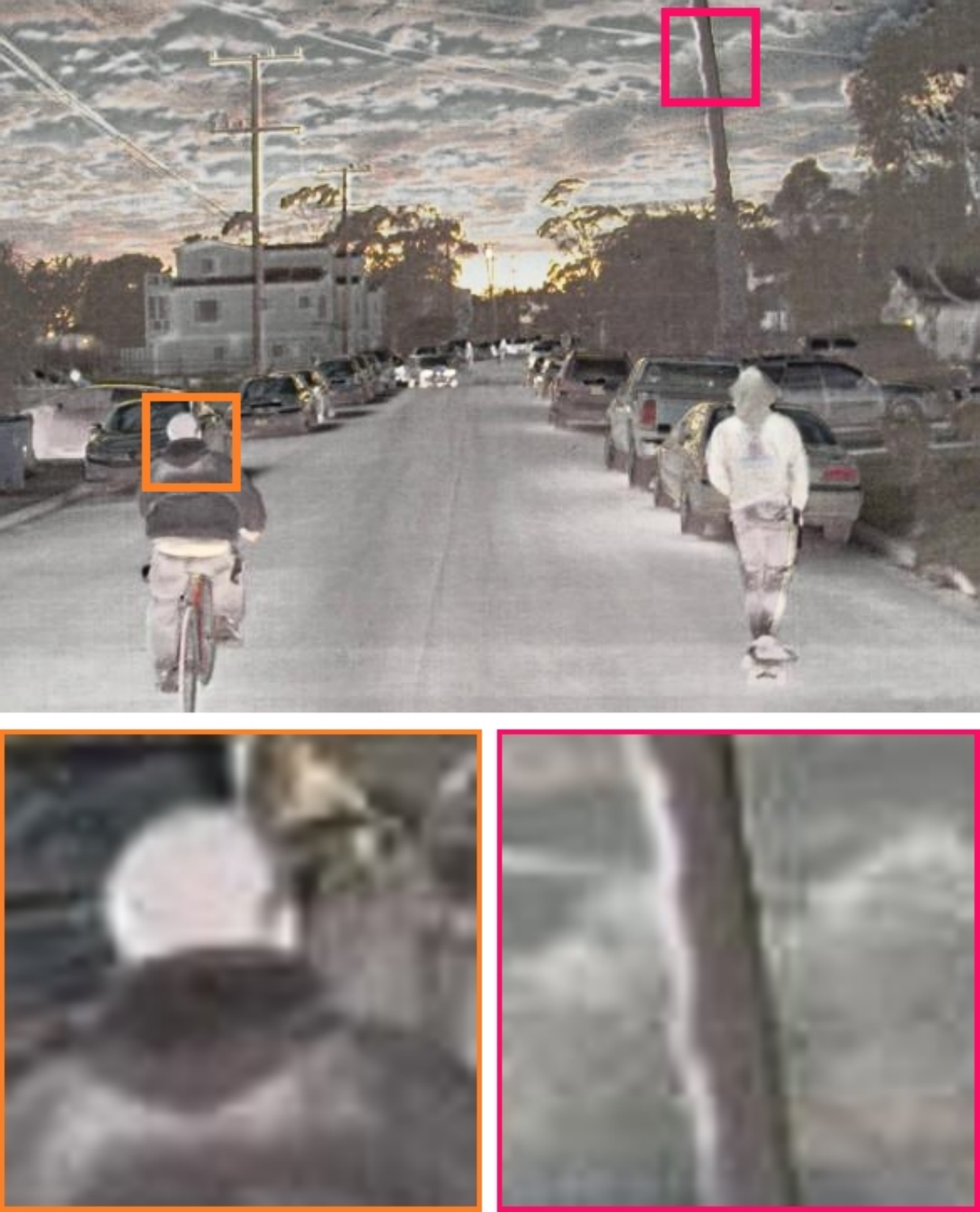}
		&\includegraphics[width=0.12\textwidth]{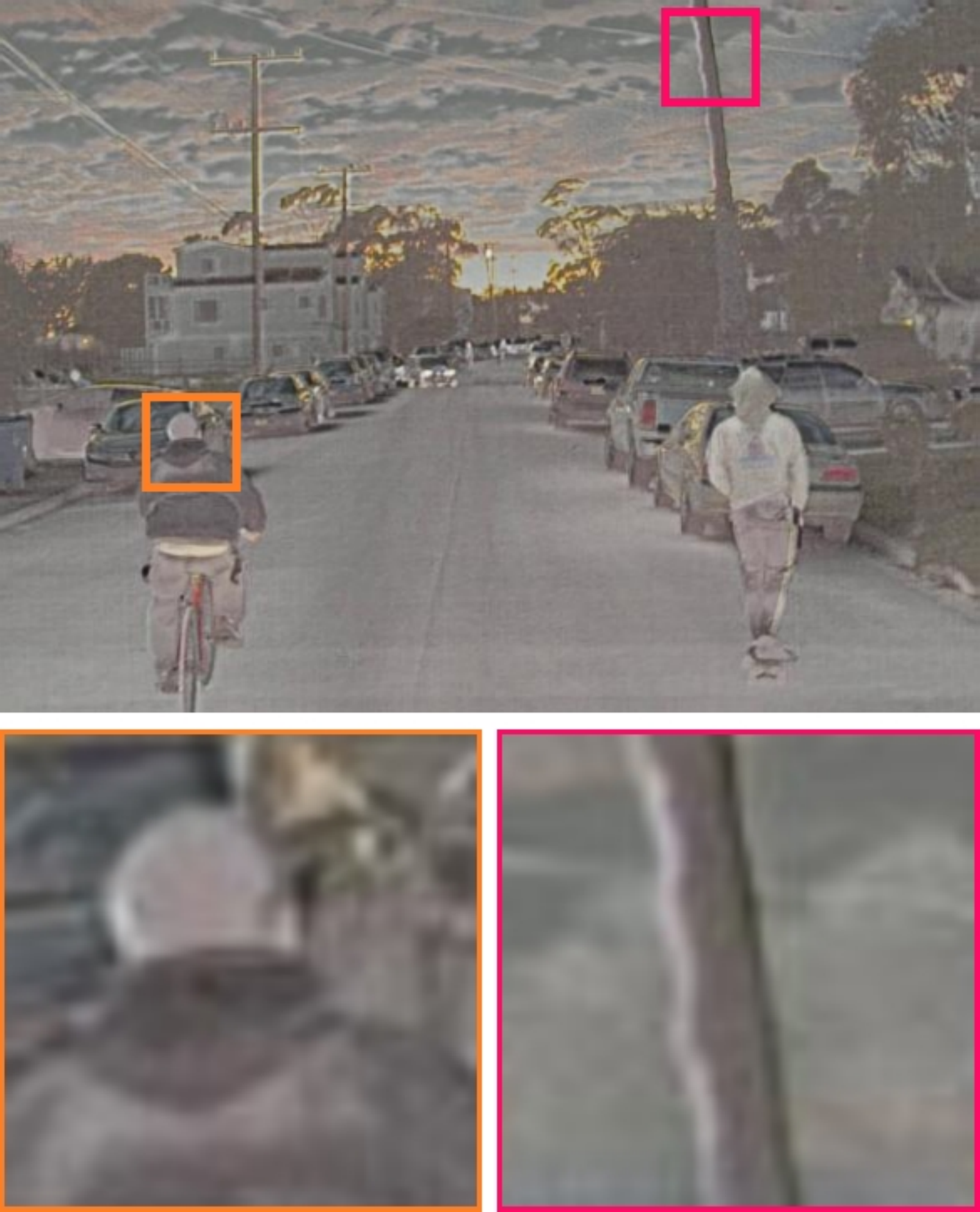}
		&\includegraphics[width=0.12\textwidth]{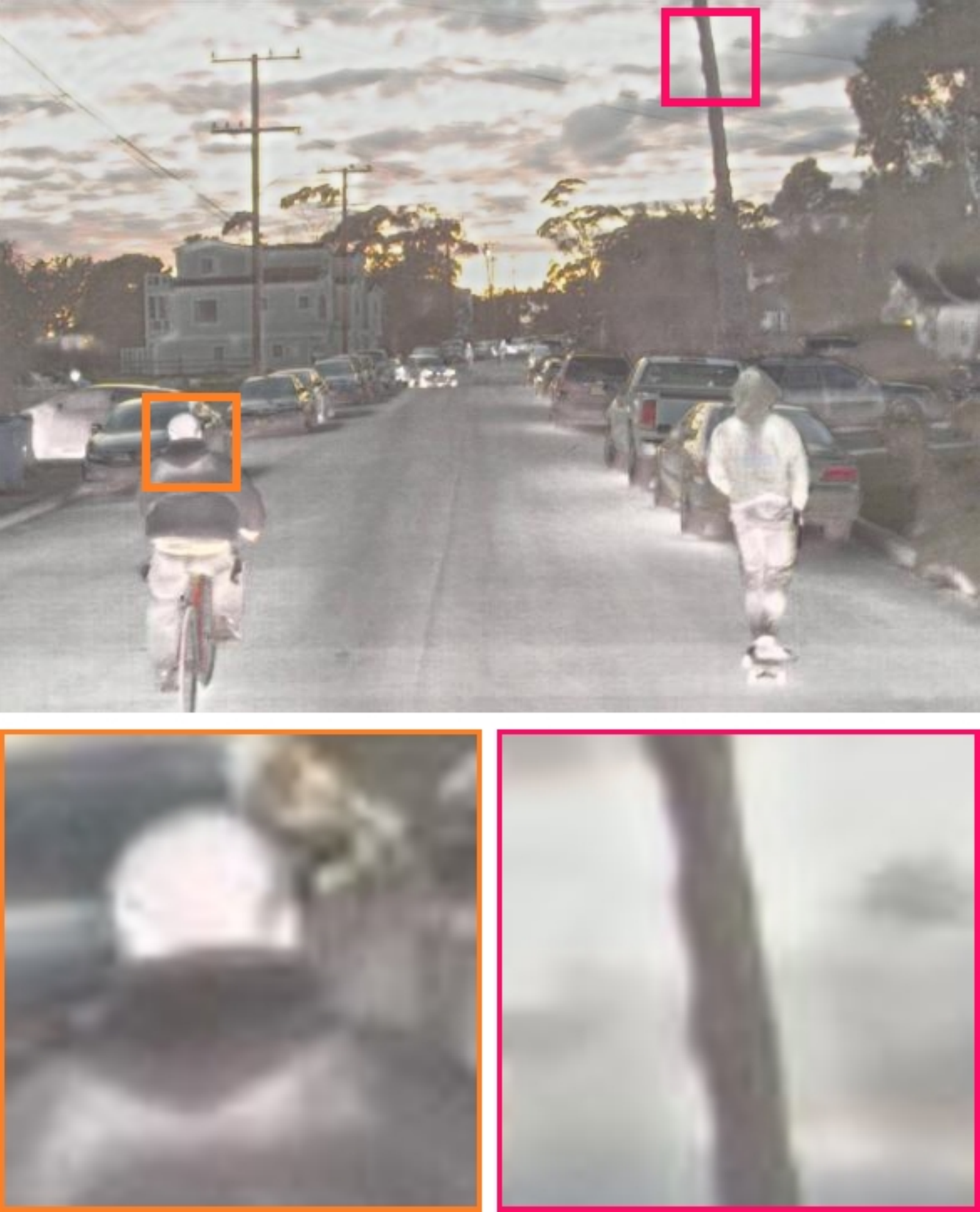}\\
		\includegraphics[width=0.12\textwidth]{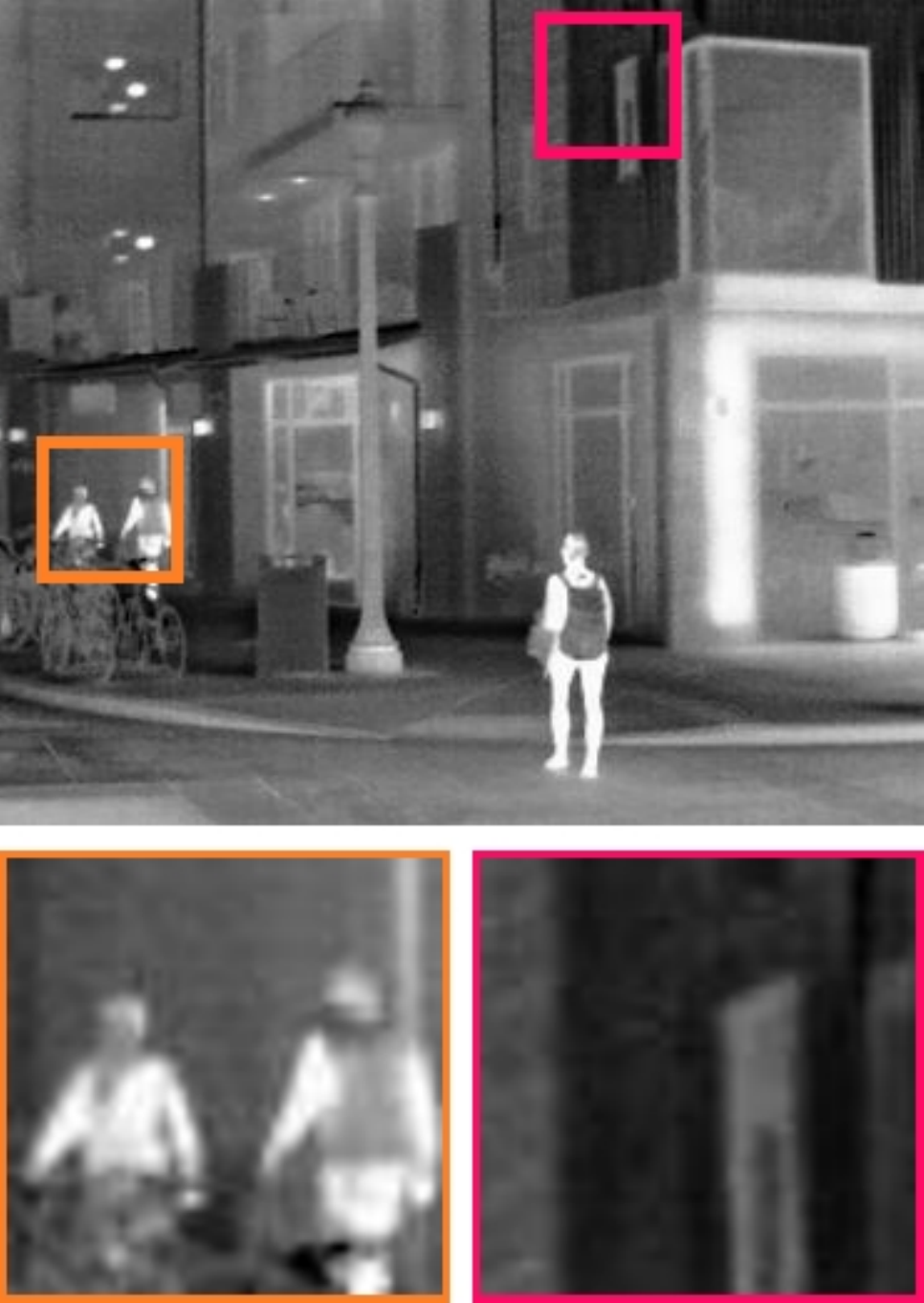}
		&\includegraphics[width=0.12\textwidth]{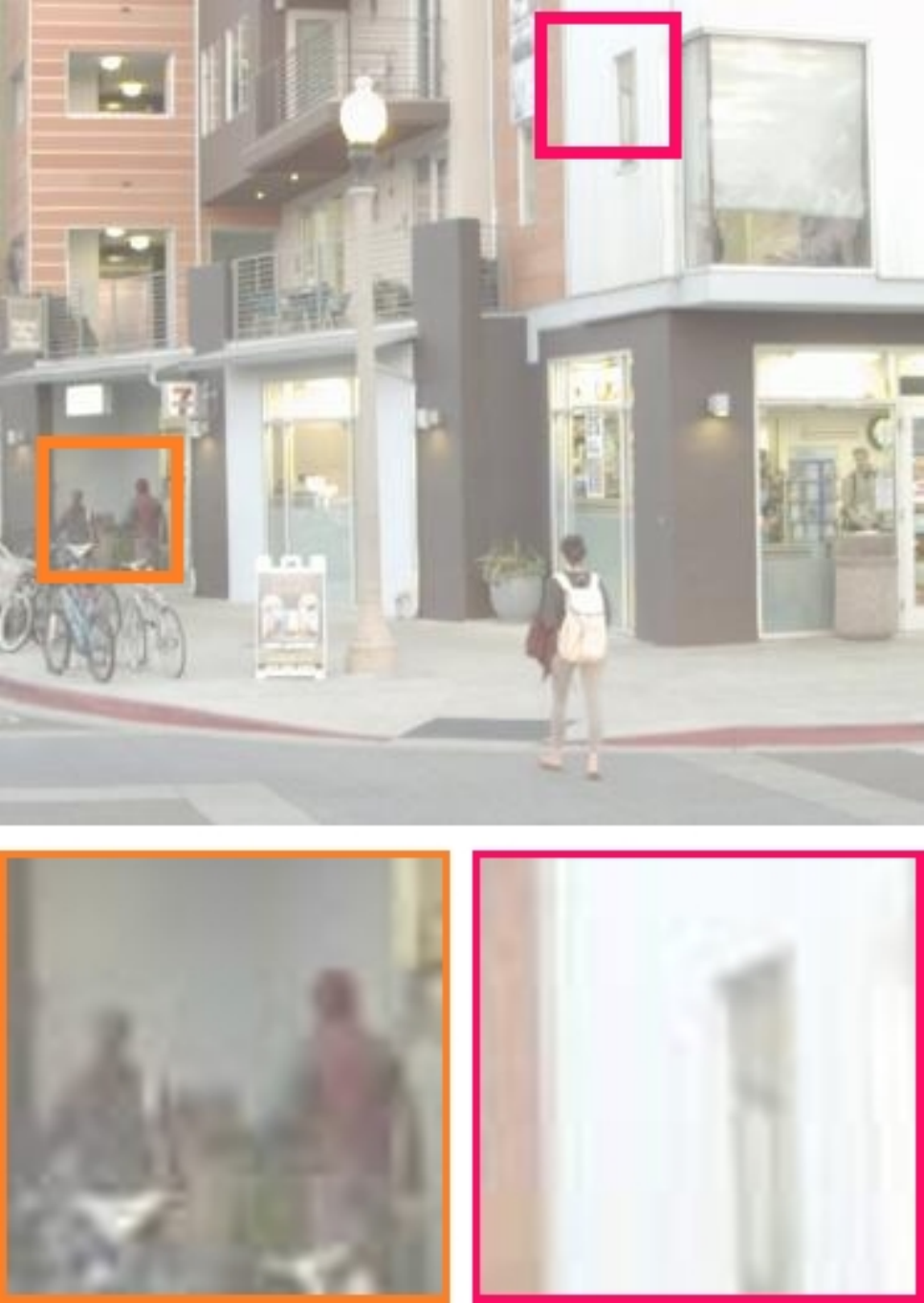}
		&\includegraphics[width=0.12\textwidth]{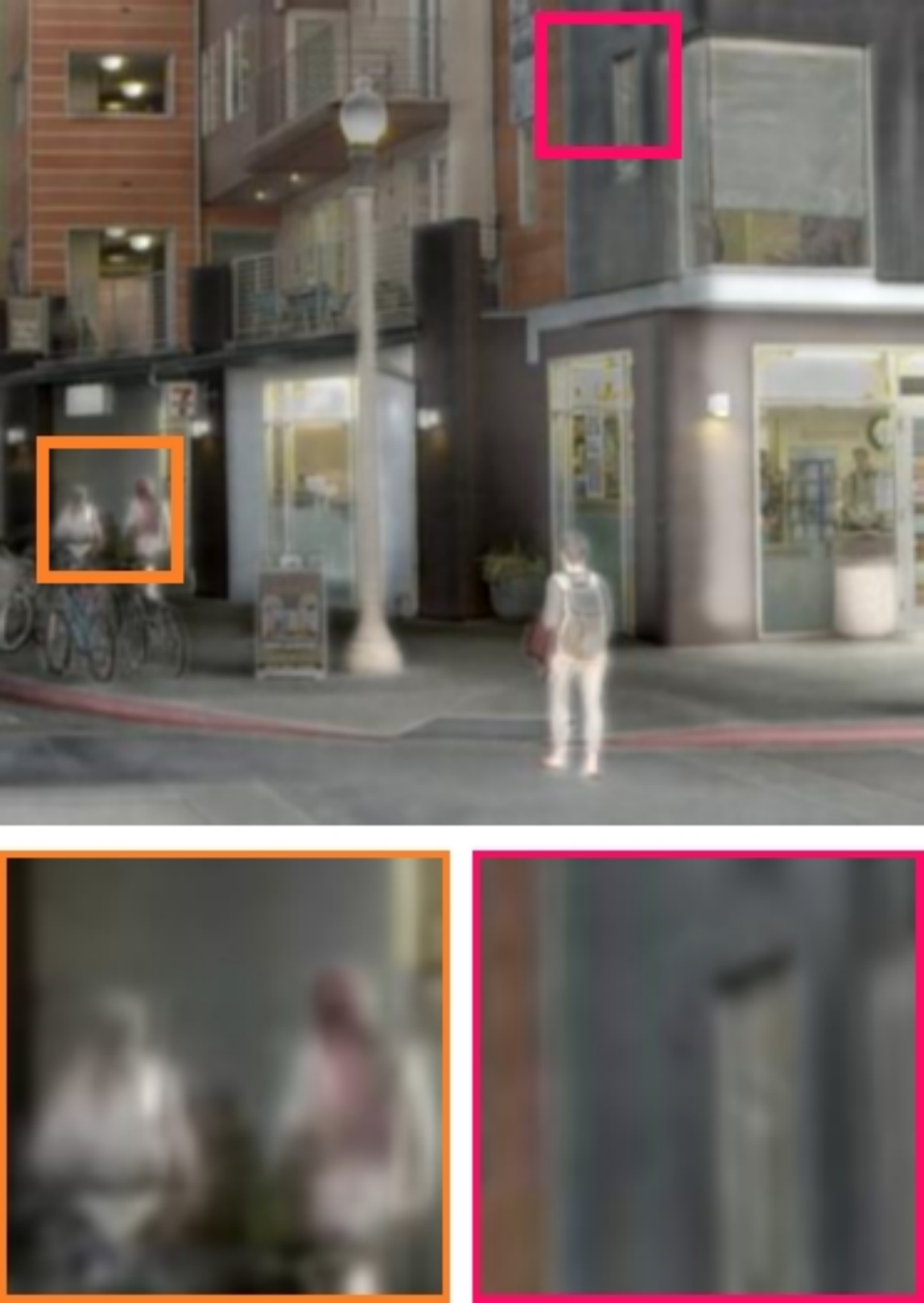}
		
		&\includegraphics[width=0.12\textwidth]{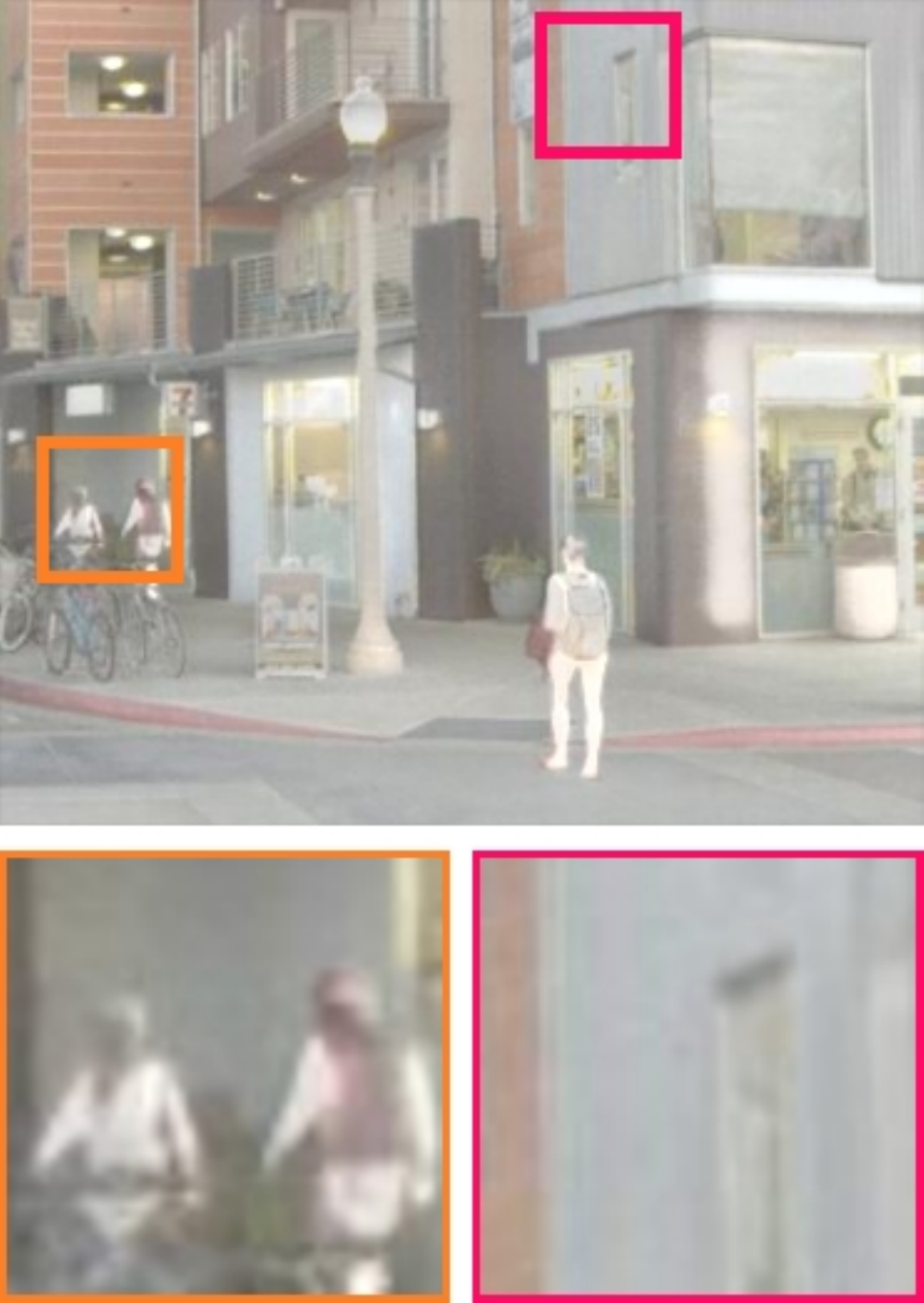}
		&\includegraphics[width=0.12\textwidth]{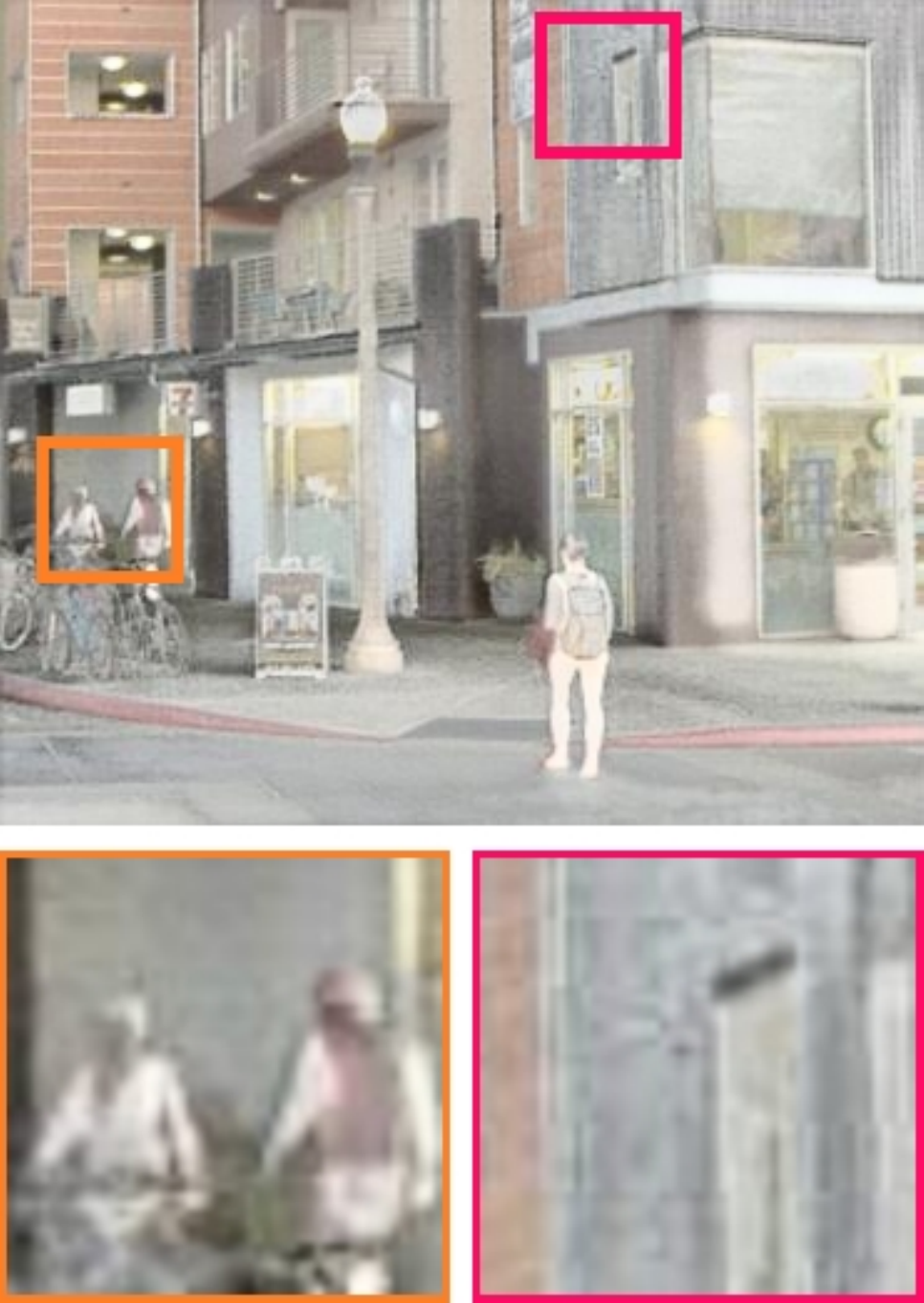}
		&\includegraphics[width=0.12\textwidth]{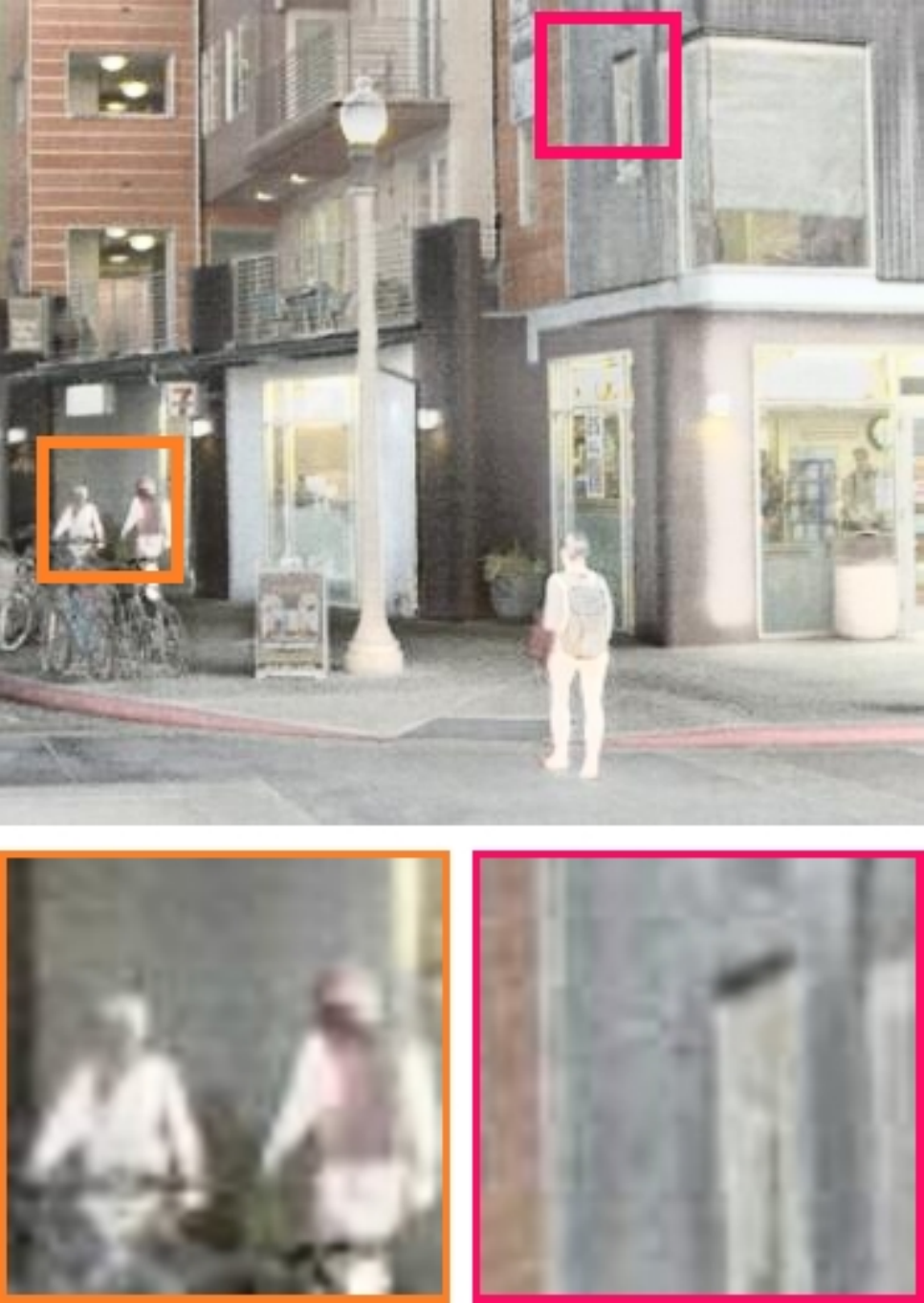}
		&\includegraphics[width=0.12\textwidth]{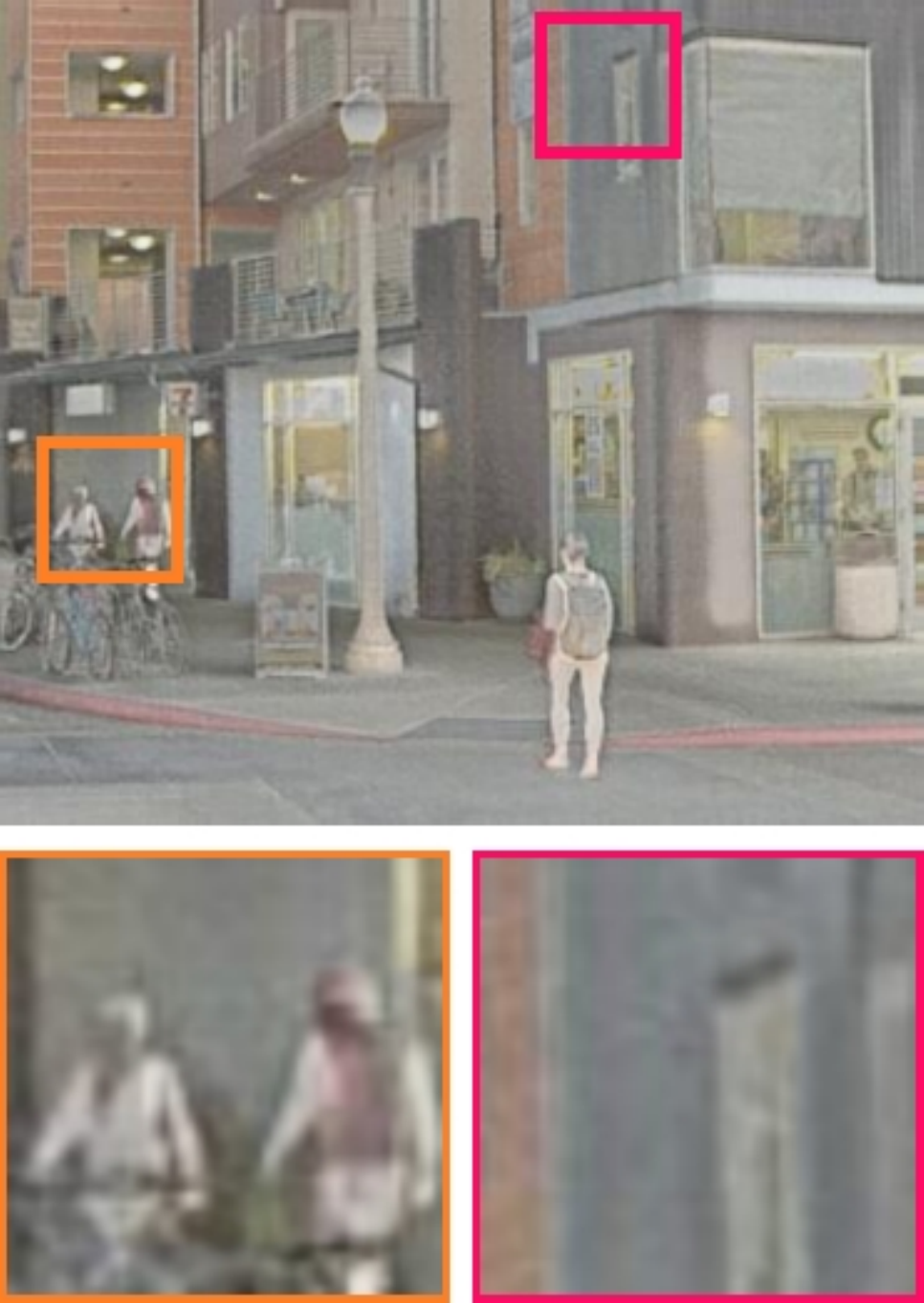}
		&\includegraphics[width=0.12\textwidth]{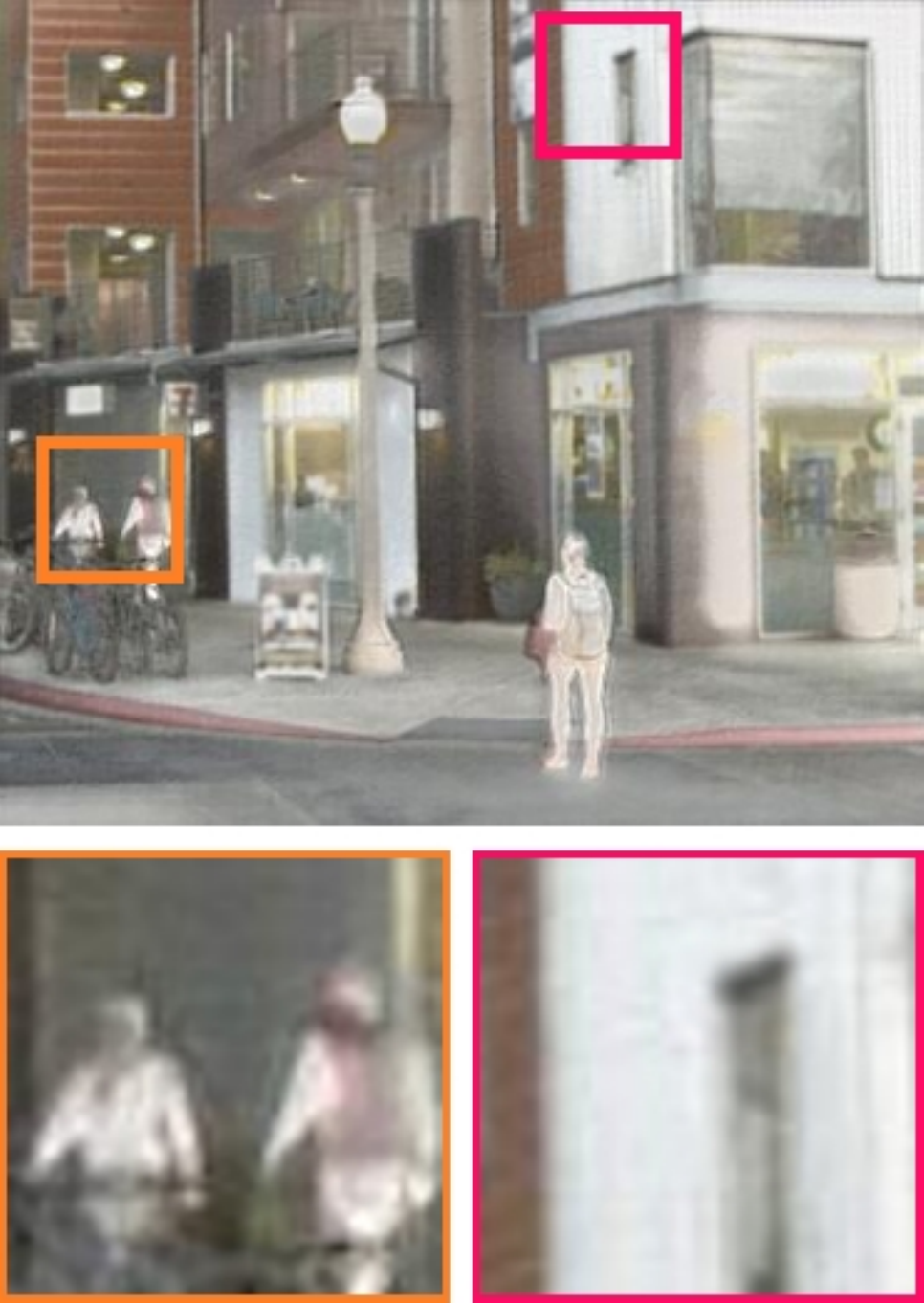}\\
		\footnotesize	Infrared&\footnotesize Visible & \footnotesize RFN    & \footnotesize MFEIF & \footnotesize  DID & \footnotesize AUIF & \footnotesize U2Fusion & \footnotesize TIM \\
	\end{tabular}
	\caption{Qualitative comparison of our method with five state-of-the-arts fusion methods on {RoadScene} dataset.}
	\label{fig:result_ir_vis_road}
\end{figure*}

\begin{table}[htb]
	\renewcommand{\arraystretch}{1.3}
	\caption{Quantitative comparison of joint image registration and fusion.}
	\label{tab:registration}
	\centering \footnotesize
	\setlength{\tabcolsep}{0.8mm}{
		\begin{tabular}{c  c c c c c c c}
			\hline
			Metrics &DenseFuse & AUIF& DID &SDNet & U2Fusion & TARDAL & TIM\\ \hline
			MI&{2.928} & 2.620& 2.654&2.622&2.202& 2.796&\textbf{2.954}\\
			FMI &\textbf{0.722}&0.707&0.708&0.711&0.713&0.710&{0.719}\\
			VIF &{0.529}&0.488&0.490&0.434&0.450&0.490&\textbf{0.560}\\
			$\mathrm{Q^{AB/F}}$&  {0.302} &0.284&0.290&0.295&0.301&0.258&\textbf{0.303}\\
			\hline
		\end{tabular}	
	}
\end{table}

\subsection{Image Fusion for  Visual Enhancement}
In this part, we performed comprehensive experiments to demonstrate our superiority based on objective and subjective evaluations on Infrared-Visible Image Fusion (IVIF). In order to verify the flexibility of our method, we extend the scheme to address Medical Image Fusion (MIF).
\subsubsection{Infrared-Visible Image Fusion}
We compared with ten  state-of-the-art learning-based competitors, which include DDcGAN~\cite{ma2020ddcgan}, RFN~\cite{li2021rfn}, DenseFuse~\cite{li2018densefuse}, FGAN~\cite{ma2019fusiongan}, DID~\cite{ZhaoDIDFuse2020}, MFEIF~\cite{liu2021learning2}, SMOA~\cite{liu2021smoa}, TARDAL~\cite{TarDAL}, SDNet~\cite{zhang2021sdnet} and U2Fusion~\cite{xu2020u2fusion}. The outer structure of our final searched architecture are $\mathbf{C}_\mathtt{{MS}}$ and $\mathbf{C}_\mathtt{{SC}}$ for fusion, $\mathbf{C}_\mathtt{{SC}}$ and $\mathbf{C}_\mathtt{{SC}}$ for enhancement. The inner operators are 3-RB, 3-DC, 3-DB, 3-DC, SA, 3-DC, CA and SA respectively.

\textbf{Qualitative Comparisons.} We perform the objective evaluation on two representation datasets {TNO} and {RoadScene} in Fig.~\ref{fig:result_ir_vis_tno} and Fig.~\ref{fig:result_ir_vis_road}. From this intuitive view, three discriminative advantages can be concluded. Firstly, our scheme can highlight significant high-contrast and clear thermal objects, as shown in the first and third row of Fig.~\ref{fig:result_ir_vis_tno}.  However DenseFuse and U2Fusion maintain abundant texture features from different modalities, the remarkable
targets of thermal radiation cannot be preserved well. Second, the proposed method effectively preserves ample texture and structure information in visible images. As shown in the first row of Fig.~\ref{fig:result_ir_vis_tno} and the last row of Fig.~\ref{fig:result_ir_vis_road}, the sky, ground material and the color of wall are sufficiently maintained in our results, which is consistent with human visual system. Suffering from the  strong pixel intensity of infrared image, most of fusion schemes have color distortion, cannot preserve rich texture structure. Furthermore, the proposed scheme can efficiently remove
artifacts coming from different modalities, such as the thermal blur and visible noise. For instance, MFEIF, DID and AUIF schemes contain obvious noises and artifacts shown in
 the second row of Fig.~\ref{fig:result_ir_vis_tno} and  Fig.~\ref{fig:result_ir_vis_road}. In contrast, our scheme not only highlights
 the distinct infrared targets but also preserves textural details, accomplishing the comprehensive  results.

\begin{figure}[htb]
	\centering \begin{tabular}{c}
		
		\includegraphics[width=0.45\textwidth]{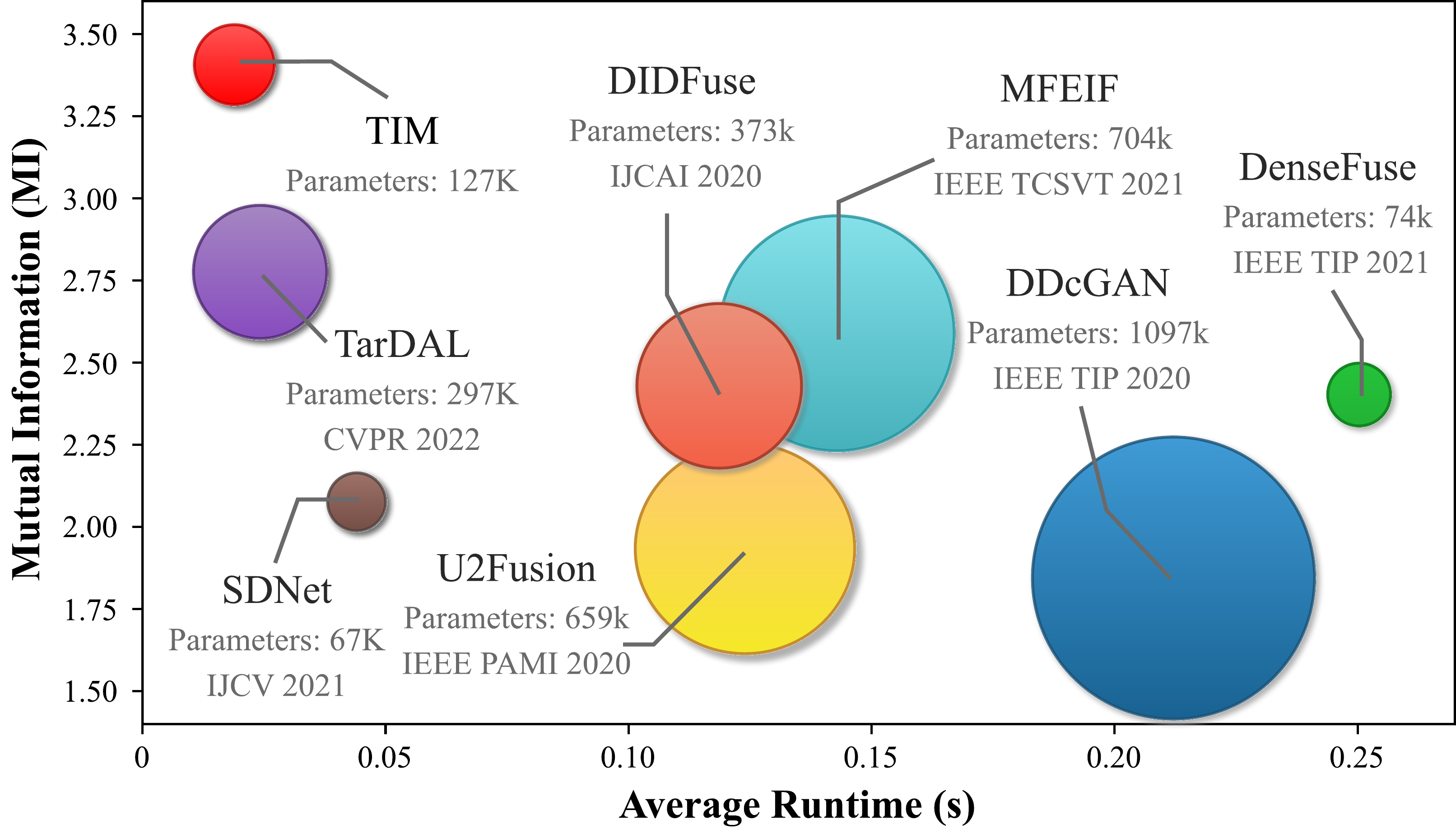} 
	\end{tabular}
	\caption{Comprehensive analysis of the proposed scheme on fusion quality, computation efficiency, and parameters for infrared-visible image fusion. The x-axis represents the average run-time, testing by images with size of $448\times 620$. Y-axis denotes the results of MI to reflect the information richness. Area of  circles represents the parameter amounts.}
	\label{fig:firstfig}
\end{figure}

\textbf{Quantitative Comparisons.} 
 We provide two version of fusion schemes for guaranteeing visual quality and fast deployment respectively, which are denoted as TIM$_{\text{w/o L}}$ and TIM$_{\text{w/ L}}$. ``L'' denotes the latency constraint. We utilize four representative reference-based numerical metrics including Mutual Information (MI), Feature Mutual Information based on image gradient (FMI), Visual Information Fidelity (VIF) and edge-based similarity measurement ($\mathrm{Q^{AB/F}}$). MI is a metric derived from information theory, which measures the transformation  from source images, represented by the mutual dependence between diverse distributions. VIF indicates the information fidelity of human visual perception by computing the loss of fidelity combing four scales. Furthermore, the higher FMI also indicates the more feature information (e.g., image edges) fused by source images. $\mathrm{Q^{AB/F}}$ is utilized to measure the textures details by a statistical scheme (i.e., computing  the amount of edge information transformed from source images). 
We report the numerical comparison in Table.~\ref{tab:irvis} and Table.~\ref{tab:irvis_latency}. Obviously, the general TIM achieves the best performance for IVIF. Moreover, the significant improvement of MI and $\mathrm{Q^{AB/F}}$ compared with the newest fusion schemes (i.e., TARDAL and SDNet) indicates that the proposed TIM achieves excellent performance with visual-pleasant, distinct but complementary information and flourish texture details. Furthermore,  TIM$_{\text{w/ L}}$ also obtains comparable performances in both datasets.
On the other hand, we also compare the computation efficiency with these competitive fusion schemes, including parameters, FLOPs and inference time under the second row of tables. We utilize ten pairs with size of $448\times 620$ to conduct these comparisons for TNO.
Meanwhile, ten pairs with size of $560\times 304$ are utilized to conduct these comparisons for RoadScene.
 DenseFuse network has  few parameters and FLOPs but the inference time is limited by $\ell_{1}$-norm fusion rule. Similarly, fusion rule-based schemes (e.g., RFN, MFEIF and SMOA) also obtain sub-optimal inference time compared with end-to-end networks.
 TIM$_{\text{w/ L}}$ achieves the fastest inference time and the lowest FLOPs between both datasets. 
 Compared with the newest fusion scheme TARDAL, TIM$_{\text{w/ L}}$ reduces 57.23\% parameters and 57.03\% of FLOPs on TNO, which can be more easily delloyed on hardware to guarantee real-time inference.
 The comprehensive analysis between visual quality and inference time is shown at Fig.~\ref{fig:firstfig}.
\begin{figure}[htb]
	\centering \begin{tabular}{c@{\extracolsep{0.15em}}c@{\extracolsep{0.15em}}c@{\extracolsep{0.15em}}c@{\extracolsep{0.15em}}c@{\extracolsep{0.15em}}c@{\extracolsep{0.15em}}c}
		\includegraphics[width=0.075\textwidth]{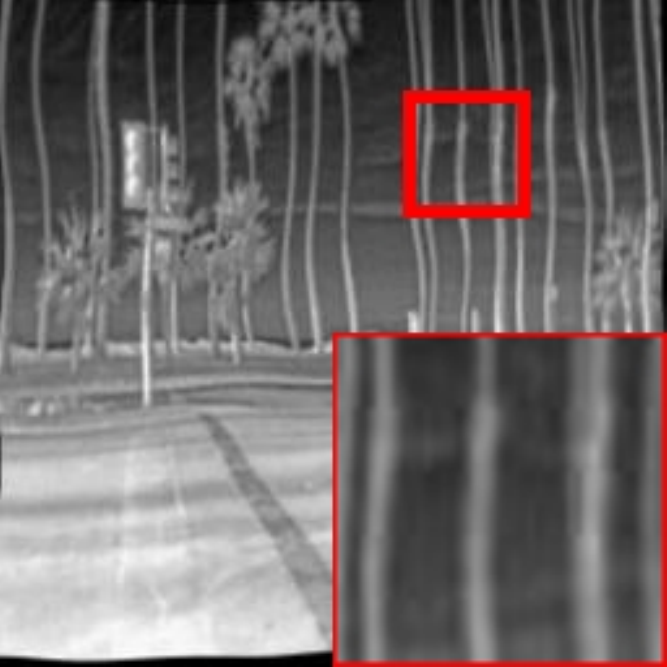}
		&\includegraphics[width=0.075\textwidth]{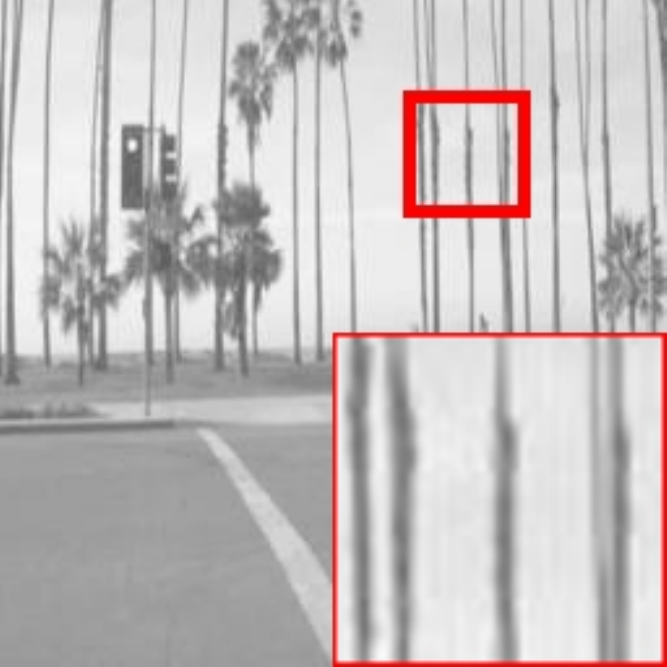}
		&\includegraphics[width=0.075\textwidth]{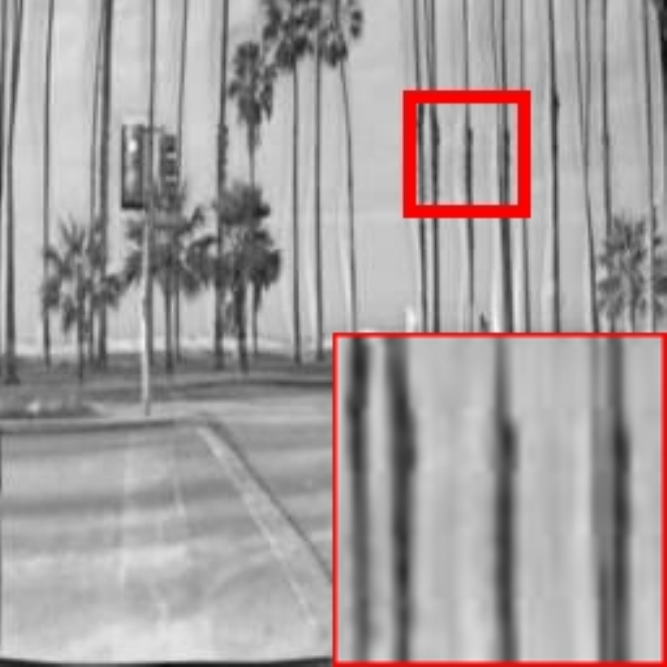}
		&\includegraphics[width=0.075\textwidth]{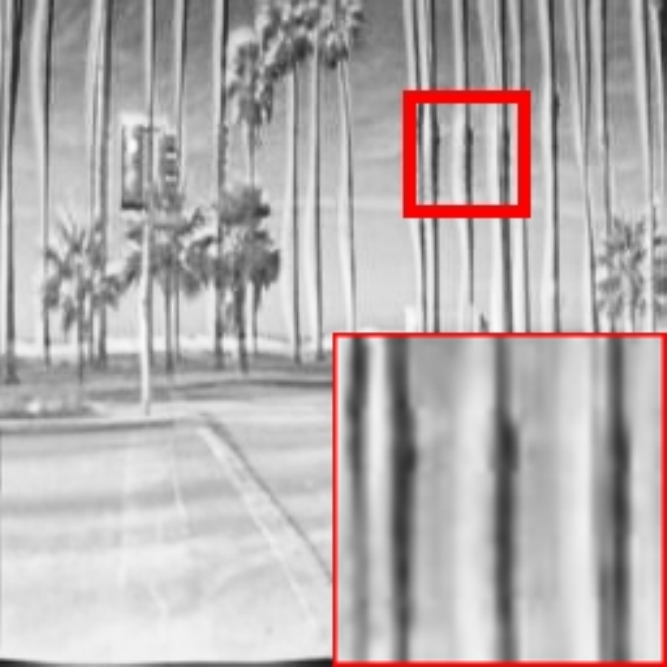}
		&\includegraphics[width=0.075\textwidth]{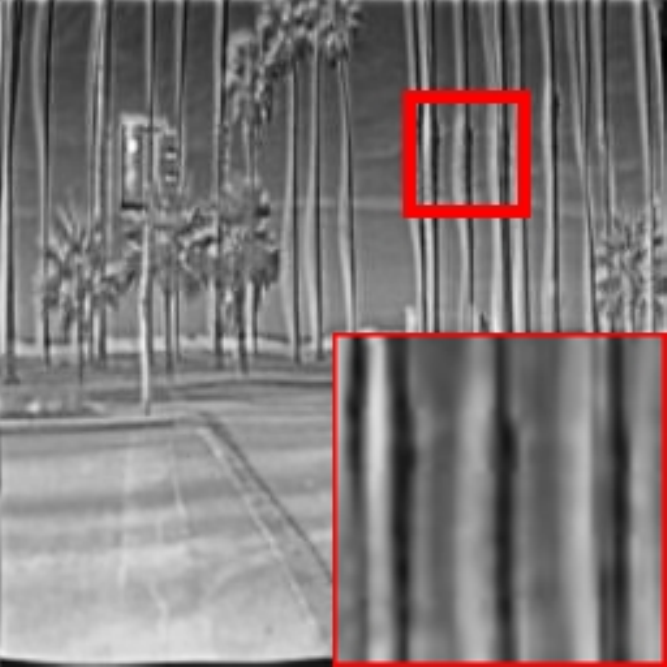}
		&\includegraphics[width=0.075\textwidth]{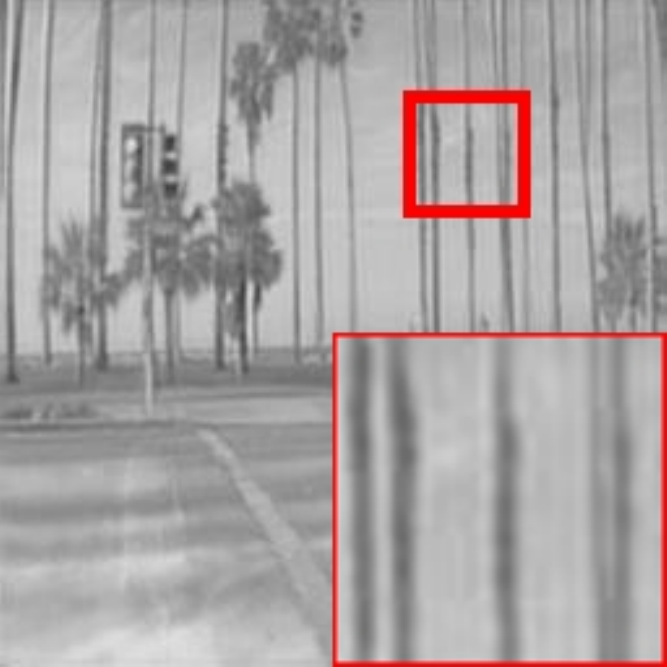}\\
		\includegraphics[width=0.075\textwidth]{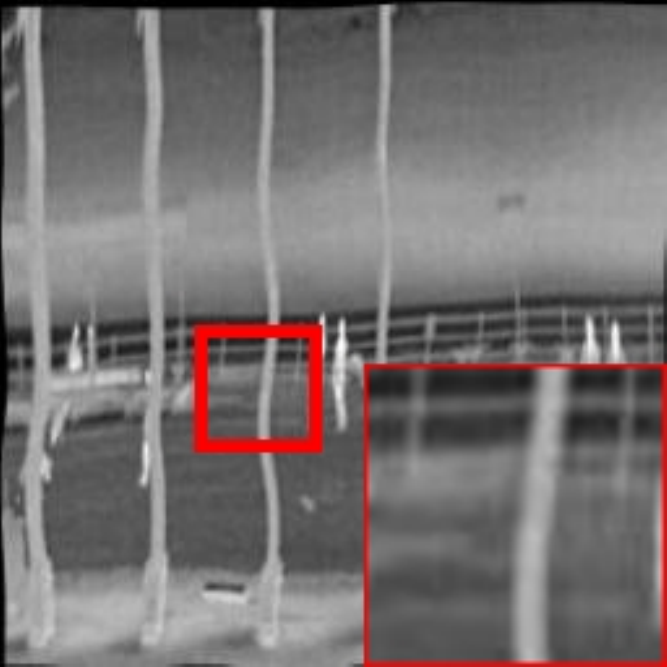}
		&\includegraphics[width=0.075\textwidth]{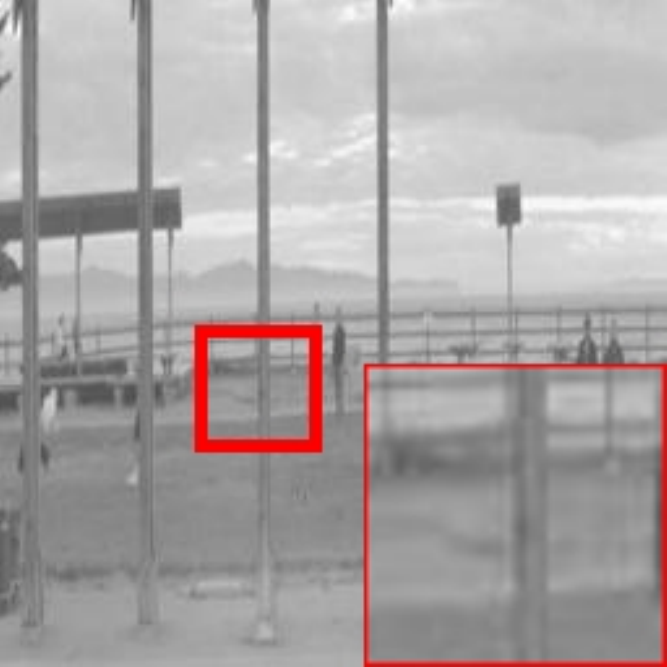}
	    &\includegraphics[width=0.075\textwidth]{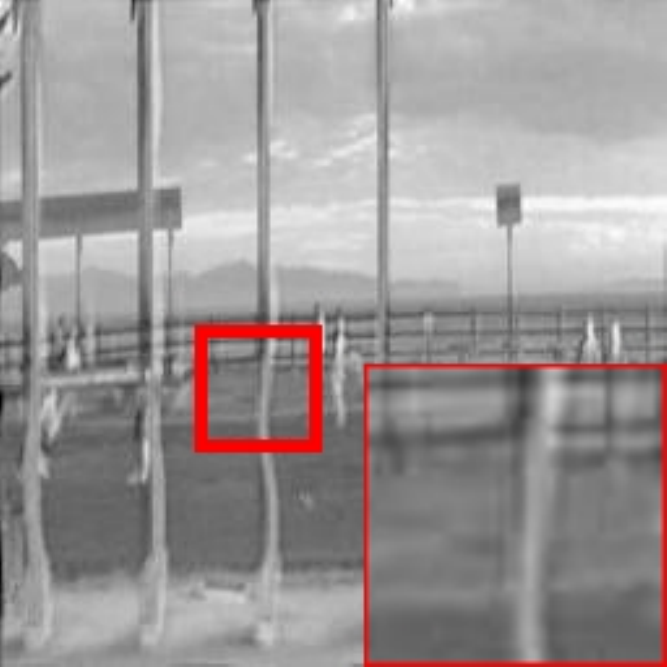}
		&\includegraphics[width=0.075\textwidth]{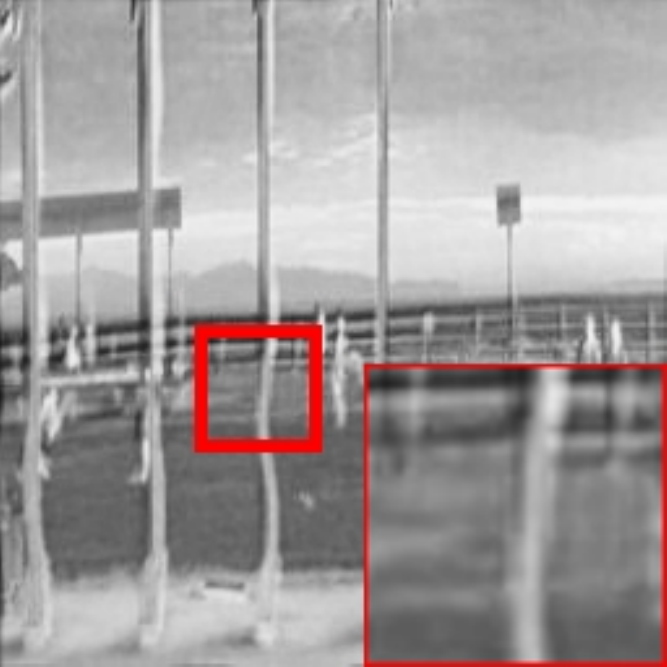}
		&\includegraphics[width=0.075\textwidth]{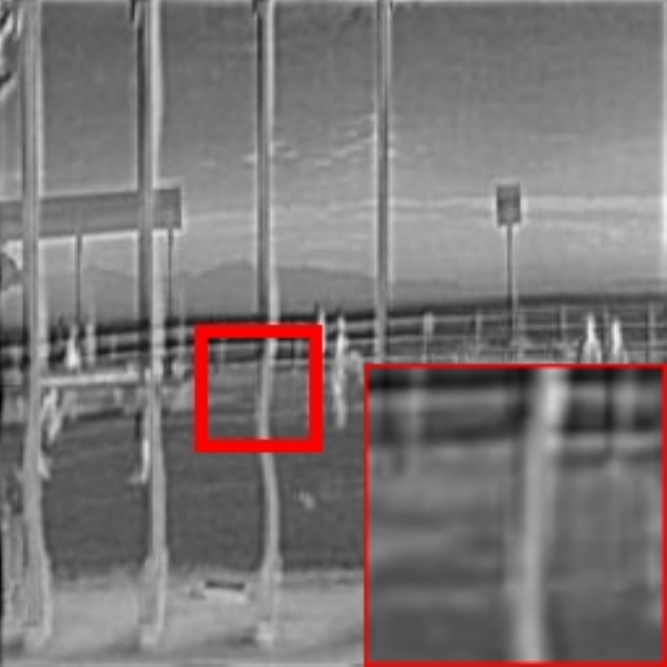}
		&\includegraphics[width=0.075\textwidth]{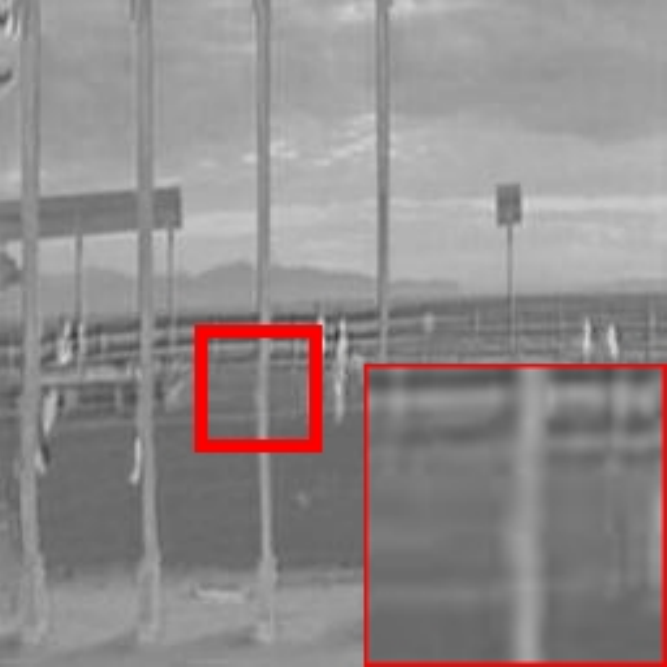}\\
		\includegraphics[width=0.075\textwidth]{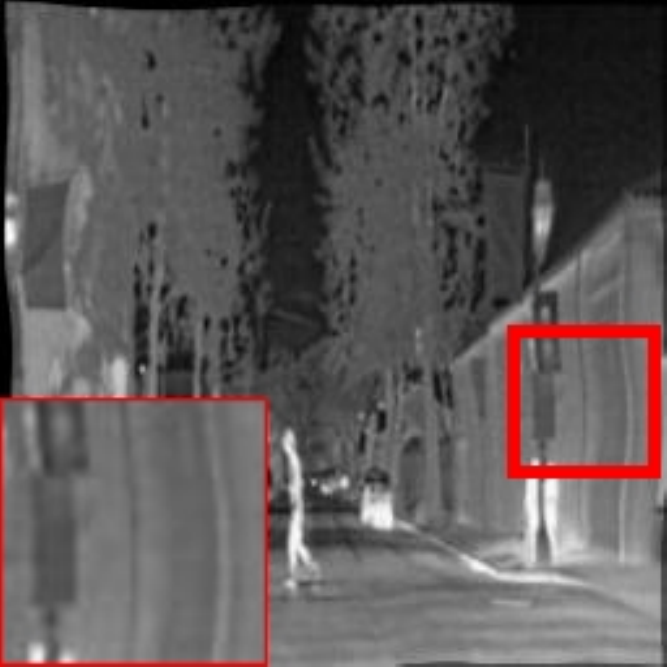}
		&\includegraphics[width=0.075\textwidth]{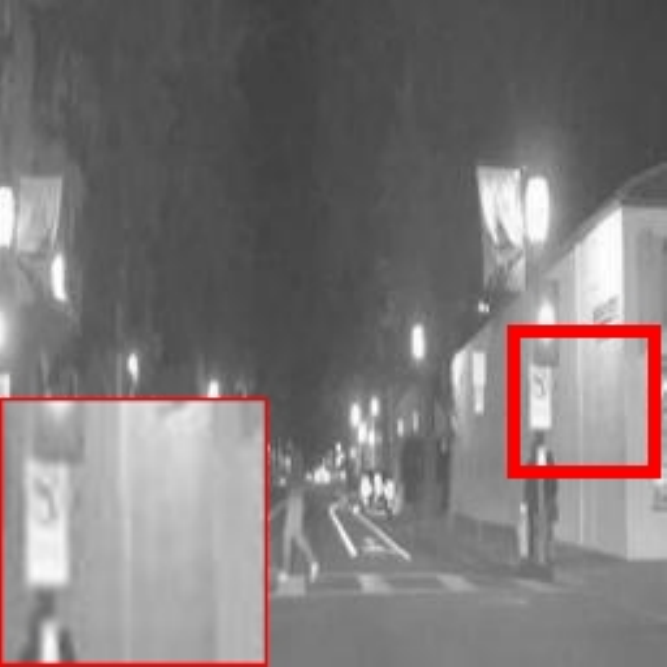}
		&\includegraphics[width=0.075\textwidth]{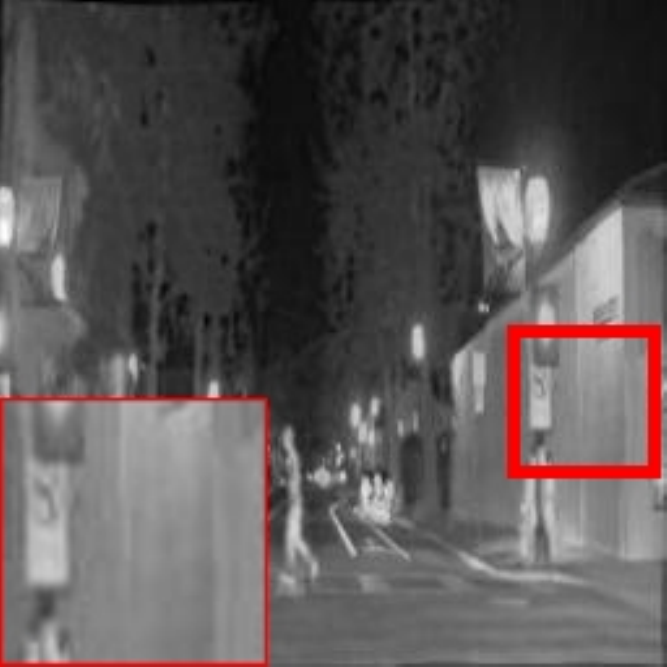}
		&\includegraphics[width=0.075\textwidth]{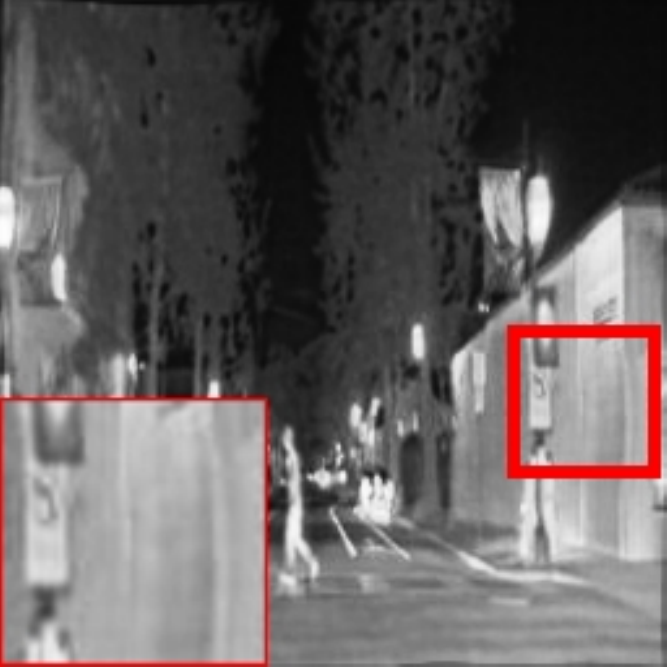}
		&\includegraphics[width=0.075\textwidth]{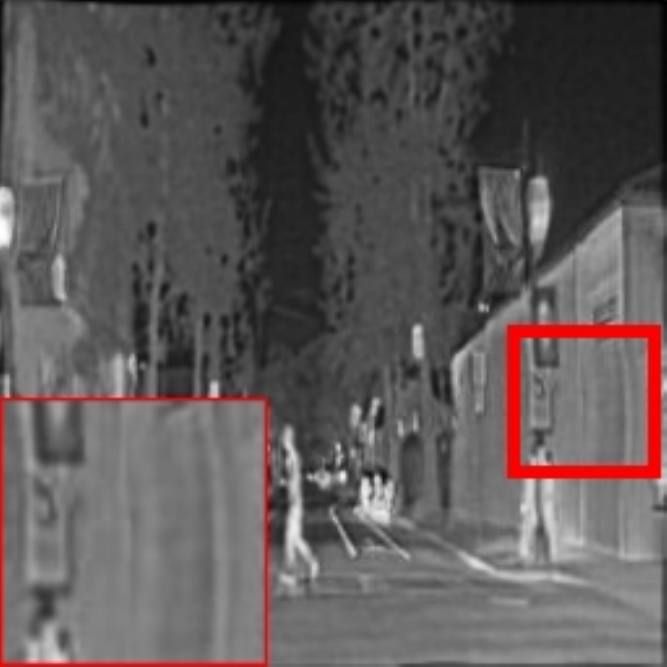}
		&\includegraphics[width=0.075\textwidth]{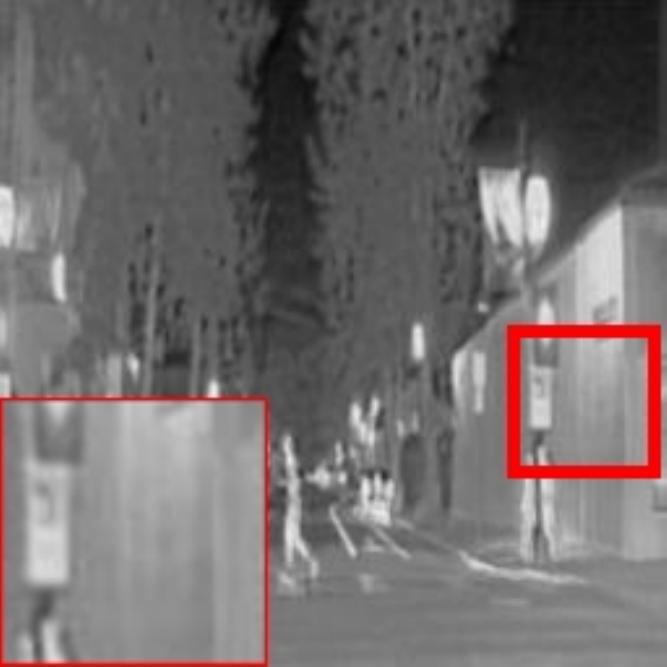}\\
		\footnotesize	Infrared &\footnotesize  Visible   &\footnotesize  DenseFuse & \footnotesize AUIF &  \footnotesize SDNet & \footnotesize TIM\\
	\end{tabular}
	\caption{Qualitative comparison of joint image registration and fusion.}
	\label{fig:result_reg_fusion}
\end{figure}
\begin{figure*}[!htb]
	\centering \begin{tabular}{c@{\extracolsep{0.4em}}c@{\extracolsep{0.4em}}c}

		\includegraphics[width=0.32\textwidth]{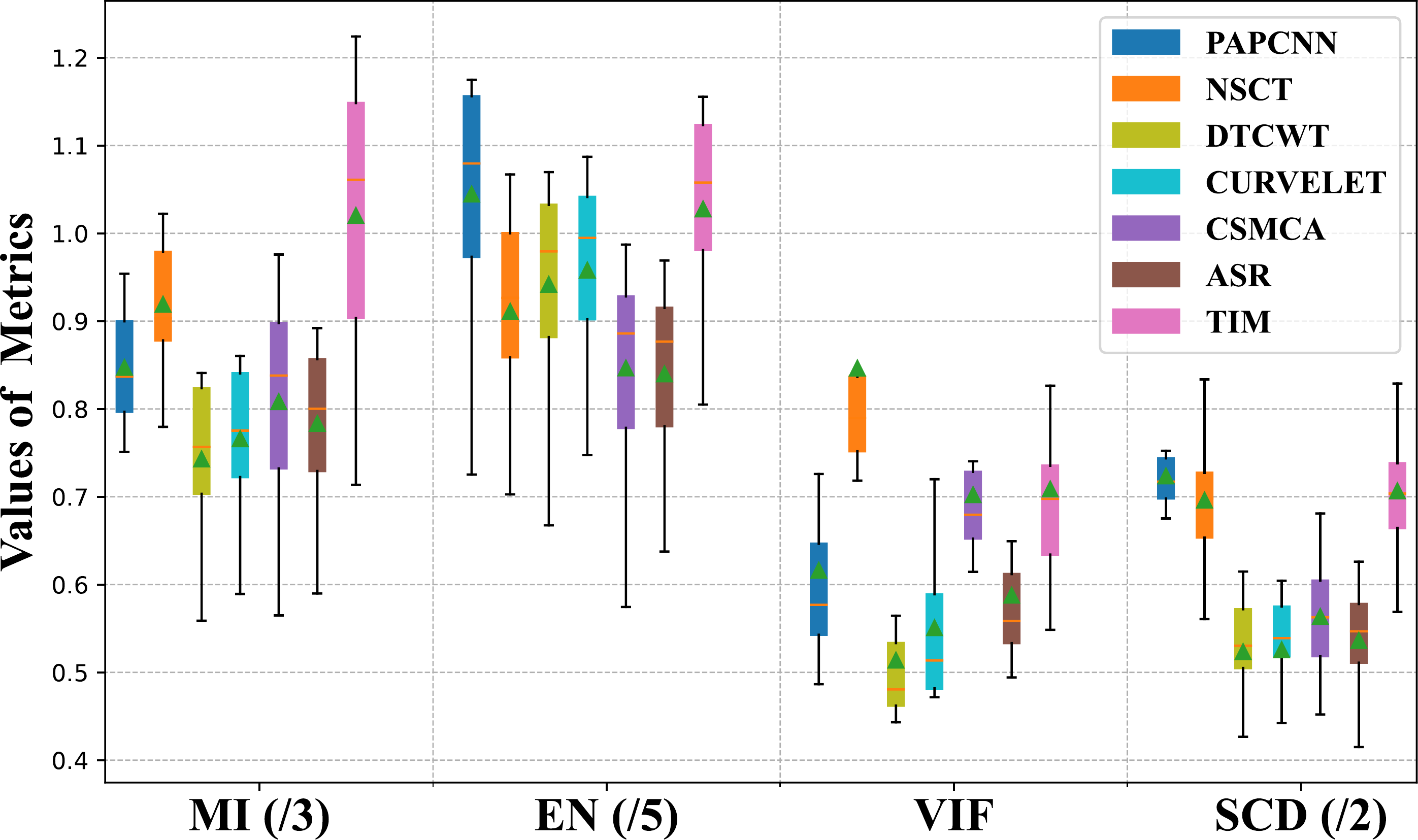}
		&\includegraphics[width=0.32\textwidth]{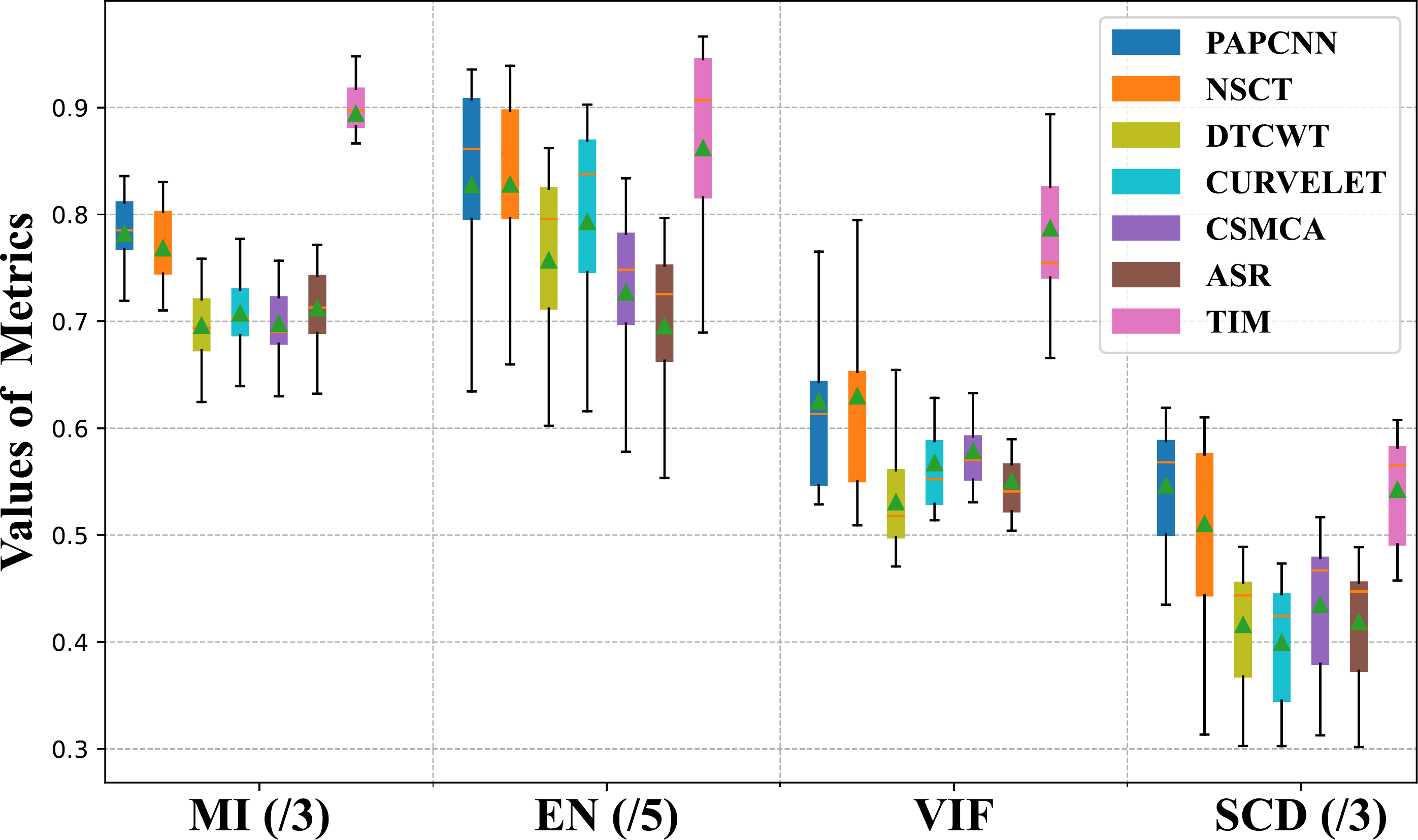}
		&\includegraphics[width=0.32\textwidth]{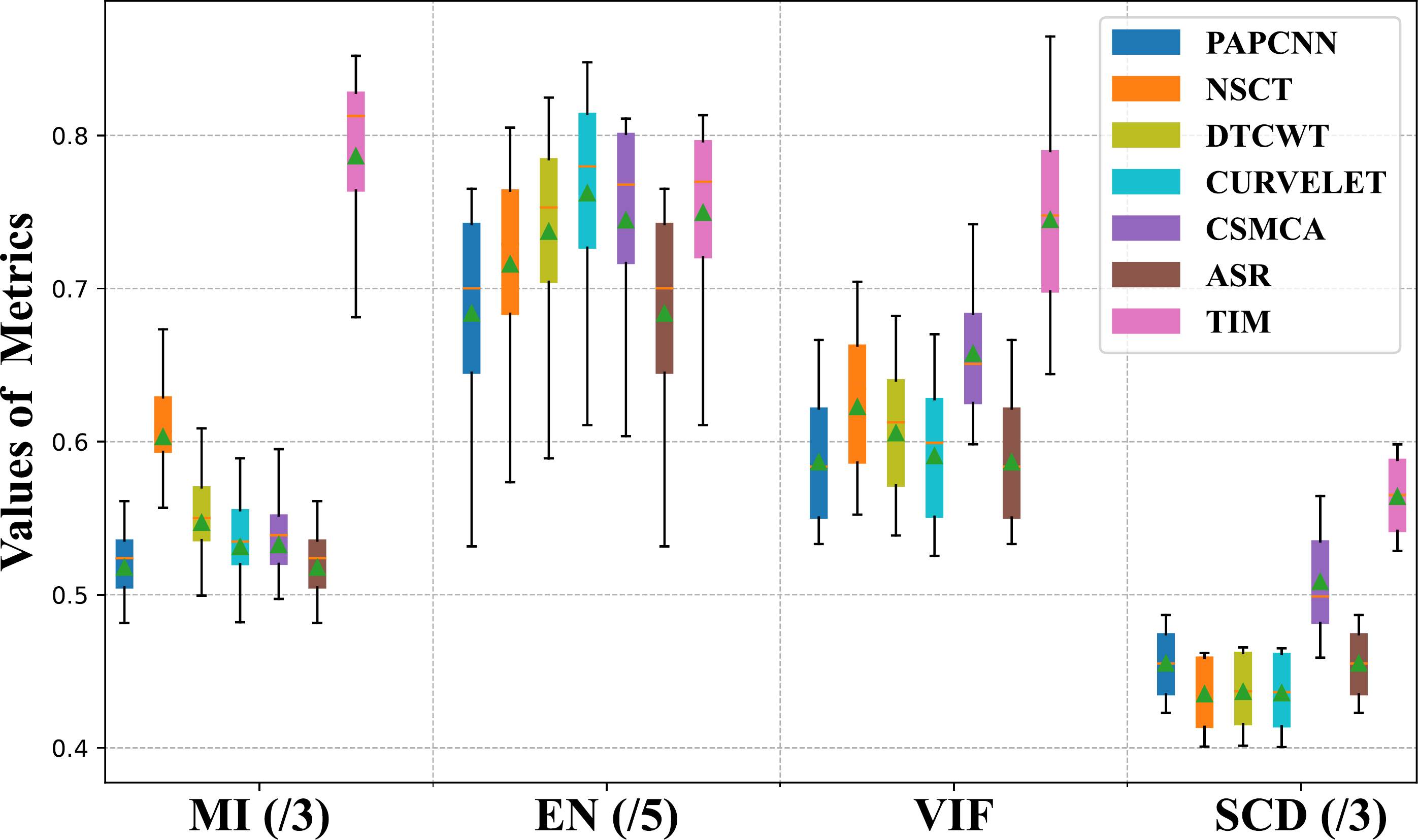}\\
		
		\footnotesize (a) MRI-CT  &\footnotesize  (b) MRI-PET  & \footnotesize (c) MRI-SPECT 
	\end{tabular}
	\caption{Quantitative comparison results of the proposed method with six competitive methods on three medical image fusion tasks, i.e., MRI-CT, MRI-PET, MRI-SPECT fusion. X-axis denotes the fusion metrics. Y-axis represents the value of metric. The green triangle and orange line are the mean and medium value.}
	\label{fig:result_medical}
\end{figure*}

\begin{figure*}[!htb]
	\centering \begin{tabular}{c@{\extracolsep{0.1em}}c@{\extracolsep{0.1em}}c@{\extracolsep{0.1em}}c@{\extracolsep{0.1em}}c@{\extracolsep{0.1em}}c@{\extracolsep{0.1em}}c@{\extracolsep{0.1em}}c@{\extracolsep{0.1em}}c}

		\includegraphics[width=0.1073\textwidth]{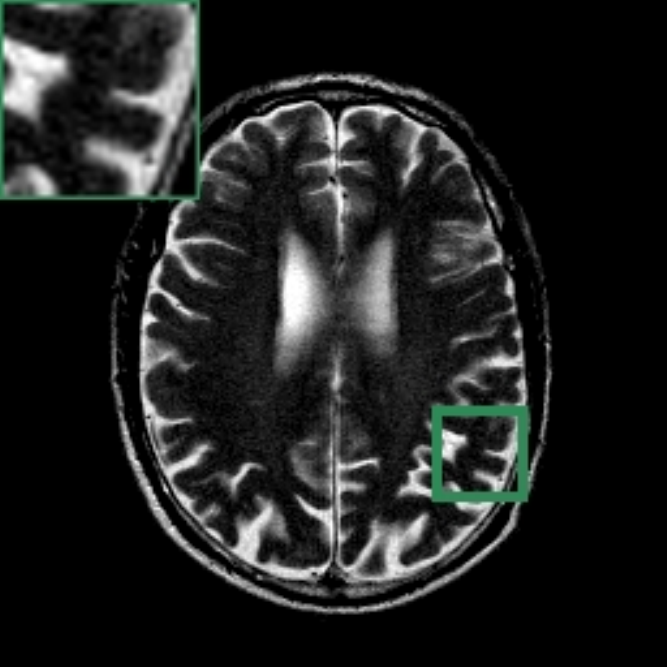}
		&\includegraphics[width=0.1073\textwidth]{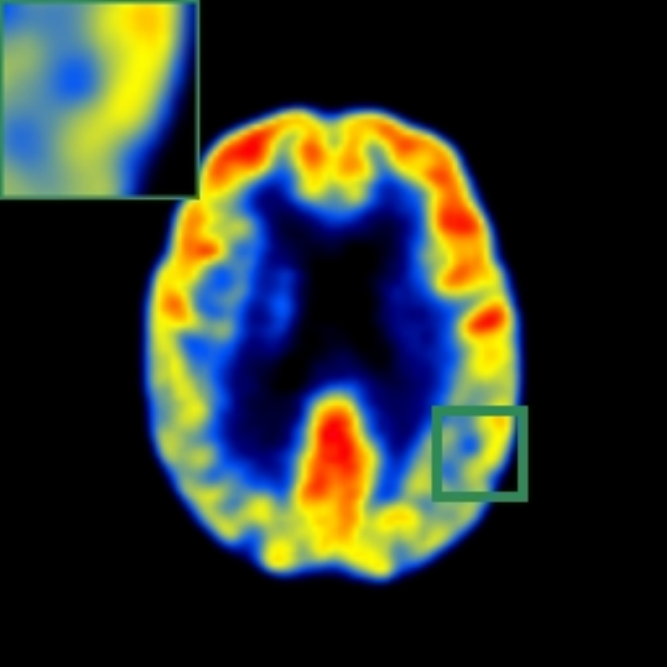}
		&\includegraphics[width=0.1073\textwidth]{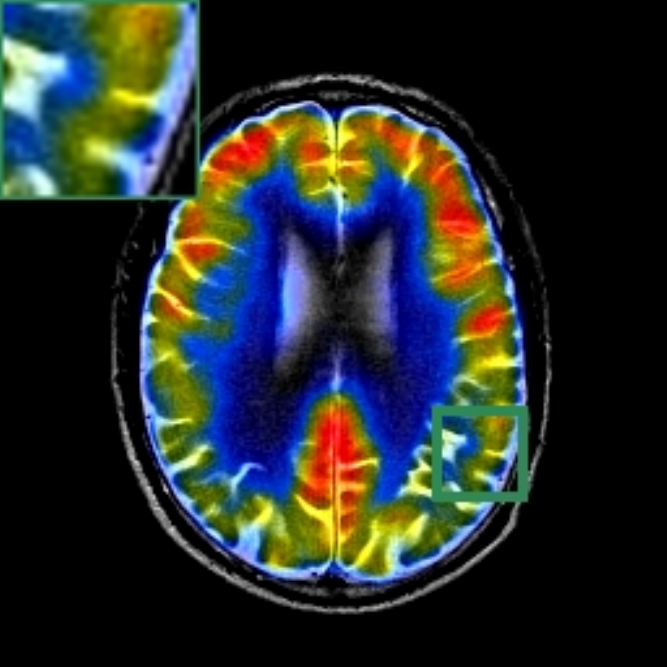}
		&\includegraphics[width=0.1073\textwidth]{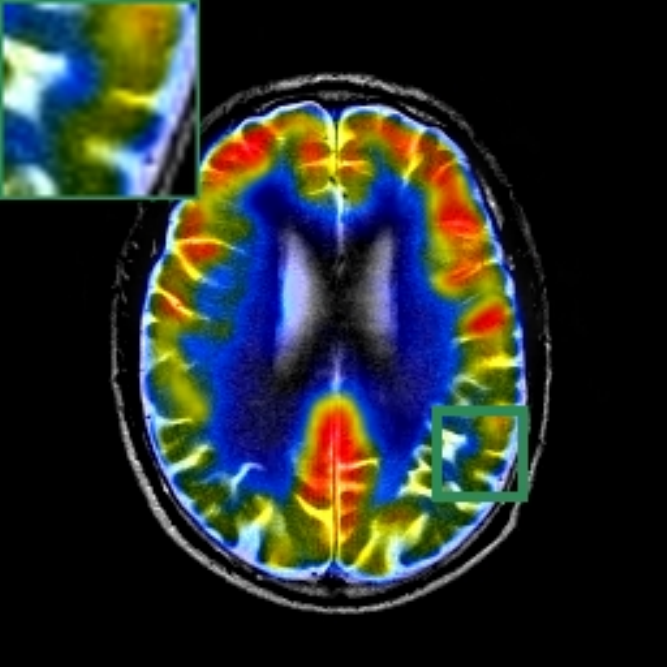}
		&\includegraphics[width=0.1073\textwidth]{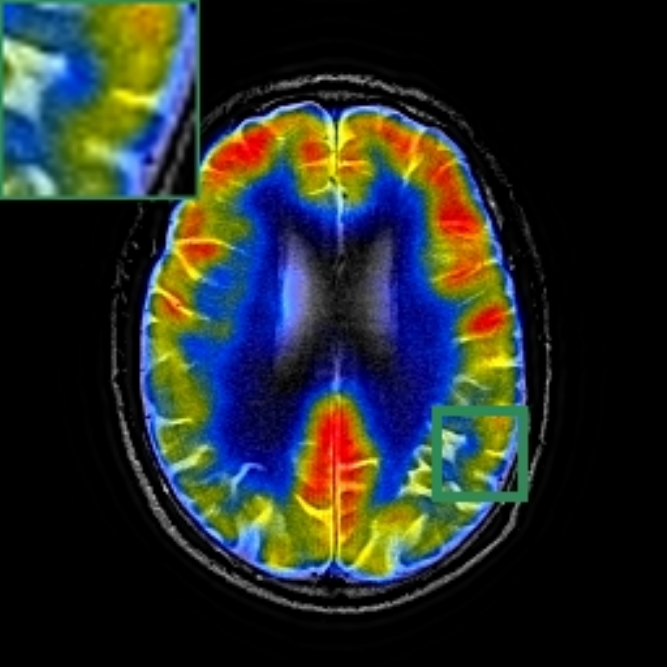}
		&\includegraphics[width=0.1073\textwidth]{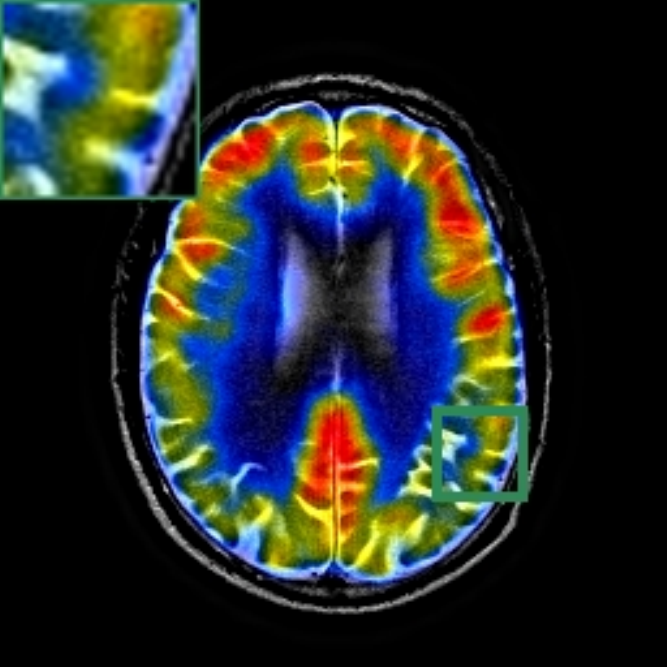}
		&\includegraphics[width=0.1073\textwidth]{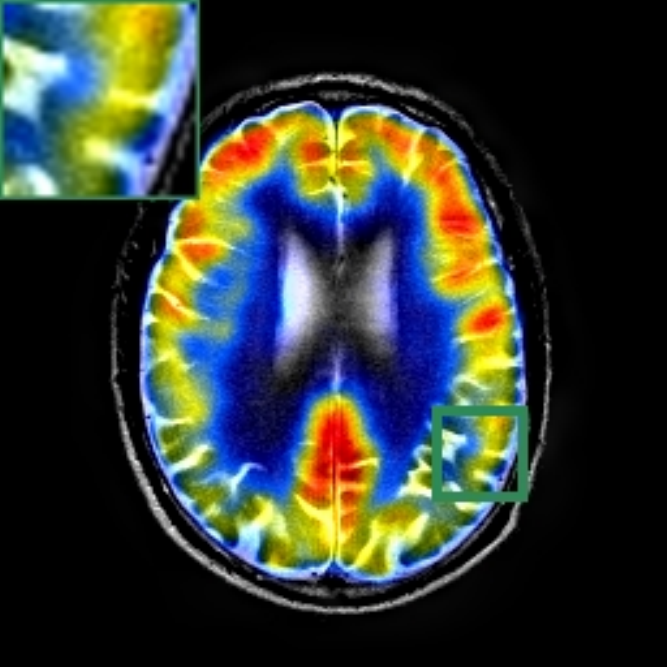}
		&\includegraphics[width=0.1073\textwidth]{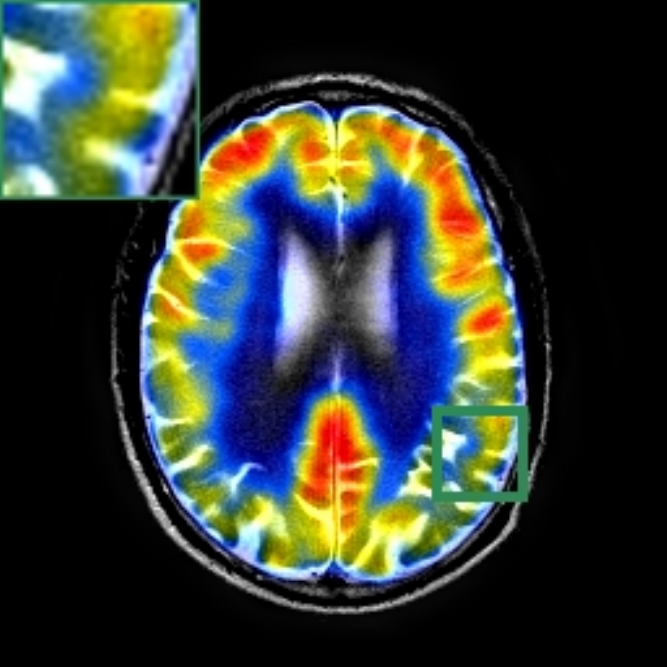}
		&\includegraphics[width=0.1073\textwidth]{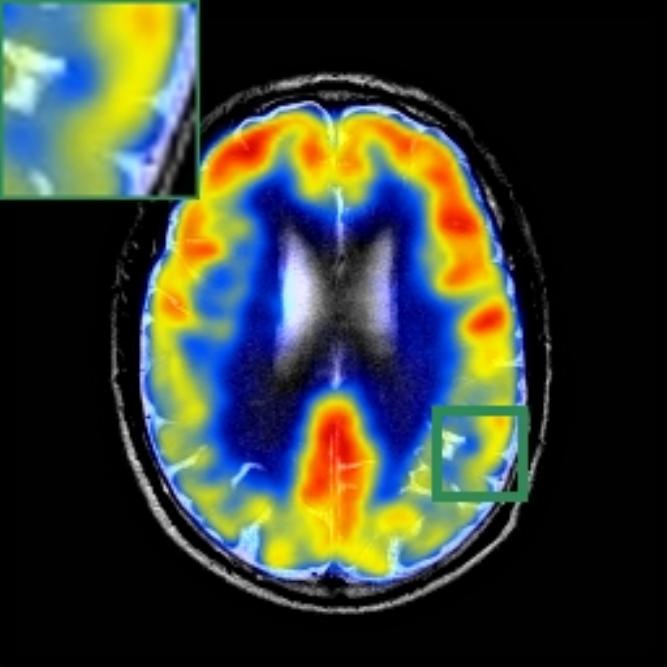}\\
		\footnotesize	MI/SCD & \footnotesize -/-&\footnotesize 2.122/1.364& \footnotesize 2.043/1.481& \footnotesize 2.066/1.301&\footnotesize 2.056/1.311& \footnotesize 2.225/1.312&\footnotesize 2.303/1.364&\footnotesize \textbf{2.660/1.680}\\

		\includegraphics[width=0.1073\textwidth]{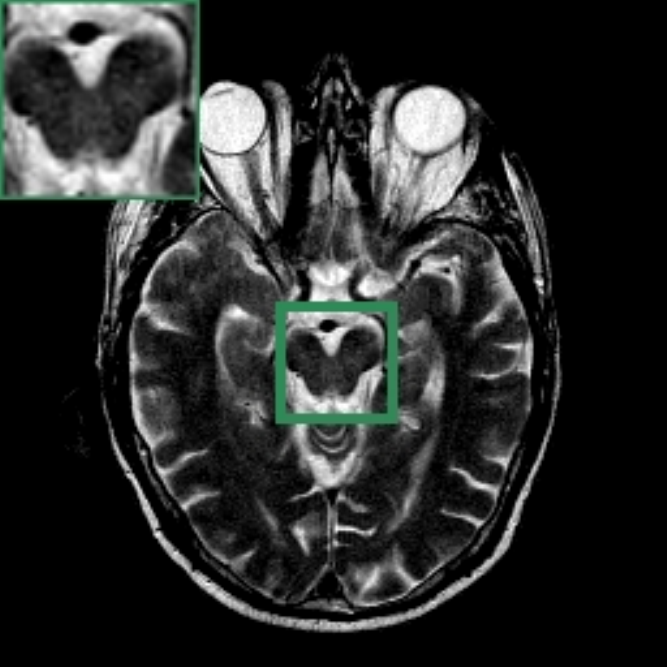}
		&\includegraphics[width=0.1073\textwidth]{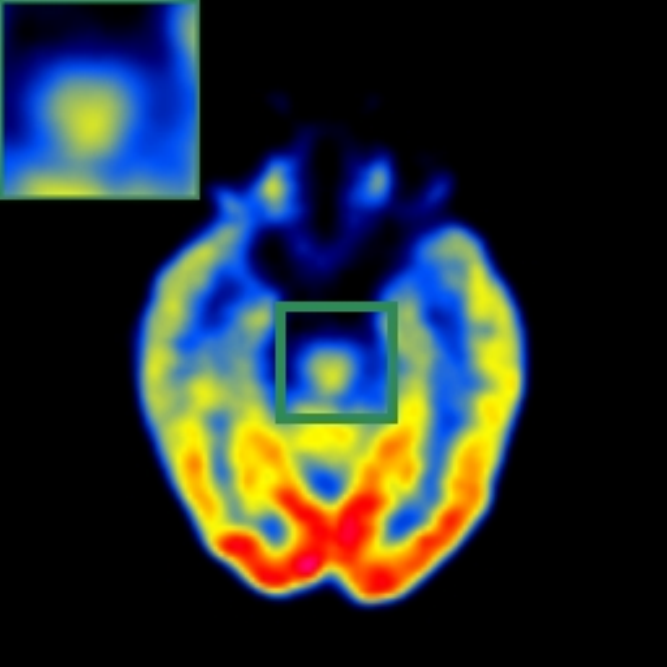}
		&\includegraphics[width=0.1073\textwidth]{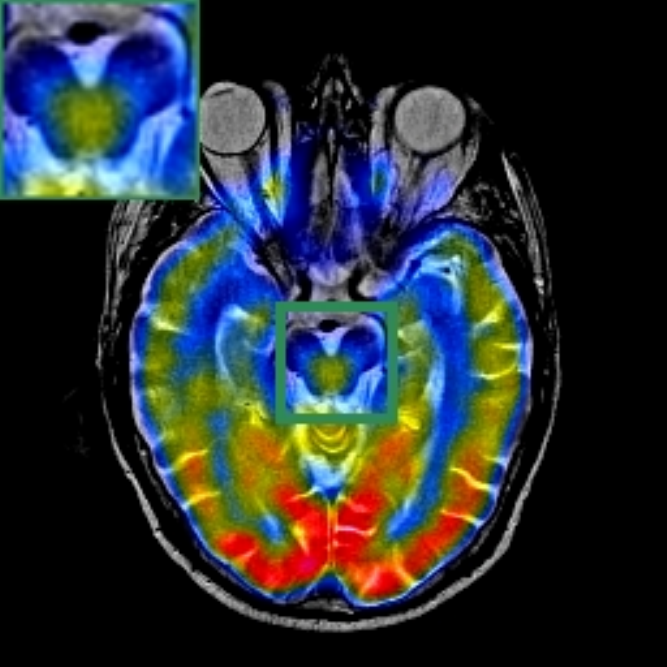}
		&\includegraphics[width=0.1073\textwidth]{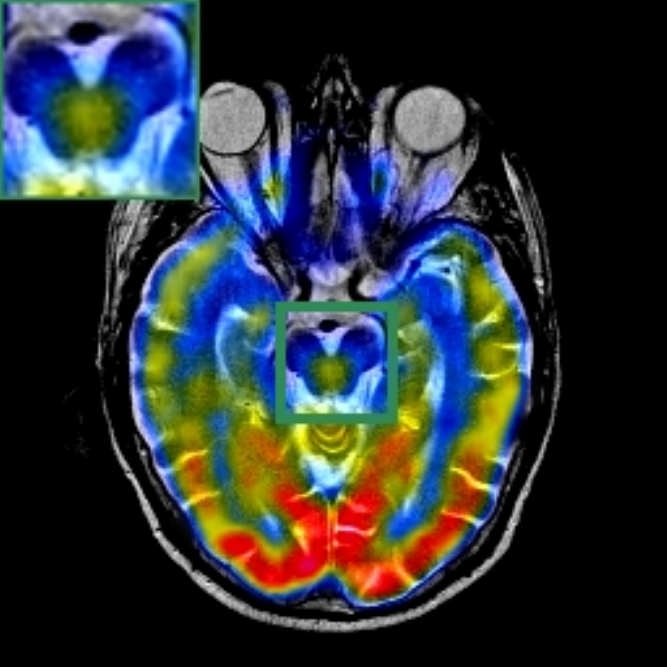}
		&\includegraphics[width=0.1073\textwidth]{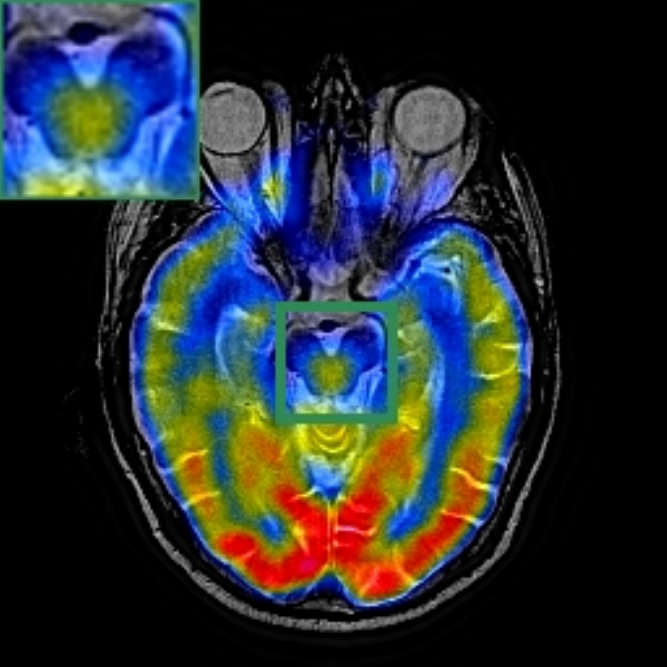}
		&\includegraphics[width=0.1073\textwidth]{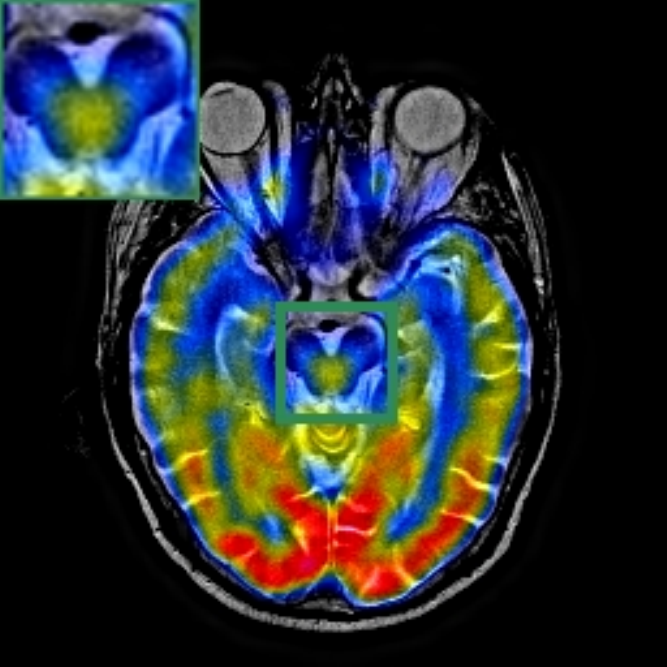}
		&\includegraphics[width=0.1073\textwidth]{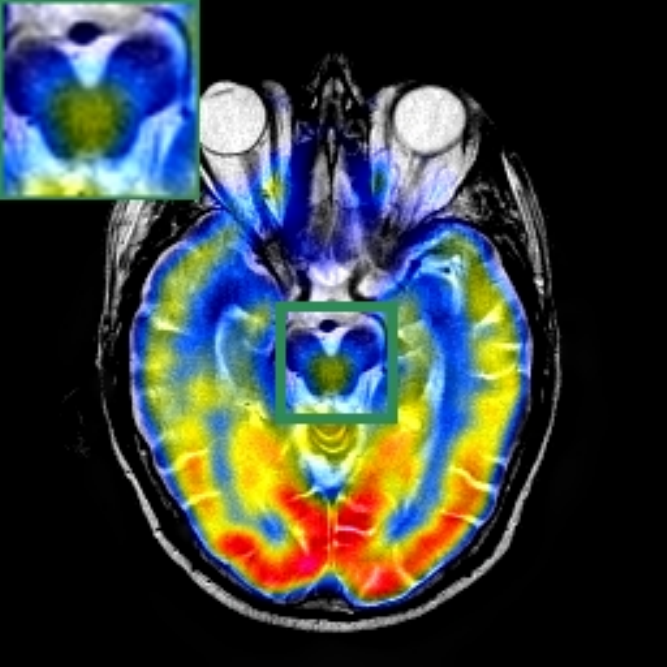}
		&\includegraphics[width=0.1073\textwidth]{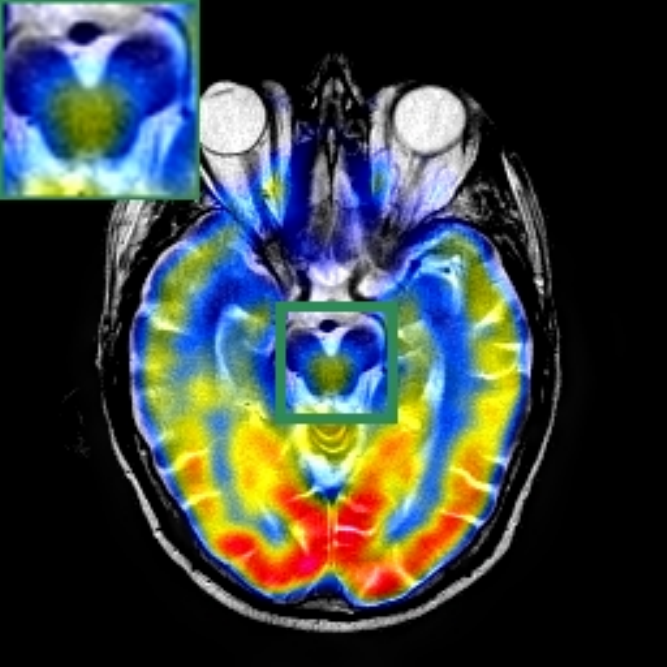}
		&\includegraphics[width=0.1073\textwidth]{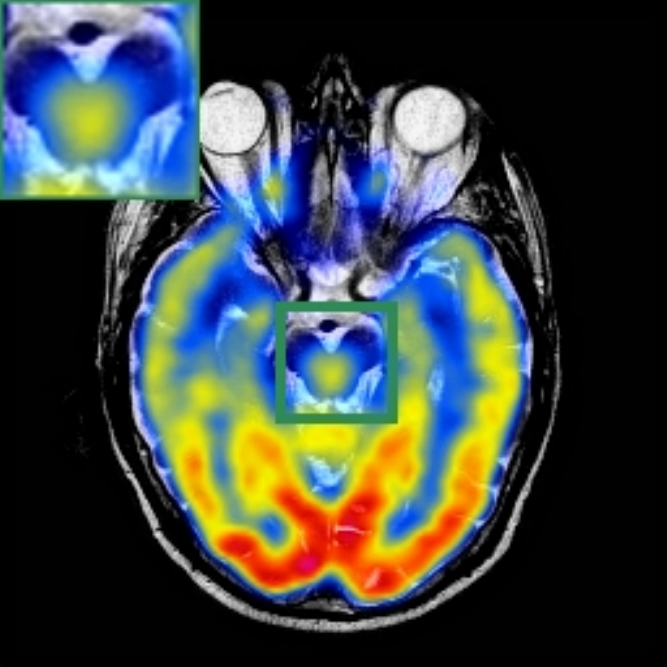}\\
		\footnotesize	MI/SCD & \footnotesize -/-&\footnotesize 2.058/1.385& \footnotesize 2.037/1.541& \footnotesize 2.049/0.326& \footnotesize 2.008/1.330& \footnotesize 2.407/1.328& \footnotesize 2.427/1.385&\footnotesize \textbf{2.739/1.712}\\
		\footnotesize	MRI& \footnotesize PET & \footnotesize ASR  & \footnotesize  CSMCA  &\footnotesize CURVELET & \footnotesize DTCWT & \footnotesize NSCT & \footnotesize PAPCNN & \footnotesize TIM \\

		\includegraphics[width=0.1073\textwidth]{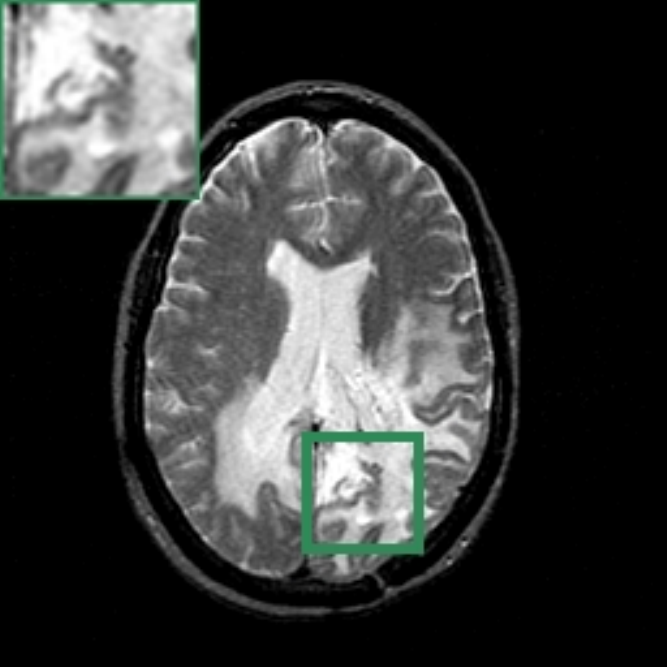}
		&\includegraphics[width=0.1073\textwidth]{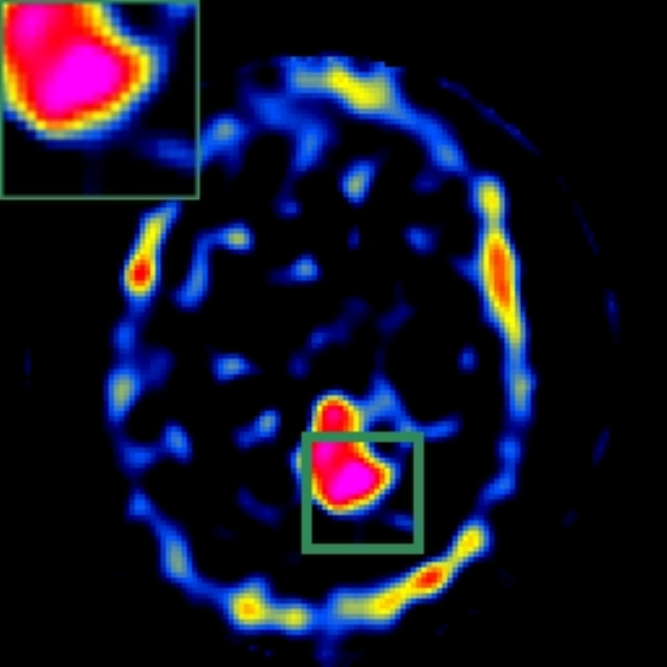}
		&\includegraphics[width=0.1073\textwidth]{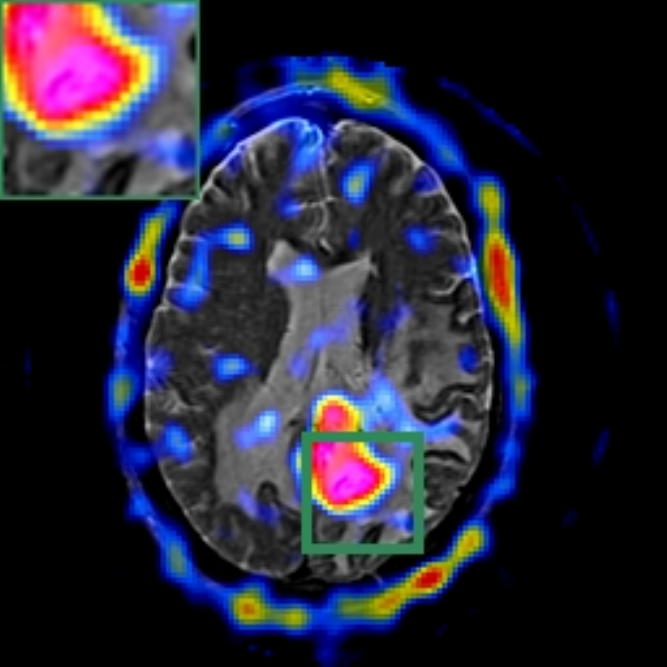}
		&\includegraphics[width=0.1073\textwidth]{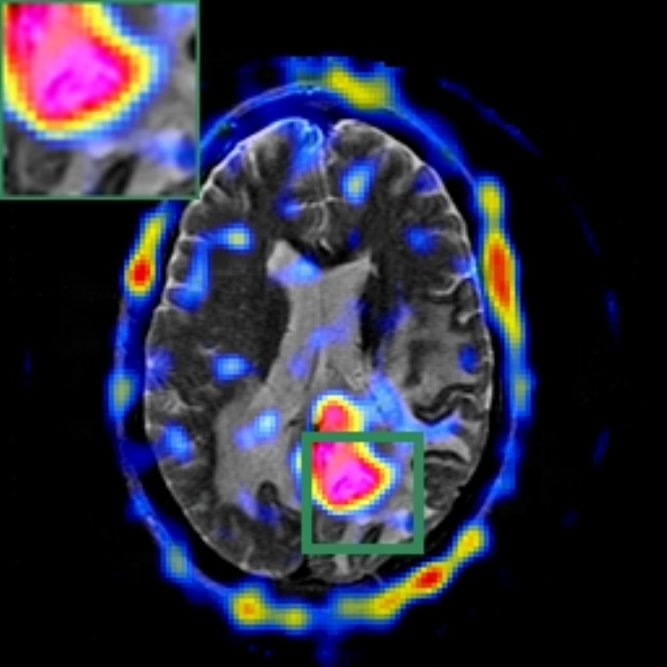}
		&\includegraphics[width=0.1073\textwidth]{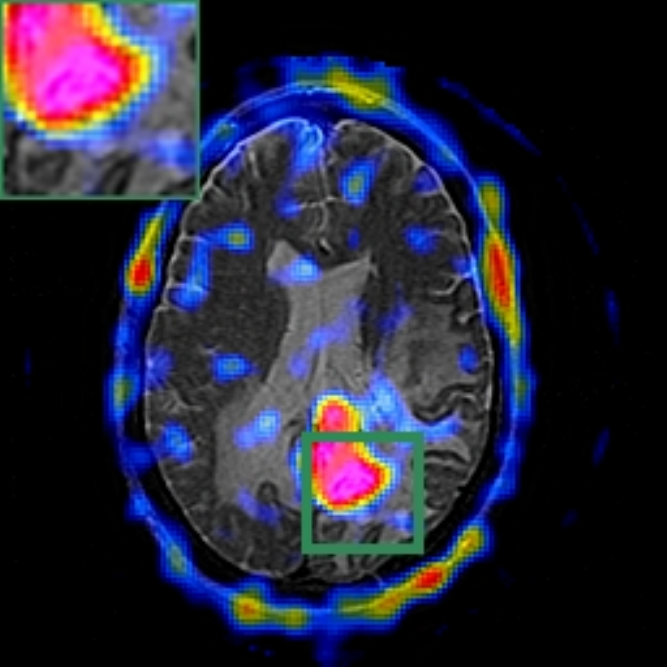}
		&\includegraphics[width=0.1073\textwidth]{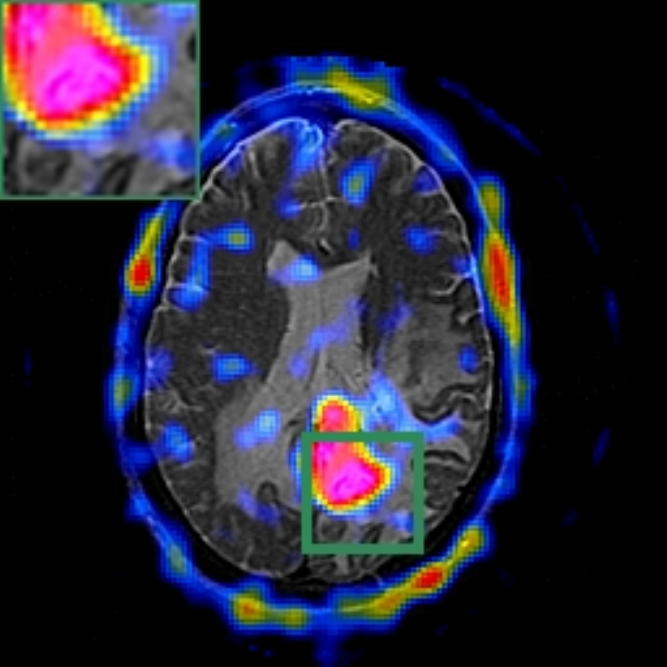}
		&\includegraphics[width=0.1073\textwidth]{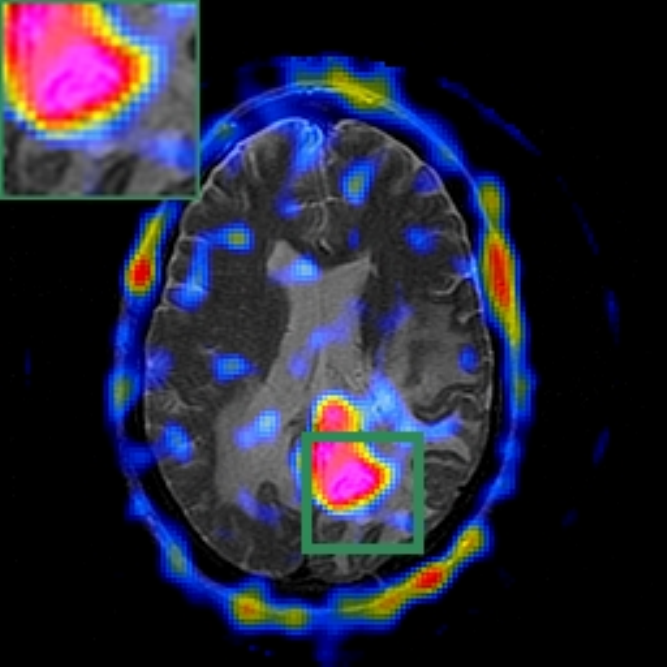}
		&\includegraphics[width=0.1073\textwidth]{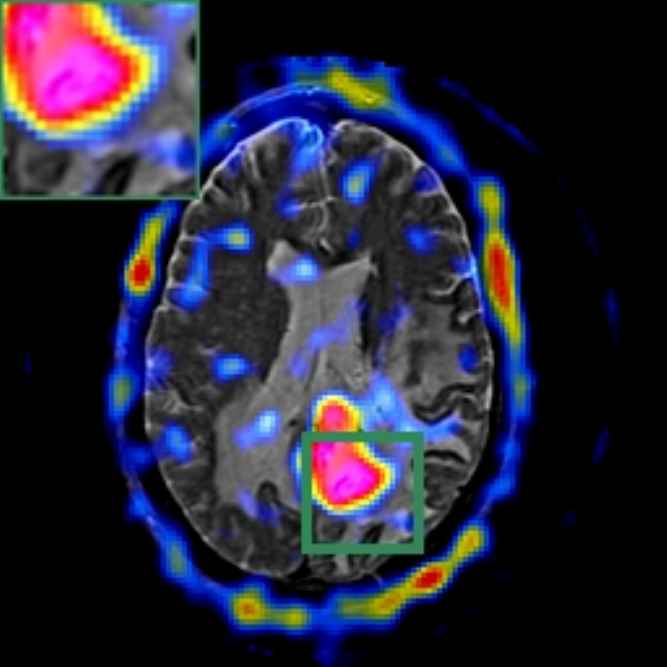}
		&\includegraphics[width=0.1073\textwidth]{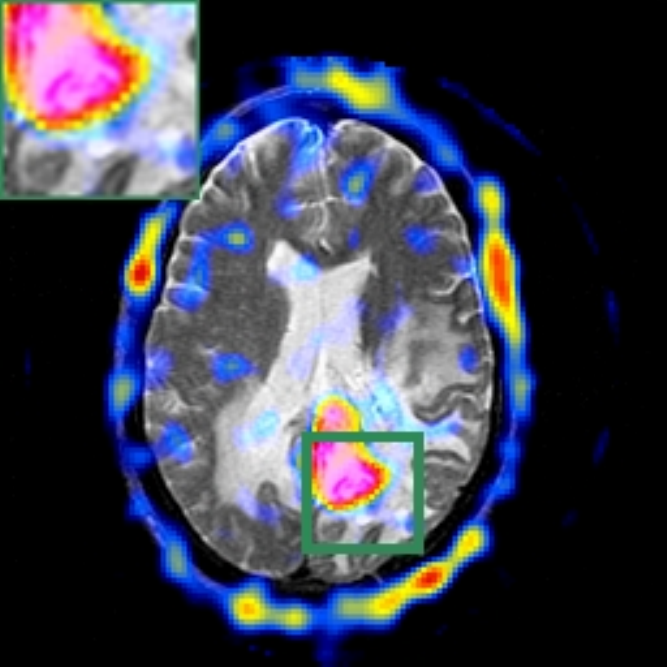}\\
		\footnotesize	MI/SCD & \footnotesize -/-& \footnotesize 1.598/1.364& \footnotesize 1.613/1.481& \footnotesize 1.685/1.301& \footnotesize 1.723/1.311&1.915/1.312&1.598/1.364&\textbf{2.481/1.680}\\

		\includegraphics[width=0.107\textwidth]{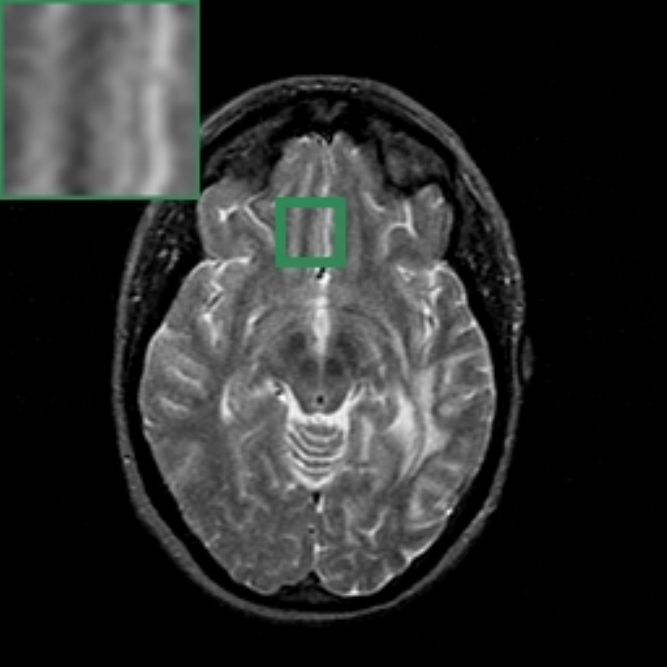}
		&\includegraphics[width=0.107\textwidth]{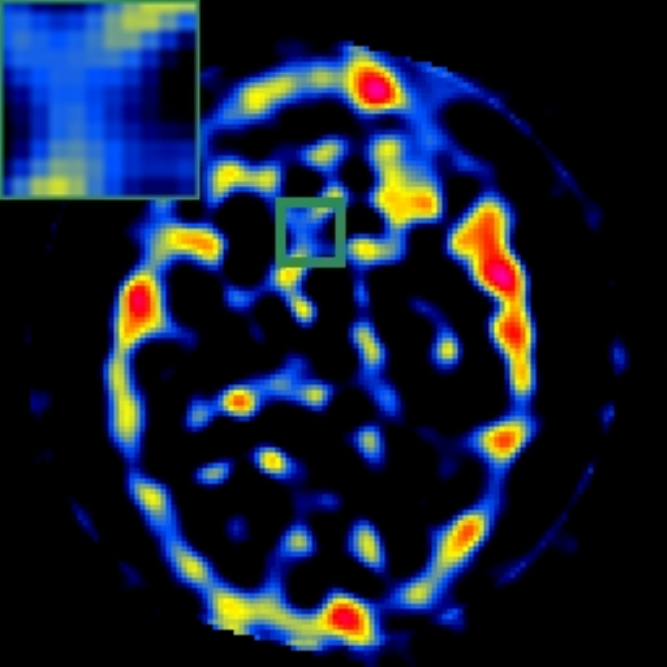}
		&\includegraphics[width=0.107\textwidth]{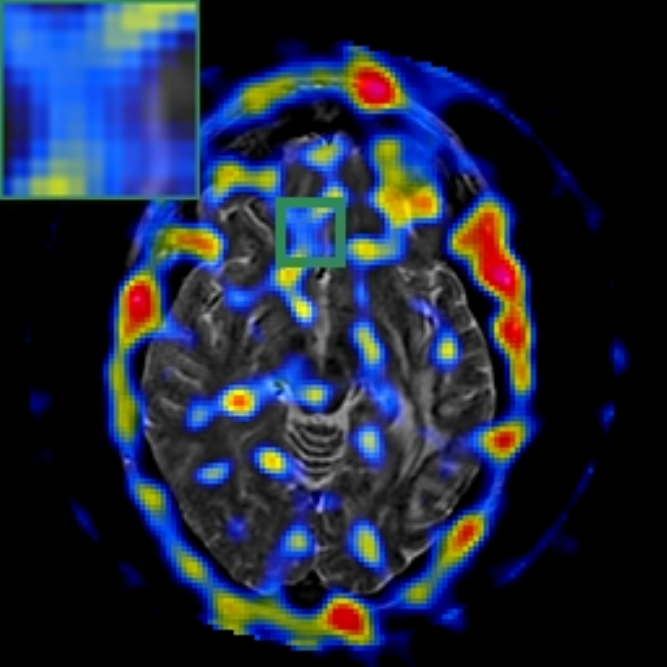}
		&\includegraphics[width=0.107\textwidth]{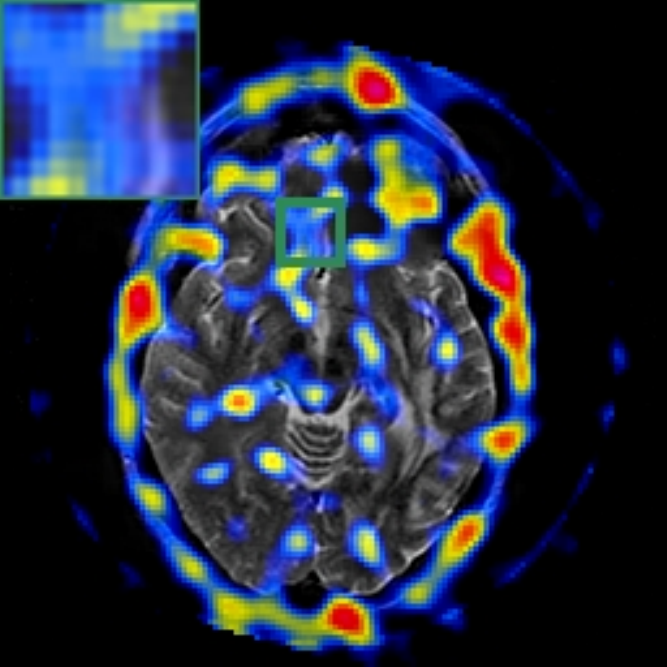}
		&\includegraphics[width=0.107\textwidth]{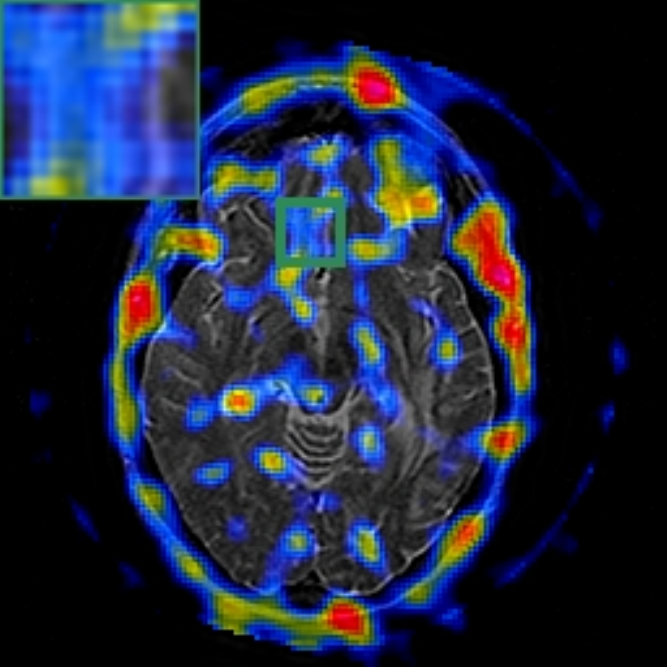}
		&\includegraphics[width=0.107\textwidth]{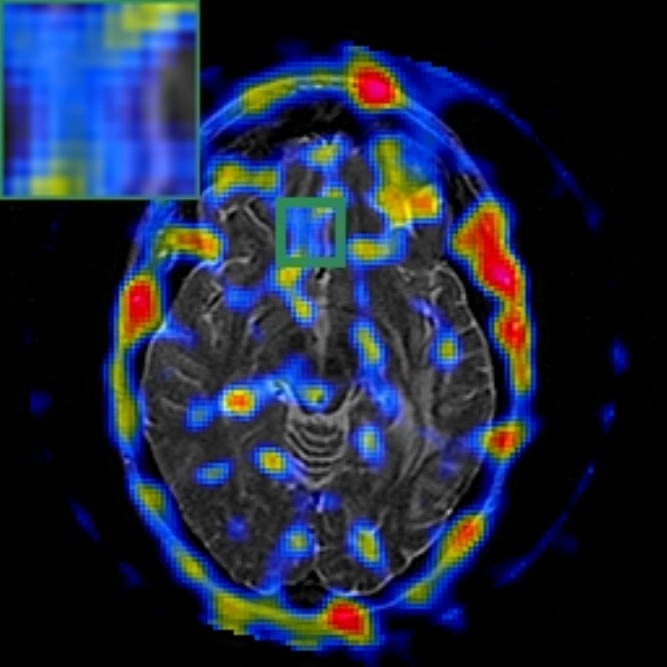}
		&\includegraphics[width=0.107\textwidth]{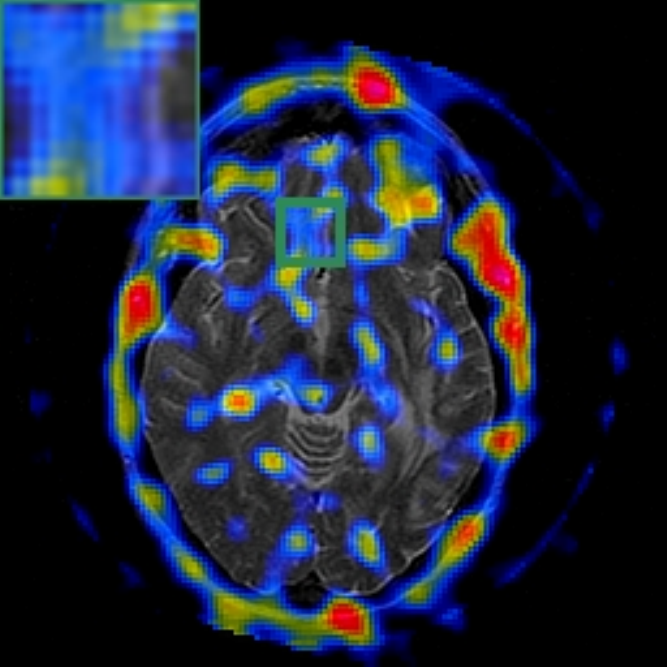}
		&\includegraphics[width=0.107\textwidth]{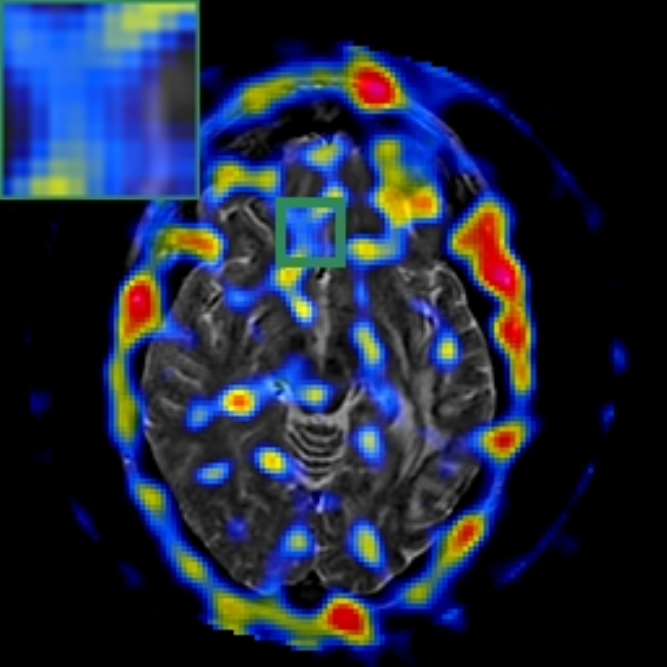}
		&\includegraphics[width=0.107\textwidth]{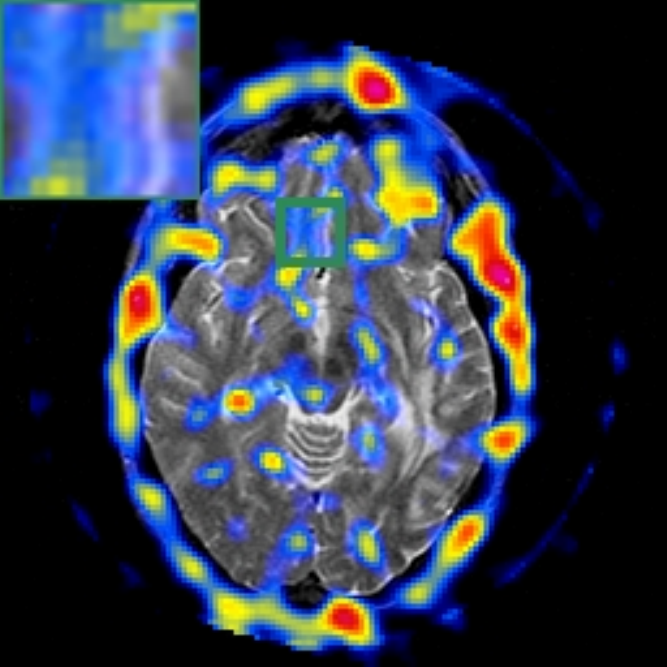}\\
		\footnotesize	MI/SCD & \footnotesize-/-&\footnotesize 1.617/1.446&\footnotesize 1.657/1.657&\footnotesize 1.617/1.382&\footnotesize 1.667/1.384&\footnotesize 1.823/1.375&\footnotesize 1.617/1.446&\footnotesize \textbf{2.491/1.772}\\
		\footnotesize	MRI&\footnotesize SPECT &\footnotesize ASR  &\footnotesize CSMCA  & \footnotesize CURVELET &\footnotesize DTCWT &\footnotesize NSCT &\footnotesize PAPCNN &\footnotesize TIM \\
	\end{tabular}
	\caption{Qualitative comparison on four groups of MRI-PET/SPECT images with various medical fusion algorithms.}
	\label{fig:result_mr-spect}
\end{figure*}

\subsubsection{Image Fusion with Registration}
In real world, due to the diverse imaging pipelines and  complex environments (e.g., temperature changes and mechanical stress), obtaining highly-accurate aligned multi-spectral images are challenging. The misalignment
of source images is easy to generate fusion results with artifacts and ghosts~\cite{reconet}.
Our method can effectively address the misaligned image fusion based on the flexible formulation. Considering the image fusion constraint (Eq.~\eqref{eq:constraint}) to connect vision task, we  introduce another constraint to align the source images $\mathbf{I}_\mathtt{B}$,  $\mathbf{I}_\mathtt{B}$ respectively, which can be written as $\mathbf{I}_\mathtt{B}=\mathcal{N}_\mathtt{R}(\mathbf{I}_\mathtt{A}, {\mathbf{I}_\mathtt{B}^{'}};\bm{\theta}_{\mathtt{R}})$.
We denote the unaligned image as $\mathbf{I}_\mathtt{B}^{'}$ and registration module as $\mathcal{N}_\mathtt{R}$.
 By the effective nested formulation, we can introduce pretrained MRRN scheme~\cite{Wang_2022_IJCAI} as $\mathcal{N}_\mathtt{R}$ to composite more general image fusion.
 For validating the robustness and flexibility of our scheme, we synthesize corrupted
infrared images utilizing random deformation fields by affine and elastic transformations. 
 Then we utilize the initialized parameters to learn a  robuster fusion scheme, which can efficiently address unregistered  multi-spectral image fusion scenarios.
The numerical and visual results are reported in Table.~\ref{tab:registration} and Fig.~\ref{fig:result_reg_fusion} respectively. Other fusion schemes are based on the pairs registered by VoxelMorph~\cite{balakrishnan2018unsupervised}.
Since 
corruption from  distortions in infrared images cannot be recovered exactly, state-of-the-art algorithms such as AUIF and SDNet still contain obvious ghosts, shown as the first row. We can conclude that our method can effectively persevere visible details and sufficient thermal information under the misaligned multi-spectral images.

\subsubsection{Extension to Medical Image Fusion}
Since the flexible formulation, we can extend our method to address other challenging fusion tasks, e.g., medical image fusion.
 Four typical images including MRI, CT, PET and SPECT, provide diverse structural and functional perception of physiological systems. 
  Utilizing {Harvard} dataset, we adopt the aforementioned search schemes and configurations to discover  suitable architectures for three tasks.
 The hierarchical structure (i.e., $\mathcal{N}_\mathtt{T}$) of MRI-CT fusion is composited by 5-RB, 5-RB, 5-RB and SA operations.
The operations for MRI-PET fusion includes 3-SC, 3-RB, 3-RB and 5-RB. Furthermore, 5-RB, 3-DB, 3-RB and SA consists of the architecture of MRI-SPECT fusion. In this part, we conduct visual and numerical comparisons with six  schemes including ASR~\cite{liu2014simultaneous}, CSMCA~\cite{liu2019medical}, CURVELET~\cite{yang2008multimodality}, DTCWT~\cite{cao2014multi}, NSCT~\cite{bhatnagar2013directive} and PAPCNN~\cite{yin2018medical}.
\begin{figure*}[htb]
	\centering \begin{tabular}{c@{\extracolsep{0.2em}}c@{\extracolsep{0.2em}}c@{\extracolsep{0.2em}}c@{\extracolsep{0.2em}}c@{\extracolsep{0.2em}}c@{\extracolsep{0.2em}}c}
		\includegraphics[width=0.142\textwidth]{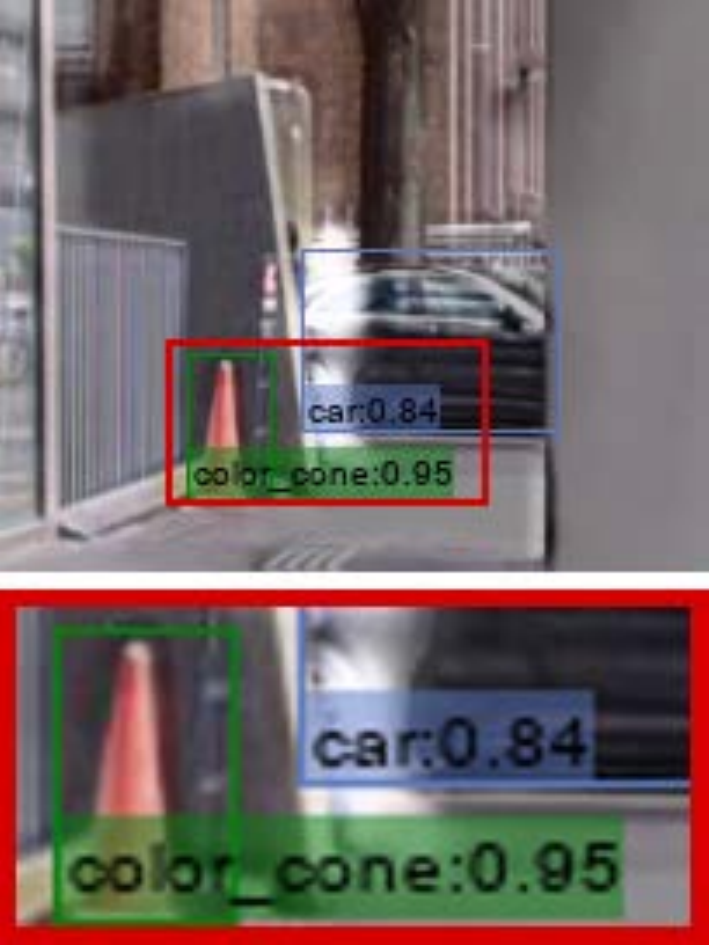}

		&\includegraphics[width=0.142\textwidth]{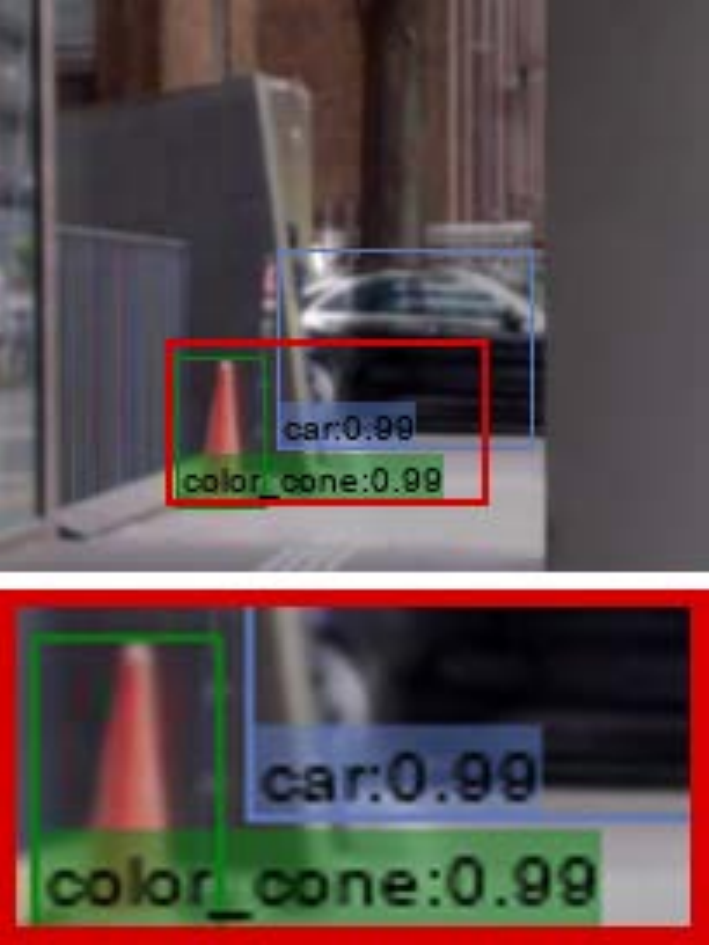}
		&\includegraphics[width=0.142\textwidth]{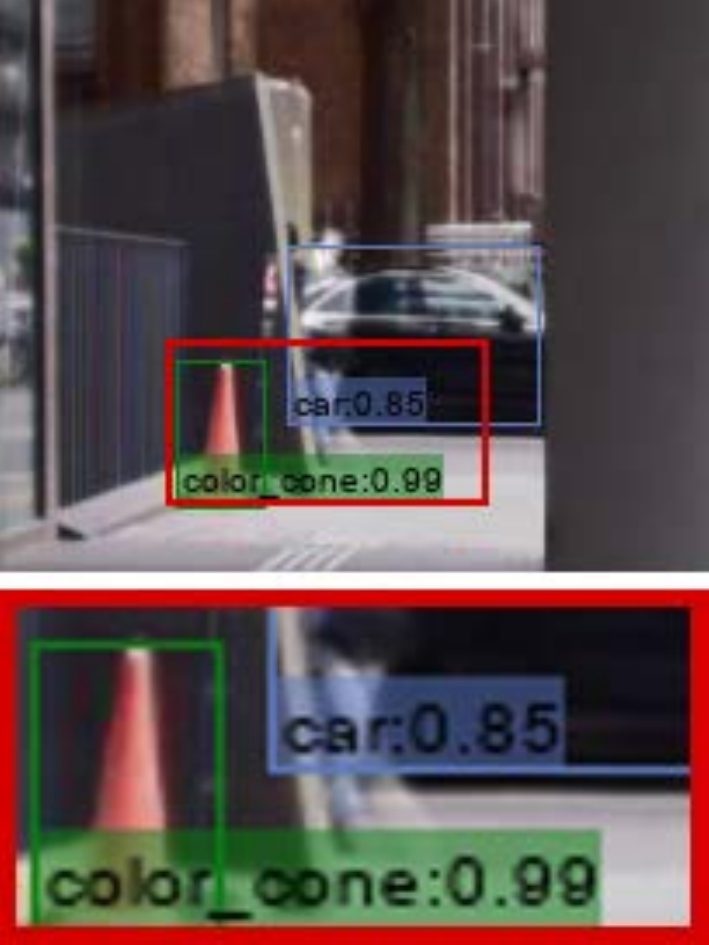}
		&\includegraphics[width=0.142\textwidth]{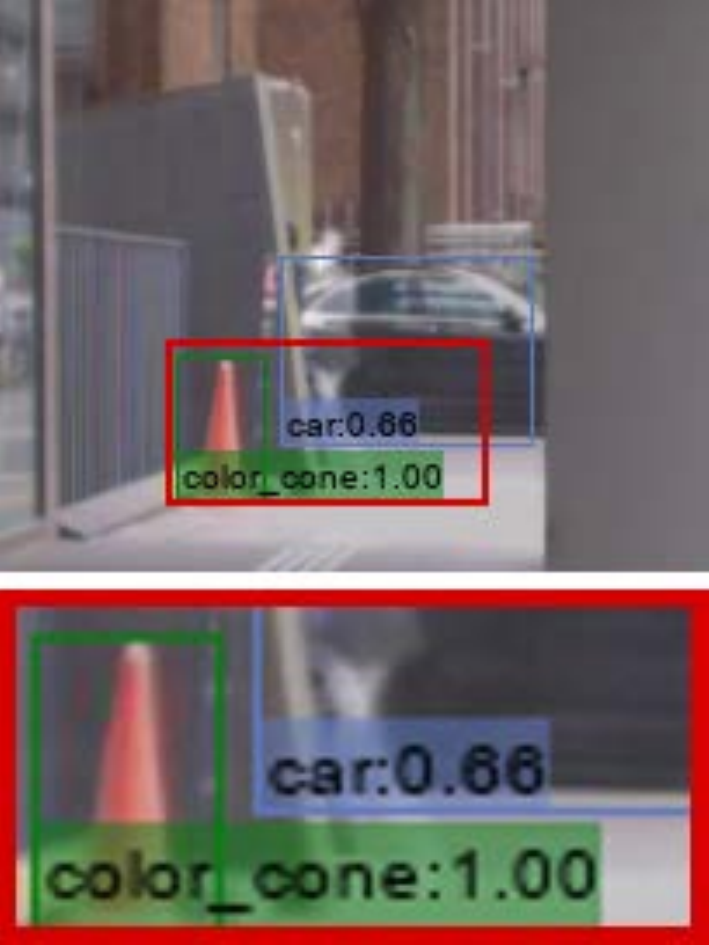}
		&\includegraphics[width=0.142\textwidth]{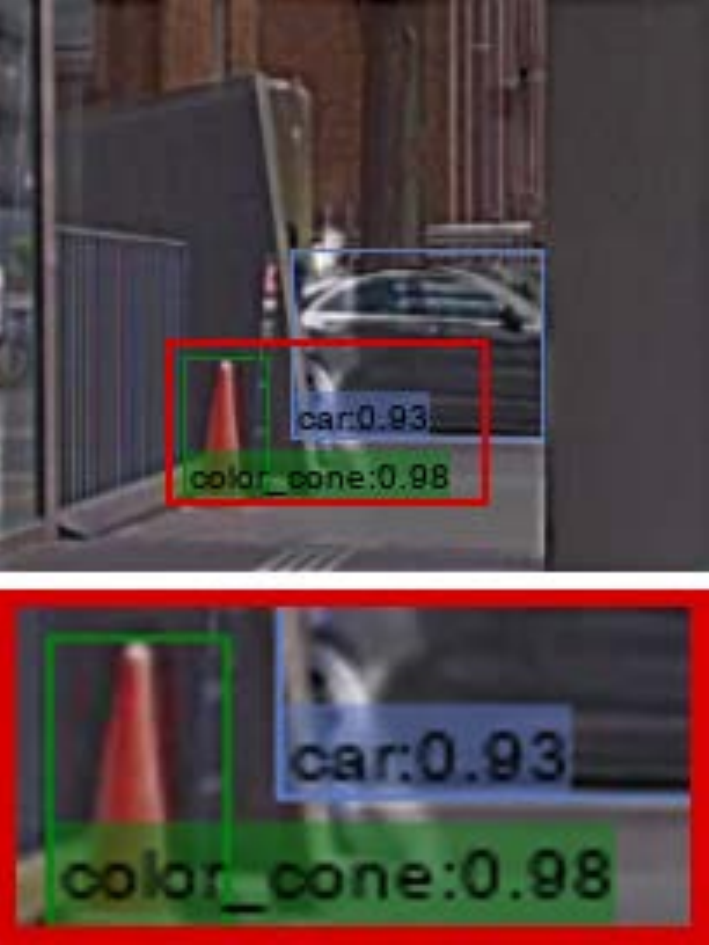}
		&\includegraphics[width=0.142\textwidth]{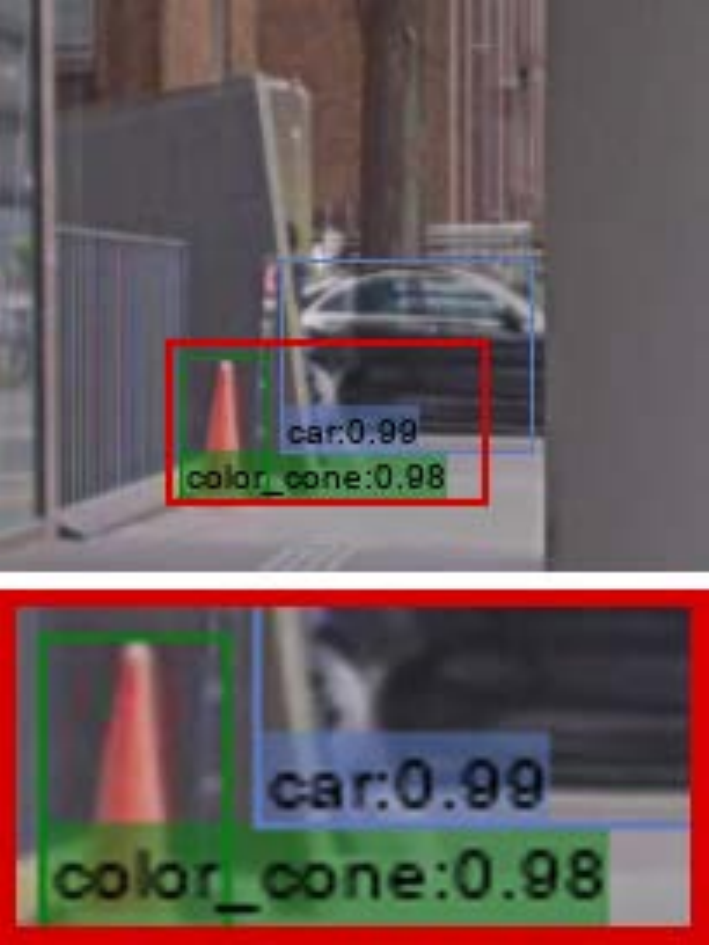}
		&\includegraphics[width=0.142\textwidth]{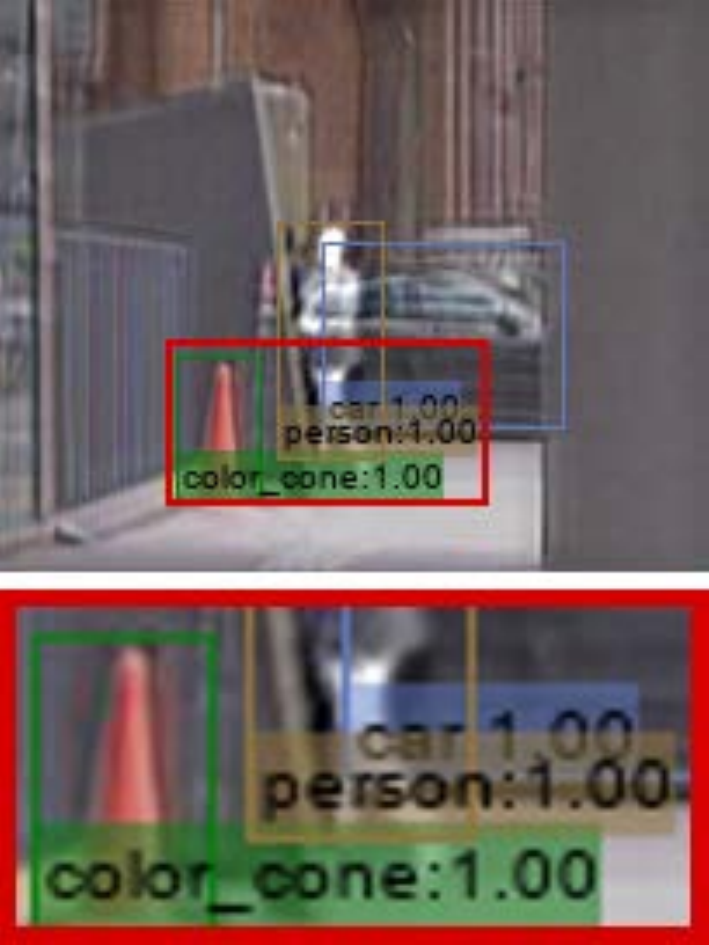}\\
		\includegraphics[width=0.142\textwidth]{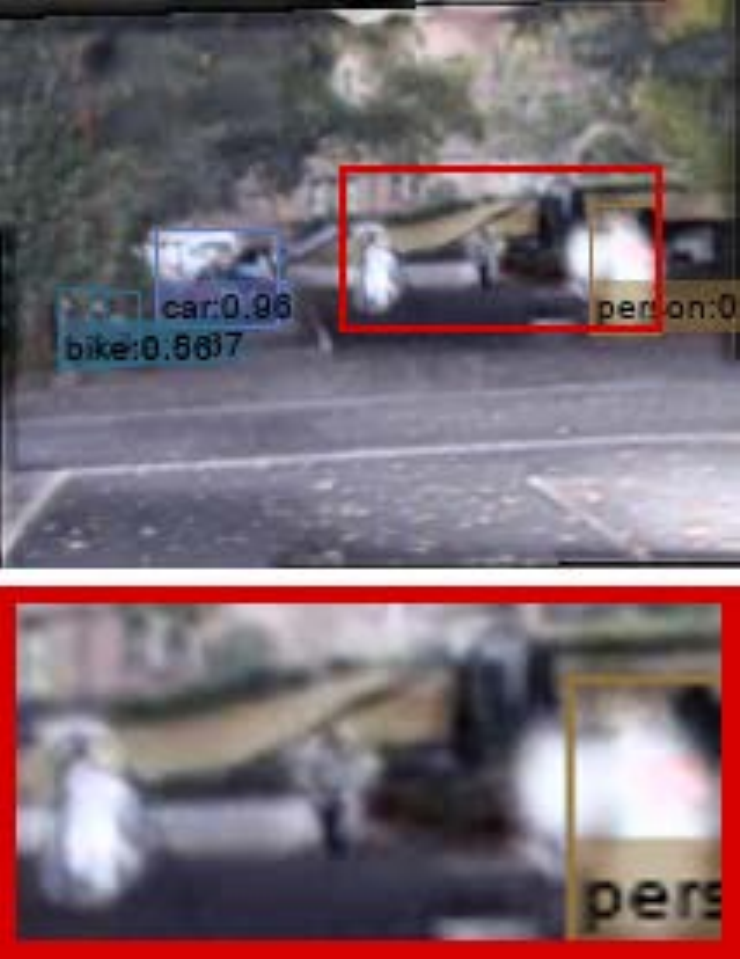}
		&\includegraphics[width=0.142\textwidth]{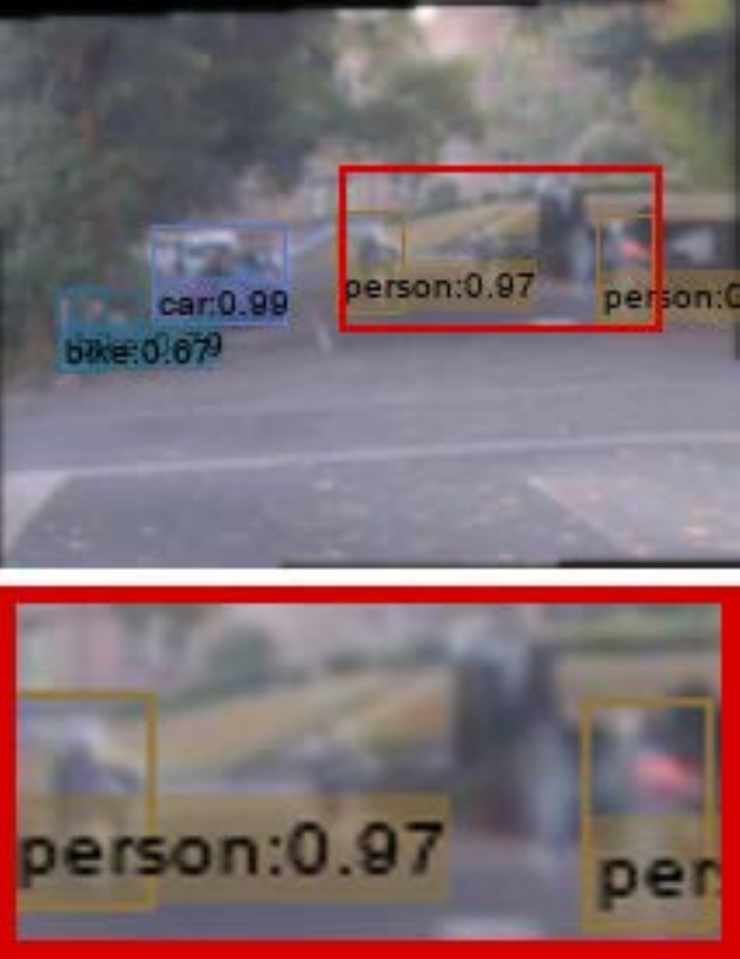}
		&\includegraphics[width=0.142\textwidth]{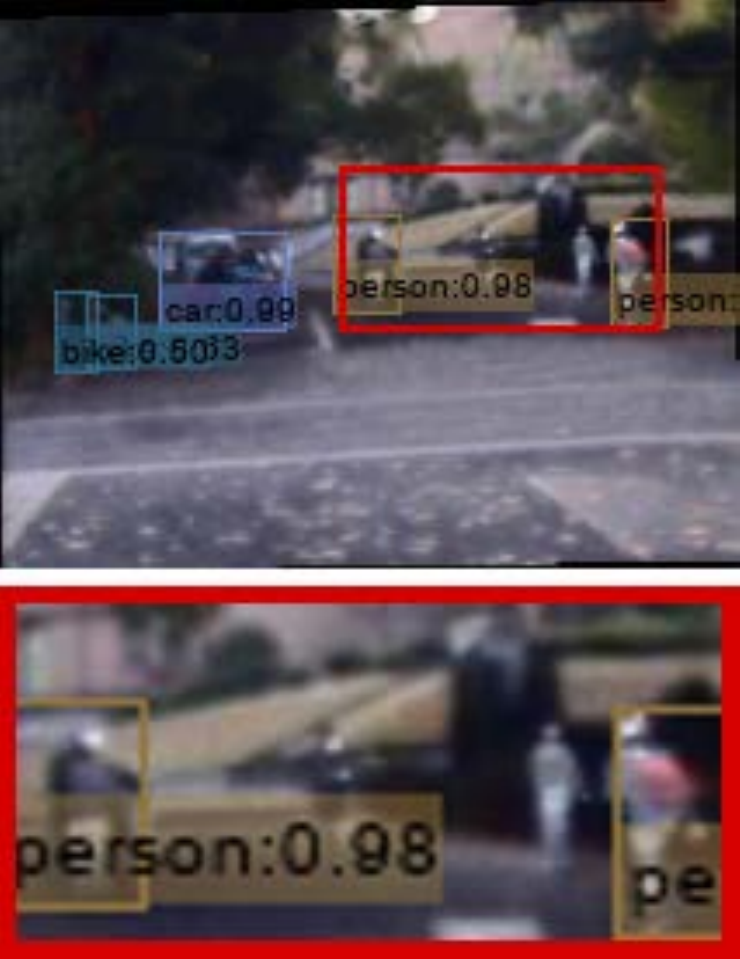}
		&\includegraphics[width=0.142\textwidth]{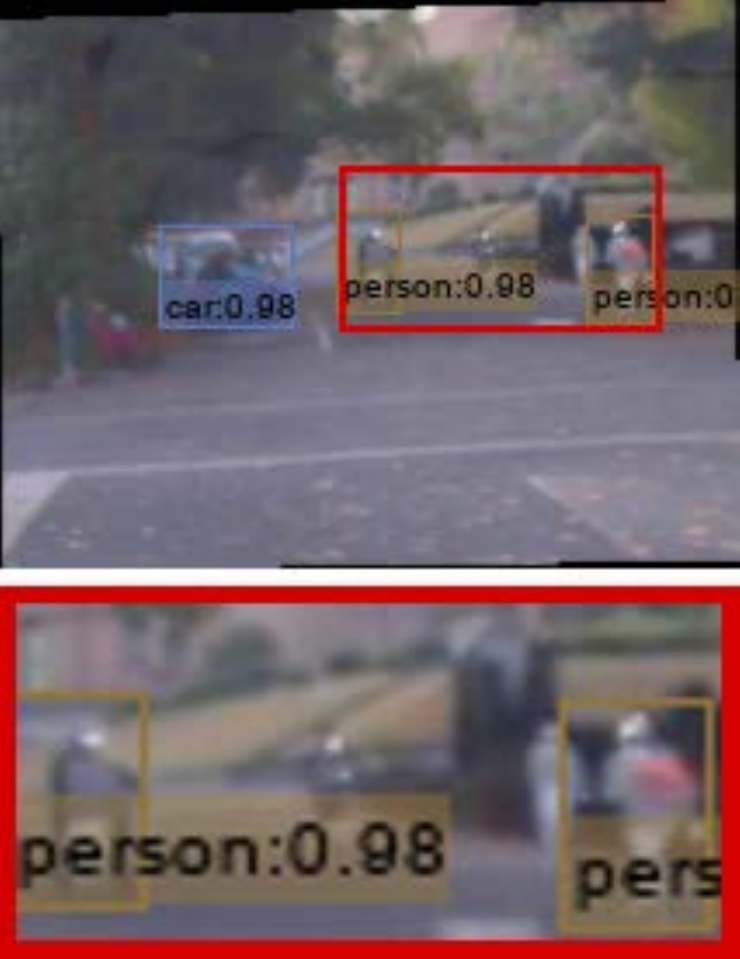}
		&\includegraphics[width=0.142\textwidth]{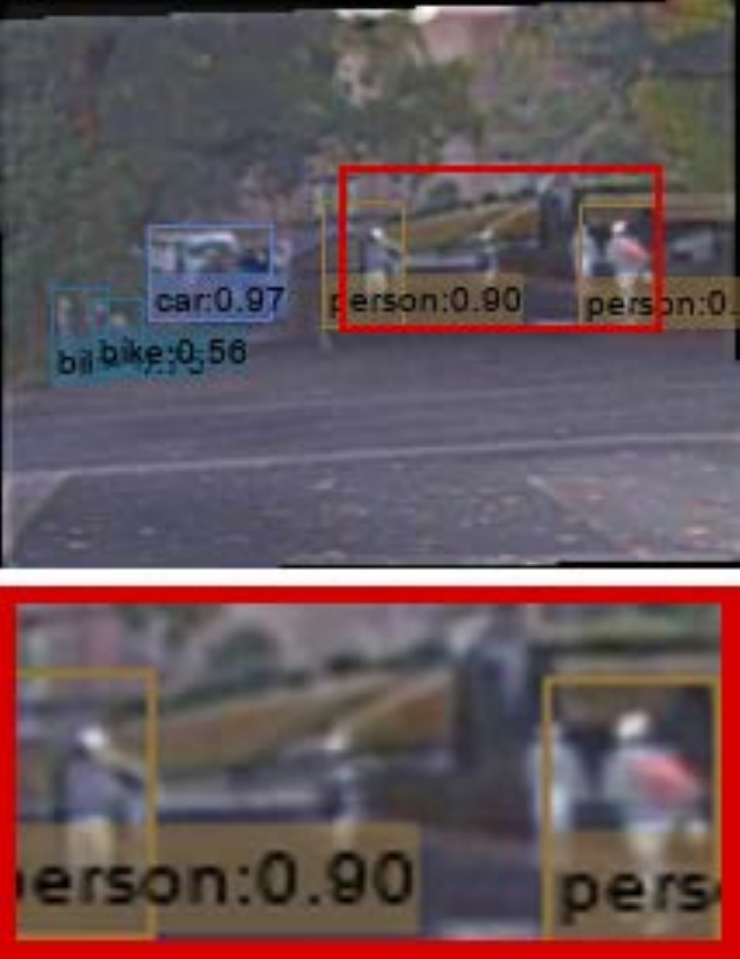}
		&\includegraphics[width=0.142\textwidth]{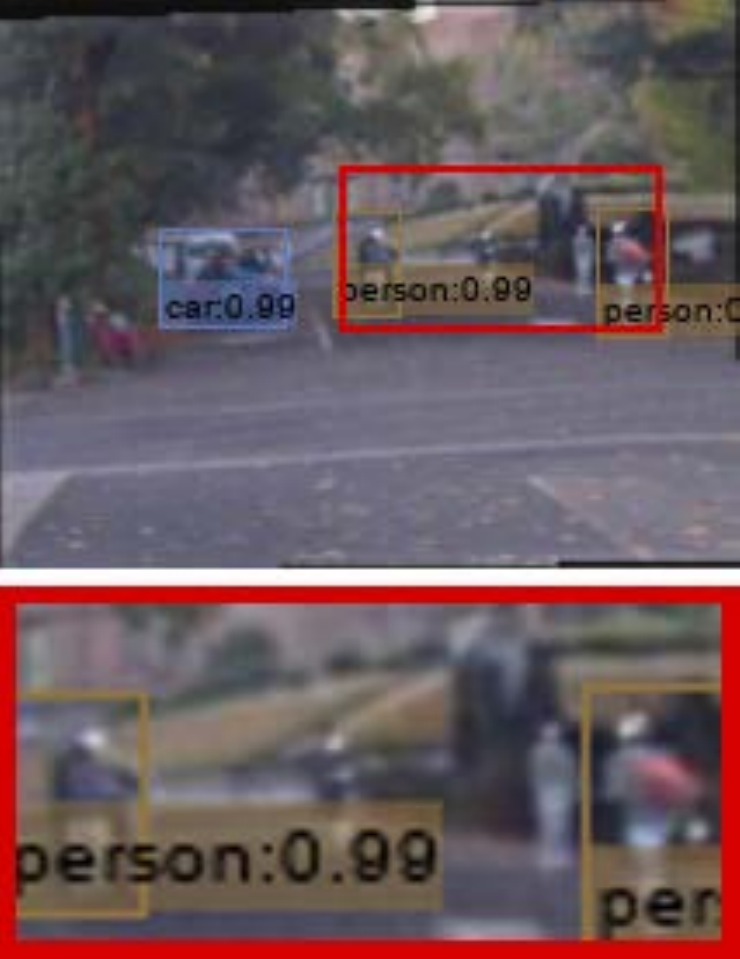}
		&\includegraphics[width=0.142\textwidth]{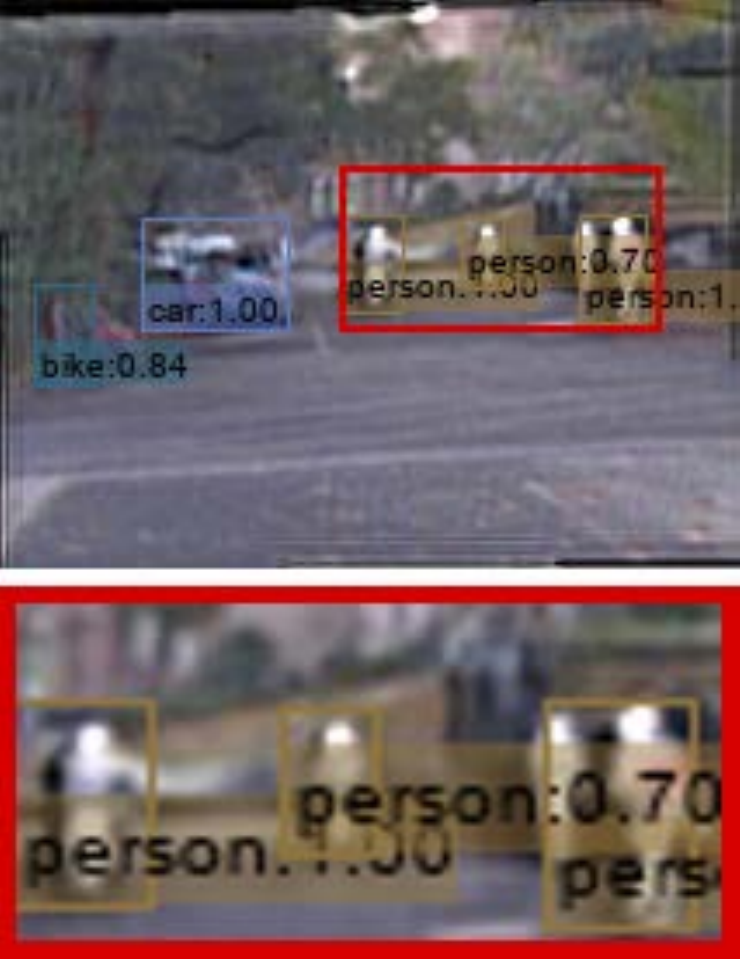}
		\\
		\includegraphics[width=0.142\textwidth]{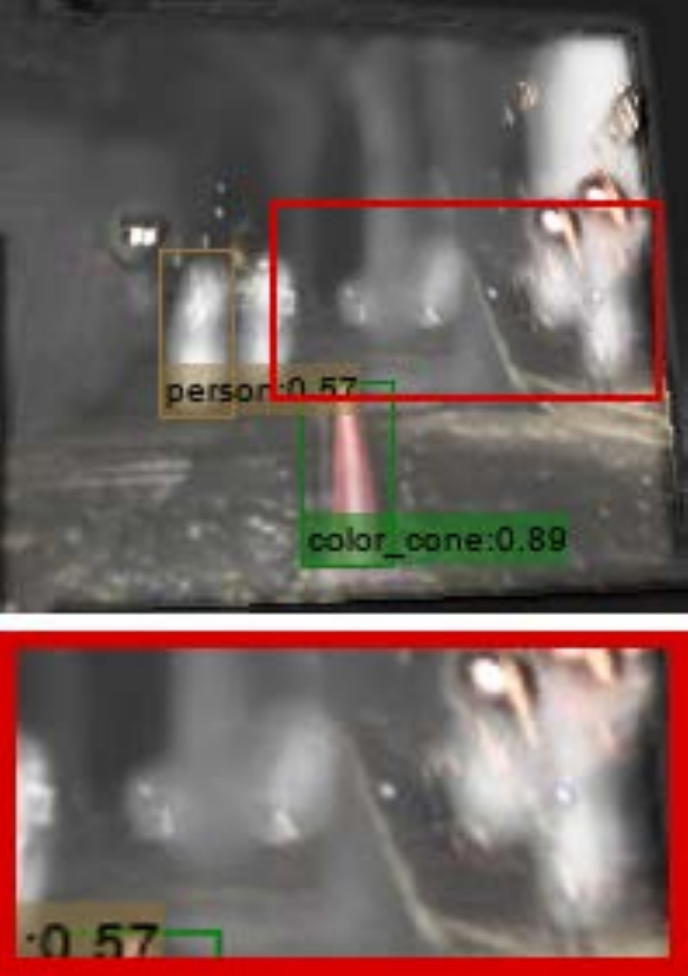}

		&\includegraphics[width=0.1428\textwidth]{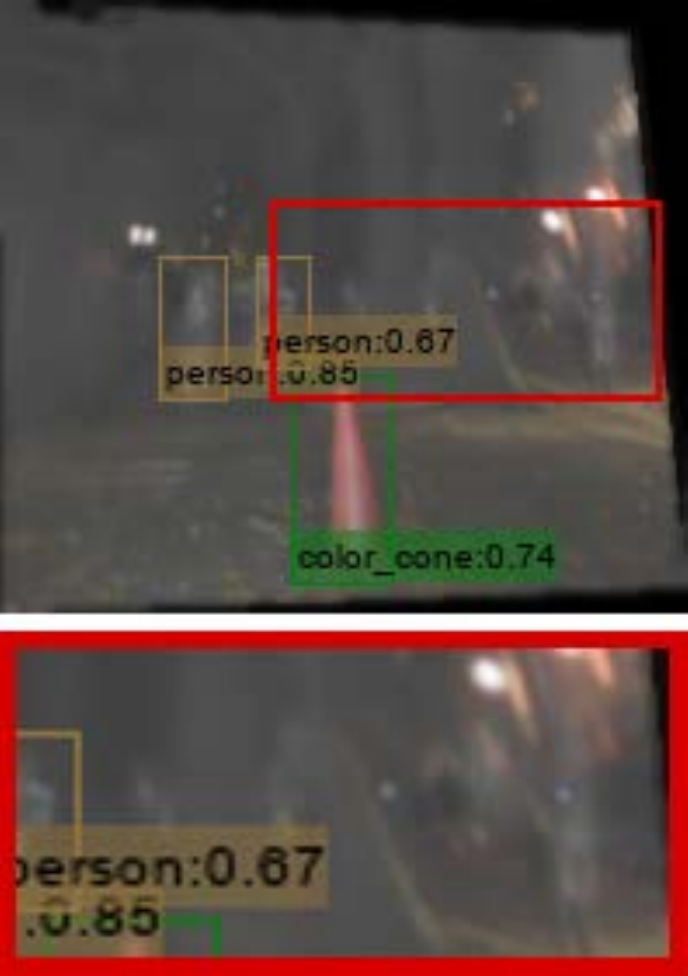}
		&\includegraphics[width=0.142\textwidth]{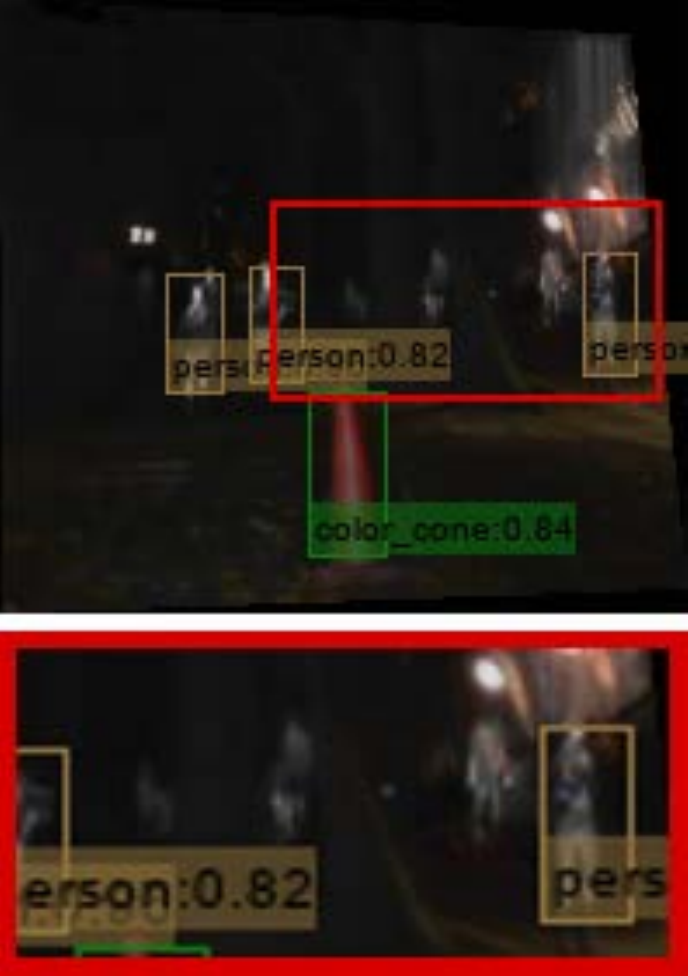}
		&\includegraphics[width=0.142\textwidth]{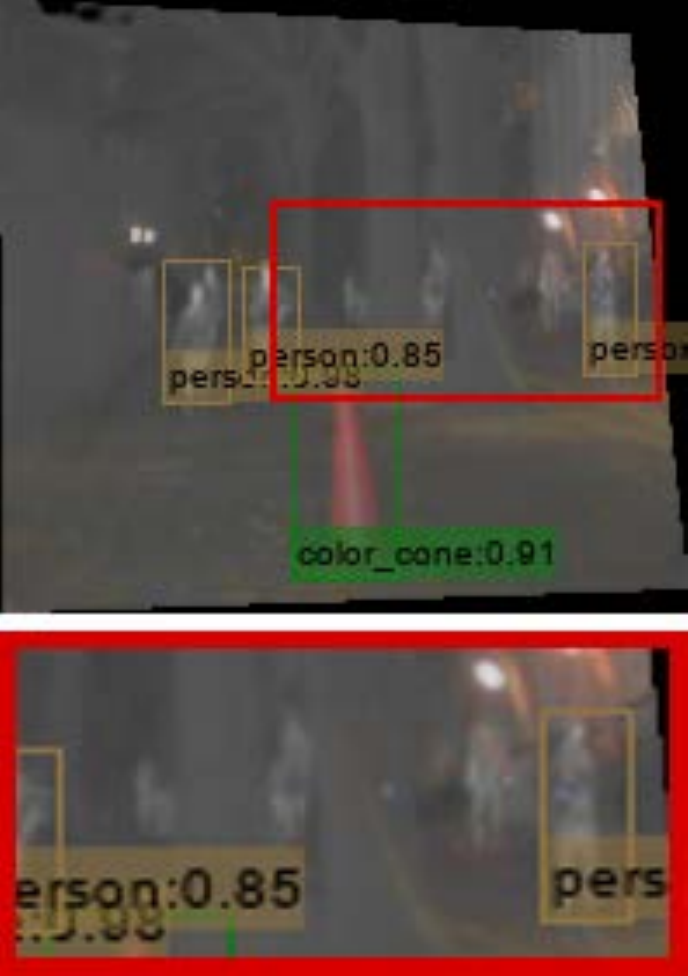}
		&\includegraphics[width=0.142\textwidth]{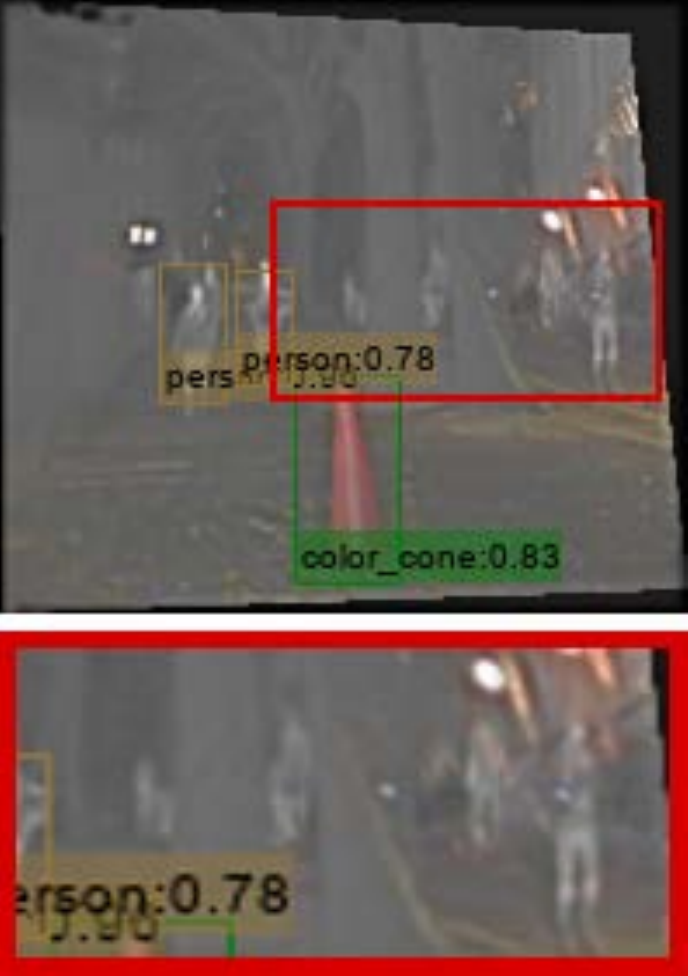}
		&\includegraphics[width=0.142\textwidth]{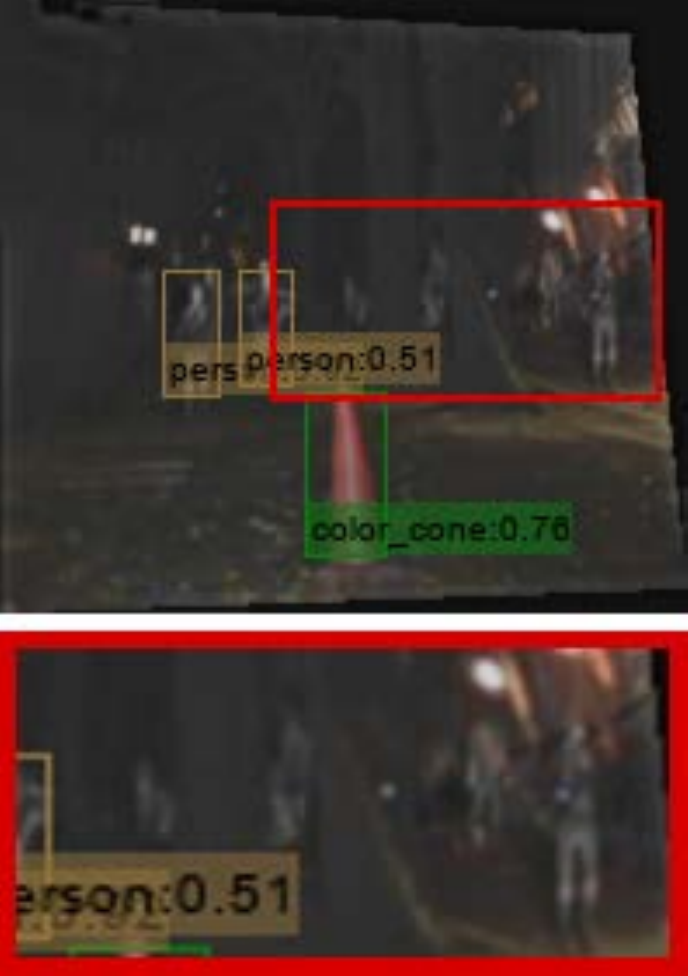}
		&\includegraphics[width=0.142\textwidth]{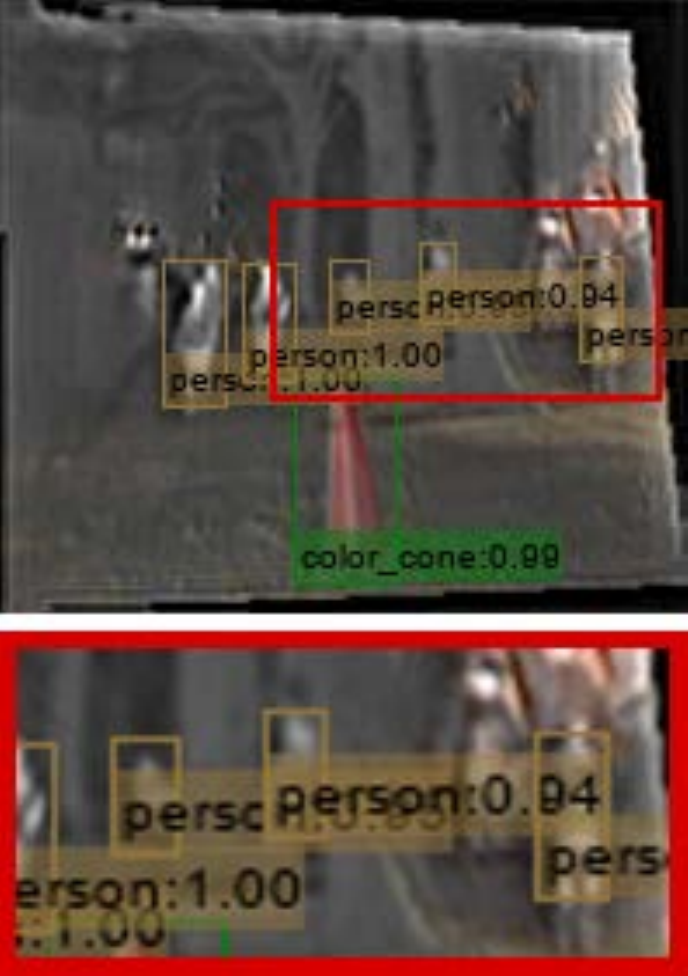}\\
		\footnotesize	DDcGAN    & \footnotesize RFN &  \footnotesize DID & \footnotesize MFEIF & \footnotesize SDNet & \footnotesize U2Fusion &\footnotesize  TIM \\
	\end{tabular}
	\caption{Object detection results based on image fusion compared with several state-of-the-art fusion methods.}
	\label{fig:result_ir_vis_od}
\end{figure*}

\textbf{Qualitative Comparisons.} 
 Intuitively, the qualitative results of MRI-PET/SPECT fusion are shown at Fig.~\ref{fig:result_mr-spect} with various brain-hemispheric transaxial sections. Essentially, limited by imaging equipment, PET/SPECT
images are with constrained resolutions and have mosaic degradation. MRI provides ample structural details. The goal of these tasks is to maintain the structural details and functional color expression. The proposed scheme can improve the visual quality by removing mosaic of PET/SPECT, shown in the last row of  Fig.~\ref{fig:result_mr-spect}. The fusion results of other  compared schemes still exists mosaic and noises. Furthermore, ASR and CSMCA cannot either maintain the significant structure of MRI or recover the color representation. Compared with these competitors, proposed scheme suppresses the generation of noised artifacts, highlights the informative structure of soft-tissue (e.g., edges) and is without color distortion. The high-contrast visual performance demonstrates the comprehensiveness.

\begin{table}[htb]
	\renewcommand{\arraystretch}{1.3}
	\caption{ Quantitative results of object detection on {Multi-Spectral} dataset.}
	\label{tab:data_set_results}
	\centering
	\setlength{\tabcolsep}{1.0mm}{
		\begin{tabular}{c c  c c c c c}
			\hline

			Methods & Color  Cone &  Car Stop &  Car & Person & Bike &  mAP \\\hline
			
			Visible &0.2454 &0.4201 &0.5782 &0.4229&0.4539&0.4241  \\
			Infrared &0.2360 &0.3671 &0.5757 &	0.5596 &0.4205&0.4318  \\
			
			DDCGAN &0.2105 &0.3591 &0.4629 &0.3086&0.4790&0.3640 \\
			RFN &0.2587 &{0.4210} &0.5258 &0.4526&0.4470&0.4210\\
			DenseFuse        &\textbf{0.2635}&0.3957&0.5541&0.5185&0.4790&0.4422  \\
			AUIF      &0.2259 &	0.3982 &	0.5318 	&0.4776 	&	0.5138 	& 0.4294  \\ 
			DID      &0.2353&0.3855&0.5176&0.5050&0.4527&0.4192 \\ 
			MFEIF        &0.2515 &	0.3999 &	0.5483& 	0.5300 	&	0.4941 &	0.4447  \\
			SMOA		&0.2592 &	0.4041& 	0.5561 &	0.5275 	&	0.5289 &	0.4551  \\
			TARDAL      &0.1504 &0.2867&0.4724&0.5594&0.3075&0.3553 \\ 
			SDNet      &0.2329 	&0.3857& 	0.5138 &	0.4470 	&	0.4595 &	0.4078 \\ 
			U2Fusion		&0.2406&{0.4138}&	0.5629&0.5113&\textbf{0.5396}&0.4537\\

			TIM   & {0.2576} &\textbf{0.4285} &\textbf{0.6253}&\textbf{0.5638} &{0.5080} &\textbf{0.4766} \\
			\hline
		\end{tabular}	
	}
\end{table}
\textbf{Quantitative Comparisons.}
 Objective evaluations are also conducted to demonstrate the superiority of our fused results based on four 
metrics, MI, Entropy (EN), VIF and Sum of Correlation of difference (SCD)~\cite{aslantas2015new}. 
Due to the
diverse imaging quality of these medical modalities (e.g., mosaic factors), we utilize EN to measure the amount of information remaining in fused images. Furthermore, the edge details is not so dense as visible images, we utilize SCD rather edge-aware metrics (FMI$_\mathtt{edge}$ and $\mathrm{Q^{AB/F}}$). SCD is leveraged to measure the correlation between the difference images. 
We depict the numerical performances for three medical image fusion tasks by plotting the box-type figures in Fig.~\ref{fig:result_medical}. 
Clearly, the proposed scheme achieves the consistent optimal mean values on these fusion tasks under four numerical metrics.

\begin{table*}[thb]
	\renewcommand{\arraystretch}{1.2}
	\caption{ Quantitative results of  semantic segmentation on {MFNet} dataset.}~\label{tab:data_set_results_seg}
	\setlength{\tabcolsep}{1.3mm}{
		\begin{tabular}{ccccccccccccccccccccccccc}
			\hline
			\multirow{2}{*}{Methods} & \multicolumn{2}{c}{Car} & \multicolumn{2}{c}{Person} & \multicolumn{2}{c}{Bike}& \multicolumn{2}{c}{Curve}  & \multicolumn{2}{c}{Car Stop}& \multicolumn{2}{c}{Guardrail}& \multicolumn{2}{c}{Color Cone}& \multicolumn{2}{c}{Bump} & \multirow{2}{*}{mAcc} &\multirow{2}{*}{mIOU} \\ \cline{2-17} 
			& Acc &IOU & Acc &IOU& Acc &IOU& Acc &IOU& Acc &IOU& Acc &IOU& Acc &IOU& Acc &IOU\\
			\hline
			Visible  &0.937	&0.829&\textbf{0.896}& 0.150&  0.725&	0.603& 0.747&	0.266& 0.748&	0.616&0.858& 	 0.635&0.776&	0.499& 0.914&	0.625& 0.842 &	0.577
			\\
			Infrared	&0.902&0.559& 0.810	&0.499& 0.736&	0.415& 0.551&	0.092& 0.728&	0.218& 0.144&	0.012& 0.815&	0.318& 0.894&	0.458& 0.727&	0.392\\
			
			DDCGAN&0.921&	0.803& 0.856&	0.388& 0.785&	0.612& 0.658&	0.194&	 0.711&0.579& 0.858&	0.474& 0.709&	0.445& 0.878	&0.530& 0.817 &	0.555 \\
			RFN   &	0.937& 0.834&0.880&	0.399& 0.694&	0.596&0.740&	0.268&0.745&	0.604& 0.820&	0.603& 0.773&	0.520&0.900&	0.608&0.830&	0.602 \\ 
			DenseFuse&\textbf{0.941}&	0.842& 0.885&	0.427& 0.693&	0.593& 0.759&	0.266& 0.753&	0.603& 0.830&	0.613& 0.798&	0.524& 0.901&	0.623& 0.838 &	0.608
			\\ 
			AUIF &	0.918&0.792& 0.888&	0.335& 0.653	&0.490	& 0.748&0.114 	& 0.779&0.552& 0.752	&0.560& 0.756	&0.450	& 0.942&0.408& 0.823 &	0.519     \\
			DID 	&0.879&0.741& 0.890&	0.274& 0.651&	0.443	& 0.690&0.111& 0.764&	0.504& 0.746&	0.500& 0.719&	0.405& \textbf{0.947}&	0.362& 0.806&	0.479	 \\
			MFEIF  &0.937& 	0.837	& 0.877 &0.429	& 0.699 &0.591 & 0.763&	0.268 & 0.764&	0.603& 0.836 &	0.639& 0.781 &	0.514& 0.915 &	0.602& 0.839  &	0.606\\
			SMOA 	&0.940&0.842 & 0.880&	0.417& 0.677 &	0.581& 0.754 &	0.246 & 0.756 &	0.602&0.829& 	0.620& 0.791 &	0.522& 0.905 &	0.611 & 0.835 &	0.602 
			\\ 
			TARDAL   &0.940 &	0.837& 0.881& 	0.444& 0.694 &	0.560 & 0.760&	0.240 & 0.782&	0.592& 0.839 &	0.585& \textbf{0.822} &	0.493& 0.898 &	0.555& 0.844 &	0.587 \\ 
			SDNet   &0.929&	0.800&0.879& 0.618 &	0.392 &	0.521& \textbf{0.796} &	0.093 & 0.773	&0.546 	&0.588& 0.826 &	0.490& 0.784 &	0.534& 0.917 & 0.834 &	0.548   \\
			U2Fusion &0.937  &	0.813 &\textbf{ 0.896}	 &0.308 & 0.725 &	0.500 & 0.747 &	0.151 & 0.748 &	0.585 & 0.858 &	0.608 & 0.776 &	0.502  & 0.914&	0.556  & 0.842 &	0.555 \\
	
			TIM  &\textbf{0.941}&	\textbf{0.880} &{0.847}&	\textbf{0.661}&\textbf{0.820} &	\textbf{0.681} &{0.771} &	\textbf{0.481}&\textbf{0.783} &	\textbf{0.659}&\textbf{0.871}   &	\textbf{0.633} &{0.809} &	\textbf{0.527} &{0.896} &\textbf{0.661} &	\textbf{0.858} &	\textbf{0.685} \\
			\hline
	\end{tabular} }
\end{table*}
\subsection{Image Fusion for Semantic Understanding}\label{sec:app}
 Benefiting from the nested optimization, we can 
facilitate the improvement for two  semantic understanding tasks (e.g.,  object detection and  segmentation) based on image fusion.

\begin{figure*}[thb]
	\centering \begin{tabular}{c@{\extracolsep{0.15em}}c@{\extracolsep{0.15em}}c@{\extracolsep{0.15em}}c@{\extracolsep{0.15em}}c@{\extracolsep{0.15em}}c@{\extracolsep{0.15em}}c@{\extracolsep{0.15em}}c@{\extracolsep{0.15em}}c}
		\includegraphics[width=0.108\textwidth]{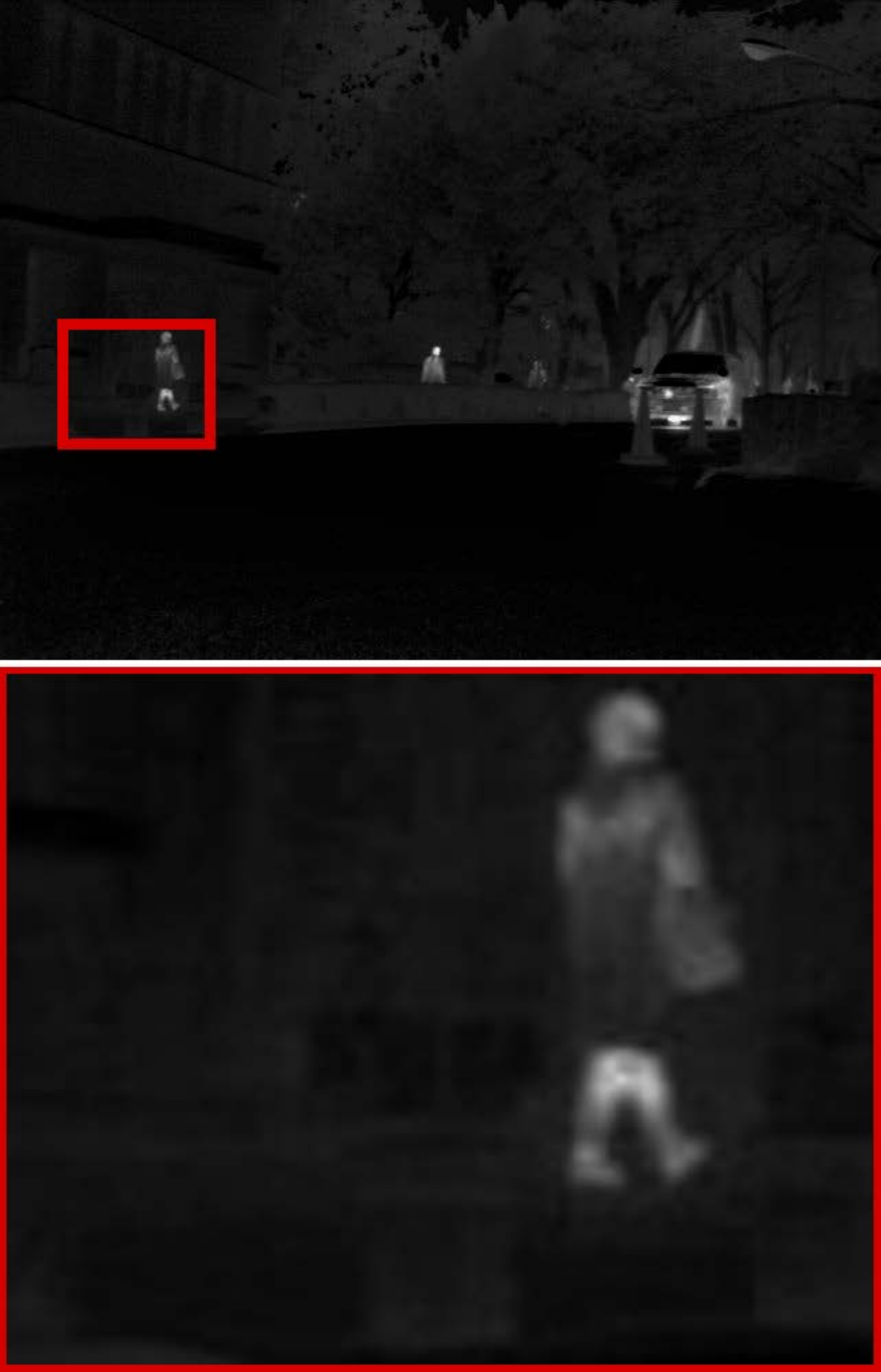}
		&	\includegraphics[width=0.108\textwidth]{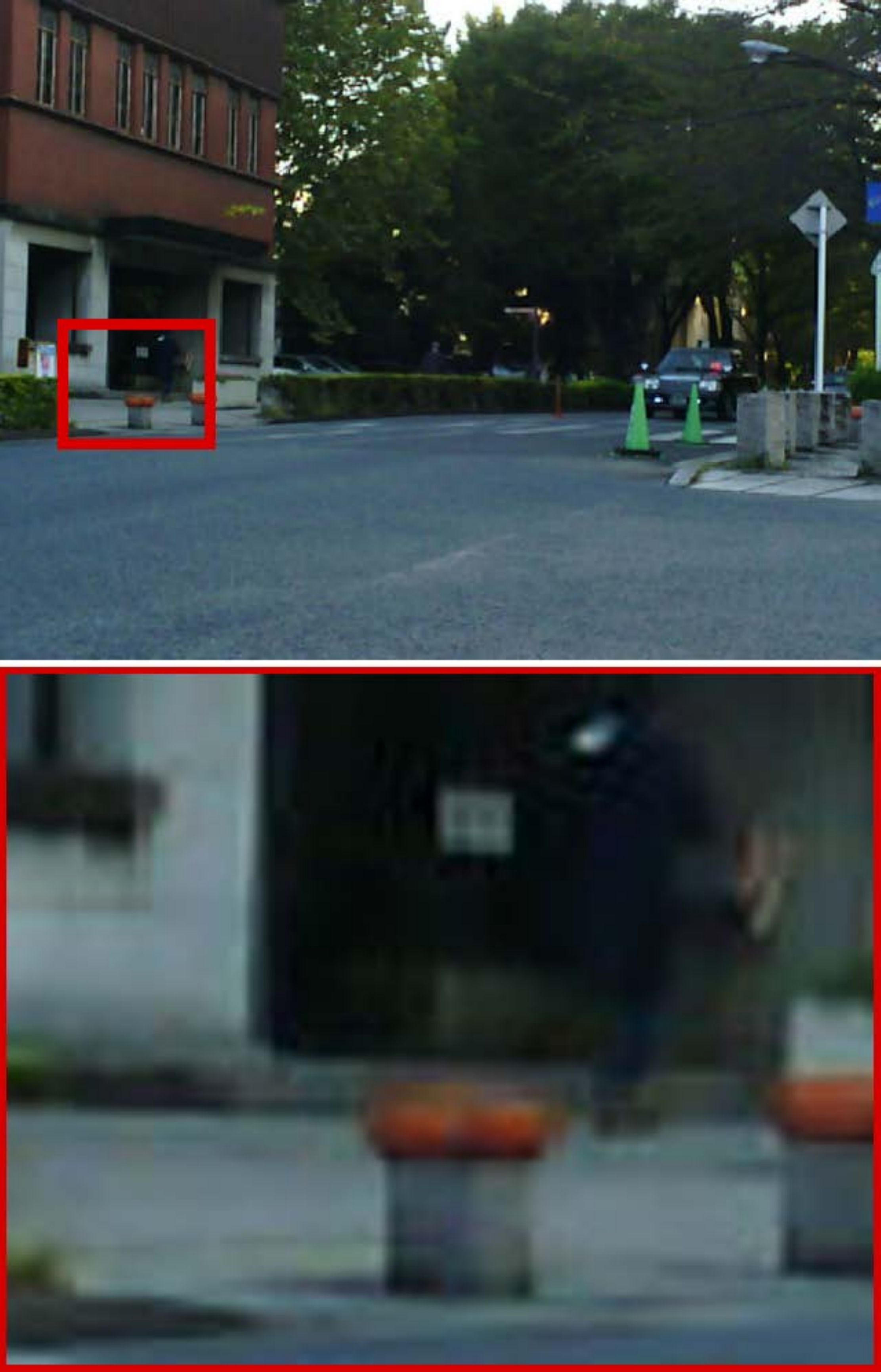}
		
		&	\includegraphics[width=0.108\textwidth]{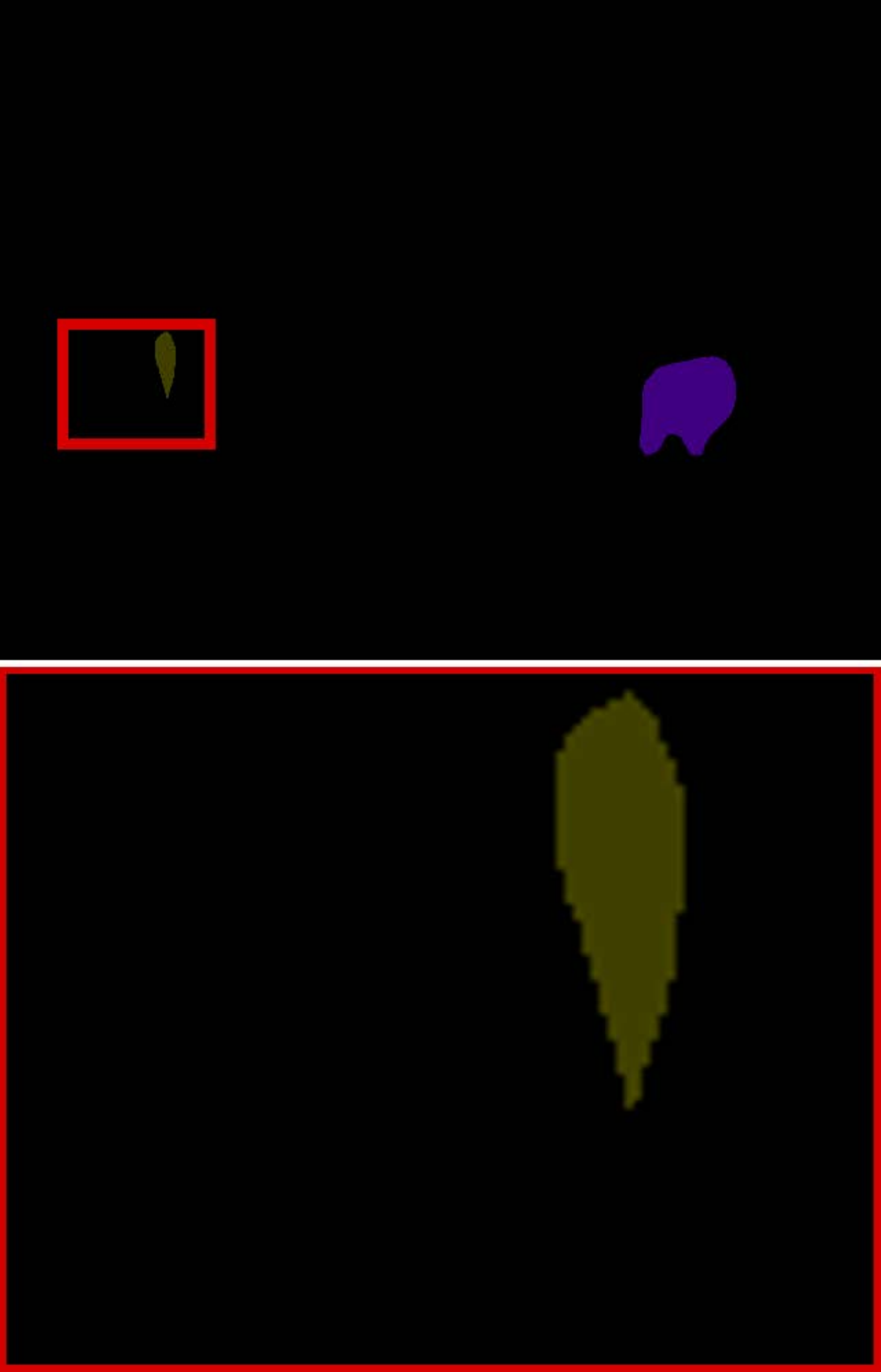}
		&	\includegraphics[width=0.108\textwidth]{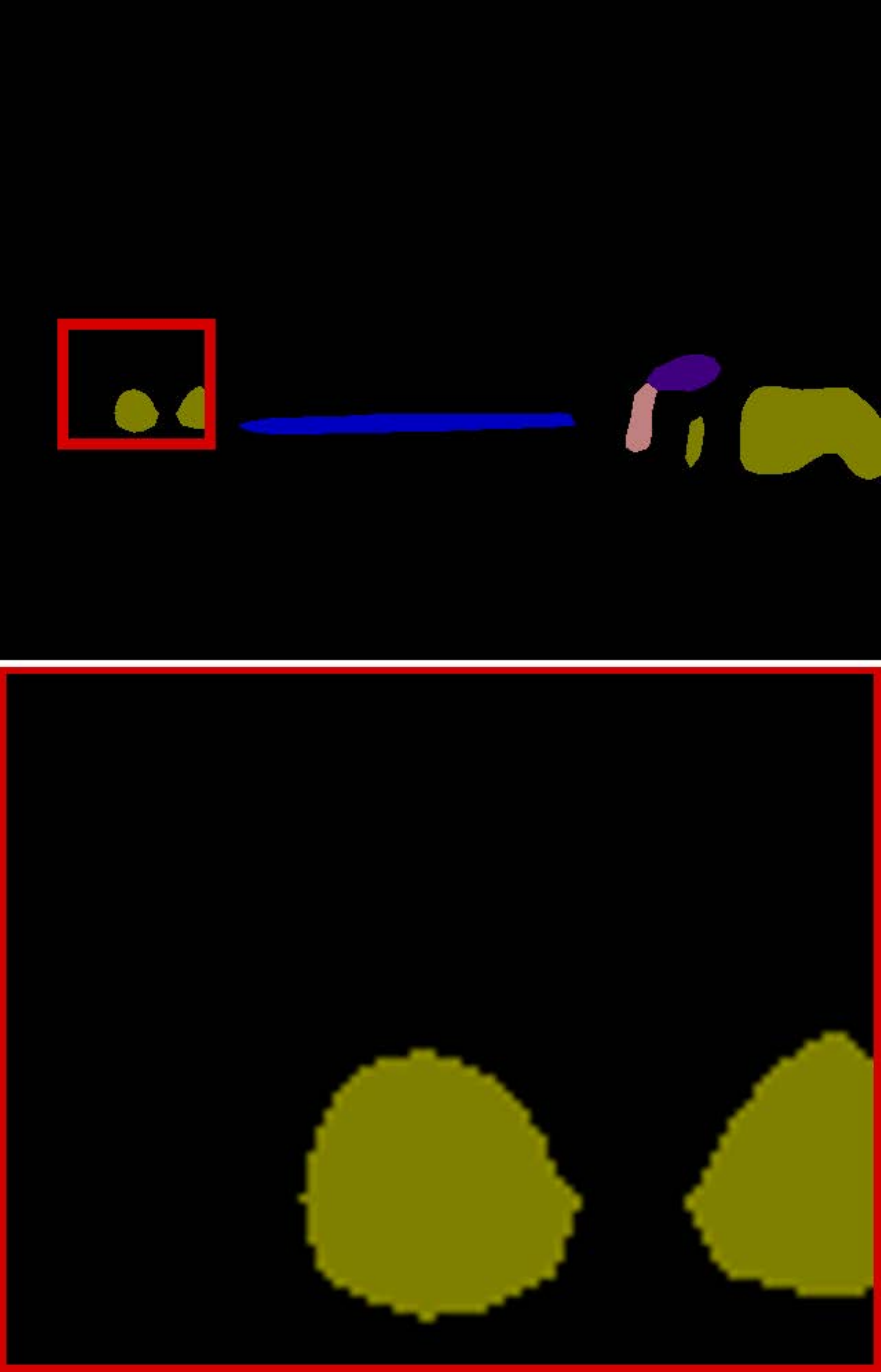}
		&	\includegraphics[width=0.108\textwidth]{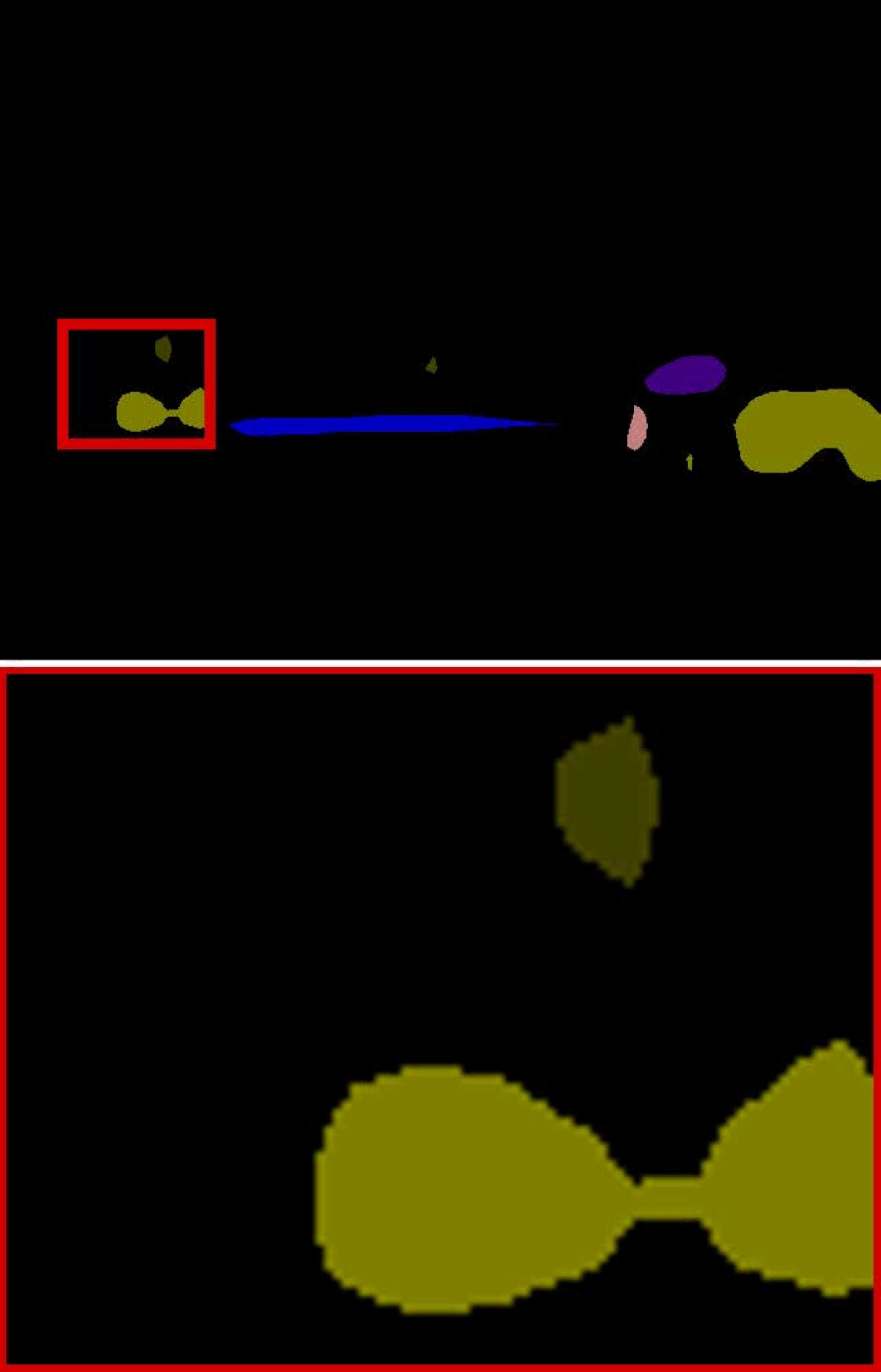}
		&	\includegraphics[width=0.108\textwidth]{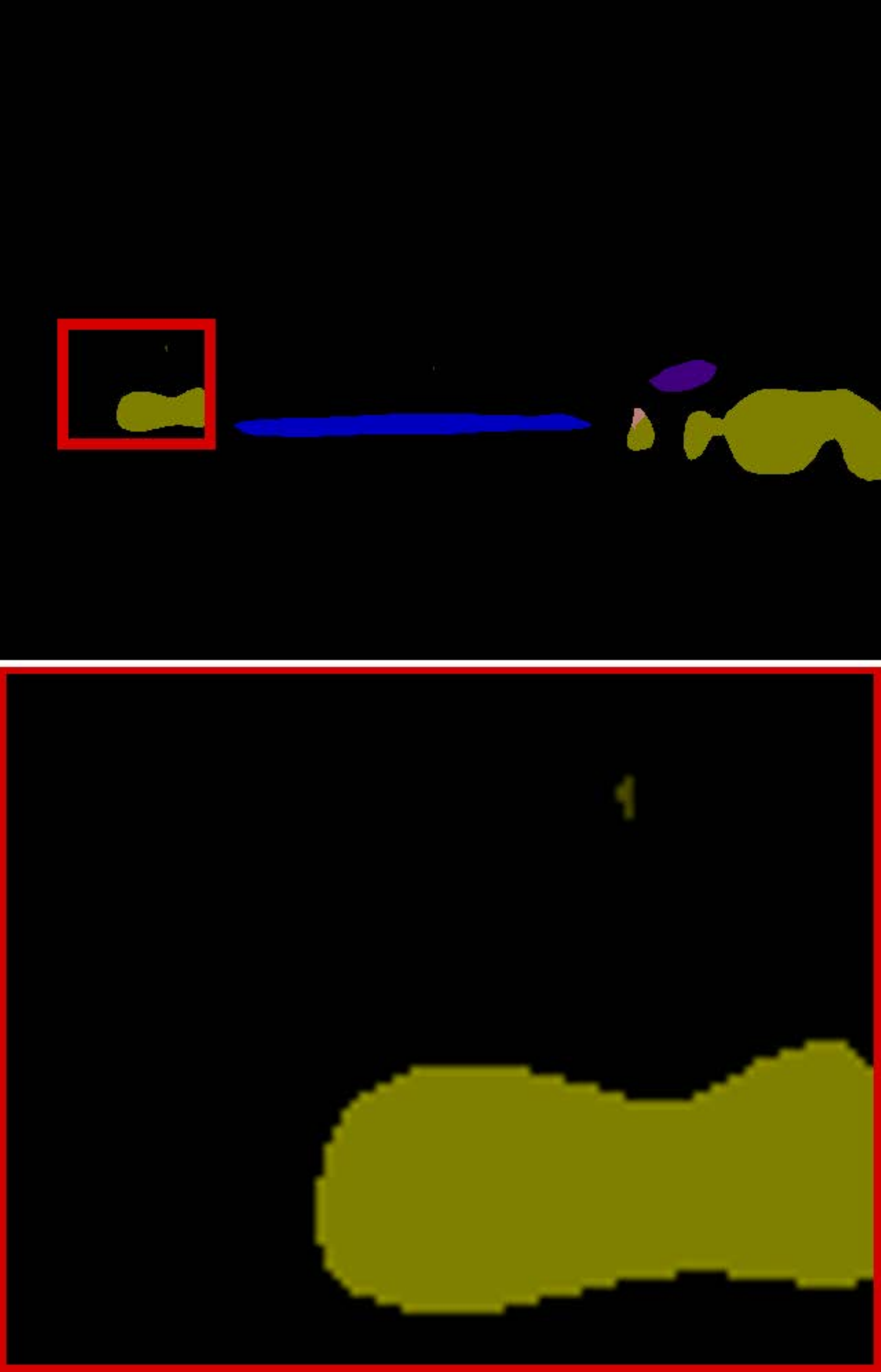}
		&	\includegraphics[width=0.108\textwidth]{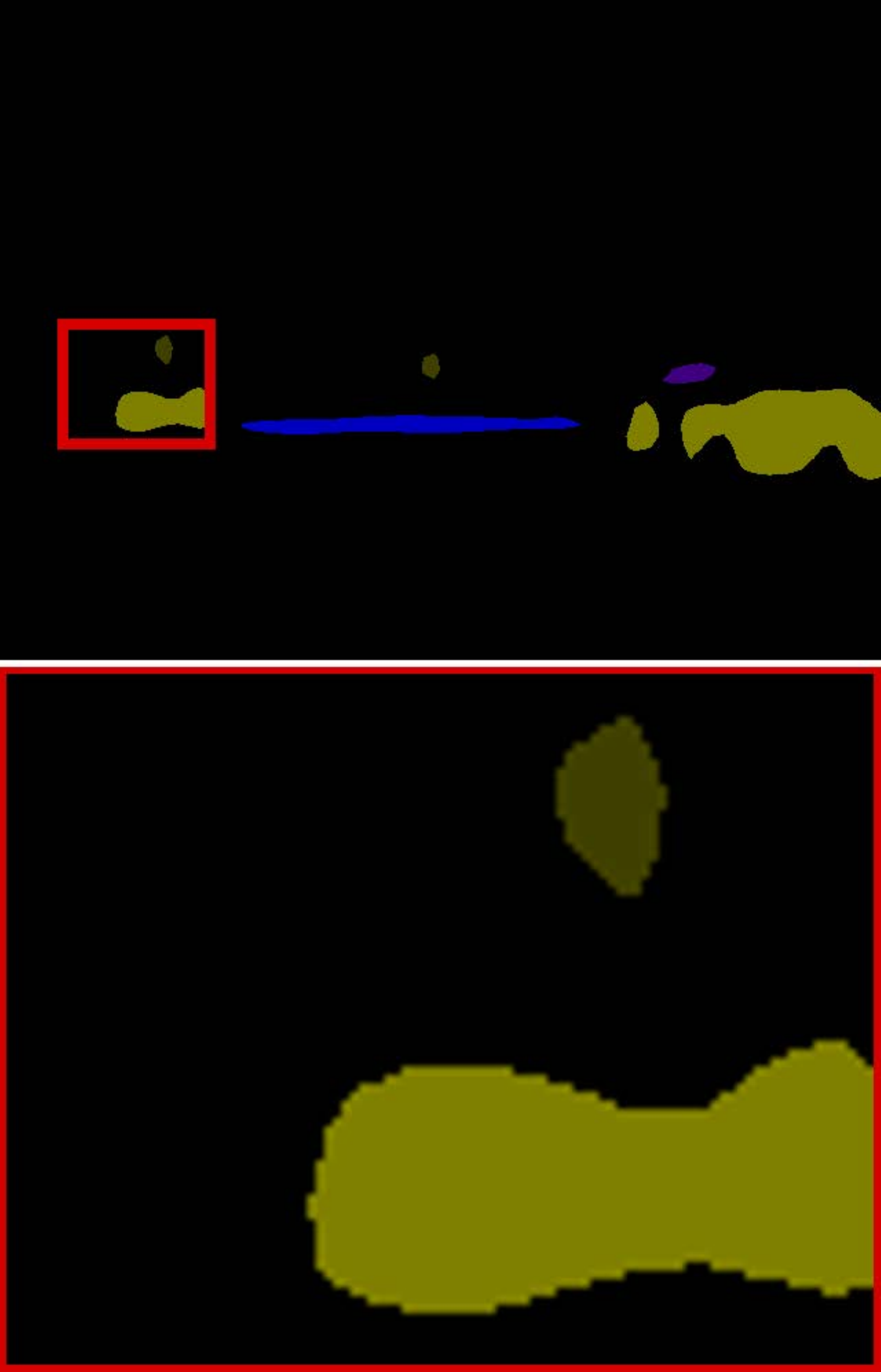}
		&	\includegraphics[width=0.108\textwidth]{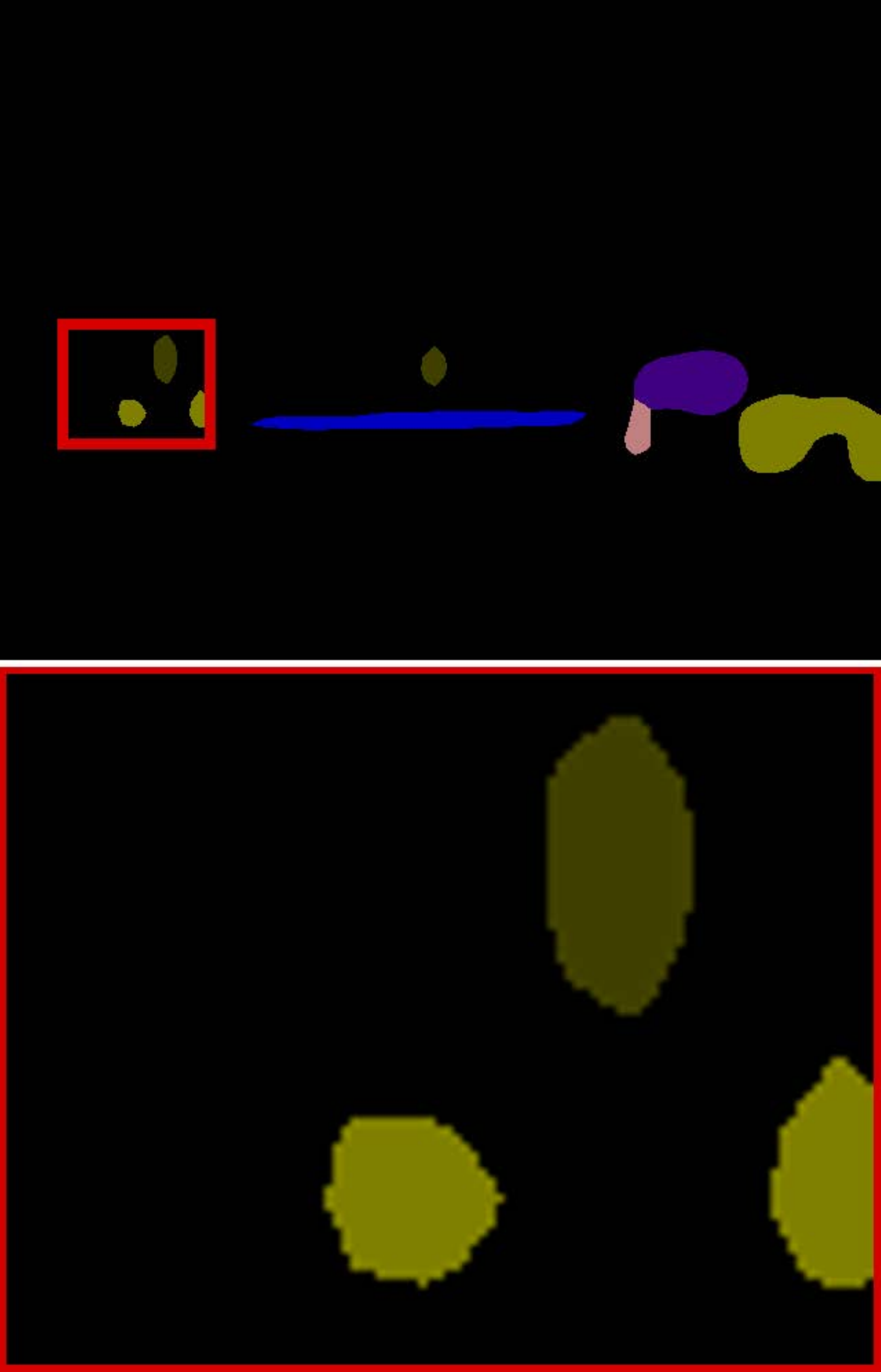}
		&	\includegraphics[width=0.108\textwidth]{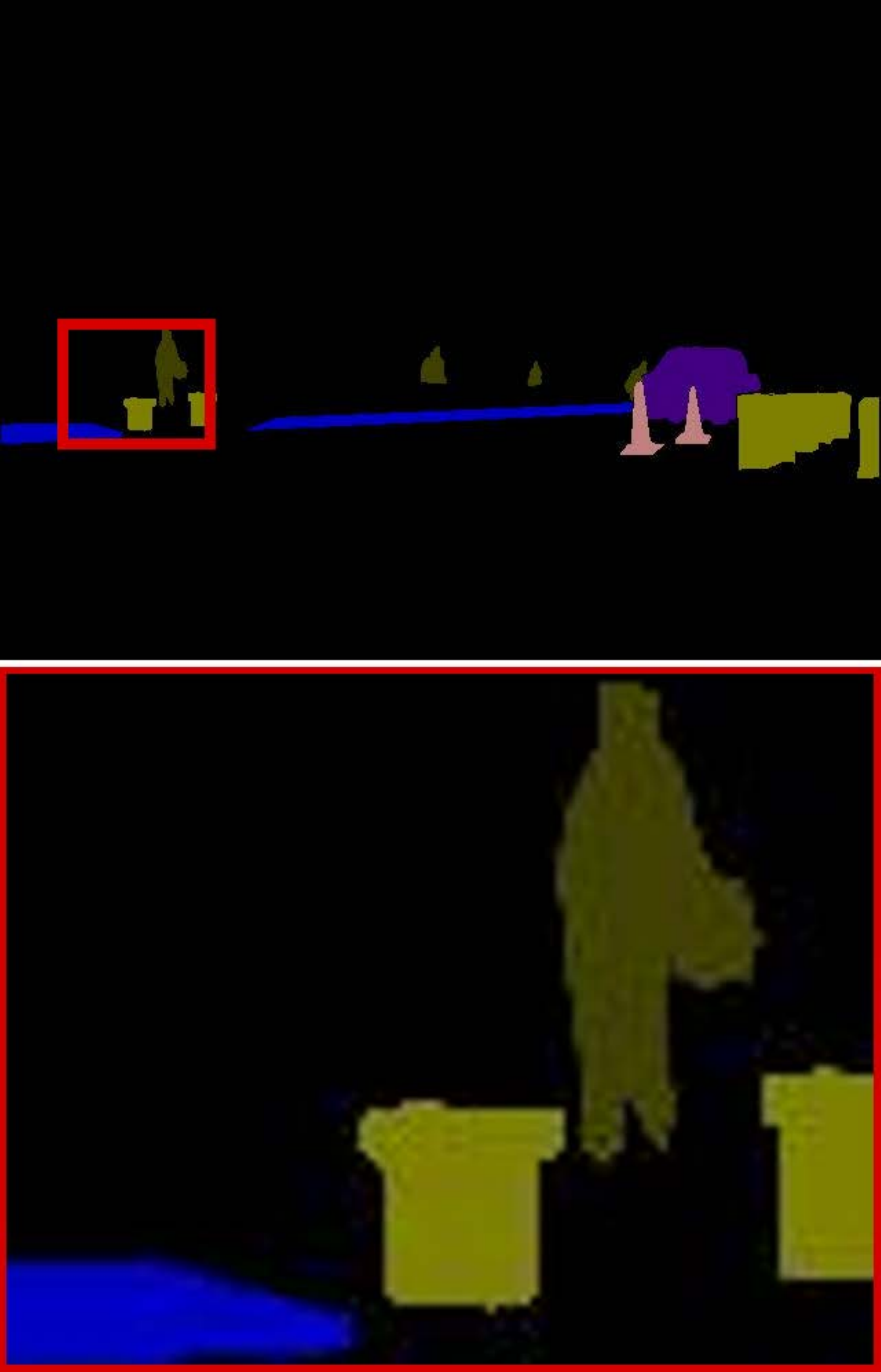}\\
\includegraphics[width=0.108\textwidth]{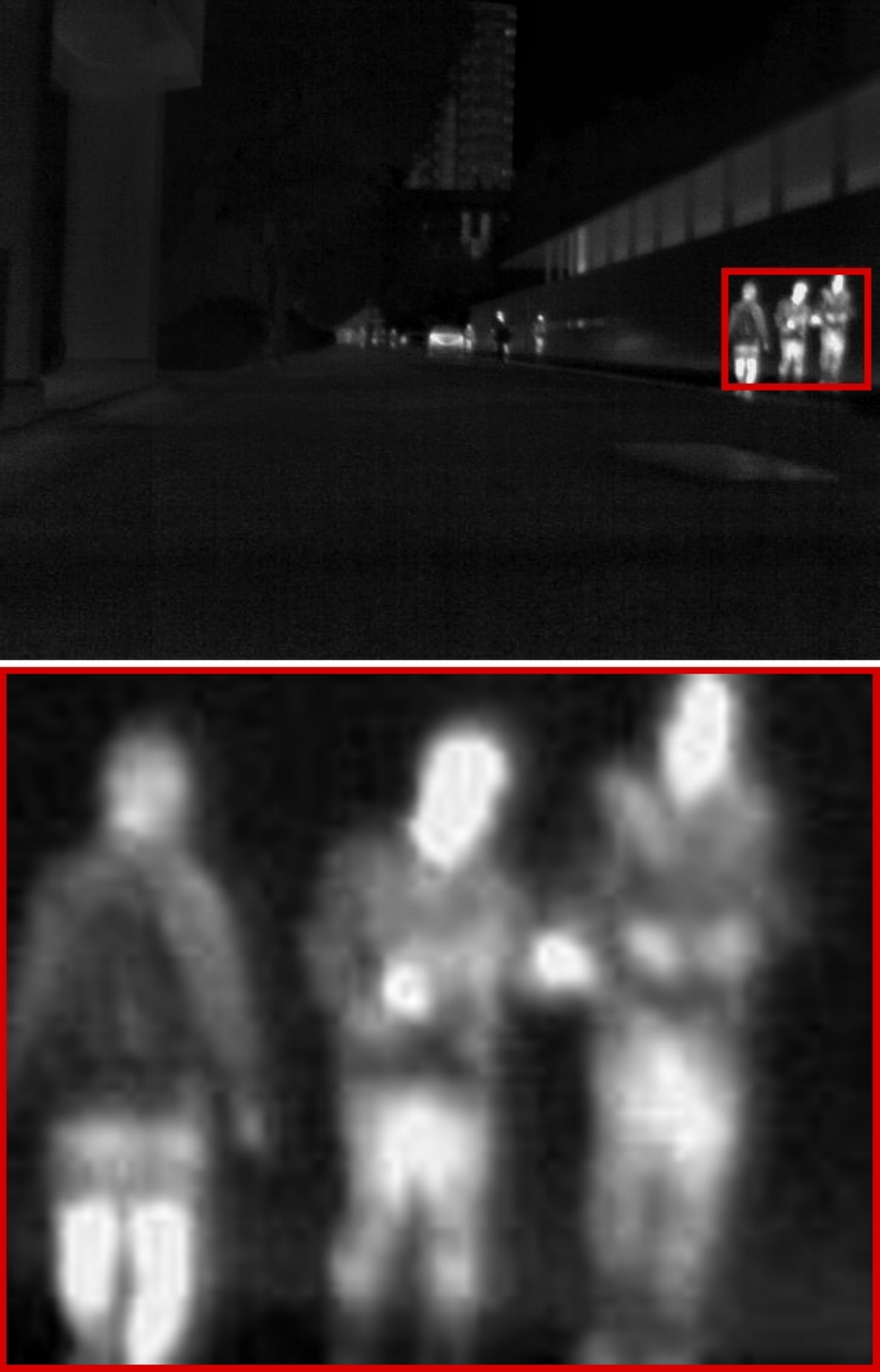}
&	\includegraphics[width=0.108\textwidth]{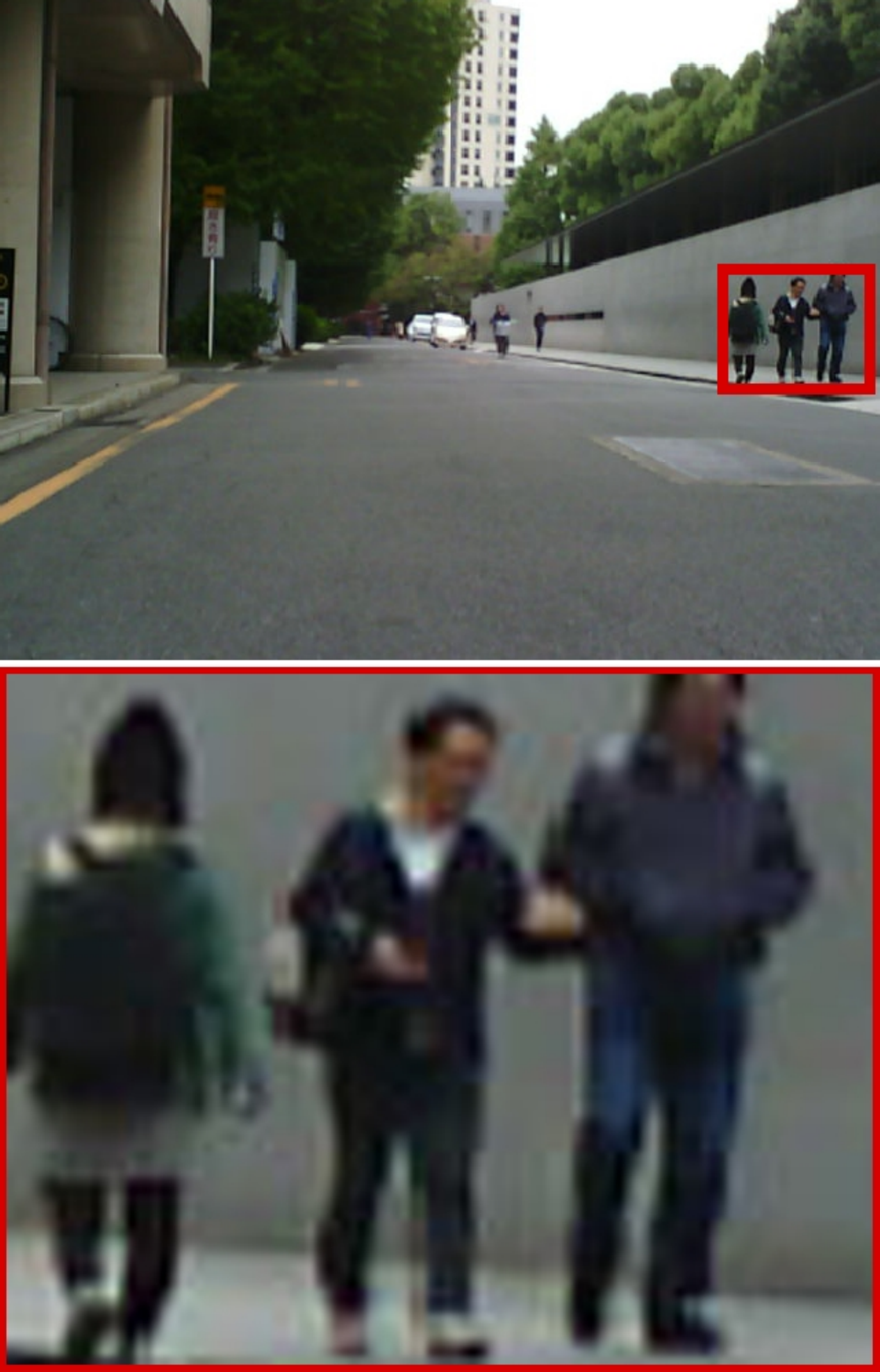}

&	\includegraphics[width=0.1080\textwidth]{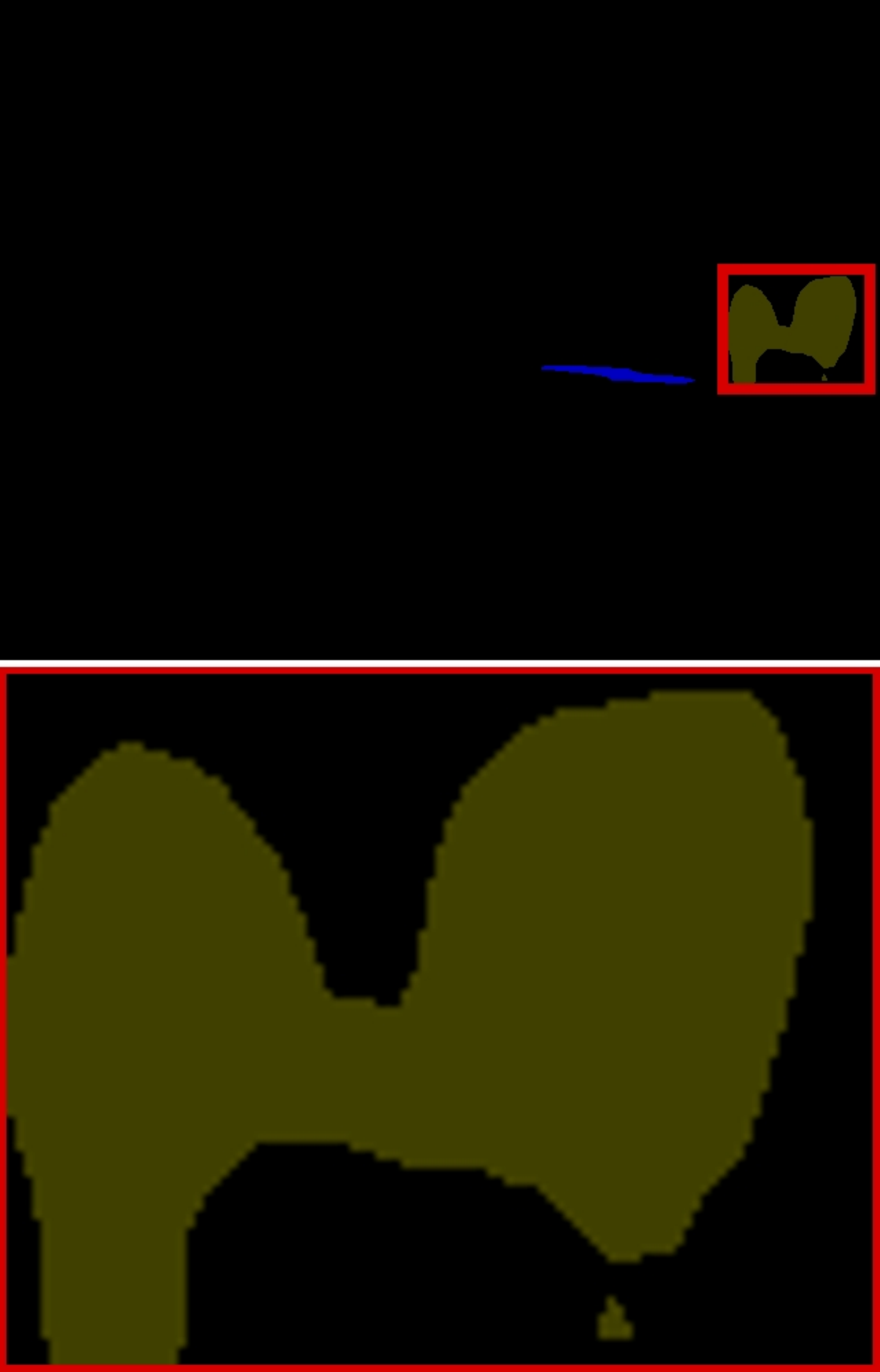}
&	\includegraphics[width=0.108\textwidth]{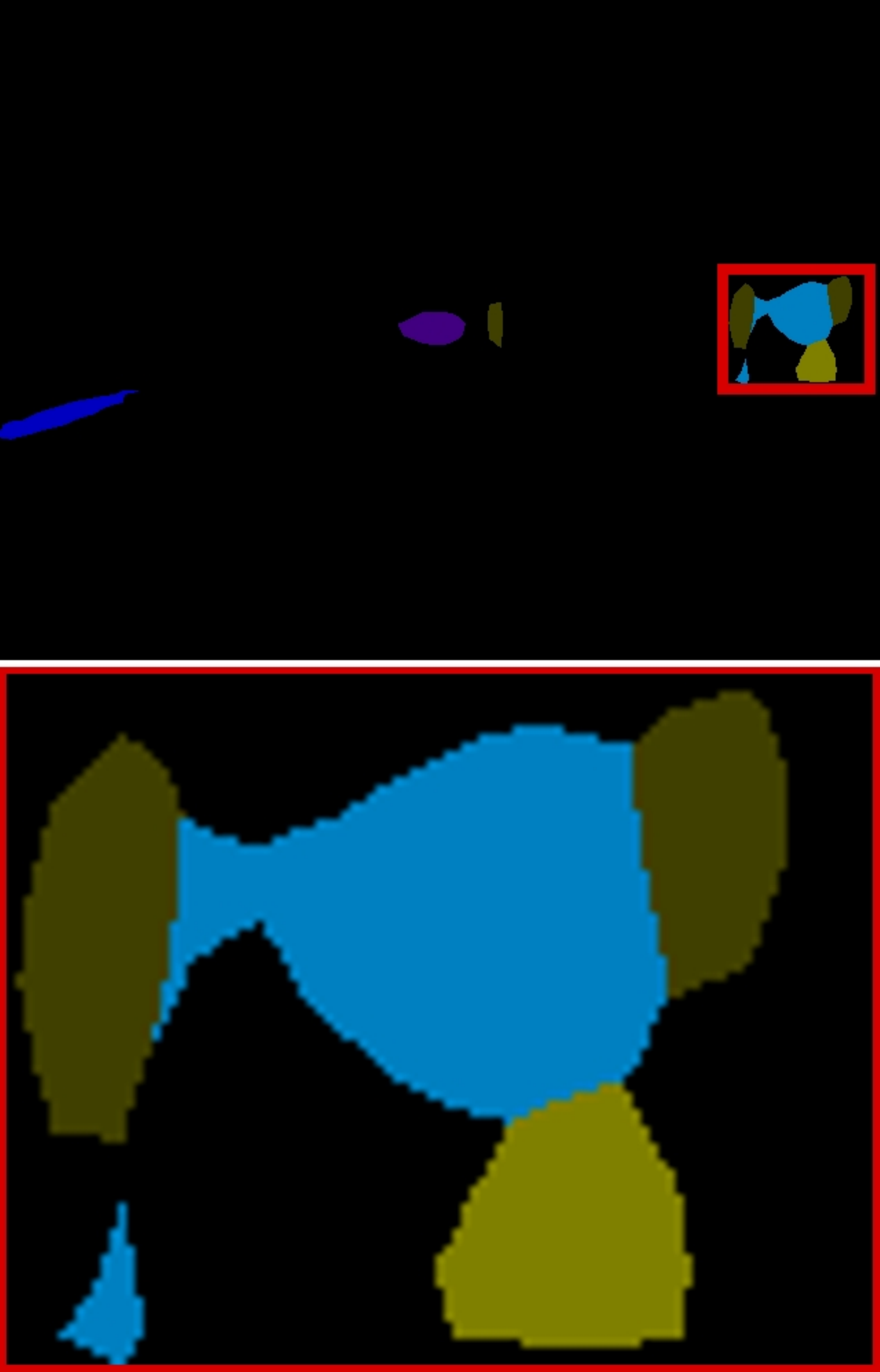}
&	\includegraphics[width=0.108\textwidth]{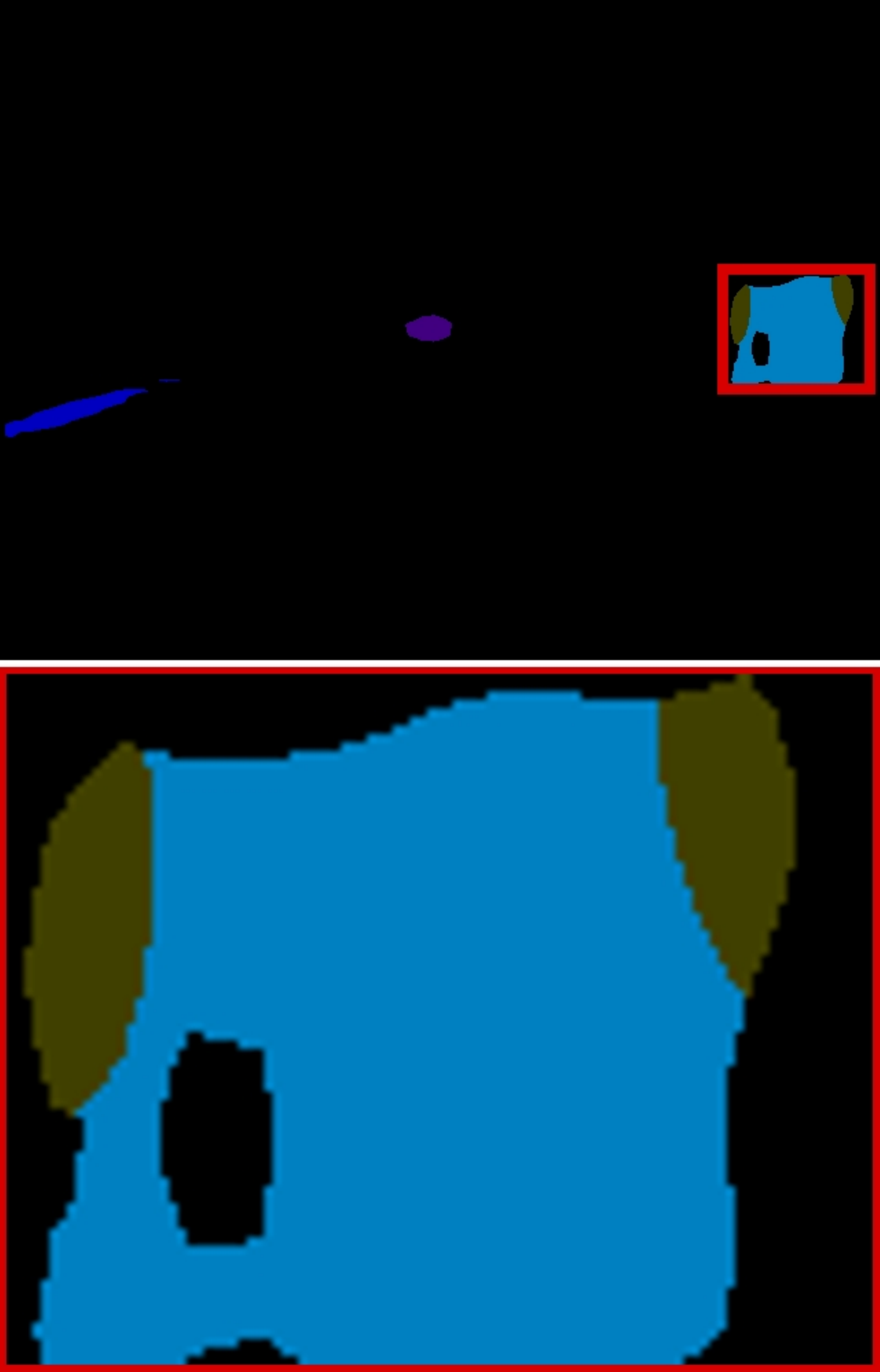}
&	\includegraphics[width=0.108\textwidth]{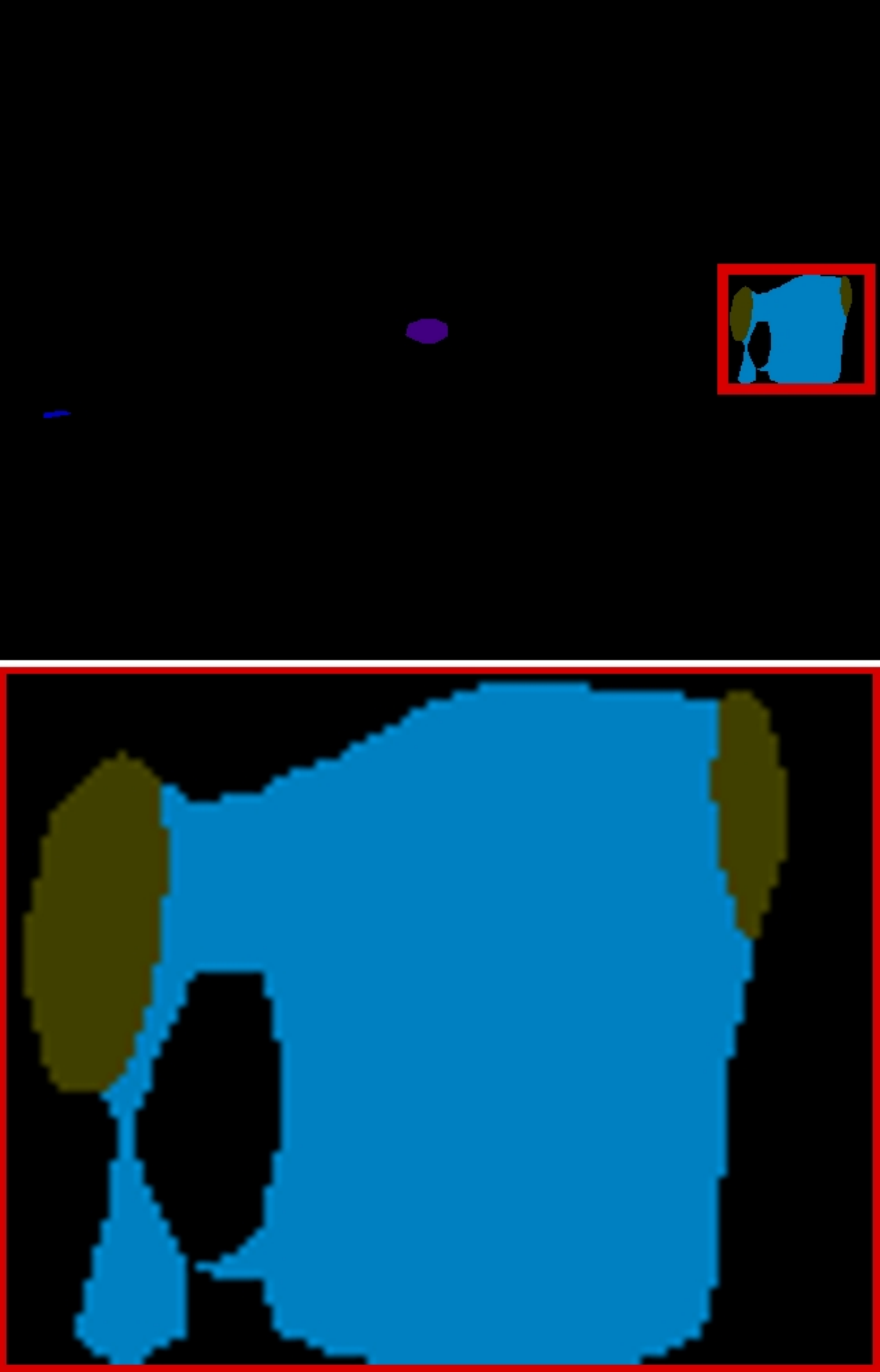}
&	\includegraphics[width=0.108\textwidth]{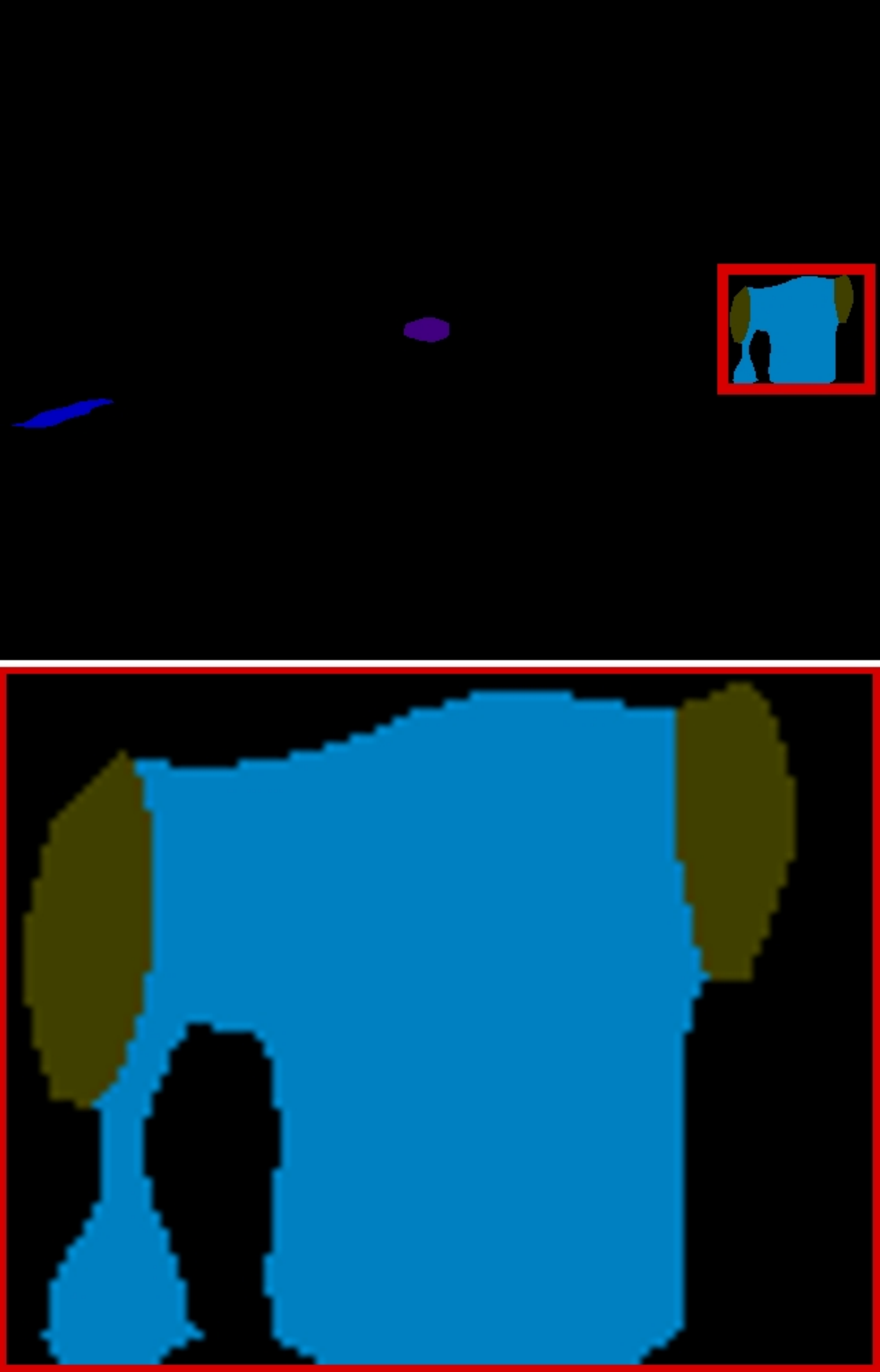}
&	\includegraphics[width=0.1080\textwidth]{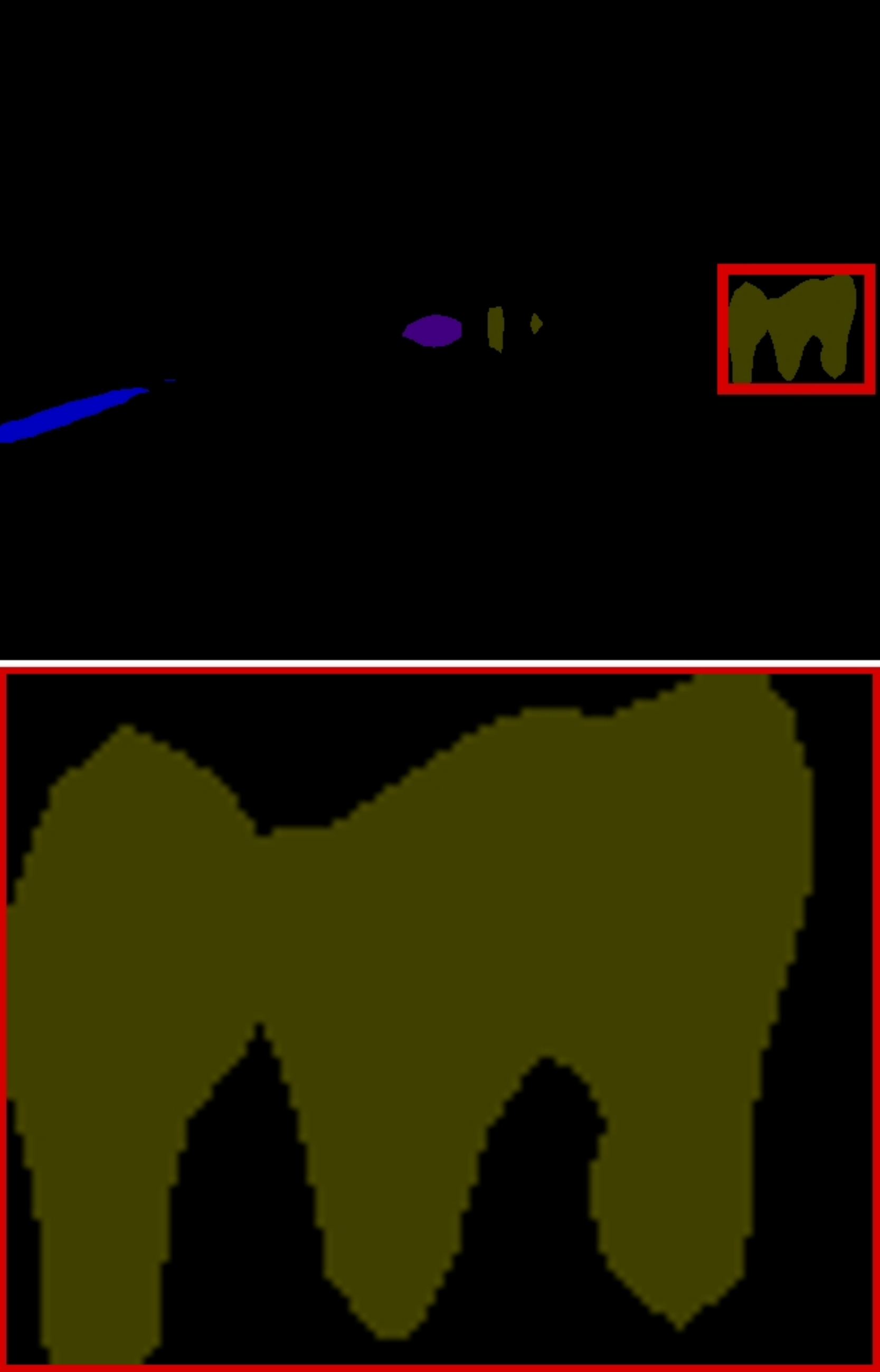}
&	\includegraphics[width=0.108\textwidth]{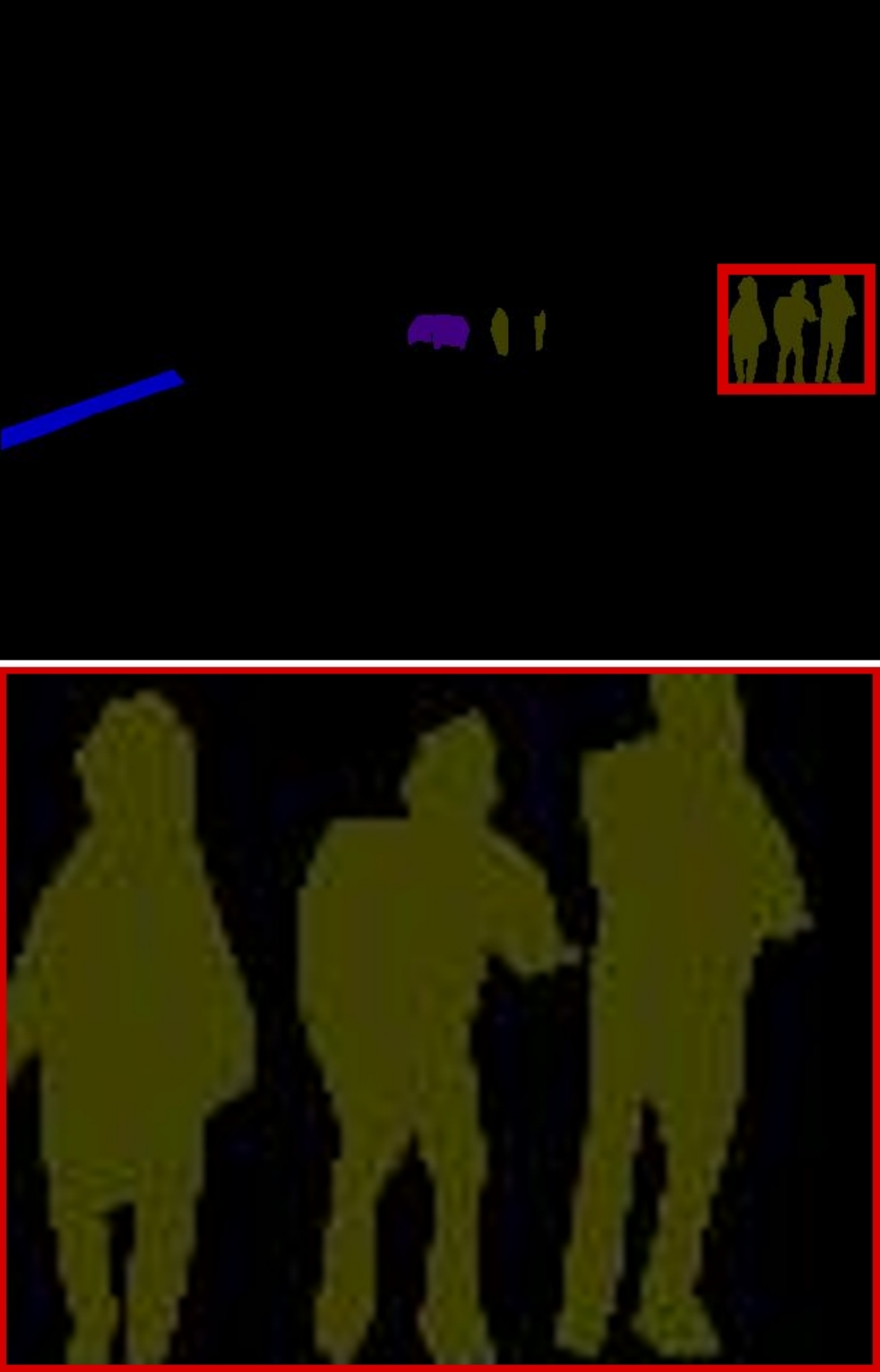}\\

	\footnotesize	Infrared &\footnotesize Visible & \footnotesize Infrared  & \footnotesize Visible &\footnotesize  \footnotesize DenseFuse & \footnotesize RFN &\footnotesize MFEIF&\footnotesize TIM &\footnotesize  Label \\
	\end{tabular}
	\caption{Semantic segmentation results based on image fusion compared with several state-of-the-art fusion methods.}
	\label{fig:result_ir_vis_seg}
\end{figure*}

\subsubsection{Object Detection}

\textbf{Quantitative Comparisons.} As shown in Table.~\ref{tab:data_set_results}, we report the qualitative results for  object detection on~{Multi-Spectral} Dataset. We illustrate the  results of separate detection using RetinaNet~\cite{ross2017focal} based on single inputs, generated by fusion networks. The RetinaNet is trained by fused images based on simple average principle.
 Our framework,  has shown significant improvements against fusion-based methods and single modality images. 
 More specific, existing detection schemes  establish the training and testing on  visible images dataset. Obviously,   the network effectively detects  visible-salient objects under the training of visible images. In contrast, 
infrared imaging contains thermal information, which is benefit for the detection of car engines and human bodies. However, this modality is insensitive for other  weak-thermal objects such as bike and color cone.  Compared with fusion-based methods, our method  fully
integrates the complementary  advantages, which achieves the best precision on person, car and stop.

\textbf{Qualitative Comparisons.} 
Subsequently, we  illustrate the visual comparisons in Fig.~\ref{fig:result_ir_vis_od}.  As shown in the first row, DDcGAN scheme cannot preserve effective edges information of infrared targets, which fails to detect any pedestrians. Our results preserve clear thermal targets (e.g., pedestrians) and ample details. For instance, the example in last row is a challenging scenarios, where the object is not salient either in infrared and visible modality. Our detector can successful detects this object that demonstrated the superiority.

\subsubsection{Semantic Segmentation}
\textbf{Quantitative Comparisons.}  We utilize the searched semantic segmentation network to test ten fusion methods  and provide the detailed  results in Table.~\ref{tab:data_set_results_seg}, which  measured by mean Intersection-over-Union (mIoU) and mean Accuracy (mACC).  From this table, we can observe that our scheme, designed with universal formulation, achieves the highest numerical performance under all eight categories.
 Moreover the pre-trained fusion schemes which have visual-pleasant results or concentrate on statistic metrics cannot be consistent with remarkable segmentation performance.  That also demonstrates the goal of not only ensuring complementary informative fusion but also assisting the improvement of semantic segmentation has been achieved uniformly.

\textbf{Qualitative Comparisons.} Furthermore, Fig.~\ref{fig:result_ir_vis_seg} depicts the visual comparisons with single modality images and other competitive fusion schemes. The results of each modality  fully reflect the complementary feature expression. Obviously, thermal-sensitive objects (e.g., pedestrian on the poor lighting condition) cannot be correctly predicted in other fusion-based schemes. That illustrates our scheme can obtain sufficient infrared salient information for target classification. On the other hand, our scheme also can preserve the texture details well to estimate other  thermal-insensitive objects, such as the bumps shown in the first row and the car in the second row. To sum up, these three representative scenarios effectively display our outperformed results compared with these advanced competitors.

\subsection{Ablation Study}
In order to evaluate the effectiveness of two proposed techniques, we perform relevant ablation analyses. All ablation experiments are based on the infrared-visible fusion vision task. We firstly evaluate the fusion performance under the proposed search  compared with the mainstream search strategy (taking DARTS~\cite{liu2018darts} as an example). Then we  verify the importance of pretext meta initialization. 

\begin{figure}[thb]
	\centering \begin{tabular}{c@{\extracolsep{0.1em}}c}
		
		\includegraphics[width=0.23\textwidth]{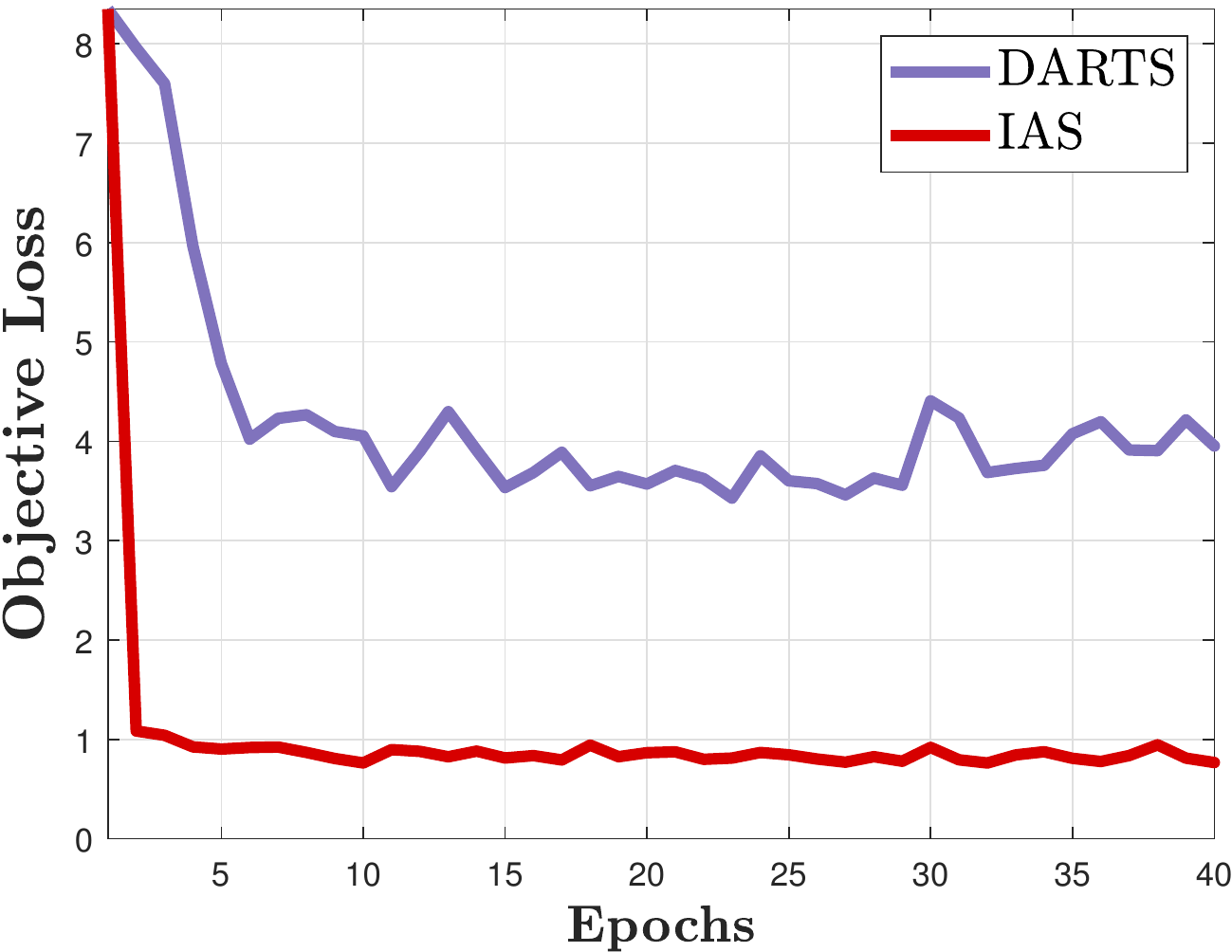}
		&\includegraphics[width=0.23\textwidth]{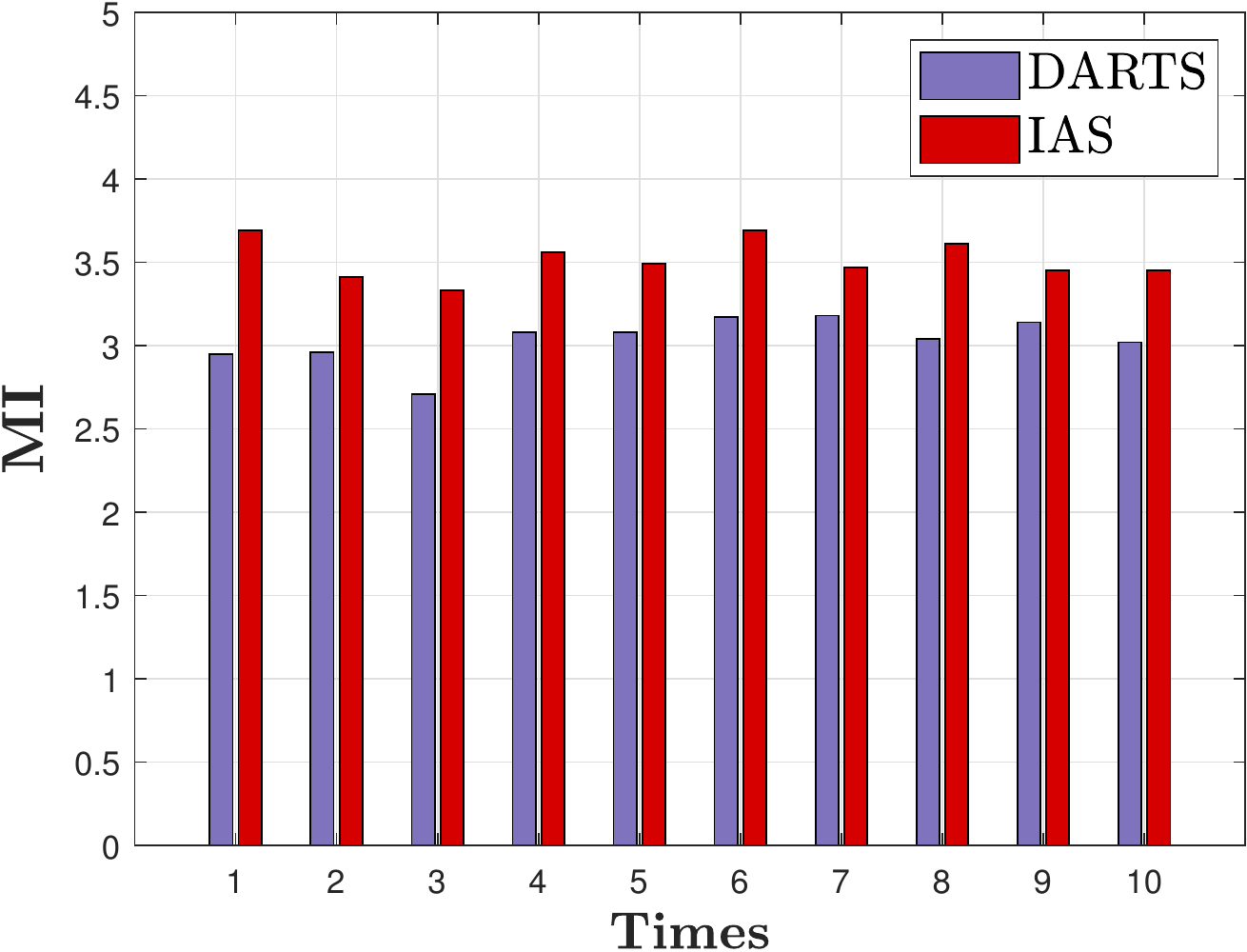}\\
		\footnotesize	(a)   & \footnotesize (b)  \\
	\end{tabular}
	\caption{Comparison with DARTS and proposed IAS. (a) plots the objective losses. (b) depicts the  performance stability of searched architectures by randomly searching 10 times.}
	\label{fig:ab0}
\end{figure}

\begin{figure}[thb]
	\centering \begin{tabular}{c@{\extracolsep{0.1em}}c}
%
		\includegraphics[width=0.23\textwidth]{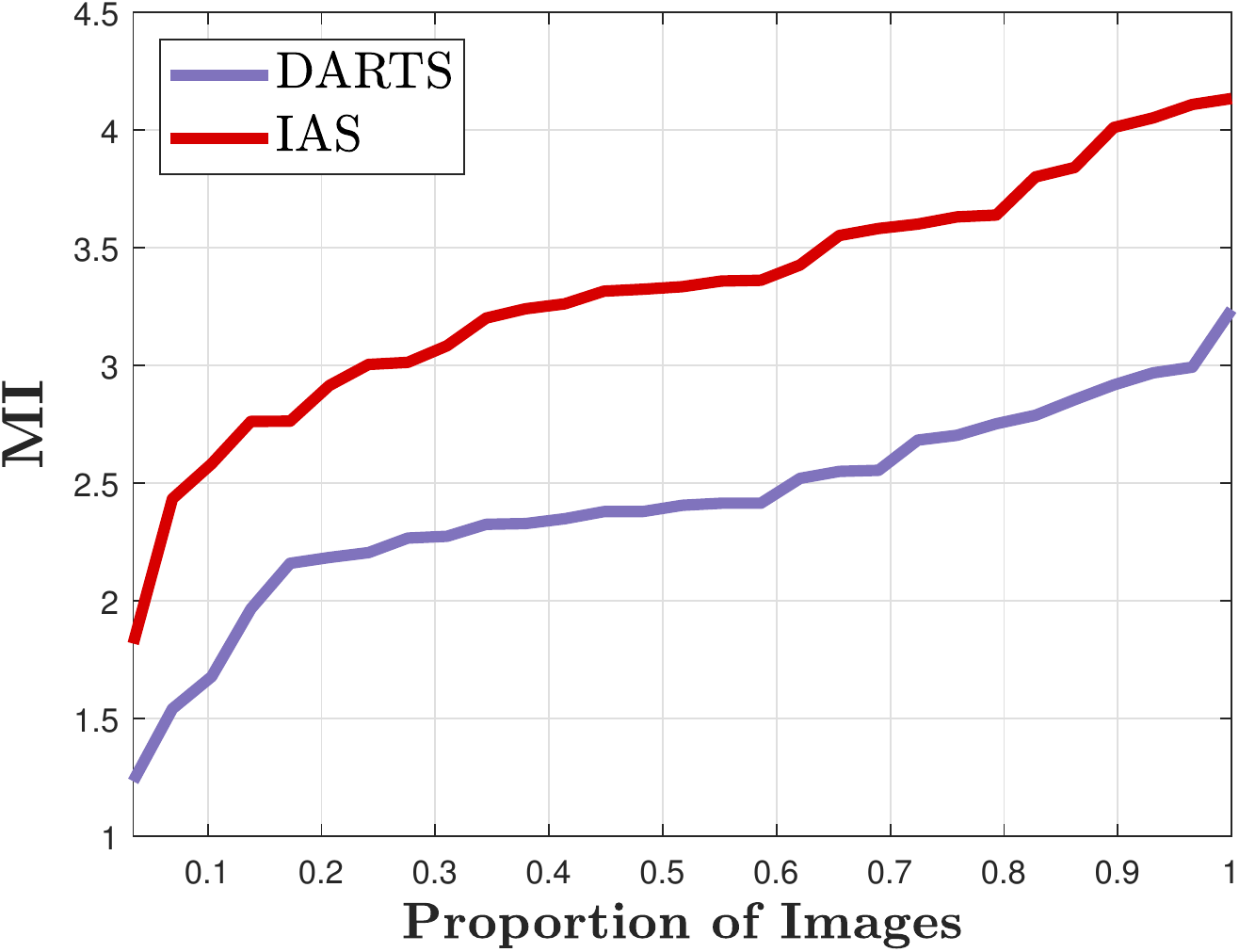}
		&\includegraphics[width=0.23\textwidth]{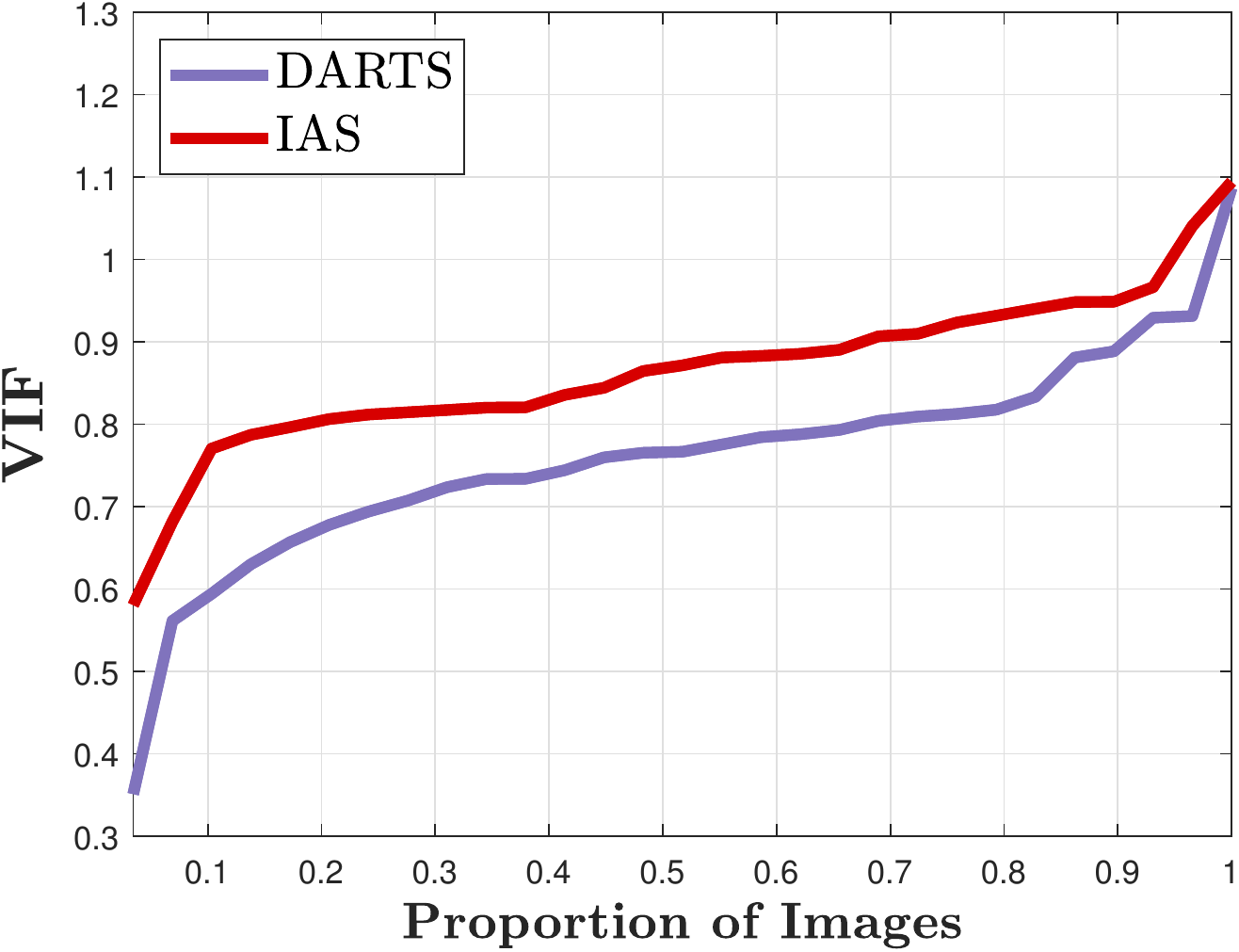}\\
		\footnotesize	(a)  & \footnotesize (b) \\
	\end{tabular}
	\caption{Comparison of the final performance with DARTS.  (a) and (b) depict the quantitative comparisons of fusion quality in terms of two metrics, where the number at x-axis represents the percent of images no more than the value of y-axis.}
	\label{fig:ab01}
\end{figure}

\textbf{Search Strategy.} We perform the comparison of search strategies (i.e., mainstream search DARTS and proposed IAS) on TNO dataset by objective losses ($\log$ formation) and numerical results in Fig.~\ref{fig:ab0}. As shown in subfigure (a), our proposed scheme achieves rapid convergence from  losses based on the accurate gradient estimation, which can support to find better solution of architecture relaxation weights. At subfigure (b), we report the performance of ten architectures based on DARTS and IAS by randomly searching ten times. We can observe that, the performances of DARTS-based architectures are not consistent, which cannot realize the stable performance. Our IAS-based architectures not only achieve the higher numerical results, but also accomplish the stable performance. The qualitative and quantitative results of final performance are reported in Fig.~\ref{fig:ab01} and Table.~\ref{tab:IAS}. Through these comparison, we can conclude  that IAS can realize the high performance, efficiency and stability.
\begin{table}[htb]
	\renewcommand{\arraystretch}{1.3}
	\caption{Evaluation of fusion performance of IAS with DARTS.}
	\label{tab:IAS}
	\centering\footnotesize
	\setlength{\tabcolsep}{4mm}{
		\begin{tabular}{c c c c c}
			\hline
			Strategy &MI& FMI&VIF&  $\mathrm{Q^{AB/F}}$\\ \hline
			DARTS&2.414&0.808&0.759&0.416\\
			IAS &3.285&0.818&0.861&0.458\\
			\hline
		\end{tabular}	
	}
\end{table}
Obviously, network discovered by the proposed search improves  performances under these metrics. 
That also demonstrates the advantages of proposed search strategy. 
 Aiming to compare the single-operator composited architecture fairly, we only leverage $\mathcal{N}_\mathtt{F}$ module to conduct experiments (training with $\ell_{\mathtt{T}}$). We leverage  widely used inner-operators to design heuristic  structures with fixed outer structure (using $\mathbf{C}_\mathtt{{MS}}$ and $\mathbf{C}_\mathtt{{SC}}$). The results are shown in Table.~\ref{tab:single_op}. As for the effectiveness of operators, 3-DB obtains the highest numerical results under MI  metric but has the second slowest inference time. Moreover, constrained by hardware latency and $\lambda=0.5$, our scheme balances  the inference time and performances. 
For the verification  about   effectiveness of trade-off parameter $\lambda$, we also provide two versions with different latency constraints. The results are reported in the Table.~\ref{tab:single_op2}. Obviously, the inference time and parameters are sensitive to the adjustment of $\lambda$. We can observe that with the increase of $\lambda$, the time is reduced with numerical performance.


\begin{table}[htb]
	\renewcommand{\arraystretch}{1.3}
	\caption{The performance of single operator-composited network on {TNO}.}
	\label{tab:single_op}
	\centering 
	\setlength{\tabcolsep}{1.6mm}{
		\begin{tabular}{c c c c c c}
			\hline
			Operator &MI& VIF&  $\mathrm{Q^{AB/F}}$ &Parameters (M)  &  Time (s)\\ \hline
			3-DC    &3.130&0.839&0.412&{0.992}&\textbf{0.0585}\\
			5-DC  &3.087&0.829&0.383&0.997&0.0963\\
			3-RB &3.207&0.874&0.422&0.999&0.0821\\
			5-RB &3.398&0.888&\textbf{0.458}&1.011&{0.0774}\\
			3-DB &\textbf{3.465}&\textbf{0.895}&0.428&\textbf{0.924}&{0.1951}\\
			5-DB &3.415&{0.869}&0.428&0.997&{0.3622}\\
			\hline
		\end{tabular}	
	}
\end{table}
\begin{table}[htb]
	\renewcommand{\arraystretch}{1.3}
	\caption{Effectiveness of hardware regularization on {TNO}.}
	\label{tab:single_op2}
	\centering 
	\setlength{\tabcolsep}{1.8mm}{
		\begin{tabular}{c c c c c c}
			\hline
			$\lambda$ &MI& VIF&  $\mathrm{Q^{AB/F}}$ &Parameters (M)  &  Time (s)\\ \hline
			$\lambda$ = 0 &\textbf{3.569}&\textbf{0.894}&\textbf{0.465}&{0.924}&{0.1532}\\
			$\lambda$ = 0.5 &{3.438}&{0.888}&{0.441}&0.995&{0.0769}\\
			$\lambda$ = 2 &{3.416}&{0.816}&{0.456}&\textbf{0.127}&\textbf{0.0012}\\
			\hline
		\end{tabular}	
	}
\end{table}
\begin{table}[htb]
	\renewcommand{\arraystretch}{1.3}
	\caption{Evaluation of update number ${K}$  for pretext meta initialization.}
	\label{tab:update_k}
	\centering\footnotesize
	\setlength{\tabcolsep}{2.8mm}{
		\begin{tabular}{c c c c c}
			\hline
			Numbers &MI& FMI&VIF&  $\mathrm{Q^{AB/F}}$\\ \hline
			w/o Initialization&3.215&0.808&0.849&0.443\\
			$K$ = 2  &3.281&0.810&0.839&0.444\\
			$K$ = 4 &\textbf{3.285}&\textbf{0.818}&\textbf{0.861}&\textbf{0.458}\\
			$K$ = 6 &3.073&{0.805}&0.797&0.422\\
			$K$ = 8 &3.093&0.810&0.795&0.434\\
			$K$ = 10 &3.128&0.809&0.833&0.442\\
			\hline
		\end{tabular}	
	}
\end{table}
\begin{table}[htb]
	\renewcommand{\arraystretch}{1.3}
	\caption{Numerical comparison between image fusion with direct fusion and proposed task-guided fusion among four representative vision tasks.}
	\label{tab:task}
	\centering\footnotesize
	\setlength{\tabcolsep}{2.1mm}{
		\begin{tabular}{c|cc|cc}
			\hline
			\multirow{2}{*}{\footnotesize Task} & \multicolumn{2}{c|}{\begin{tabular}[c]{@{}c@{}} \footnotesize  Visual Enhancement\end{tabular}} & \multicolumn{2}{c}{\begin{tabular}[c]{@{}c@{}} \footnotesize Semantic Understanding\end{tabular}} \\ \cline{2-5} 
			& \multicolumn{1}{c|}{\footnotesize IVIF}   & \footnotesize MIF      & \multicolumn{1}{c|}{\footnotesize Detection}      & \footnotesize Segmentation    \\ \hline
			\footnotesize	Metrics               & \multicolumn{1}{c|}{MI}                            & \footnotesize MI                           & \multicolumn{1}{c|}{mAP}    & \footnotesize mIOU\\ \hline
			\footnotesize	Direct Fusion   & \multicolumn{1}{c|}{2.141}                              &        1.852                      & \multicolumn{1}{c|}{0.447}                               &       0.575                         \\\hline
			
			TIM                  & \multicolumn{1}{c|}{\textbf{3.285}}                              &   \textbf{2.359}                           & \multicolumn{1}{c|}{\textbf{0.476}}                               &            \textbf{0.685}                 \\ \hline
		\end{tabular} 
	}
\end{table}
\begin{figure}[thb]
	\centering \begin{tabular}{c@{\extracolsep{0.1em}}c}
		
		\includegraphics[width=0.24\textwidth]{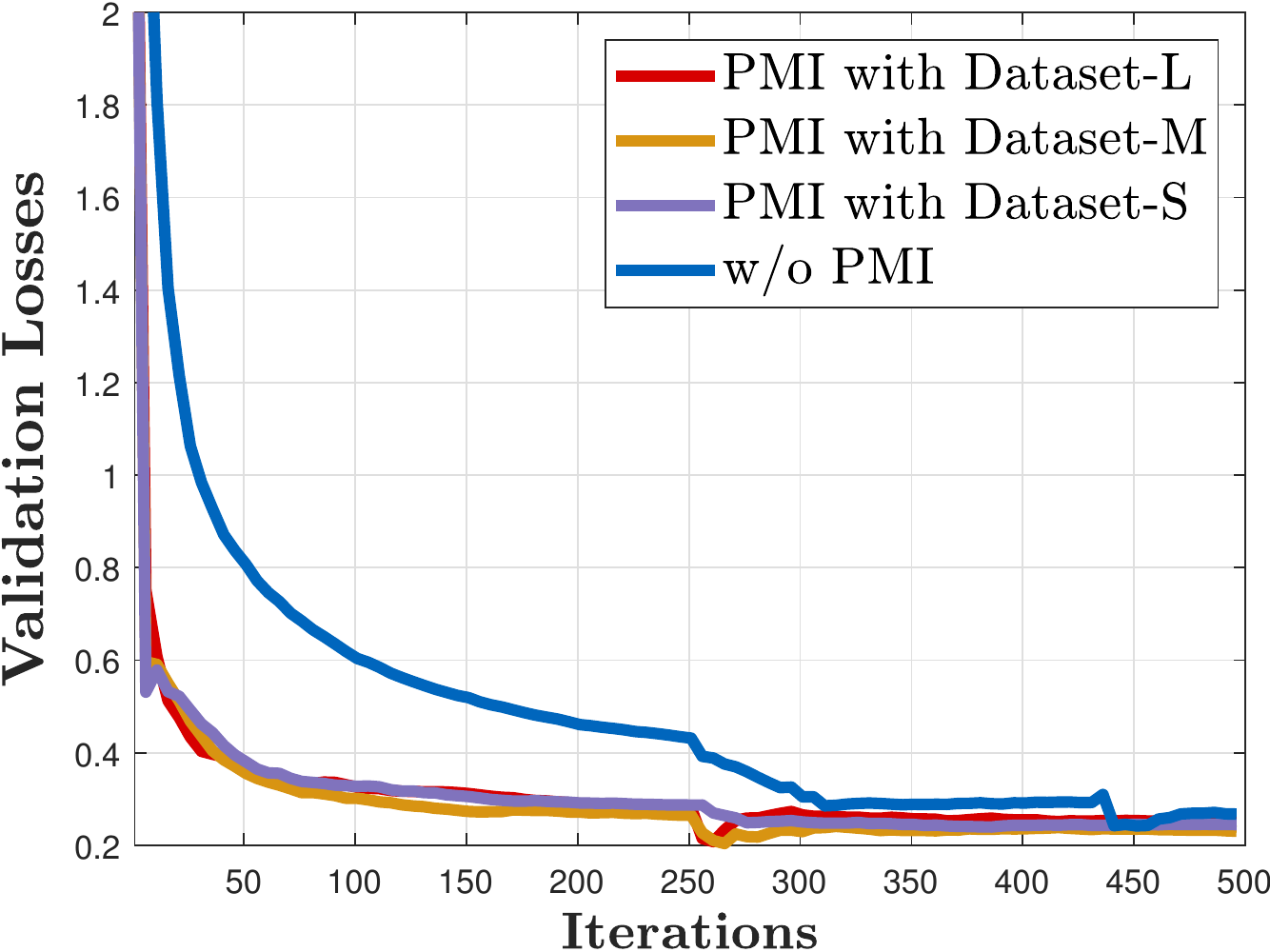}
		&\includegraphics[width=0.235\textwidth]{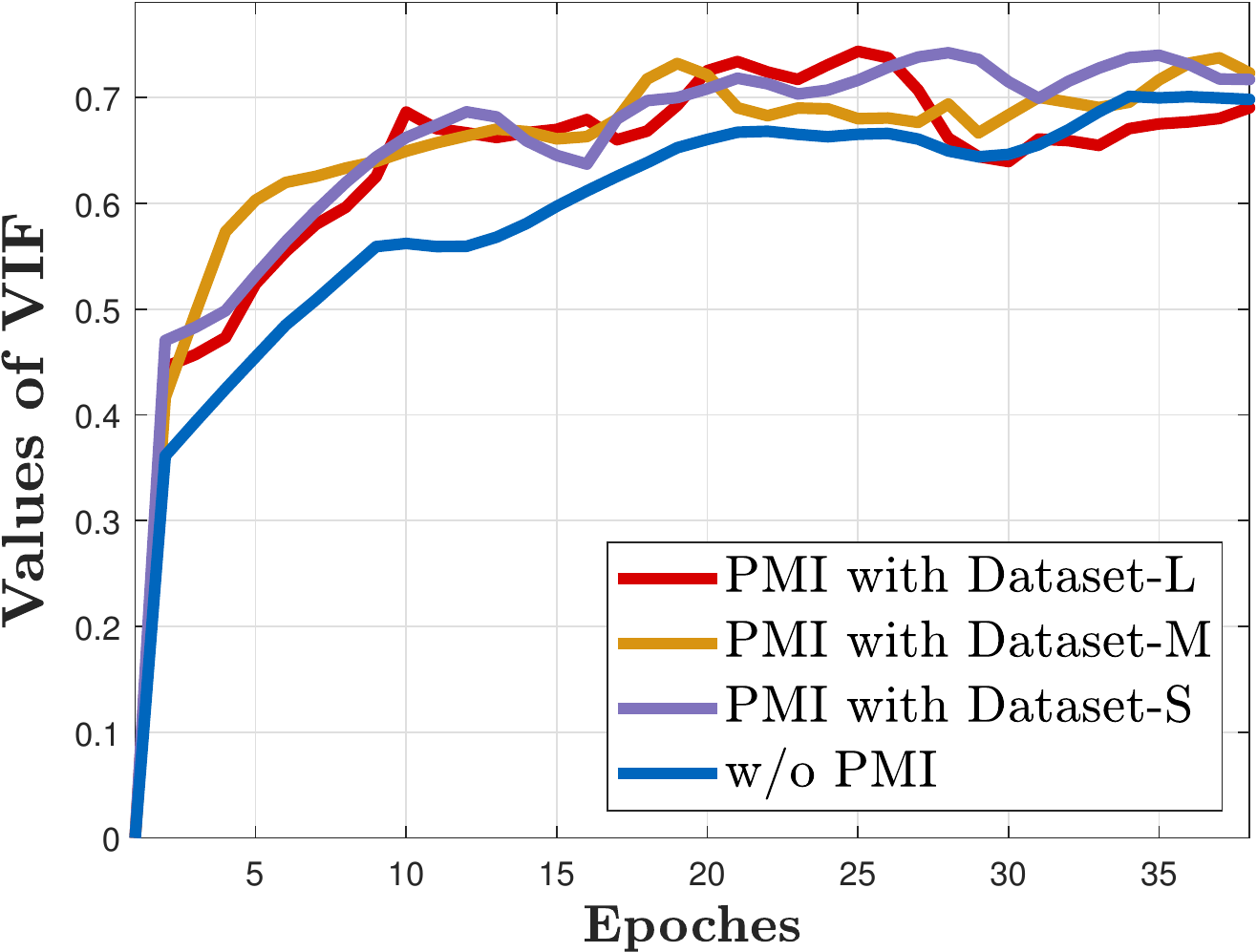}\\

		\footnotesize   (a)  &  \footnotesize (b)  \\		
	\end{tabular}
	\caption{Illustration of the effectiveness of PMI with diverse scale of training data. The  curves of losses and VIF metrics on the validation datasets are plotted at (a) and (b) respectively.}
	\label{fig:ab3}
\end{figure}

\textbf{Meta Initialization.} PMI strategy is proposed to achieve the generalization parameters for the fast adaptation of image fusion.
We conducted the experiments to verify the influences  of proposed training strategy and discuss the optimal inner updates (i.e., $K$) of initialization in Table.~\ref{tab:update_k}. ``w/o Initialization'' denotes the version that is performed with end-to-end training directly with specific data.
Evidently, suitable $K$ can improve  final numerical performances significantly. When $K=4$, we can obtain the comprehensive numerical results, benefiting from the intrinsic fusion feature learning from multi-tasks and multi-data distribution. Especially,  increasing the number of inner updates cannot always strength the performance. From this table, we can conclude that pretext meta initialization can effectively learn the inhere fusion features, which can  improve the performance of image fusion remarkably. Furthermore, we also plot the curve of losses and related fusion quality (measured by VIF) in Fig.~\ref{fig:ab3}. We can observe that the variant with PMI has more lower validation loss and converges faster to the 
stable stage compared with the version ``w/o PMI''. On the other hand, our scheme with PMI can quickly achieve the best VIF metric, which represents the robust visual quality. More importantly, we also demonstrate that, only using partially few training data of specific tasks, we also can obtain the significant numerical results. As shown in  Fig.~\ref{fig:ab3}, we also illustrate the remarkable results based on PMI with 
different scales of training data. Dataset-L, Dataset-M and Dataset-S include 6195, 3097 and 1548 pairs of patches for IVIF. As shown at subfigure (b), PMI with utilizing large dataset has the slower convergence. The variant with small dataset cannot maintain the stable stage and is easy to generate oscillation.  Thus, considering the training efficiency and quality, we select 3097 pairs of patches to train the fusion.
For further evaluating the role of initialization, we compare with the version based on the direct fusion. This version  is to utilize the original training strategy (only with $\ell_{\mathtt{F}}$) to generate  fused images. 
  Numerical comparison among four tasks is reported in Table.~\ref{tab:task}. Obviously, task-oriented fusion can effectively improve the performance of different  tasks. We can summarize that PMI is benefit for the task-guided fusion of visual effects and semantic understanding.
\section{Concluding Remarks}
In this paper, we developed a generic task-guided image fusion. Based on a constrained strategy, we realized the flexible learning paradigm to guide image fusion, incorporating information from downstream vision tasks.   The implicit architecture search strategy was proposed to discover nimble and effective fusion networks. We also introduced the pretext meta initialization strategy to endow the fast adaptation of image fusion with multiple fusion scenarios.  Comprehensive qualitative and quantitative results among various visual enhancement and semantic understanding tasks demonstrated the superiority. Furthermore,  implicit  search strategy is also capable of the architecture construction for more unsupervised vision tasks. This constrained paradigm can be extended to other visual applications (e.g., image restoration and semantic understanding).

\ifCLASSOPTIONcompsoc
  \section*{Acknowledgments}
\else
  \section*{Acknowledgment}
\fi

This work is partially supported by the National Key R\&D Program of China (2020YFB1313503), the National Natural Science Foundation of China (Nos. U22B2052  and 61922019),   the Fundamental Research Funds for the Central Universities and the Major Key Project of PCL (PCL2021A12).

\ifCLASSOPTIONcaptionsoff
  \newpage
\fi

\bibliographystyle{IEEEtran}
\bibliography{reference}

\end{document}